%% file: main.tex
\documentclass{article}

\usepackage[final]{corl_2022} 

\include{imports/math}
\include{imports/tikz}
\include{imports/references}

\title{Deep Black-Box Reinforcement Learning with Movement Primitives}

\author{
Fabian Otto$^{1,2}$, Onur Celik$^{3}$, Hongyi Zhou$^{3}$, Hanna Ziesche$^{1}$, Ngo Anh Vien$^{1}$,\\ and \textbf{Gerhard Neumann$^{3}$} \\
\\
$^{1}$ Bosch Center for AI, Germany \\
$^{2}$ University of Tübingen, Germany \\
$^{3}$ Autonomous Learning Robots, KIT, Germany \\
\texttt{fabian.otto@bosch.com} \\
}

\begin{document}
\maketitle


\begin{abstract}
    \textit{Episode-based reinforcement learning} (ERL) algorithms treat \textit{reinforcement learning} (RL) as a black-box optimization problem where we learn to select a parameter vector of a controller, often represented as a movement primitive, for a given task descriptor called a context.
    ERL offers several distinct benefits in comparison to step-based RL. 
    It generates smooth control trajectories, can handle non-Markovian reward definitions, and the resulting exploration in parameter space is well suited for solving sparse reward settings. 
    Yet, the high dimensionality of the movement primitive parameters has so far hampered the effective use of deep RL methods. 
    In this paper, we present a new algorithm for deep ERL. 
    It is based on differentiable trust region layers, a successful on-policy deep RL algorithm. 
    These layers allow us to specify trust regions for the policy update that are solved exactly for each state using convex optimization, which enables policies learning with the high precision required for the ERL. 
    We compare our ERL algorithm to state-of-the-art step-based algorithms in many complex simulated robotic control tasks. 
    In doing so, we investigate different reward formulations - dense, sparse, and non-Markovian. 
    While step-based algorithms perform well only on dense rewards, ERL performs favorably on sparse and non-Markovian rewards. 
    Moreover, our results show that the sparse and the non-Markovian rewards are also often better suited to define the desired behavior, allowing us to obtain considerably higher quality policies compared to step-based RL.
\end{abstract}



\section{Introduction}

\textit{Reinforcement learning} (RL) problems can be viewed from a step-based \citep{Haarnoja2018,Schulman2017,Fujimoto2018} and an episode-based perspective \citep{Deisenroth2013,Abdolmaleki2015,Daniel2012}. 
In the former, most commonly found in deep RL, a policy selects an action for each state of the trajectory. 
Consequently, step-based RL requires Markovian reward definitions. 
Moreover, the exploration in action space often induces a random walk that inadequately explores the entire behavior space of the agent. 
As a result, most step-based RL algorithms only work well with dense rewards, where the agent receives a reward signal at each time step, rather than only at the last step. 
We refer to the latter as a sparse reward setting.

In the \textit{episode-based reinforcement learning} (ERL) perspective, we choose the complete behavior during the episode with respect to the controller parameters in the beginning of the episode. 
Typically, simple controllers are used that are valid only for executing a single trajectory, such as \textit{movement primitives} (MPs). 
At the beginning of an episode, the MP parameters are adapted to a context vector that serves as task descriptor and may contain \eg different initial joint configurations, the goal to reach, or start positions as well as object and obstacles locations. 
ERL allows for efficient exploration of the behavior space because exploration is implemented in the parameter space of the MP, allowing these algorithms to learn from sparse or even non-Markovian rewards.
In addition, ERL inherently generates smooth control trajectories due to the use of MPs, which often leads to more energy efficient behavior.
However, ERL has so far not gained much popularity as it does not exploit the time-series structure of the RL problem and is therefore considered less data efficient, at least in the dense reward setting. 
So far, there is no deep learning method to adapt the MP parameters to the context, which limits ERL to linear adaptation strategies \citep{Abdolmaleki2017,Tangkaratt2017}.

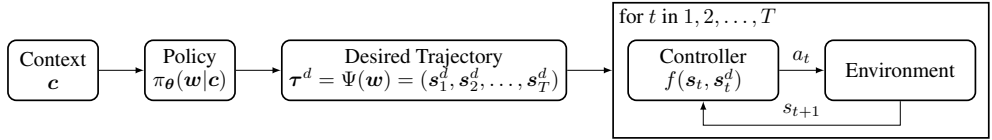
\begin{figure}[t]
    \centering
	\resizebox{0.95\textwidth}{!}{%
		\input{tikz_plots/diagrams/overview}
	}%
	\caption{Overview of the proposed Black-Box Reinforcement Learning (BBRL) framework. The normal distributed policy predicts, given the context $\bm{c}$, the parameters of a movement primitive that translates to a desired trajectory $\tau^d$. 
	A trajectory tracking controller $f$ generates low-level actions $a_t$ given the current $s_t$ and the desired state $s_t^d$ from the trajectory $\tau^d$.
	\label{fig:schema}}
\end{figure}

This paper offers two main contributions: (i) We propose a new algorithm for deep ERL.
The high-dimensional action space imposed by the MP parameters requires highly accurate action selection, hence robust and precise policy updates are needed.
We adapt a recent step-based RL algorithm based on differentiable \textit{trust region projection layers} (TRPLs) \citep{Otto2021} to the ERL setting.
Unlike methods such as \textit{proximal policy optimization} (PPO) \citep{Schulman2017}, that are based on approximated trust regions, TRPL offers exact trust regions for the policy updates which is essential for learning precise policies in high-dimensional action spaces.
As second contribution, (ii) we show that ERL can be superior to step-based RL for tasks that are more easily defined by sparse or non-Markovian rewards than by a dense reward. 
To this end, we have investigated different reward formulations for a variety of complex simulated robot control tasks. 
Our experiments show that step-based algorithms have difficulties learning high-quality policies in sparse or non-Markovian reward settings, while our proposed algorithm scales well to these reward settings. 
We further show that dense Markovian rewards can be very restrictive in terms of specifying a desired behavior in our domains. 
For example, if we want to achieve energy efficient, time-optimal reaching motions, sparse rewards can be used that only penalize the distance to the goal in the last time step.
Non-Markovian rewards are useful for tasks with inherent non-Markovian optimality descriptors, \eg, the maximum height during a jump or the minimum distance of a bat to a ball. 
As a consequence, our algorithm can master tasks where step-based RL performs poorly even after intensive engineering of the used dense rewards. 

In summary, our findings show that our algorithm has the following benefits: (a) it produces inherently smooth control policies, (b) offers efficient exploration 
in parameter space, which in turn enables learning with (c) sparse and non-Markovian rewards. 
These rewards also (d) allow for a more direct specification of the desired behavior, which leads to (e) higher quality of the learned behavior. 
These benefits come at the cost of an increased data complexity (a factor of 2 to 3).


\section{Background and Related Work}
{\color{black}\textbf{Black-Box Reinforcement Learning.} }
Contextual episode-based policy search \cite{Deisenroth2013,Tangkaratt2017} treats RL as black-box optimization problem where a contextual search distribution $\pi(\bm{w}|\bm{c})$ over the controller parameters $\bm{w}$ is optimized to maximize the expected return $R(\bm{w},\bm{c}),$ \ie, 
\begin{align*}
    \argmax_{\pi(\bm{w}|\bm{c})} \mathbb{E}_{p(\bm{c})}\left[\mathbb{E}_{\pi(\bm{w}|\bm{c})}[R(\bm{w},\bm{c})] \right],
    \label{eq:eps_obj_context}
\end{align*}
where $p(\bm{c})$ denotes the context distribution given by the task. 
Due to the black-box nature of the problem, no structural assumptions are made for the return function $R(\bm{w},\bm{c})$, \ie, the return can be any non-Markovian function of the resulting trajectory. 
Most ERL algorithms only consider the non-contextual setting, where different optimization techniques have been used, such as policy gradients \cite{Sehnke2010}, natural gradients \cite{Wierstra2014}, stochastic search strategies \cite{Hansen2001,Mannor2003,Abdolmaleki2019}, or trust-region optimization techniques \cite{Abdolmaleki2015,Daniel2012,Tangkaratt2017}. 
The only methods that incorporate context adaptation \cite{Tangkaratt2017,Abdolmaleki2019} consider a linear mapping from context to parameter space. 
We, in contrast, consider highly nonlinear context-parameter relationships using deep learning. 

{\color{black} As alternative to gradient- and step-based methods recent works \citep{Mania2018,Salimans2017,Chrabaszcz2018} also proposed using a full gradient-free black-box approach for finding the optimal neural network parameters.
They typically consider learning the neural network policy parameters as the black-box optimization problem and leverage the episode performance for evaluating network configurations. 
While these methods can be competitive for some tasks, they are still difficult to use for most contextual setups as they require rollouts with different neural network parameters to be comparable. 
Yet, in the contextual case the performance of the rollout does not only depend on the neural network parameters but also on the context (\eg the goal) which makes the evaluation much more noisy. 
For example, a good neural network parametrization could still yield a poor performance because it has been evaluated in a hard context. 
The above approaches are completely ignorant to the context and, therefore, cannot attribute the performance differences to the context. 

In contrast, our setting does not perform black-box optimization on the level of a global neural network control policy with several thousand parameters but on the level of local control parameters (in the range of 20-50 dimensions). 
While we, unlike the above linear approaches, still use a neural network policy with a high number of parameters to adapt the local control parameters to the context, parameters are \textit{not} treated as a black-box. 
Instead, the DNN policy is updated using policy gradient, which utilizes the context information as well as the derivatives of the policy.

}

\textbf{Step-Based Reinforcement Learning.}
Unlike episode-based methods, step-based approaches maximize the expected return by optimizing an action-selection policy that chooses a new action in each time step of the trajectory. 
While the goal remains to learn optimal trajectories, they interact directly with the environment based on raw actions, such as positions, velocities, or torques generated from the current state information. 
One such method is the trust region policy optimization algorithm \citep{Schulman2015}, which first introduced trust region methods to deep RL, but is rather complex and difficult to scale.
PPO \citep{Schulman2017} simplifies the update by introducing a clipping heuristic to the policy gradient objective. 
An alternative to the above policy gradient methods are policy iteration based approaches \citep{Song2020}.


\textbf{Differentiable Trust Region Layer.}
Differentiable \textit{trust region projection layers} (TRPLs) \cite{Otto2021} present a scalable and mathematically sound approach for enforcing trust regions in step-based deep RL.
While PPO \cite{Schulman2017} is also motivated by trust-region updates of the policy, it cannot enforce the trust region exactly. 
TRPL provides an efficient way to enforce a trust region as well as more stability and control during training, while reducing the dependency on code level choices \citep{Engstrom2020}.

The layer takes the output of a standard Gaussian policy as input in terms of mean $\bm{\mu}$ and variance $\bm{\Sigma}$ and projects it into the trust region if the given mean and variance violate their respective bounds. 
This projection is done for each input state individually. 
Subsequently, the projected Gaussian policy distribution with parameters $\tilde{\bm{\mu}}$, $\tilde{\bm{\Sigma}}$ is used for any further steps, \eg for sampling and/or loss computation.
Formally, the layer solves the following two optimization problems for each state $\bm{s}$
\begin{equation*}
    \label{eq:generic_projection}
    \begin{split}
        \argmin_{\til{\bm{\mu}}_s} d_\textrm{mean} \left(\til{\bm{\mu}}_s, \bm{\mu}(s) \right),  \quad &\st \quad d_\textrm{mean} \left(\til{\bm{\mu}}_s,  \old{\bm{\mu}}(s) \right) \leq \epsilon_{\bm{\mu}}, \quad \textrm{and}\\ 
        \argmin_{\til{\bm{\Sigma}}_s} d_\textrm{cov} \left(\til{\bm{\Sigma}}_s, \bm{\Sigma}(\bm{s}) \right), \quad &\st \quad d_\textrm{cov} \left(\til{\bm{\Sigma}}_s, \old{\bm{\Sigma}}(\bm{s}) \right) \leq \epsilon_\Sigma,
    \end{split}
\end{equation*}
where $\tilde{\bm{\mu}}_s$ and $\tilde{ \bm{\Sigma}}_s$ are the optimization variables for input state $\bm{s}$ and $\epsilon_\mu$ and $\epsilon_\Sigma$ are the trust region bounds for mean and covariance, respectively.
Finally,  $d_\textrm{mean} $ and $d_\textrm{cov}$ are the similarity metrics for the mean and covariance of a decomposable distance or divergence measure. 
\citet{Otto2021} proposed three such measures, we will, however, only leverage the decomposed KL-divergence in this work.
The trust region for the KL-divergence measure can be made fully differentiable as shown in \cite{Otto2021} and is also explained in more detail in \Apxref{app:kl_trpl}. 

\textbf{Reinforcement Learning with Movement Primitives.}
While most work on RL with movement primitives \citep{Abdolmaleki2015,Kober2008,Stulp2012,Stulp2012b} concentrates on learning a single MP parameter vector for a single task configuration, a few methods allow linear adaptation \citep{Daniel2012, Kupcsik2017,celik2022specializing} of the MP's parameter vector to the context. 
In terms of step-based deep RL, \citet{Bahl2020} propose \textit{neural dynamic processes} (NDP). 
Their goal is to embed the structure of \textit{dynamical movement primitives} (DMPs) into deep policies by reparametrizing action spaces via second-order differential equations. 
This can be seen as intersection between step-based and trajectory methods by learning subtrajectories via DMPs spanning multiple timesteps.
Their approach allows for effective replanning, but unlike our approach, their main exploration still takes place at the action level rather than at the trajectory level. 
Hence, equivalent to standard step-based approaches, they have difficulties with sparse and non-Markovian rewards. 
Lastly, using DMPs requires several numerical integration steps that also need to be differentiated, which is computationally expensive.


\section{Deep Black-Box Reinforcement Learning}

Formally, we learn a policy $\pi_{\bm{\theta}}(\bm{w} \vert \bm{c})$ that models a distribution over the MP parameters $\bm{w}$ based on the context information $\bm{c}$.
For most traditional RL tasks, the context space is often a subset of the observations space.
As we learn a policy in the parameter space $\bm{w}$, exploration is also taking place at this level.
Instead of sampling actions in each time step, we sample a parameter vector $\bm{w}$ only in the beginning of each episode. 
This leads to exploration at the level of desired trajectories and thus to much smoother and temporally correlated exploration.
An overview is shown in \Figref{fig:schema}.

\subsection{Episode-Based Reinforcement Learning Objective}
Similar to step-based policy gradient \citep{Williams1992,Schulman2015}, we can also optimize the advantage function using the likelihood ratio gradient and an importance sampling estimator in the ERL setting. 
Our objective has thus the following form
\begin{align*}
     \hat{J}(\pi_{\bm{\theta}}, \pi_{\old{\bm{\theta}}}) = \mathbb{E}_{({\bm{c}},{\bm{w}})\sim p(\bm{c}),\pi_{\old{\bm{\theta}}}} \left[\frac{{{\pi}_{\bm{\theta}}}({\bm{w}}\vert{\bm{c}})}{\pi_{\old{\bm{\theta}}}({\bm{w}}\vert{\bm{c}})} A^{\pi_{\old{\bm{\theta}}}}({\bm{c}}, {\bm{w}})\right],
    \label{eq:objective}
\end{align*} 
which is maximized w.r.t $\bm{\theta}$.
The term $A^\pi(\bm{c},\bm{w}) = \mathbb{E}\left[R|\bm{c}, \bm{w};\pi \right] - \mathbb{E}\left[R|\bm{c};\pi \right]$ denotes the advantage function and $\pi_{\old{\bm{\theta}}}$ is the old policy used for sampling. 
We can further make use of a learned context-value function $V_\phi(\bm{c}) \approx \mathbb{E}\left[R|\bm{c};\pi \right]$ for the advantage estimator, which is approximated by optimizing
\begin{equation*}
    \argmin_\phi \mathbb{E}_{(\bm{c},\bm{w})\sim p(\bm{c}),\pi_{\old{\bm{\theta}}}} \left[(V_\phi(\bm{c}) - R(\bm{w}, \bm{c}))^2\right].
\end{equation*}
Unlike step-based methods, our approach leverages samples of the form $(\bm{c}, \bm{w}, R)$ that are generated per trajectory, where $R$ is the trajectory performance. 

\subsection{Choice of the Deep Reinforcement Learning Algorithm}
Training this type of policy can theoretically be done with most existing deep RL methods. 
Yet, learning in the parameter space requires to learn policies with a higher level of precision than in the step-based case. 
In the step-based case, smaller errors during action selection can still be corrected at a later time step.
In ERL, we only select one parameter vector $\bm{w}$ per episode, hence no error correction of these parameters is possible. 
As a consequence, we chose TRPL since this method has shown to be more stable and precise than other RL methods as the trust regions for the policy update are implemented exactly.
Furthermore, the trust regions are enforced per state while most other deep RL methods \citep{Schulman2015, Schulman2017, Akrour2019} only offer approximate trust region updates for which the trust region is enforced for the mean policy change over all states. 
We will now present the objective based on the adaptation of the TRPL algorithm to the ERL setup.  


\subsection{Movement Primitives as Parametrized Controllers}
So far we only discussed how to learn the distribution in the parameter space, but not how to create actions. 
For this purpose, we use a trajectory generator $\Psi(\bm{w})$ that is typically translating the parameters $\bm{w}$ to a desired trajectory $\bm{\tau}^d=(\bm{s}_1^d, \bm{s}_2^d, \ldots, \bm{s}_T^d)$.
Common choices of trajectory generators are DMPs \citep{Schaal2006, Ijspeert2013}, \textit{probabilistic movement primitives} (ProMPs) \citep{Paraschos2013}, or Viapoint MPs \citep{Zhou2019}.
The step-based control signal is then retrieved by employing a trajectory tracking controller $f(\bm{s}_t, \bm{s}_t^d)$, such as a simple PD-controller that outputs torques given the current and desired position as well as velocity. 
Here, \textit{no noise} is applied to the raw actions for exploration.
Even more complex controllers such as Cartesian impedance controllers can be readily used. 

As a result, trajectory generation and execution is independent of the policy training, enabling the use of any trajectory generator and tracking controller.
For this work, we will use ProMPs \citep{Paraschos2013} for trajectory generation and torque-based PD-controllers for action execution.
In the case of a ProMP, we can optimize its weight vector that linearly influences the generated trajectories. 
Further, we can include additional parameters, such as the trajectory starting time as well as the execution speed of the motion (represented as velocity of the ProMP's phase variable) to our ERL optimization process. 
As our experiments show, this offers a powerful parametrization for tasks where exact timing is crucial, such as robot table tennis and throwing.

\subsection{Advantages of Episode-Based Versus Step-Based Reinforcement Learning}
Step-based RL is currently by far the more prominent perspective on RL. 
This seems a natural choice as step-based RL exploits the temporal structure of RL problems and is therefore expected to outperform ERL in terms of sample efficiency. 
Yet, step-based RL also comes with a few disadvantages.
First, exploration in action space often results in a very jerky random walk behavior that does not fully explore the trajectory space of the agent.
In addition, the stochastic action selection also results in noisy returns, which translates into high variance of the policy gradient estimate. 
Second, the step-based exploration process also complicates the use of sparse or non-Markovian rewards in the step-based setting.
While there do exist specialized solutions for some of these problems, they often require much more complex algorithms.

In contrast, ERL offers a simple framework for learning complex behavior with sparse and non-Markovian rewards, without special treatment of these cases.
ERL only uses a fraction of data points to update the policy, since each trajectory is abstracted as only a single data point. 
While the small number of available training samples may initially appear to be a disadvantage, it offers three distinct benefits: 
(i) As the used controllers are deterministic, the returns used for the policy updates contain less variance than in the step-based case, which simplifies finding good policy updates.
(ii) The exploration in parameter space allows for learning with sparse and non-Markovian rewards which can be used to define the desired behavior in a more direct way.
Finally, (iii) the controller parametrization allows the inclusion of temporal parameters such as shifting or scaling of the desired trajectories. 
We believe that all these features constitute to superior performance of learned ERL policies in comparison to the step-based variant in most of our experiments. 
In conclusion, we consider ERL as a promising alternative to step-based RL in particularly for tasks where the desired behavior is hard to define with dense rewards. 


\section{Experimental Results}
For our evaluation, we first demonstrate the benefits of deep ERL in terms of sparse rewards, precision, and energy efficiency. 
Afterwards, we conduct a large scale study on all 50 Meta-World tasks \citep{Yu2019} to show that we {\color{black} achieve a competitive performance on a variety of robot manipulation tasks with highly shaped dense rewards.} 
Lastly, we investigate multiple challenging control problems that are hard to solve in the step-based setting. 
{\color{black}We compare our method, which we call \textit{BBRL-TRPL}, to the step-based methods PPO \citep{Schulman2017}, TRPL \citep{Otto2021}, SAC \citep{Haarnoja2018}, and NDP \citep{Bahl2020} as well as to a deep \textit{evolution strategies} (ES) approach \citep{Salimans2017}, the linear adaption method CMORE with ProMPs \citep{Tangkaratt2017} and BBRL-PPO, which is equivalent to BBRL-TRPL but trained with PPO instead of TRPL.}

Note that for NDP, the authors report the performance in terms of the used samples and not in terms of environment interactions (the original work only uses every fifth interaction).
We report the total number of environment interactions because we think this leads to a fairer comparison and also explains the rather poor performance of NDP in our experiments.
{\color{black} For all tasks the context information $\bm{c}$ is represented as a subset of the full observation of the first time step including only the stochastic elements, \ie the parts that are randomly initialized, such as goal or object positions.}
The trajectory performance $R(\bm{w}, \bm{c})$ is, if not specified otherwise, the full undiscounted trajectory return. 
We always evaluate on 20 different seeds and compute 10 evaluation runs after each iteration.
Following \citet{Agarwal2021}, we report the \textit{interquartile mean} (IQM) with a 95\% stratified bootstrap confidence interval and performance profiles where feasible (for hyperparameters see \Apxref{app:parameters}).

\subsection{Sparse Rewards Induce Energy Efficient Behavior}
\label{sec:effectiveness}
\begin{figure}[t]
    \centering
    \begin{minipage}{0.25\textwidth}
    \resizebox{\textwidth}{!}{\input{tikz_plots/reacher/reacher_iqm_sample_efficiency}}
    \end{minipage}%
    \begin{minipage}{0.25\textwidth}
    \resizebox{\textwidth}{!}{\input{tikz_plots/reacher/reacher_sparse_iqm_sample_efficiency}}%
    \end{minipage}%
    \begin{minipage}{0.25\textwidth}
    \resizebox{\textwidth}{!}{\input{tikz_plots/box_pushing/box_pushing_iqm_sample_efficiency}}%
    \end{minipage}%
    \begin{minipage}{0.25\textwidth}
    \centering
    \resizebox{\textwidth}{!}{\input{tikz_plots/box_pushing/box_pushing_sparse_iqm_sample_efficiency}}%
    \end{minipage}%
    \\
    \resizebox{0.75\textwidth}{!}{
    \input{tikz_plots/reacher/legend}
    }%
    \caption{
    The learning curve for the 5D reacher with dense (left) and sparse reward (center left) signal. 
    The box pushing task is trained with three different rewards. 
    The dense reward (center right) has the best performance for step-based PPO, while it performs poorly with sparsity in time (right, solid line) and sparsity in time and space (right, dashed line). 
    BBRL-TRPL solves the task in the first two settings and is even partially able to do so for the more difficult sparse reward.
    }
    \label{fig:reacher}
\end{figure}
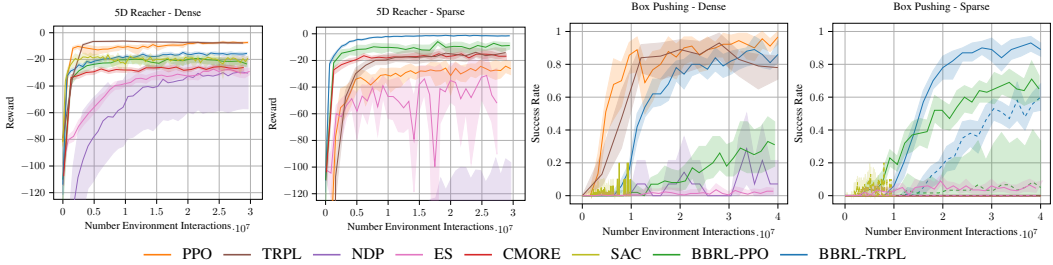

As introductory task we use an extension of the reacher from OpenAI gym \citep{Brockman2016}. 
Instead of two actuated joints, we use five, but limit the context space, \ie the location of the goal, to $y \geq 0$.
This results in an increased control complexity with a slightly decreased task complexity.  
We investigate two types of rewards, a dense reward equivalent to the original reacher and a sparse reward that only provides the distance to the goal in the last episode time step (see \Apxref{app:reacher}).
While BBRL-TRPL and BBRL-PPO can solve the task for both rewards, NDP {\color{black}and ES} do not succeed at either (\Figref{fig:reacher} left and center left). 
PPO {\color{black}and TRPL} achieve a slightly better asymptotic performance than BBRL-TRPL in the dense setting, but are not able to consistently reach the goal for the sparse reward signal. 
{\color{black}SAC achieves a comparable performance to BBRL-TRPL in the dense setting, however, cannot leverage the sparse reward (see \Apxref{app:evaluations})
While, CMORE performs reasonably well, it is only able to cover part of context space the due to its linear adaption.}

To demonstrate these learning capabilities for sparse rewards in a more complex setting, we also evaluate the algorithms on a box pushing task. 
The goal is to move a box to a given goal location and orientation using a simulated Franka Emika Panda (see \Apxref{app:box_pushing}).
We consider three rewards, a dense reward based on the goal and rotation distance for each time step and two sparse rewards. 
The first induces time-dependent sparsity by returning the reward signal only in the final time step. 
The second extends this to include space-dependent sparsity, and additionally requires the box to move near the goal in order to receive the reward.
PPO {\color{black}and TRPL} can only solve this task in the dense setting (\Figref{fig:reacher} center right and right).
{\color{black}SAC and ES even struggle with the dense reward. 
For SAC, penalties due to constraint violations in particular lead to this unstable behavior.}
The same applies for NDP, thus we did not evaluate it for the sparse rewards.
BBRL-TRPL, on the other hand, learns the task in the dense and the first sparse setting without any problems.
Even for the complex time and space sparsity, BBRL-TRPL is able to partially solve the task. 
BBRL-PPO receives merely moderate results for the dense and first sparse setting. 

Yet, why would we use sparse rewards when dense rewards work well with step-based RL? 
A dense reward typically induces a rather fast motion with poor energy efficiency. 
As shown in our experiments, dense rewards do not work well if we want to generate accurate motions that reach the goal after a certain time and are energy efficient as well.
Here, we can use sparse rewards instead, \ie, the goal distance is penalized only in the final time step, while the energy cost is accounted for in each time step. 

To illustrate the trade-off between precision and energy efficiency, we analyzed the final behaviors in both reward setups with different action penalty factors in the reward function. 
For each of these factors, we computed the average precision and energy consumption (\Figref{fig:metaworld} left and center left).
For all methods, decreasing the action penalty factor leads to a higher task precision. 
However, for the dense reward, a high task precision can only be achieved with high energy costs, while the sparse reward generates behavior of similar precision with one (box pushing) or even two (reacher) factors of magnitude less energy consumption. 
When analyzing the behaviors, it is visible that the dense reward behavior quickly moves to the target and stays there while the sparse reward behavior reaches the target only slightly before the specified end of the episode, resulting in a much slower, smoother, and more energy efficient motion. 


\subsection{Large Scale Robot Manipulation}
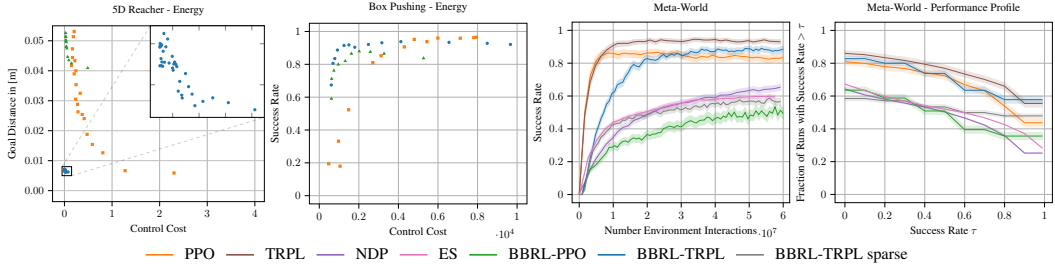
\begin{figure}
    \centering
    \begin{minipage}{0.25\textwidth}
    \resizebox{\textwidth}{!}{\input{tikz_plots/reacher/reacher_pareto}}%
    \end{minipage}%
    \begin{minipage}{0.25\textwidth}
    \resizebox{\textwidth}{!}{\input{tikz_plots/box_pushing/box_pushing_pareto}}%
    \end{minipage}%
    \begin{minipage}{0.25\textwidth}
    \resizebox{\textwidth}{!}{\input{tikz_plots/metaworld/metaworld_IQM_sample_efficiency}}%
    \end{minipage}%
    \begin{minipage}{0.25\textwidth}
    \resizebox{\textwidth}{!}{\input{tikz_plots/metaworld/metaworld_performance_profile_6e7}}%
    \end{minipage}%
    \\
    \resizebox{0.75\textwidth}{!}{
    \input{tikz_plots/metaworld/legend}
    }%
    \caption{
    Energy efficiency (sum of squared control action costs) and task performance (distance to the target in the last time step or success rate) trade-off for reacher (left) and box pushing (center left).
    We average over 100 evaluation runs and all seeds and choose action penalty factors in the intervals $(0,100]$ for the reacher and $(0,1]$ for box pushing. 
    For both tasks, BBRL-TRPL with sparse rewards can achieve a much higher energy efficiency (1 to 2 factors of magnitude) while achieving a similar target precision as PPO in the dense reward setting.
    The success rate (center right) for all 50 Meta-World tasks and the corresponding performance profile (right), \ie the fraction of runs that perform above the threshold given by the x axis. 
    While the sample efficiency is lower for BBRL-TRPL, the final quality of the policy is higher {\color{black}than PPO}.
    }
    \label{fig:metaworld}
\end{figure}

On the Meta-World benchmark suite \citep{Yu2019}, we showcase our ability to learn high quality policies.
We train individual policies for each environment, but use the same hyperparameters. 
{\color{black}PPO and TRPL achieve the best sample complexity (\Figref{fig:metaworld} center right). 
Nevertheless, BBRL-TRPL has competitive asymptotic performance and even converges slightly higher than PPO.
While in the aggregated view the gap is rather small, the corresponding performance profiles (\Figref{fig:metaworld} right) show that BBRL-TRPL performs better above the 80\% threshold.}
This means that BBRL-TRPL finds more consistent solutions than PPO with higher precision and solves these tasks without failures. 
{\color{black}NDP, ES, and BBRL-PPO are not achieving a competitive performance.}

As an additional ablation study, we trained BBRL-TRPL using a sparse reward (only the final step reward of each episode is used), denoted as BBRL-TRPL sparse. 
While the IQM score is lower, it still completes 50\% of the tasks with 100\% success rate, which is higher than PPO using the dense reward.
Moreover, the slope of the performance profile is rather small, \ie almost all tasks that can be solved are solved to perfection. 
After further investigation, we found that poorly performing tasks all require behavior that involves sub-goals, \eg bin picking, pick and place, etc. 
This could be addressed \eg by sequencing several MPs and also giving the corresponding sub-goal as intermediate rewards, which is an interesting direction for future research.

\subsection{Dealing with non-Markovian Rewards}
\begin{figure}
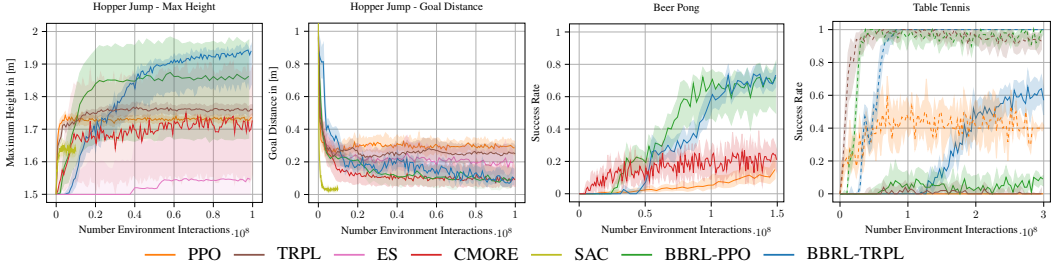

    \centering
    \begin{minipage}{0.25\textwidth}
     \resizebox{\textwidth}{!}{\input{tikz_plots/hopper_jump/hj_iqm_sample_efficiency_max_heights}}%
    \end{minipage}%
    \begin{minipage}{0.25\textwidth}
    \resizebox{\textwidth}{!}{\input{tikz_plots/hopper_jump/hj_iqm_sample_efficiency_goal_dists}}%
    \end{minipage}%
    \begin{minipage}{0.25\textwidth}
    \resizebox{\textwidth}{!}{\input{tikz_plots/beer_pong/bp_iqm_sample_efficiency}}%
    \end{minipage}%
    \begin{minipage}{0.25\textwidth}
    \resizebox{\textwidth}{!}{\input{tikz_plots/table_tennis/table_tennis_iqm_sample_efficiency}}%
    \end{minipage}%
            \\
    \resizebox{0.75\textwidth}{!}{
    \input{tikz_plots/hopper_jump/legend}
    }%
    \caption{
    The maximum jumping height of the hopper's center of mass (left) and the target distance (center left).
    The non-Markovian reward allows to jump approximately 20cm higher with increased goal precision. 
    The beer pong task (center right) demonstrates a similar behavior. 
    PPO has difficulties throwing the ball into the cup even with the fixed optimal release point, whereas BBRL-TRPL can consistently succeed in the tasks with dynamic release points.
    The success rate of the table tennis task (right) for hitting the ball (dashed line) and successfully returning the ball near the target position (solid line) shows BBRL-TRPL consistently hits the ball and returns it in most of cases.
    }
    \label{fig:hopper_jump}
\end{figure}

As final evaluation, we demonstrate the effectiveness of our method in the presence of non-Markovian rewards.
These rewards are particularly useful for complex robot learning tasks, where the whole trajectory history is needed to provide feedback to the agent.
We first use a modification of the hopper from Open AI gym \citep{Brockman2016}, where we aim to jump as high as possible and land at a target location (see \Apxref{app:hopper_jump}).
The non-Markovian reward returns the maximum height and the minimum distance to the target during the episode.
We compare our results to CMORE and ES, which also use the non-Markovian reward.
PPO, SAC, and TRPL are trained with a Markovian version that provides height and goal distance in each time step. 
Please note that we conducted extensive reward shaping to gain the highest performance (\ie maximum height and minimum goal distance).
Overall, BBRL-TRPL achieves a higher jump height than {\color{black} most other methods} (\Figref{fig:hopper_jump} left) accompanied with a smaller target distance (\Figref{fig:hopper_jump} center left).
BBRL-PPO and CMORE can match the target distance, SAC can even exceed it, but none can reliably learn a good jump height.
While BBRL-TRPL charges energy and then jumps just once, {\color{black}the step-based methods} try to maximize the height in each time step leading to multiple jumps in one episode (see \Apxref{app:evaluations}). 

{\color{black}Beer pong \cite{celik2022specializing} is another example where non-Markovian rewards are beneficial.
The goal is to throw a ball into a cup at various locations on a table.}
The return is computed from the entire trajectory of the ball, \eg by leveraging table contacts or the minimum distance to the cup {\color{black}(see \Apxref{app:beer_pong})}. 
Since we cannot directly train PPO on such a reward and designing a Markovian version is particularly difficult here, we fix the ball release time and consider the time between release of the ball and the end of the ball trajectory as final time step.
This allows to compute the reward in a similar manner as for the non-Markovian setting. 
We evaluate BBRL-TRPL {\color{black}and CMORE} on the non-Markovian reward with a learned ball release time as additional controller parameter as well as PPO for the setting described above (\Figref{fig:metaworld} right).
BBRL-TRPL and BBRL-PPO both manage to throw the ball into the cup, while PPO struggles.
{\color{black}CMORE can reliably throw the ball, however only for a subset of the context space.}
Similar to the jumping task, BBRL-PPO exhibits a larger confidence interval as it is not able to consistently solve the task. 

Lastly, we train an agent for simulated table tennis \cite{celik2022specializing} where the context is four dimensional and given by features of the initial ball trajectory and the desired location for returning the ball {\color{black}(see \Apxref{app:table_tennis})}. 
Intuitively, the agent should be rewarded if it hits the ball and returns the ball near the designated goal position. 
Similar to the throwing task, for {\color{black}the step-based methods} we consider the time after hitting the ball as one time step. 
For the BBRL approaches, we also learn the trajectory starting time as well as the speed of the desired trajectory (which is a parameter of the ProMP). 
Both parameters help to learn the precise timing required to play table tennis. 
BBRL-TRPL always manages to hit the ball and successfully returns it within the vicinity of the goal in more than 60\% of the cases.
While BBRL-PPO {\color{black}and TRPL} can hit the ball, they cannot return it with enough precision. 
We also tested BBRL-TRPL with fixed timing parameters (start time and velocity) and observed a similar behavior - the algorithm requires these timing parameters to learn how to put the ball at the desired location. 
PPO fails to even hit the ball consistently. 

Summarizing, BBRL-TRPL is able to successfully deal with non-Markovian reward structures, which can be more intuitive, easier to define, and achieve better learned behaviors than engineered dense rewards.  

\section{Conclusion and Limitations}
\label{sec:conclusion}
In this work, we proposed a new algorithm for ERL. 
Our method leverages the recently introduced TRPLs \cite{Otto2021} to guarantee precise and robust policy updates in high-dimensional parameter spaces.
In thorough empirical evaluations, we have shown that we can indeed learn policies with high precision even in the presence of sparse and non-Markovian rewards, where step-based approaches typically fail. 
Due to the use of sparse rewards, our learned behaviors are more energy efficient and less sensitive to high action costs, which leads to almost universally well-performing policies. 

The main limitation of our method and ERL in general is that they typically require more interaction time than step-based RL in the dense reward setting. 
While we agree this can be critical in some scenarios, we still think the benefits outweigh this drawback in many scenarios. 
To address the sample complexity issue, we plan to investigate off-policy approaches of our method as well as re-planning of the motion primitive to obtain more training data for the policy update in future work.
Another limitation of our method is that the desired trajectory is planned in advance in the beginning of the episode, and hence, cannot be altered during the execution. 
This might be problematic for unforeseen events or perturbations, \ie highly complex or reactive behavior that cannot directly be modeled with the current motion representation. 
We believe that this limitation can also be tackled by incorporating re-planning the MP trajectory in the RL algorithm. 
For future work, we will also investigate sequencing multiple MPs to solve more complex tasks which are composed of sub-goals. 



\clearpage


\bibliography{bibliography}  

\appendix
\section{KL-Divergence Trust Region Projection Layer}
\label{app:kl_trpl}
\input{appendix/kl_trpl}

\section{Environment Details}
\label{app:details}
\input{appendix/environments}

\section{Additional Evaluations}
\label{app:evaluations}
\input{appendix/evaluations}

\newpage
\section{Hyperparameters}
\label{app:parameters}
\input{appendix/hyperparameters}

\end{document}

%% file: imports/math.tex
\usepackage{amsmath, amssymb}
\usepackage{mathtools}
\usepackage{algorithm,algorithmic}

\usepackage{bm}

\newcommand{\vecT}[1]{#1^\mathrm{T}}
\newcommand{\til}[1]{\tilde{#1}}
\newcommand{\inv}[1]{{#1}^{-1}}

\newcommand{\old}[1]{{#1}_{\textrm{old}}}
\newcommand{\oldInv}[1]{{#1}_{\textrm{old}}^{-1}}

\newcommand{\st}{\textrm{s.t.}}

\DeclareMathOperator*{\argmax}{arg\,max}
\DeclareMathOperator*{\argmin}{arg\,min}

%% file: imports/tikz.tex
\usepackage{xcolor}

\usepackage{tikz}
\usetikzlibrary{positioning, arrows, automata}
\usepackage{pgfplots}
\pgfplotsset{compat=1.8}
\usepackage{pgfplotstable}
\usepgfplotslibrary{groupplots}
\usetikzlibrary{external}
\usetikzlibrary{calc}
\usetikzlibrary{backgrounds, calc}
\usetikzlibrary{spy}
\usetikzlibrary{fit}
\usepackage{adjustbox}

\definecolor{C0}{HTML}{1f77b4}  
\definecolor{C1}{HTML}{ff7f0e}  
\definecolor{C4}{HTML}{2ca02c}  
\definecolor{C7}{HTML}{d62728}  
\definecolor{C2}{HTML}{9467bd}  
\definecolor{C5}{HTML}{8c564b}  
\definecolor{C6}{HTML}{e377c2}  
\definecolor{C3}{HTML}{7f7f7f}  
\definecolor{C8}{HTML}{bcbd22}  
\definecolor{C9}{HTML}{17becf}

\definecolor{darkgray176}{RGB}{176,176,176}
\definecolor{lightgray204}{RGB}{204,204,204}

%% file: imports/references.tex
\def\Figref#1{Figure~\ref{#1}}

\def\Apxref#1{Appendix~\ref{#1}}

\def\eqref#1{equation~\ref{#1}}
\def\Eqref#1{Equation~\ref{#1}}

\newcommand{\ie}{i.\,e.\,}
\newcommand{\eg}{e.\,g.\,}

%% file: tikz_plots/diagrams/overview.tex
\tikzset{node style/.style={draw, thick, rounded corners, minimum height=1cm, minimum width=1.5cm, align=center}}

\begin{tikzpicture}
    
    \node[node style] (c) at (0,0) {Context\\$\bm{c}$};  
    \node[node style, right = 0.75cm of c] (pi) {Policy\\$\pi_{\bm{\theta}}(\bm{w} \vert \bm{c})$};
    \node[node style, right = 0.75cm of pi, minimum width=2.5cm] (tau) {Desired Trajectory\\$\bm{\tau}^d = \Psi(\bm{w}) = (\bm{s}_1^d, \bm{s}_2^d, \ldots, \bm{s}_T^d)$};
    
    
    \node[node style, right = of tau, minimum width=2.5cm] (controller) {Controller\\$f(\bm{s}_t, \bm{s}_t^d)$};
    \node[node style, right = 0.75cm of controller, minimum width=2.5cm] (env) {Environment};
    \node[draw, thick, minimum height=2.25cm, minimum width=6.25cm, right = 0.75cm of tau] (loop) {};
    \node[anchor=north west,inner sep=3pt] at (loop.north west) {for $t$ in $1,2,\ldots, T$};
    
    \draw[-latex]
        (c) edge (pi) 
        (pi) edge (tau)
        (tau) edge (loop)
        (controller) edge node[node distance=2.5cm, auto] {$a_t$} (env)
        (env.south) -- +(0.0,-0.4) -| node[above, pos=0.25] {$s_{t+1}$} (controller.south)
        ;

\end{tikzpicture}

%% file: tikz_plots/reacher/reacher_iqm_sample_efficiency.tex
\begin{tikzpicture}
\begin{axis}[
legend cell align={left},
legend style={
  fill opacity=0.8,
  draw opacity=1,
  text opacity=1,
  at={(0.97,0.03)},
  anchor=south east,
  draw=lightgray204
},
title={5D Reacher - Dense},
tick align=outside,
tick pos=left,
x grid style={darkgray176},
xlabel={Number Environment Interactions},
xmajorgrids,
xmin=-1472000, xmax=31912000,
xtick style={color=black},
y grid style={darkgray176},
ylabel={Reward},
ymajorgrids,
ymin=-125, ymax=5,
ytick style={color=black}
]
\path [draw=C1, fill=C1, opacity=0.2]
(axis cs:0,-79.05324711725)
--(axis cs:0,-84.63037864475)
--(axis cs:800000,-32.8115140615)
--(axis cs:1600000,-13.2322147985)
--(axis cs:2400000,-11.68837910825)
--(axis cs:3200000,-15.2848050775)
--(axis cs:4000000,-17.106422825)
--(axis cs:4800000,-16.95078287075)
--(axis cs:5600000,-15.94174291075)
--(axis cs:6400000,-15.191033705)
--(axis cs:7200000,-14.615818328)
--(axis cs:8000000,-13.765861217)
--(axis cs:8800000,-13.3994899125)
--(axis cs:9600000,-12.047429106)
--(axis cs:10400000,-12.645014972)
--(axis cs:11200000,-14.7315653545)
--(axis cs:12000000,-16.3544475585)
--(axis cs:12800000,-11.50100277425)
--(axis cs:13600000,-12.5599155155)
--(axis cs:14400000,-10.13073311975)
--(axis cs:15200000,-13.18609032375)
--(axis cs:16000000,-11.64194035525)
--(axis cs:16800000,-10.18474978175)
--(axis cs:17600000,-10.3924906705)
--(axis cs:18400000,-10.186464589)
--(axis cs:19200000,-11.15898969375)
--(axis cs:20000000,-8.57888589025)
--(axis cs:20800000,-8.92528216075)
--(axis cs:21600000,-7.79273712575)
--(axis cs:22400000,-8.730254349)
--(axis cs:23200000,-8.40544673225)
--(axis cs:24000000,-8.391081865)
--(axis cs:24800000,-8.30460871825)
--(axis cs:25600000,-7.79877392525)
--(axis cs:26400000,-8.240331879)
--(axis cs:27200000,-7.7151417615)
--(axis cs:28000000,-7.76464915275)
--(axis cs:28800000,-8.181919909)
--(axis cs:29600000,-8.05651712675)
--(axis cs:29600000,-6.6406088775)
--(axis cs:29600000,-6.6406088775)
--(axis cs:28800000,-6.84796712275)
--(axis cs:28000000,-6.99363814825)
--(axis cs:27200000,-6.5987920985)
--(axis cs:26400000,-6.83475914)
--(axis cs:25600000,-6.361669385)
--(axis cs:24800000,-6.99440352825)
--(axis cs:24000000,-7.27568808675)
--(axis cs:23200000,-7.33852466175)
--(axis cs:22400000,-7.63640670625)
--(axis cs:21600000,-6.75571718625)
--(axis cs:20800000,-7.4206419405)
--(axis cs:20000000,-7.622106888)
--(axis cs:19200000,-8.1786296475)
--(axis cs:18400000,-8.5252213025)
--(axis cs:17600000,-8.2632582065)
--(axis cs:16800000,-8.572088146)
--(axis cs:16000000,-9.1099391085)
--(axis cs:15200000,-8.911866829)
--(axis cs:14400000,-7.804627223)
--(axis cs:13600000,-9.243398305)
--(axis cs:12800000,-8.75279544775)
--(axis cs:12000000,-9.60059092575)
--(axis cs:11200000,-9.547792074)
--(axis cs:10400000,-9.063023324)
--(axis cs:9600000,-9.8434259305)
--(axis cs:8800000,-9.198404566)
--(axis cs:8000000,-10.20512101275)
--(axis cs:7200000,-10.2296991)
--(axis cs:6400000,-10.25934769075)
--(axis cs:5600000,-10.68962128925)
--(axis cs:4800000,-10.10028268575)
--(axis cs:4000000,-9.4324426695)
--(axis cs:3200000,-9.519567145)
--(axis cs:2400000,-9.201319376)
--(axis cs:1600000,-10.364684022)
--(axis cs:800000,-28.2645140145)
--(axis cs:0,-79.05324711725)
--cycle;

\addplot [thick, C1, mark=*, mark size=0, mark options={solid}]
table {%
0 -81.29115461
800000 -30.44954982
1600000 -11.59825969
2400000 -10.27448889
3200000 -11.01075121
4000000 -11.67191702
4800000 -12.28817495
5600000 -12.8725522
6400000 -12.31591147
7200000 -12.27361792
8000000 -11.44556578
8800000 -10.48940334
9600000 -10.98147513
10400000 -10.39212016
11200000 -11.23088767
12000000 -11.82384373
12800000 -10.08772845
13600000 -11.03447467
14400000 -8.89906159
15200000 -10.45312094
16000000 -9.90723062
16800000 -9.38423345
17600000 -9.24622341
18400000 -9.34440619
19200000 -9.22216463
20000000 -8.15281924
20800000 -8.07822166
21600000 -7.18655618
22400000 -8.14538416
23200000 -7.89523865
24000000 -7.7371999
24800000 -7.51529004
25600000 -7.05006767
26400000 -7.48002035
27200000 -7.07621374
28000000 -7.35634714
28800000 -7.57307732
29600000 -7.2717984
};

\path [draw=C2, fill=C2, opacity=0.2]
(axis cs:16000,-524.41720758475)
--(axis cs:16000,-590.722545549)
--(axis cs:816000,-247.650543129)
--(axis cs:1616000,-164.7975073115)
--(axis cs:2416000,-143.16365511)
--(axis cs:3216000,-133.39715536975)
--(axis cs:4016000,-116.80760174625)
--(axis cs:4816000,-111.89185983375)
--(axis cs:5616000,-105.2711387755)
--(axis cs:6416000,-100.05690822775)
--(axis cs:7216000,-95.37596741775)
--(axis cs:8016000,-90.46219070425)
--(axis cs:8816000,-86.800957992)
--(axis cs:9616000,-85.234039062)
--(axis cs:10416000,-79.1640826665)
--(axis cs:11216000,-78.404046493)
--(axis cs:12016000,-75.5763082745)
--(axis cs:12816000,-72.94152765225)
--(axis cs:13616000,-71.81310615475)
--(axis cs:14416000,-73.40973958675)
--(axis cs:15216000,-69.26838184125)
--(axis cs:16016000,-65.24313816875)
--(axis cs:16816000,-65.76290524825)
--(axis cs:17616000,-64.3293822595)
--(axis cs:18416000,-65.91695998775)
--(axis cs:19216000,-63.015601451)
--(axis cs:20016000,-59.51773016575)
--(axis cs:20816000,-60.7832998505)
--(axis cs:21616000,-57.7134562665)
--(axis cs:22416000,-60.47647770975)
--(axis cs:23216000,-59.347551069)
--(axis cs:24016000,-59.63206923725)
--(axis cs:24816000,-58.49161196175)
--(axis cs:25616000,-58.60750700625)
--(axis cs:26416000,-59.779392494)
--(axis cs:27216000,-58.515217444)
--(axis cs:28016000,-57.47420636325)
--(axis cs:28816000,-56.964218768)
--(axis cs:29616000,-57.0865545745)
--(axis cs:29616000,-18.0987433055)
--(axis cs:29616000,-18.0987433055)
--(axis cs:28816000,-16.9489908455)
--(axis cs:28016000,-17.280957752)
--(axis cs:27216000,-18.02128354875)
--(axis cs:26416000,-19.45192854125)
--(axis cs:25616000,-17.30727741575)
--(axis cs:24816000,-19.78518170325)
--(axis cs:24016000,-20.6083357505)
--(axis cs:23216000,-20.38215668275)
--(axis cs:22416000,-19.789055469)
--(axis cs:21616000,-19.352603426)
--(axis cs:20816000,-20.58325815)
--(axis cs:20016000,-19.72396109225)
--(axis cs:19216000,-21.503201882)
--(axis cs:18416000,-20.94745574)
--(axis cs:17616000,-22.6953913715)
--(axis cs:16816000,-23.1679417925)
--(axis cs:16016000,-20.33303161975)
--(axis cs:15216000,-24.139735398)
--(axis cs:14416000,-25.8086497455)
--(axis cs:13616000,-23.57459776925)
--(axis cs:12816000,-25.793725683)
--(axis cs:12016000,-27.9337606725)
--(axis cs:11216000,-30.066736534)
--(axis cs:10416000,-30.0644982725)
--(axis cs:9616000,-35.796111426)
--(axis cs:8816000,-36.46393918725)
--(axis cs:8016000,-42.1957502375)
--(axis cs:7216000,-37.53196742075)
--(axis cs:6416000,-41.371320021)
--(axis cs:5616000,-45.61242991825)
--(axis cs:4816000,-49.16218151375)
--(axis cs:4016000,-53.7073196765)
--(axis cs:3216000,-58.600117997)
--(axis cs:2416000,-75.5254784185)
--(axis cs:1616000,-100.5990122195)
--(axis cs:816000,-149.15423413675)
--(axis cs:16000,-524.41720758475)
--cycle;

\addplot [thick, C2, mark=*, mark size=0, mark options={solid}]
table {%
16000 -555.06350312
816000 -200.8295831
1616000 -132.83299638
2416000 -108.52098944
3216000 -95.41731789
4016000 -85.06981331
4816000 -78.87815079
5616000 -74.05565526
6416000 -66.1689331
7216000 -62.90497837
8016000 -63.31658192
8816000 -58.44371041
9616000 -55.96727097
10416000 -47.79812264
11216000 -47.45076776
12016000 -44.27652403
12816000 -40.89664002
13616000 -40.34627008
14416000 -41.26055441
15216000 -40.10857771
16016000 -34.87121144
16816000 -37.54798386
17616000 -36.94770856
18416000 -35.99376864
19216000 -35.4441751
20016000 -32.56261911
20816000 -33.62665481
21616000 -30.49091506
22416000 -32.58163577
23216000 -31.97089699
24016000 -32.53597899
24816000 -31.49123055
25616000 -29.92823403
26416000 -32.31755456
27216000 -30.32309103
28016000 -29.70289871
28816000 -29.57210064
29616000 -30.35598481
};

\path [draw=C6, fill=C6, opacity=0.2]
(axis cs:40000,-93.40240038875)
--(axis cs:40000,-102.71471367625)
--(axis cs:869600,-83.27830982625)
--(axis cs:1702000,-81.3575156825)
--(axis cs:2536800,-73.95006505375)
--(axis cs:3362800,-68.784588505)
--(axis cs:4195200,-64.65009325)
--(axis cs:5025200,-60.1615524)
--(axis cs:5860000,-55.82868753875)
--(axis cs:6691200,-50.34205127125)
--(axis cs:7520800,-45.65923158625)
--(axis cs:8349600,-42.41499699)
--(axis cs:9177600,-42.17874125125)
--(axis cs:10009600,-40.1580719825)
--(axis cs:10837200,-40.10012065625)
--(axis cs:11662400,-37.252737745)
--(axis cs:12492000,-37.11992521375)
--(axis cs:13322000,-37.90622292)
--(axis cs:14151600,-35.25931770875)
--(axis cs:14984800,-35.25531696875)
--(axis cs:15814800,-35.27500466375)
--(axis cs:16645200,-36.93335170625)
--(axis cs:17478800,-35.40087718375)
--(axis cs:18308800,-34.0500024125)
--(axis cs:19148800,-33.72036527625)
--(axis cs:19970400,-35.48734755625)
--(axis cs:20793200,-34.9663227625)
--(axis cs:21616800,-35.42848392875)
--(axis cs:22442000,-32.2999810025)
--(axis cs:23267600,-32.27250215)
--(axis cs:24089200,-31.68355104125)
--(axis cs:24916800,-32.1796179)
--(axis cs:25740800,-31.03567708125)
--(axis cs:26582000,-30.44629995625)
--(axis cs:27404800,-32.00276341)
--(axis cs:28230400,-33.32871781625)
--(axis cs:29051600,-33.79524576875)
--(axis cs:29874400,-31.520631435)
--(axis cs:29874400,-26.6287056025)
--(axis cs:29874400,-26.6287056025)
--(axis cs:29051600,-27.11928028375)
--(axis cs:28230400,-26.83696428875)
--(axis cs:27404800,-24.999661555)
--(axis cs:26582000,-24.46598723375)
--(axis cs:25740800,-25.0229992)
--(axis cs:24916800,-25.78280152)
--(axis cs:24089200,-23.66144709375)
--(axis cs:23267600,-25.67291054375)
--(axis cs:22442000,-25.84001334375)
--(axis cs:21616800,-28.63873576125)
--(axis cs:20793200,-29.48341506875)
--(axis cs:19970400,-29.52329685625)
--(axis cs:19148800,-29.75990891)
--(axis cs:18308800,-28.0569320775)
--(axis cs:17478800,-29.29063654)
--(axis cs:16645200,-29.1757596375)
--(axis cs:15814800,-28.510941465)
--(axis cs:14984800,-29.3990493975)
--(axis cs:14151600,-30.75634753125)
--(axis cs:13322000,-32.1727562475)
--(axis cs:12492000,-31.17427549625)
--(axis cs:11662400,-31.62044648875)
--(axis cs:10837200,-34.588859735)
--(axis cs:10009600,-36.09974353375)
--(axis cs:9177600,-36.3243020125)
--(axis cs:8349600,-36.86386571)
--(axis cs:7520800,-39.99140985)
--(axis cs:6691200,-44.79858999125)
--(axis cs:5860000,-47.159158125)
--(axis cs:5025200,-51.01601880625)
--(axis cs:4195200,-55.41118438125)
--(axis cs:3362800,-60.99809511625)
--(axis cs:2536800,-65.62585353625)
--(axis cs:1702000,-73.71470698875)
--(axis cs:869600,-77.144968715)
--(axis cs:40000,-93.40240038875)
--cycle;

\addplot [thick, C6, mark=*, mark size=0, mark options={solid}]
table {%
40000 -98.03823615
869600 -80.30221085
1702000 -77.7589808
2536800 -70.0001597
3362800 -64.90621025
4195200 -59.91296215
5025200 -55.7332474
5860000 -51.55856595
6691200 -47.50510775
7520800 -42.69194965
8349600 -39.624137
9177600 -39.1272012
10009600 -38.13569015
10837200 -37.26832325
11662400 -34.5196266
12492000 -34.16187015
13322000 -35.03417465
14151600 -32.94741265
14984800 -32.20802805
15814800 -31.84749805
16645200 -32.86444065
17478800 -32.37651755
18308800 -31.0935787
19148800 -31.7644897
19970400 -32.4998472
20793200 -32.2094244
21616800 -31.7149336
22442000 -28.9589726
23267600 -28.9899369
24089200 -27.3388858
24916800 -29.0116248
25740800 -27.99183935
26582000 -27.4144216
27404800 -28.5169702
28230400 -30.1007912
29051600 -30.38443685
29874400 -29.00412755
};

\path [draw=C4, fill=C4, opacity=0.2]
(axis cs:0,-90.5506495742556)
--(axis cs:0,-129.650763883801)
--(axis cs:1280000,-39.1865308339224)
--(axis cs:2560000,-31.3106747461979)
--(axis cs:3840000,-28.9898556582979)
--(axis cs:5120000,-25.4226017766956)
--(axis cs:6400000,-26.4089895874003)
--(axis cs:7680000,-25.4033575661297)
--(axis cs:8960000,-23.3695026602758)
--(axis cs:10240000,-22.9807407502324)
--(axis cs:11520000,-23.2321711407608)
--(axis cs:12800000,-23.4769446910625)
--(axis cs:14080000,-25.9082416992667)
--(axis cs:15360000,-26.9781747829757)
--(axis cs:16640000,-21.6558096081002)
--(axis cs:17920000,-23.8359418161485)
--(axis cs:19200000,-24.3224722244434)
--(axis cs:20480000,-22.4621227177545)
--(axis cs:21760000,-22.4061046787525)
--(axis cs:23040000,-23.3369769239337)
--(axis cs:24320000,-24.1010694461075)
--(axis cs:25600000,-25.3544579937076)
--(axis cs:26880000,-24.4938301178556)
--(axis cs:28160000,-23.9148317699142)
--(axis cs:29440000,-25.6257596001822)
--(axis cs:29440000,-20.5913325938463)
--(axis cs:29440000,-20.5913325938463)
--(axis cs:28160000,-19.8915576540685)
--(axis cs:26880000,-18.993032990139)
--(axis cs:25600000,-19.0969493817248)
--(axis cs:24320000,-18.4921297777347)
--(axis cs:23040000,-18.3283357338246)
--(axis cs:21760000,-17.4050495293924)
--(axis cs:20480000,-17.5937424318514)
--(axis cs:19200000,-18.5468443776832)
--(axis cs:17920000,-18.9199427034529)
--(axis cs:16640000,-18.6329189769775)
--(axis cs:15360000,-19.5812616062859)
--(axis cs:14080000,-18.6584831379783)
--(axis cs:12800000,-17.735820551571)
--(axis cs:11520000,-17.5339645323426)
--(axis cs:10240000,-18.5599499892508)
--(axis cs:8960000,-18.4479016911886)
--(axis cs:7680000,-20.3104636608799)
--(axis cs:6400000,-19.1538742396432)
--(axis cs:5120000,-21.6051088272989)
--(axis cs:3840000,-20.5708300718653)
--(axis cs:2560000,-26.690956800616)
--(axis cs:1280000,-32.5067218782709)
--(axis cs:0,-90.5506495742556)
--cycle;

\addplot [thick, C4, mark=*, mark size=0, mark options={solid}]
table {%
0 -110.254882819678
1280000 -35.1404691099435
2560000 -28.5816787646744
3840000 -24.7241744121467
5120000 -23.5547309453525
6400000 -22.7343581411903
7680000 -22.5839383013582
8960000 -20.3980584314209
10240000 -20.3191247125363
11520000 -19.8408682243256
12800000 -20.2200611329189
14080000 -22.0794832746305
15360000 -22.9166875899129
16640000 -20.076705402943
17920000 -20.7521114059131
19200000 -20.9800382776795
20480000 -19.5980556451524
21760000 -19.2291571076081
23040000 -20.2843661581412
24320000 -20.5467448680433
25600000 -21.4094418002792
26880000 -21.5331740927988
28160000 -21.2528654868961
29440000 -23.4953842402717
};

\path [draw=C0, fill=C0, opacity=0.2]
(axis cs:0,-105.975249623745)
--(axis cs:0,-121.521295745532)
--(axis cs:640000,-35.23311324152)
--(axis cs:1280000,-29.6336175520152)
--(axis cs:1920000,-28.6223969093843)
--(axis cs:2560000,-25.5335534455092)
--(axis cs:3200000,-27.0375999102163)
--(axis cs:3840000,-22.3177006195893)
--(axis cs:4480000,-23.5475646582328)
--(axis cs:5120000,-22.7080193855059)
--(axis cs:5760000,-22.5981865835894)
--(axis cs:6400000,-20.9410312854338)
--(axis cs:7040000,-20.4190669495036)
--(axis cs:7680000,-20.1888101765741)
--(axis cs:8320000,-19.1736311124779)
--(axis cs:8960000,-18.8365278086663)
--(axis cs:9600000,-20.1948203379057)
--(axis cs:10240000,-19.6574339171333)
--(axis cs:10880000,-19.8442657296715)
--(axis cs:11520000,-19.6946736769037)
--(axis cs:12160000,-18.5047608388294)
--(axis cs:12800000,-17.6778067706583)
--(axis cs:13440000,-18.6115697997995)
--(axis cs:14080000,-18.7922548917385)
--(axis cs:14720000,-19.1077759179039)
--(axis cs:15360000,-18.7627898478747)
--(axis cs:16000000,-17.6134402201054)
--(axis cs:16640000,-18.097956557094)
--(axis cs:17280000,-17.205230895839)
--(axis cs:17920000,-18.2108496239647)
--(axis cs:18560000,-17.6165099647073)
--(axis cs:19200000,-16.1064428366865)
--(axis cs:19840000,-17.702497002784)
--(axis cs:20480000,-17.8958088003918)
--(axis cs:21120000,-16.7643020324199)
--(axis cs:21760000,-16.636483411665)
--(axis cs:22400000,-17.0407074339504)
--(axis cs:23040000,-18.1399555078678)
--(axis cs:23680000,-16.0334708289854)
--(axis cs:24320000,-17.0415004707809)
--(axis cs:24960000,-17.0089342352471)
--(axis cs:25600000,-16.7758593786928)
--(axis cs:26240000,-17.4460254211489)
--(axis cs:26880000,-17.4957343496353)
--(axis cs:27520000,-17.0688395957993)
--(axis cs:28160000,-17.0601015482693)
--(axis cs:28800000,-17.5249493236117)
--(axis cs:29440000,-16.2715475694409)
--(axis cs:29440000,-14.8470929233907)
--(axis cs:29440000,-14.8470929233907)
--(axis cs:28800000,-14.6977090456836)
--(axis cs:28160000,-14.6056896639193)
--(axis cs:27520000,-14.9992230948546)
--(axis cs:26880000,-14.267204489548)
--(axis cs:26240000,-14.7811018637729)
--(axis cs:25600000,-14.8093001737485)
--(axis cs:24960000,-14.7894487182753)
--(axis cs:24320000,-14.3727632586699)
--(axis cs:23680000,-14.6206099806891)
--(axis cs:23040000,-15.4923752981252)
--(axis cs:22400000,-14.7860613890712)
--(axis cs:21760000,-14.5007800543833)
--(axis cs:21120000,-14.3991262550134)
--(axis cs:20480000,-15.2313707074982)
--(axis cs:19840000,-14.523320768731)
--(axis cs:19200000,-13.8350444523718)
--(axis cs:18560000,-15.5764788999328)
--(axis cs:17920000,-15.6990094196873)
--(axis cs:17280000,-14.6051667488812)
--(axis cs:16640000,-15.0158527587624)
--(axis cs:16000000,-14.9690208336036)
--(axis cs:15360000,-16.2488889491617)
--(axis cs:14720000,-16.9154851692866)
--(axis cs:14080000,-16.8396814994852)
--(axis cs:13440000,-16.3556602970405)
--(axis cs:12800000,-14.8525895157645)
--(axis cs:12160000,-16.5738513662987)
--(axis cs:11520000,-16.7781132172681)
--(axis cs:10880000,-17.7739151260646)
--(axis cs:10240000,-16.3803813511126)
--(axis cs:9600000,-17.7394347312634)
--(axis cs:8960000,-16.9511225705658)
--(axis cs:8320000,-18.1292062794858)
--(axis cs:7680000,-18.6092403683847)
--(axis cs:7040000,-17.8154641587961)
--(axis cs:6400000,-18.8253734618227)
--(axis cs:5760000,-18.9901179340686)
--(axis cs:5120000,-19.7397324394887)
--(axis cs:4480000,-21.4858816454568)
--(axis cs:3840000,-20.8186035832968)
--(axis cs:3200000,-23.5463275875953)
--(axis cs:2560000,-22.7203584871049)
--(axis cs:1920000,-26.1052834617733)
--(axis cs:1280000,-27.1399669822538)
--(axis cs:640000,-30.9805643063827)
--(axis cs:0,-105.975249623745)
--cycle;

\addplot [thick, C0, mark=*, mark size=0, mark options={solid}]
table {%
0 -113.965753788549
640000 -32.6372088538183
1280000 -28.6097848871899
1920000 -27.2887746672996
2560000 -24.0244659667733
3200000 -25.4435787376178
3840000 -21.564794615298
4480000 -22.4400226865221
5120000 -21.1188647048268
5760000 -20.6276264109638
6400000 -19.8224443280626
7040000 -19.0569048020311
7680000 -19.5054471842752
8320000 -18.6189813291994
8960000 -17.8238613382247
9600000 -18.7279563507236
10240000 -17.6171868338754
10880000 -18.8256863894924
11520000 -18.3329765020674
12160000 -17.3579120014141
12800000 -16.0800744622004
13440000 -17.4053569415664
14080000 -17.7596498423216
14720000 -17.8183977143982
15360000 -17.5532552395323
16000000 -16.4410290193874
16640000 -16.4699716535266
17280000 -16.0649958820751
17920000 -16.8847224109908
18560000 -16.5788912995912
19200000 -14.8214606594733
19840000 -16.0953089740623
20480000 -16.3644169560751
21120000 -15.669825976469
21760000 -15.2975295349404
22400000 -15.8452411471551
23040000 -17.0129531069215
23680000 -15.1718337599647
24320000 -15.6774648657243
24960000 -15.8426464746607
25600000 -15.6416250200406
26240000 -16.3393874124373
26880000 -15.7786819589159
27520000 -15.786802071726
28160000 -15.9851353980675
28800000 -15.9896225168086
29440000 -15.51337354213
};
\path [draw=C7, fill=C7, opacity=0.2]
(axis cs:120000,-100.878935952369)
--(axis cs:120000,-114.635281627823)
--(axis cs:1320000,-35.3783485713438)
--(axis cs:2520000,-34.1935819849449)
--(axis cs:3720000,-31.5348749935718)
--(axis cs:4920000,-31.3173256586367)
--(axis cs:6120000,-28.7961535585281)
--(axis cs:7320000,-30.655496339064)
--(axis cs:8520000,-29.5143244102836)
--(axis cs:9720000,-29.8450987475536)
--(axis cs:10920000,-29.2696508764533)
--(axis cs:12120000,-30.293389974425)
--(axis cs:13320000,-27.5626494745228)
--(axis cs:14520000,-27.9302868357061)
--(axis cs:15720000,-27.127807411049)
--(axis cs:16920000,-29.3259818064061)
--(axis cs:18120000,-27.4906982108967)
--(axis cs:19320000,-29.4522528001584)
--(axis cs:20520000,-28.9091154683864)
--(axis cs:21720000,-26.5630784463001)
--(axis cs:22920000,-28.5461155928452)
--(axis cs:24120000,-27.0252672855476)
--(axis cs:25320000,-27.5677728458031)
--(axis cs:26520000,-28.2698446167357)
--(axis cs:27720000,-29.2237718625544)
--(axis cs:28920000,-28.3615674809166)
--(axis cs:28920000,-24.2534952975506)
--(axis cs:28920000,-24.2534952975506)
--(axis cs:27720000,-25.30934012478)
--(axis cs:26520000,-24.4269067829809)
--(axis cs:25320000,-23.758710878157)
--(axis cs:24120000,-24.7305585073662)
--(axis cs:22920000,-26.3826037827303)
--(axis cs:21720000,-24.9709990654373)
--(axis cs:20520000,-25.8327273664012)
--(axis cs:19320000,-26.7746644811414)
--(axis cs:18120000,-24.5189048068647)
--(axis cs:16920000,-25.0043415872893)
--(axis cs:15720000,-25.2672292380341)
--(axis cs:14520000,-24.6291061455313)
--(axis cs:13320000,-24.8701565996323)
--(axis cs:12120000,-27.2384902171097)
--(axis cs:10920000,-26.3405092379432)
--(axis cs:9720000,-26.2435514264615)
--(axis cs:8520000,-25.7774382154631)
--(axis cs:7320000,-27.4982143093558)
--(axis cs:6120000,-25.7888131195994)
--(axis cs:4920000,-28.1802533182799)
--(axis cs:3720000,-28.178908529621)
--(axis cs:2520000,-30.2313489764142)
--(axis cs:1320000,-31.406777751324)
--(axis cs:120000,-100.878935952369)
--cycle;

\addplot [thick, C7, mark=*, mark size=0, mark options={solid}]
table {%
120000 -107.593696457684
1320000 -33.5400907799063
2520000 -32.0438813428006
3720000 -29.8419883756385
4920000 -29.3217164259596
6120000 -27.1884614060859
7320000 -28.9251735530112
8520000 -27.5944232162365
9720000 -27.9322743319613
10920000 -28.0263633032567
12120000 -28.9373664530664
13320000 -26.2682055832005
14520000 -26.4217325970271
15720000 -26.0176653498634
16920000 -26.905649749588
18120000 -25.7637855151197
19320000 -28.1880870299133
20520000 -27.5946674926184
21720000 -25.7662219970146
22920000 -27.5250303273218
24120000 -25.785393820541
25320000 -25.3969947043774
26520000 -26.3811653398569
27720000 -26.8868519595693
28920000 -26.0621161669802
};

\path [draw=C5, fill=C5, opacity=0.2]
(axis cs:0,-75.33036760275)
--(axis cs:0,-81.82432121875)
--(axis cs:1600000,-34.19354396925)
--(axis cs:3200000,-9.49829455975)
--(axis cs:4800000,-6.983098321)
--(axis cs:6400000,-6.850647577)
--(axis cs:8000000,-6.85330476275)
--(axis cs:9600000,-6.51522705225)
--(axis cs:11200000,-6.778518714)
--(axis cs:12800000,-7.100520893)
--(axis cs:14400000,-7.379891908)
--(axis cs:16000000,-7.33931100575)
--(axis cs:17600000,-7.63855244)
--(axis cs:19200000,-7.68761892925)
--(axis cs:20800000,-7.70304709475)
--(axis cs:22400000,-7.96381529725)
--(axis cs:24000000,-8.163025499)
--(axis cs:25600000,-8.04363475125)
--(axis cs:27200000,-8.17628977675)
--(axis cs:28800000,-8.19994152675)
--(axis cs:28800000,-7.2324384955)
--(axis cs:28800000,-7.2324384955)
--(axis cs:27200000,-7.179605746)
--(axis cs:25600000,-6.9799863005)
--(axis cs:24000000,-7.27824883675)
--(axis cs:22400000,-6.78918735675)
--(axis cs:20800000,-6.72912496325)
--(axis cs:19200000,-6.78557707025)
--(axis cs:17600000,-6.54858921)
--(axis cs:16000000,-6.63377321775)
--(axis cs:14400000,-6.315725352)
--(axis cs:12800000,-6.343486354)
--(axis cs:11200000,-6.1337877515)
--(axis cs:9600000,-5.78678454175)
--(axis cs:8000000,-5.94622877775)
--(axis cs:6400000,-6.1363691815)
--(axis cs:4800000,-6.14260806725)
--(axis cs:3200000,-8.33001056475)
--(axis cs:1600000,-30.65855575425)
--(axis cs:0,-75.33036760275)
--cycle;

\addplot [thick, C5, mark=*, mark size=0, mark options={solid}]
table {%
0 -79.00454459
1600000 -32.24266839
3200000 -8.99628744
4800000 -6.58845829
6400000 -6.5139348
8000000 -6.44535738
9600000 -6.22843917
11200000 -6.51144539
12800000 -6.75860554
14400000 -6.93146127
16000000 -6.98857994
17600000 -7.03030769
19200000 -7.2673816
20800000 -7.12079663
22400000 -7.41772029
24000000 -7.6166994
25600000 -7.51272468
27200000 -7.73691681
28800000 -7.70281474
};

\path [draw=C8, fill=C8, opacity=0.2]
(axis cs:0,-78.79149964675)
--(axis cs:0,-85.90704297775)
--(axis cs:400000,-39.51112612475)
--(axis cs:800000,-31.20423370125)
--(axis cs:1200000,-26.14212420875)
--(axis cs:1600000,-26.47944918)
--(axis cs:2000000,-23.109687951)
--(axis cs:2400000,-24.4652093115)
--(axis cs:2800000,-22.377064581)
--(axis cs:3200000,-22.13559921975)
--(axis cs:3600000,-22.28794645325)
--(axis cs:4000000,-23.06730810725)
--(axis cs:4400000,-20.40011833775)
--(axis cs:4800000,-20.6199140255)
--(axis cs:5200000,-20.408310599)
--(axis cs:5600000,-20.9913161855)
--(axis cs:6000000,-21.798932095)
--(axis cs:6400000,-19.53260317725)
--(axis cs:6800000,-22.4109096845)
--(axis cs:7200000,-19.1589935205)
--(axis cs:7600000,-23.0969458345)
--(axis cs:8000000,-25.9635252255)
--(axis cs:8400000,-24.3621759835)
--(axis cs:8800000,-25.5364306825)
--(axis cs:9200000,-27.8634294215)
--(axis cs:9600000,-24.74521602675)
--(axis cs:10000000,-28.80359979725)
--(axis cs:10400000,-23.70007816525)
--(axis cs:10800000,-19.665591137)
--(axis cs:11200000,-23.364980249)
--(axis cs:11600000,-22.46583403925)
--(axis cs:12000000,-22.81929785175)
--(axis cs:12400000,-24.89701827725)
--(axis cs:12800000,-21.396868658)
--(axis cs:13200000,-24.4090694815)
--(axis cs:13600000,-26.26941103425)
--(axis cs:14000000,-22.7253406665)
--(axis cs:14400000,-25.671371496)
--(axis cs:14800000,-22.4133236285)
--(axis cs:15200000,-23.26200177225)
--(axis cs:15600000,-24.1388881085)
--(axis cs:16000000,-26.13487763275)
--(axis cs:16400000,-23.20506106775)
--(axis cs:16800000,-21.4853831405)
--(axis cs:17200000,-24.6806816245)
--(axis cs:17600000,-23.55224661925)
--(axis cs:18000000,-22.253590063)
--(axis cs:18400000,-26.51146896875)
--(axis cs:18800000,-22.798051345)
--(axis cs:19200000,-24.71814329425)
--(axis cs:19600000,-20.884046144)
--(axis cs:20000000,-24.09952173025)
--(axis cs:20400000,-21.9510286385)
--(axis cs:20800000,-24.698812307)
--(axis cs:21200000,-24.3148143685)
--(axis cs:21600000,-21.16660128425)
--(axis cs:22000000,-24.895718091)
--(axis cs:22400000,-22.32230786875)
--(axis cs:22800000,-25.9453319935)
--(axis cs:23200000,-23.57119550925)
--(axis cs:23600000,-22.31713220625)
--(axis cs:24000000,-25.21052241875)
--(axis cs:24400000,-22.28872935525)
--(axis cs:24800000,-24.80307191775)
--(axis cs:25200000,-26.6792838525)
--(axis cs:25600000,-33.615133929)
--(axis cs:26000000,-23.294537331)
--(axis cs:26400000,-21.257977822)
--(axis cs:26800000,-24.527136355)
--(axis cs:27200000,-26.6784466845)
--(axis cs:27600000,-22.9799564585)
--(axis cs:28000000,-25.28378677525)
--(axis cs:28400000,-30.81580250325)
--(axis cs:28800000,-21.88523357575)
--(axis cs:29200000,-25.4214952155)
--(axis cs:29600000,-20.479135723)
--(axis cs:29600000,-17.37276370625)
--(axis cs:29600000,-17.37276370625)
--(axis cs:29200000,-18.73788619925)
--(axis cs:28800000,-17.48229967325)
--(axis cs:28400000,-18.91083976)
--(axis cs:28000000,-17.9660484355)
--(axis cs:27600000,-17.37147233625)
--(axis cs:27200000,-16.57223655875)
--(axis cs:26800000,-18.31037259475)
--(axis cs:26400000,-16.38102238)
--(axis cs:26000000,-18.2973559975)
--(axis cs:25600000,-18.88394983325)
--(axis cs:25200000,-18.49614176575)
--(axis cs:24800000,-17.5795839945)
--(axis cs:24400000,-17.82174711975)
--(axis cs:24000000,-16.42771479775)
--(axis cs:23600000,-16.73405412925)
--(axis cs:23200000,-15.37404252225)
--(axis cs:22800000,-17.69315123775)
--(axis cs:22400000,-17.6106822165)
--(axis cs:22000000,-20.03661031575)
--(axis cs:21600000,-17.112730966)
--(axis cs:21200000,-18.9191594595)
--(axis cs:20800000,-17.18060802025)
--(axis cs:20400000,-16.90372975925)
--(axis cs:20000000,-16.87805006825)
--(axis cs:19600000,-16.41307557125)
--(axis cs:19200000,-18.26787591675)
--(axis cs:18800000,-17.48587650075)
--(axis cs:18400000,-17.71136512725)
--(axis cs:18000000,-16.46236767)
--(axis cs:17600000,-16.989220129)
--(axis cs:17200000,-17.96778606475)
--(axis cs:16800000,-17.68717705675)
--(axis cs:16400000,-17.02916516925)
--(axis cs:16000000,-17.03068465225)
--(axis cs:15600000,-16.4760491135)
--(axis cs:15200000,-17.13081018625)
--(axis cs:14800000,-17.90935832425)
--(axis cs:14400000,-18.0254191785)
--(axis cs:14000000,-15.669730318)
--(axis cs:13600000,-17.683833273)
--(axis cs:13200000,-19.41510340325)
--(axis cs:12800000,-16.90275953675)
--(axis cs:12400000,-16.878441911)
--(axis cs:12000000,-16.83308618275)
--(axis cs:11600000,-16.99636749)
--(axis cs:11200000,-17.407485507)
--(axis cs:10800000,-16.194055644)
--(axis cs:10400000,-18.2102531805)
--(axis cs:10000000,-18.65599601725)
--(axis cs:9600000,-19.09863612525)
--(axis cs:9200000,-18.464011891)
--(axis cs:8800000,-15.512652458)
--(axis cs:8400000,-19.07976879425)
--(axis cs:8000000,-16.91415189)
--(axis cs:7600000,-17.34414981425)
--(axis cs:7200000,-15.364103922)
--(axis cs:6800000,-17.47550995375)
--(axis cs:6400000,-16.59098122075)
--(axis cs:6000000,-17.24313659)
--(axis cs:5600000,-16.15098925825)
--(axis cs:5200000,-16.28883860575)
--(axis cs:4800000,-16.62339258)
--(axis cs:4400000,-15.150303691)
--(axis cs:4000000,-16.64104300025)
--(axis cs:3600000,-15.40904499025)
--(axis cs:3200000,-15.40538560025)
--(axis cs:2800000,-17.415246191)
--(axis cs:2400000,-16.89209121)
--(axis cs:2000000,-15.8478941275)
--(axis cs:1600000,-18.21852818875)
--(axis cs:1200000,-22.21821622575)
--(axis cs:800000,-26.111904424)
--(axis cs:400000,-33.6051650455)
--(axis cs:0,-78.79149964675)
--cycle;

\addplot [thick, C8, mark=*, mark size=0, mark options={solid}]
table {%
0 -82.207614
400000 -36.0846397
800000 -28.37600711
1200000 -24.34359356
1600000 -21.68137796
2000000 -19.25828815
2400000 -20.4727873
2800000 -19.90017081
3200000 -18.1862397
3600000 -18.24988778
4000000 -18.74111089
4400000 -17.74641451
4800000 -18.92077459
5200000 -18.40057556
5600000 -18.0212515
6000000 -19.05198812
6400000 -17.94739231
6800000 -19.85512016
7200000 -16.69094479
7600000 -19.89969022
8000000 -20.51475452
8400000 -21.53044195
8800000 -19.75057922
9200000 -22.49861644
9600000 -22.04866159
10000000 -23.57442469
10400000 -20.03168133
10800000 -17.61753564
11200000 -18.94933798
11600000 -19.29964114
12000000 -18.89424714
12400000 -19.41205475
12800000 -18.81625403
13200000 -21.70598568
13600000 -20.28093938
14000000 -18.25681174
14400000 -21.32144176
14800000 -20.18616788
15200000 -19.95347828
15600000 -20.1010645
16000000 -19.82952327
16400000 -19.64606916
16800000 -18.9916723
17200000 -21.17706067
17600000 -20.22189116
18000000 -19.197764
18400000 -20.97418046
18800000 -19.96288438
19200000 -20.8738946
19600000 -18.45461743
20000000 -19.28820307
20400000 -18.82085543
20800000 -18.80719629
21200000 -21.27669268
21600000 -18.99655908
22000000 -22.5032572
22400000 -19.95181414
22800000 -19.44067763
23200000 -18.42595506
23600000 -18.99832926
24000000 -19.75866448
24400000 -19.5948365
24800000 -20.39769533
25200000 -22.32114416
25600000 -23.35066594
26000000 -20.44493513
26400000 -18.66564826
26800000 -20.35658497
27200000 -18.90462825
27600000 -19.35918882
28000000 -21.24477745
28400000 -22.49455025
28800000 -19.4490495
29200000 -21.631935
29600000 -18.96668016
};

\end{axis}

\end{tikzpicture}

%% file: tikz_plots/reacher/reacher_sparse_iqm_sample_efficiency.tex
\begin{tikzpicture}
\begin{axis}[
legend cell align={left},
legend style={
  fill opacity=0.8,
  draw opacity=1,
  text opacity=1,
  at={(0.97,0.03)},
  anchor=south east,
  draw=lightgray204
},
title={5D Reacher - Sparse},
tick align=outside,
tick pos=left,
x grid style={darkgray176},
xlabel={Number Environment Interactions},
xmajorgrids,
xmin=-1472000, xmax=31912000,
xtick style={color=black},
y grid style={darkgray176},
ylabel={Reward},
ymajorgrids,
ymin=-125, ymax=5,
ytick style={color=black}
]
\path [draw=C0, fill=C0, opacity=0.2]
(axis cs:0,-102.580555960714)
--(axis cs:0,-118.76971750849)
--(axis cs:640000,-24.8537360506726)
--(axis cs:1280000,-18.1856220986577)
--(axis cs:1920000,-13.6104849000828)
--(axis cs:2560000,-10.4237160053574)
--(axis cs:3200000,-8.99557520679147)
--(axis cs:3840000,-6.48790765826594)
--(axis cs:4480000,-5.85028174304192)
--(axis cs:5120000,-4.79407925623378)
--(axis cs:5760000,-5.10106480599175)
--(axis cs:6400000,-4.24987292603042)
--(axis cs:7040000,-3.59890413234501)
--(axis cs:7680000,-2.90555823817469)
--(axis cs:8320000,-2.95437684097964)
--(axis cs:8960000,-2.95173915359711)
--(axis cs:9600000,-2.32947739513634)
--(axis cs:10240000,-2.62996595190685)
--(axis cs:10880000,-2.18123360696393)
--(axis cs:11520000,-2.14825780434834)
--(axis cs:12160000,-2.32968789078343)
--(axis cs:12800000,-1.94072054677848)
--(axis cs:13440000,-2.07448388681231)
--(axis cs:14080000,-2.16085880382432)
--(axis cs:14720000,-1.95270573497629)
--(axis cs:15360000,-1.918978674904)
--(axis cs:16000000,-1.56371169996908)
--(axis cs:16640000,-1.78070881880984)
--(axis cs:17280000,-1.52398883721383)
--(axis cs:17920000,-1.96718140579942)
--(axis cs:18560000,-1.69824933871197)
--(axis cs:19200000,-1.40828233878042)
--(axis cs:19840000,-1.59775981555558)
--(axis cs:20480000,-1.98563140101485)
--(axis cs:21120000,-1.59174379950224)
--(axis cs:21760000,-1.46288350399775)
--(axis cs:22400000,-1.37344725995206)
--(axis cs:23040000,-1.90443762580937)
--(axis cs:23680000,-1.46804140093914)
--(axis cs:24320000,-1.54536803576294)
--(axis cs:24960000,-1.87319400347657)
--(axis cs:25600000,-1.71157191813956)
--(axis cs:26240000,-1.84850547785089)
--(axis cs:26880000,-1.85661570639884)
--(axis cs:27520000,-2.1540952462786)
--(axis cs:28160000,-1.79098497136347)
--(axis cs:28800000,-1.91514601438773)
--(axis cs:29440000,-1.81245170980871)
--(axis cs:29440000,-1.22999793392381)
--(axis cs:29440000,-1.22999793392381)
--(axis cs:28800000,-1.25381231505005)
--(axis cs:28160000,-1.18748275182826)
--(axis cs:27520000,-1.28701743460027)
--(axis cs:26880000,-1.35955984073811)
--(axis cs:26240000,-1.21147644366557)
--(axis cs:25600000,-1.09564117301307)
--(axis cs:24960000,-1.13571935862397)
--(axis cs:24320000,-1.07404183973358)
--(axis cs:23680000,-1.00189173518314)
--(axis cs:23040000,-1.17312561994621)
--(axis cs:22400000,-0.95526666534692)
--(axis cs:21760000,-1.07090289820821)
--(axis cs:21120000,-1.05115059866126)
--(axis cs:20480000,-1.30810716458762)
--(axis cs:19840000,-1.21163281996339)
--(axis cs:19200000,-0.948635158316741)
--(axis cs:18560000,-1.22793714914317)
--(axis cs:17920000,-1.27446177015447)
--(axis cs:17280000,-1.15452802873507)
--(axis cs:16640000,-1.19858165523795)
--(axis cs:16000000,-1.17813618216561)
--(axis cs:15360000,-1.2779324470806)
--(axis cs:14720000,-1.37110645852216)
--(axis cs:14080000,-1.51770206567279)
--(axis cs:13440000,-1.44690449719705)
--(axis cs:12800000,-1.47791080114637)
--(axis cs:12160000,-1.72361059630364)
--(axis cs:11520000,-1.66306563484022)
--(axis cs:10880000,-1.65553321678944)
--(axis cs:10240000,-1.94529381995673)
--(axis cs:9600000,-1.82767292603873)
--(axis cs:8960000,-2.35558040767295)
--(axis cs:8320000,-2.2115842309846)
--(axis cs:7680000,-2.47311441728713)
--(axis cs:7040000,-2.70204678722479)
--(axis cs:6400000,-3.51328571317469)
--(axis cs:5760000,-3.90078057925662)
--(axis cs:5120000,-3.76161621022445)
--(axis cs:4480000,-4.94168669849114)
--(axis cs:3840000,-5.37566050726666)
--(axis cs:3200000,-7.37913689635145)
--(axis cs:2560000,-8.60339974937336)
--(axis cs:1920000,-11.8785071384255)
--(axis cs:1280000,-15.5021009931008)
--(axis cs:640000,-20.0922523547992)
--(axis cs:0,-102.580555960714)
--cycle;

\addplot [thick, C0, mark=*, mark size=0, mark options={solid}]
table {%
0 -110.103809931734
640000 -22.5783719439843
1280000 -16.8224745899867
1920000 -12.7485518088624
2560000 -9.31041901788333
3200000 -8.06597732388567
3840000 -5.78734259305789
4480000 -5.31778546810841
5120000 -4.2074431232624
5760000 -4.44439557531722
6400000 -3.87919807026901
7040000 -3.12504533964971
7680000 -2.69340076846944
8320000 -2.52832922143538
8960000 -2.67698471430748
9600000 -2.0470337520222
10240000 -2.18784405868604
10880000 -1.92138335639022
11520000 -1.90831747262455
12160000 -2.01019310552202
12800000 -1.69447830267964
13440000 -1.73920693762851
14080000 -1.81242175867444
14720000 -1.62402180642166
15360000 -1.5836091431314
16000000 -1.37559715877191
16640000 -1.44554726642315
17280000 -1.35348699317283
17920000 -1.49340187116918
18560000 -1.39681790219763
19200000 -1.1470190409546
19840000 -1.39027900906519
20480000 -1.64718373892329
21120000 -1.28338160263164
21760000 -1.26546411123121
22400000 -1.15704641192748
23040000 -1.51766407684758
23680000 -1.19526788818549
24320000 -1.26668728558739
24960000 -1.46201651909865
25600000 -1.36970476793559
26240000 -1.46235171024555
26880000 -1.5488087687783
27520000 -1.67195324646318
28160000 -1.53952077047819
28800000 -1.60163518264851
29440000 -1.43701636657618
};

\path [draw=C1, fill=C1, opacity=0.2]
(axis cs:0,-1233.22492706375)
--(axis cs:0,-1975.8282964095)
--(axis cs:800000,-192.26201389775)
--(axis cs:1600000,-88.9737674065)
--(axis cs:2400000,-68.023081966)
--(axis cs:3200000,-62.01525578175)
--(axis cs:4000000,-54.26616625025)
--(axis cs:4800000,-44.854208288)
--(axis cs:5600000,-39.13595829775)
--(axis cs:6400000,-53.64720977875)
--(axis cs:7200000,-45.14812309825)
--(axis cs:8000000,-40.8774312015)
--(axis cs:8800000,-43.611659891)
--(axis cs:9600000,-35.59425513875)
--(axis cs:10400000,-43.6645996335)
--(axis cs:11200000,-40.35065571325)
--(axis cs:12000000,-39.6984393305)
--(axis cs:12800000,-35.5833456345)
--(axis cs:13600000,-32.7749893715)
--(axis cs:14400000,-34.24772617425)
--(axis cs:15200000,-32.01586984925)
--(axis cs:16000000,-31.458535595)
--(axis cs:16800000,-31.1663340735)
--(axis cs:17600000,-38.97219989025)
--(axis cs:18400000,-37.9490979735)
--(axis cs:19200000,-35.0584121985)
--(axis cs:20000000,-45.5447440495)
--(axis cs:20800000,-33.469654539)
--(axis cs:21600000,-35.568964282)
--(axis cs:22400000,-35.9963616285)
--(axis cs:23200000,-31.6175507995)
--(axis cs:24000000,-32.94011212925)
--(axis cs:24800000,-30.3019194515)
--(axis cs:25600000,-30.37896885775)
--(axis cs:26400000,-32.7055418635)
--(axis cs:27200000,-34.2337312255)
--(axis cs:28000000,-33.115884899)
--(axis cs:28800000,-29.04616046425)
--(axis cs:29600000,-30.58872274625)
--(axis cs:29600000,-20.90564165625)
--(axis cs:29600000,-20.90564165625)
--(axis cs:28800000,-20.17645706675)
--(axis cs:28000000,-21.28226049025)
--(axis cs:27200000,-20.144937273)
--(axis cs:26400000,-21.52346611025)
--(axis cs:25600000,-21.62233774575)
--(axis cs:24800000,-19.6795585945)
--(axis cs:24000000,-22.7015236685)
--(axis cs:23200000,-22.591099545)
--(axis cs:22400000,-23.32310333225)
--(axis cs:21600000,-24.0520702645)
--(axis cs:20800000,-21.10054653875)
--(axis cs:20000000,-25.617830085)
--(axis cs:19200000,-20.2939814505)
--(axis cs:18400000,-21.92295942325)
--(axis cs:17600000,-21.67452427675)
--(axis cs:16800000,-21.547533223)
--(axis cs:16000000,-24.42332975225)
--(axis cs:15200000,-24.3998541965)
--(axis cs:14400000,-24.92449601025)
--(axis cs:13600000,-22.65551848275)
--(axis cs:12800000,-23.91089746325)
--(axis cs:12000000,-26.0622401675)
--(axis cs:11200000,-25.785326573)
--(axis cs:10400000,-25.75894404575)
--(axis cs:9600000,-26.34552781925)
--(axis cs:8800000,-25.7816256185)
--(axis cs:8000000,-30.434036328)
--(axis cs:7200000,-31.0795451705)
--(axis cs:6400000,-28.127026205)
--(axis cs:5600000,-27.15455199925)
--(axis cs:4800000,-27.426777363)
--(axis cs:4000000,-35.9240665445)
--(axis cs:3200000,-39.37794853575)
--(axis cs:2400000,-41.22501126625)
--(axis cs:1600000,-65.42098248575)
--(axis cs:800000,-133.684880422)
--(axis cs:0,-1233.22492706375)
--cycle;

\addplot [thick, C1, mark=*, mark size=0, mark options={solid}]
table {%
0 -1618.87501635
800000 -161.25423804
1600000 -76.95257208
2400000 -55.40487248
3200000 -51.73321746
4000000 -43.60125941
4800000 -34.12536485
5600000 -32.83467456
6400000 -35.24979075
7200000 -38.40169788
8000000 -34.88408664
8800000 -31.19070863
9600000 -30.62893872
10400000 -31.68109773
11200000 -29.62193869
12000000 -31.98558061
12800000 -29.11610751
13600000 -27.68165932
14400000 -29.75672858
15200000 -27.8755656
16000000 -28.1904063
16800000 -27.06080225
17600000 -27.84810586
18400000 -27.70628676
19200000 -25.75441093
20000000 -31.32367162
20800000 -25.96584874
21600000 -30.11796088
22400000 -28.2553759
23200000 -26.01279499
24000000 -27.71797081
24800000 -24.41757784
25600000 -26.05547179
26400000 -27.93371804
27200000 -27.21542671
28000000 -27.57696618
28800000 -24.27631279
29600000 -26.19810486
};
\path [draw=C2, fill=C2, opacity=0.2]
(axis cs:6416000,-996.760465308)
--(axis cs:7216000,-1188.62521421875)
--(axis cs:8016000,-990.33978851275)
--(axis cs:8816000,-914.09750830225)
--(axis cs:9616000,-886.482414763)
--(axis cs:10416000,-954.87457275275)
--(axis cs:11216000,-815.1324615105)
--(axis cs:12016000,-779.702020571)
--(axis cs:12816000,-733.66176910225)
--(axis cs:13616000,-688.1269586585)
--(axis cs:14416000,-674.6948913215)
--(axis cs:15216000,-651.46190441075)
--(axis cs:16016000,-629.9694678545)
--(axis cs:16816000,-629.3665233645)
--(axis cs:17616000,-582.69514659675)
--(axis cs:18416000,-564.553303346)
--(axis cs:19216000,-557.705622194)
--(axis cs:20016000,-528.6306987345)
--(axis cs:20816000,-541.43169350675)
--(axis cs:21616000,-522.771248445)
--(axis cs:22416000,-517.9439595465)
--(axis cs:23216000,-507.42758763325)
--(axis cs:24016000,-498.13881358325)
--(axis cs:24816000,-494.35196967)
--(axis cs:25616000,-499.050244981)
--(axis cs:26416000,-472.24032710225)
--(axis cs:27216000,-476.37463018625)
--(axis cs:28016000,-478.32482385075)
--(axis cs:28816000,-473.84976648825)
--(axis cs:29616000,-470.3687130705)
--(axis cs:29616000,-96.6345448072501)
--(axis cs:29616000,-96.6345448072501)
--(axis cs:28816000,-93.9127164765)
--(axis cs:28016000,-106.0180931865)
--(axis cs:27216000,-93.61470866125)
--(axis cs:26416000,-104.78689452025)
--(axis cs:25616000,-111.39044553325)
--(axis cs:24816000,-102.663256423)
--(axis cs:24016000,-102.2310970155)
--(axis cs:23216000,-108.884659985)
--(axis cs:22416000,-100.29796249625)
--(axis cs:21616000,-100.1461108245)
--(axis cs:20816000,-102.9438240235)
--(axis cs:20016000,-109.34568983125)
--(axis cs:19216000,-117.1644313295)
--(axis cs:18416000,-108.37116183025)
--(axis cs:17616000,-123.2321269485)
--(axis cs:16816000,-124.21782601125)
--(axis cs:16016000,-127.8566940515)
--(axis cs:15216000,-134.96922940375)
--(axis cs:14416000,-129.57735867925)
--(axis cs:13616000,-140.40989229575)
--(axis cs:12816000,-143.5219404415)
--(axis cs:12016000,-141.42966810725)
--(axis cs:11216000,-153.37819051375)
--(axis cs:10416000,-155.30516515425)
--(axis cs:9616000,-193.034359529)
--(axis cs:8816000,-204.07734426475)
--(axis cs:8016000,-201.72546026425)
--(axis cs:7216000,-224.14430557825)
--(axis cs:6416000,-227.11054894275)
--(axis cs:5616000,-296.34253836675)
--(axis cs:4816000,-290.101108795501)
--(axis cs:4016000,-362.76029769225)
--(axis cs:3216000,-450.958547666)
--(axis cs:2416000,-655.523202635)
--cycle;

\addplot [thick, C2, mark=*, mark size=0, mark options={solid}]
table {%
3216000 -888.08319107
4016000 -747.1967382
4816000 -683.73164416
5616000 -616.59805296
6416000 -578.29637899
7216000 -589.30810556
8016000 -499.44326529
8816000 -475.47165405
9616000 -457.98300831
10416000 -417.72225101
11216000 -418.07163843
12016000 -393.62306166
12816000 -382.18107156
13616000 -367.3137079
14416000 -351.30107486
15216000 -344.33635326
16016000 -335.5577415
16816000 -329.84695631
17616000 -311.36123798
18416000 -288.0816708
19216000 -295.50934975
20016000 -279.18906416
20816000 -272.81065879
21616000 -269.75070327
22416000 -263.59521728
23216000 -269.1639867
24016000 -256.73841641
24816000 -250.80803856
25616000 -259.77949451
26416000 -249.0738373
27216000 -243.65254177
28016000 -243.41681231
28816000 -241.43372605
29616000 -240.15788125
};

\path [draw=C6, fill=C6, opacity=0.2]
(axis cs:41200,-96.22297011125)
--(axis cs:41200,-108.22129079375)
--(axis cs:863600,-122.77037149875)
--(axis cs:1700000,-66.74242055125)
--(axis cs:2522000,-80.17543023125)
--(axis cs:3348000,-63.00120088625)
--(axis cs:4186400,-56.36629841375)
--(axis cs:5020400,-78.54847586875)
--(axis cs:5847600,-58.1328518675)
--(axis cs:6675200,-67.211693515)
--(axis cs:7505200,-48.941466815)
--(axis cs:8330000,-75.84489787)
--(axis cs:9161600,-47.5757650975)
--(axis cs:9989200,-63.43443767375)
--(axis cs:10821600,-64.4992519625)
--(axis cs:11648000,-73.409794295)
--(axis cs:12479600,-73.88162828625)
--(axis cs:13311600,-70.410041439375)
--(axis cs:14140400,-118.7504023725)
--(axis cs:14974000,-57.5680315425)
--(axis cs:15806400,-45.1930339)
--(axis cs:16639200,-74.9991764375)
--(axis cs:17472800,-235.078612825)
--(axis cs:18298800,-67.45415132125)
--(axis cs:19136000,-58.61726976875)
--(axis cs:19958000,-51.208665880125)
--(axis cs:20783200,-54.1984312755)
--(axis cs:21616000,-82.84704027075)
--(axis cs:22445200,-58.60813299875)
--(axis cs:23276800,-97.324052594375)
--(axis cs:24103600,-71.0555624425)
--(axis cs:24927200,-48.697114279125)
--(axis cs:25754400,-57.0816406305)
--(axis cs:26578000,-66.0284295985)
--(axis cs:27408400,-90.37153057825)
--(axis cs:27408400,-25.89866229975)
--(axis cs:27408400,-25.89866229975)
--(axis cs:26578000,-27.895376474125)
--(axis cs:25754400,-15.33966604575)
--(axis cs:24927200,-19.687703798875)
--(axis cs:24103600,-19.92791080125)
--(axis cs:23276800,-16.907269835625)
--(axis cs:22445200,-25.40604110125)
--(axis cs:21616000,-24.282635950375)
--(axis cs:20783200,-21.824307731375)
--(axis cs:19958000,-22.394493515)
--(axis cs:19136000,-30.17825346125)
--(axis cs:18298800,-22.97394558625)
--(axis cs:17472800,-25.20764170875)
--(axis cs:16639200,-24.0879123025)
--(axis cs:15806400,-24.1194810775)
--(axis cs:14974000,-24.032108527)
--(axis cs:14140400,-47.24793063375)
--(axis cs:13311600,-33.401315219375)
--(axis cs:12479600,-39.087895925)
--(axis cs:11648000,-32.73577699375)
--(axis cs:10821600,-34.026813735)
--(axis cs:9989200,-35.22190696125)
--(axis cs:9161600,-31.70409089)
--(axis cs:8330000,-37.67584936875)
--(axis cs:7505200,-32.79046596625)
--(axis cs:6675200,-38.78091758)
--(axis cs:5847600,-38.29787153)
--(axis cs:5020400,-44.5034336275)
--(axis cs:4186400,-41.47261826)
--(axis cs:3348000,-45.46937670125)
--(axis cs:2522000,-48.7966925925)
--(axis cs:1700000,-53.750182955)
--(axis cs:863600,-87.89200526375)
--(axis cs:41200,-96.22297011125)
--cycle;

\addplot [thick, C6, mark=*, mark size=0, mark options={solid}]
table {%
41200 -102.0666363
863600 -104.52086915
1700000 -59.8728618
2522000 -62.2194617
3348000 -53.7286059
4186400 -48.64508225
5020400 -58.05640055
5847600 -46.74099685
6675200 -50.35780515
7505200 -40.19704515
8330000 -54.7528279
9161600 -38.99229795
9989200 -47.0066104
10821600 -47.10737885
11648000 -50.67562185
12479600 -55.5481872
13311600 -50.707618325
14140400 -79.13863485
14974000 -38.140971445
15806400 -33.79160095
16639200 -45.59075285
17472800 -99.7966942
18298800 -41.164181
19136000 -43.50043485
19958000 -35.053063815
20783200 -36.48723118
21616000 -48.706014705
22445200 -41.19262485
23276800 -47.820927
24103600 -40.251637135
24927200 -32.63388102
25754400 -31.26943963
26578000 -45.991348975
27408400 -52.03368353
};

\path [draw=C4, fill=C4, opacity=0.2]
(axis cs:12800,-92.5461608550286)
--(axis cs:12800,-129.679489505239)
--(axis cs:1292800,-25.8100353078756)
--(axis cs:2572800,-19.1762455050454)
--(axis cs:3852800,-15.7044608715739)
--(axis cs:5132800,-16.2084672344551)
--(axis cs:6412800,-14.3109130125772)
--(axis cs:7692800,-13.4270770997847)
--(axis cs:8972800,-14.5632346331471)
--(axis cs:10252800,-12.7886822680592)
--(axis cs:11532800,-14.7748553357038)
--(axis cs:12812800,-13.8847569965757)
--(axis cs:14092800,-13.4208604798537)
--(axis cs:15372800,-15.8926452597058)
--(axis cs:16652800,-16.7608967423751)
--(axis cs:17932800,-8.32936240246671)
--(axis cs:19212800,-13.0033111153689)
--(axis cs:20492800,-14.4560679063042)
--(axis cs:21772800,-13.2562208022342)
--(axis cs:23052800,-14.6670369649199)
--(axis cs:24332800,-12.8642247669418)
--(axis cs:25612800,-12.4843082958058)
--(axis cs:26892800,-11.4395135949265)
--(axis cs:28172800,-13.2931494086909)
--(axis cs:29452800,-11.4907614562282)
--(axis cs:29452800,-6.81631585119059)
--(axis cs:29452800,-6.81631585119059)
--(axis cs:28172800,-6.40163151759735)
--(axis cs:26892800,-5.92417091725651)
--(axis cs:25612800,-7.81390821338587)
--(axis cs:24332800,-5.79138472607406)
--(axis cs:23052800,-8.10349965238765)
--(axis cs:21772800,-6.33993308025917)
--(axis cs:20492800,-6.2418974916779)
--(axis cs:19212800,-8.09965440701397)
--(axis cs:17932800,-4.40195637822387)
--(axis cs:16652800,-8.44174900591085)
--(axis cs:15372800,-8.53559098564235)
--(axis cs:14092800,-7.94397024129264)
--(axis cs:12812800,-9.44864854533045)
--(axis cs:11532800,-7.8851311271626)
--(axis cs:10252800,-7.61382717143751)
--(axis cs:8972800,-7.442868841952)
--(axis cs:7692800,-7.07999839848616)
--(axis cs:6412800,-8.73431138283031)
--(axis cs:5132800,-8.54307073855751)
--(axis cs:3852800,-11.3830744721563)
--(axis cs:2572800,-11.935262519277)
--(axis cs:1292800,-17.9845450218585)
--(axis cs:12800,-92.5461608550286)
--cycle;

\addplot [thick, C4, mark=*, mark size=0, mark options={solid}]
table {%
12800 -109.624021391876
1292800 -21.7420956436938
2572800 -14.5311178583
3852800 -13.2101494467144
5132800 -11.6868589265178
6412800 -11.4416141222293
7692800 -9.73905394993923
8972800 -10.2170584319859
10252800 -10.1948893254106
11532800 -10.5832176821178
12812800 -11.7467313944959
14092800 -10.4000650681322
15372800 -12.0607732557247
16652800 -12.3074167335109
17932800 -5.88024015045429
19212800 -10.1401479781539
20492800 -9.43478993465618
21772800 -9.59083508117152
23052800 -11.3513279206631
24332800 -8.77598759440872
25612800 -10.0612902745955
26892800 -7.15116960742507
28172800 -9.07090747006746
29452800 -8.77132173994418
};

<\path [draw=C7, fill=C7, opacity=0.2]
(axis cs:120000,-96.9534127402004)
--(axis cs:120000,-110.199354783297)
--(axis cs:1320000,-30.5368304628427)
--(axis cs:2520000,-26.2883652049538)
--(axis cs:3720000,-23.0171887452816)
--(axis cs:4920000,-21.2441827838459)
--(axis cs:6120000,-18.838267277395)
--(axis cs:7320000,-19.9061225847521)
--(axis cs:8520000,-19.9004802905736)
--(axis cs:9720000,-19.8760941683491)
--(axis cs:10920000,-19.6679458317666)
--(axis cs:12120000,-19.1942641818355)
--(axis cs:13320000,-18.6868513321961)
--(axis cs:14520000,-19.53604285769)
--(axis cs:15720000,-17.4656372975223)
--(axis cs:16920000,-19.2995046501027)
--(axis cs:18120000,-17.3651743990009)
--(axis cs:19320000,-19.0403860626927)
--(axis cs:20520000,-18.9129586929809)
--(axis cs:21720000,-16.6011860113772)
--(axis cs:22920000,-19.5142267477535)
--(axis cs:24120000,-16.9336807659342)
--(axis cs:25320000,-18.6382730090572)
--(axis cs:26520000,-17.6773745416543)
--(axis cs:27720000,-19.0401404184716)
--(axis cs:28920000,-19.4595196070063)
--(axis cs:28920000,-15.0071894200028)
--(axis cs:28920000,-15.0071894200028)
--(axis cs:27720000,-15.0406574585075)
--(axis cs:26520000,-13.264343023825)
--(axis cs:25320000,-12.8593468276954)
--(axis cs:24120000,-14.3096198655651)
--(axis cs:22920000,-16.3656495717378)
--(axis cs:21720000,-13.5732605025297)
--(axis cs:20520000,-14.6103711938754)
--(axis cs:19320000,-15.6046153430199)
--(axis cs:18120000,-13.2162234364703)
--(axis cs:16920000,-14.8731657540105)
--(axis cs:15720000,-15.4494590704486)
--(axis cs:14520000,-14.2873792498626)
--(axis cs:13320000,-15.300543548281)
--(axis cs:12120000,-15.8360250361627)
--(axis cs:10920000,-15.7887983678064)
--(axis cs:9720000,-16.386999574621)
--(axis cs:8520000,-15.9224292003888)
--(axis cs:7320000,-16.858503347747)
--(axis cs:6120000,-14.8606821939897)
--(axis cs:4920000,-17.8003245037475)
--(axis cs:3720000,-18.7870448893237)
--(axis cs:2520000,-21.3210467917098)
--(axis cs:1320000,-24.7872380714841)
--(axis cs:120000,-96.9534127402004)
--cycle;

\addplot [thick, C7, mark=*, mark size=0, mark options={solid}]
table {%
120000 -103.940541889479
1320000 -26.9993336198246
2520000 -23.1935194938102
3720000 -21.1295897241459
4920000 -19.3457149493713
6120000 -16.4686781274257
7320000 -18.186464787214
8520000 -17.5860513815509
9720000 -18.1240556720399
10920000 -17.2957877677369
12120000 -17.5216780227307
13320000 -17.0801863756709
14520000 -17.1226195692877
15720000 -16.3331576737321
16920000 -17.2000206277584
18120000 -14.9833770437023
19320000 -17.1873587708049
20520000 -16.4537151139861
21720000 -15.0407430396366
22920000 -18.0680724519519
24120000 -15.300284787324
25320000 -15.2418491094564
26520000 -15.5213188092058
27720000 -16.8527913627188
28920000 -16.8283819249577
};

\path [draw=C5, fill=C5, opacity=0.2]
(axis cs:0,-1226.684258954)
--(axis cs:0,-2080.33147683075)
--(axis cs:1600000,-123.410569892)
--(axis cs:3200000,-54.1054241325)
--(axis cs:4800000,-34.7401974)
--(axis cs:6400000,-26.9855102525)
--(axis cs:8000000,-23.17397464325)
--(axis cs:9600000,-23.048392996)
--(axis cs:11200000,-20.8130964705)
--(axis cs:12800000,-21.130581853)
--(axis cs:14400000,-22.306504992)
--(axis cs:16000000,-21.23398616025)
--(axis cs:17600000,-20.81882680625)
--(axis cs:19200000,-18.16250127975)
--(axis cs:20800000,-19.70766600675)
--(axis cs:22400000,-21.0160171905)
--(axis cs:24000000,-17.8334497195)
--(axis cs:25600000,-19.7878144265)
--(axis cs:27200000,-19.261026207)
--(axis cs:28800000,-17.44055900675)
--(axis cs:28800000,-11.13719780525)
--(axis cs:28800000,-11.13719780525)
--(axis cs:27200000,-10.498404362)
--(axis cs:25600000,-12.109488554)
--(axis cs:24000000,-10.6798414885)
--(axis cs:22400000,-11.14657149675)
--(axis cs:20800000,-10.59962612075)
--(axis cs:19200000,-10.588897293)
--(axis cs:17600000,-12.13512022375)
--(axis cs:16000000,-12.50364729075)
--(axis cs:14400000,-12.7108785675)
--(axis cs:12800000,-12.78110308275)
--(axis cs:11200000,-14.5528376465)
--(axis cs:9600000,-15.2104995505)
--(axis cs:8000000,-17.5992498215)
--(axis cs:6400000,-20.77623155925)
--(axis cs:4800000,-25.617975888)
--(axis cs:3200000,-42.45072384)
--(axis cs:1600000,-95.55091064775)
--(axis cs:0,-1226.684258954)
--cycle;

\addplot [thick, C5, mark=*, mark size=0, mark options={solid}]
table {%
0 -1600.33797412
1600000 -107.16267348
3200000 -47.4457137
4800000 -29.92845559
6400000 -23.62940457
8000000 -19.8602451
9600000 -19.71486898
11200000 -17.63365531
12800000 -17.29637819
14400000 -17.76328023
16000000 -17.16065826
17600000 -17.01151253
19200000 -14.52984836
20800000 -15.48611766
22400000 -16.43473771
24000000 -13.9060871
25600000 -16.37577843
27200000 -15.51013976
28800000 -14.07539819
};

\end{axis}

\end{tikzpicture}

%% file: tikz_plots/box_pushing/box_pushing_iqm_sample_efficiency.tex
\begin{tikzpicture}

\begin{axis}[
legend cell align={left},
legend style={fill opacity=0.8, draw opacity=1, text opacity=1, draw=lightgray204, at={(0.03,0.03)},  anchor=north west},
title={Box Pushing - Dense},
tick align=outside,
tick pos=left,
x grid style={darkgray176},
xlabel={Number Environment Interactions},
xmajorgrids,
xmin=-2500000, xmax=42000000.05,
xtick style={color=black},
y grid style={darkgray176},
ylabel={Success Rate},
ymajorgrids,
ymin=-0.05, ymax=1.05,
ytick style={color=black}
]
\path [draw=C1, fill=C1, opacity=0.2]
(axis cs:0,0)
--(axis cs:0,0)
--(axis cs:1600000,0)
--(axis cs:3200000,0.0444444444444444)
--(axis cs:4800000,0.377777777777778)
--(axis cs:6400000,0.466666666666667)
--(axis cs:8000000,0.544444444444444)
--(axis cs:9600000,0.744444444444444)
--(axis cs:11200000,0.755555555555556)
--(axis cs:12800000,0.5)
--(axis cs:14400000,0.577777777777778)
--(axis cs:16000000,0.7)
--(axis cs:17600000,0.733333333333333)
--(axis cs:19200000,0.711111111111111)
--(axis cs:20800000,0.8)
--(axis cs:22400000,0.822222222222222)
--(axis cs:24000000,0.755555555555556)
--(axis cs:25600000,0.766666666666667)
--(axis cs:27200000,0.844444444444445)
--(axis cs:28800000,0.844444444444445)
--(axis cs:30400000,0.755555555555556)
--(axis cs:32000000,0.9)
--(axis cs:33600000,0.833333333333333)
--(axis cs:35200000,0.822222222222222)
--(axis cs:36800000,0.9)
--(axis cs:38400000,0.877777777777778)
--(axis cs:40000000,0.922222222222222)
--(axis cs:40000000,1)
--(axis cs:38400000,0.955555555555555)
--(axis cs:36800000,1)
--(axis cs:35200000,0.966666666666667)
--(axis cs:33600000,0.988888888888889)
--(axis cs:32000000,0.988888888888889)
--(axis cs:30400000,0.977777777777778)
--(axis cs:28800000,0.988888888888889)
--(axis cs:27200000,0.933333333333333)
--(axis cs:25600000,0.966666666666667)
--(axis cs:24000000,0.933333333333333)
--(axis cs:22400000,0.96694444444444)
--(axis cs:20800000,0.966666666666667)
--(axis cs:19200000,0.922222222222222)
--(axis cs:17600000,0.966666666666667)
--(axis cs:16000000,0.888888888888889)
--(axis cs:14400000,0.877777777777778)
--(axis cs:12800000,0.811111111111111)
--(axis cs:11200000,0.966666666666667)
--(axis cs:9600000,0.933333333333333)
--(axis cs:8000000,0.811111111111111)
--(axis cs:6400000,0.833333333333333)
--(axis cs:4800000,0.6)
--(axis cs:3200000,0.255555555555556)
--(axis cs:1600000,0.0333333333333333)
--(axis cs:0,0)
--cycle;

\addplot [thick, C1, mark=*, mark size=0, mark options={solid}]
table {%
0 0
1600000 0
3200000 0.133333333333333
4800000 0.511111111111111
6400000 0.677777777777778
8000000 0.7
9600000 0.855555555555556
11200000 0.888888888888889
12800000 0.688888888888889
14400000 0.766666666666667
16000000 0.8
17600000 0.877777777777778
19200000 0.844444444444445
20800000 0.9
22400000 0.911111111111111
24000000 0.855555555555556
25600000 0.888888888888889
27200000 0.9
28800000 0.922222222222222
30400000 0.922222222222222
32000000 0.944444444444444
33600000 0.922222222222222
35200000 0.9
36800000 0.955555555555555
38400000 0.911111111111111
40000000 0.966666666666667
};

\path [draw=C2, fill=C2, opacity=0.2]
(axis cs:16000,0)
--(axis cs:16000,0)
--(axis cs:1616000,0)
--(axis cs:3216000,0)
--(axis cs:4816000,0)
--(axis cs:6416000,0)
--(axis cs:8016000,0)
--(axis cs:9616000,0)
--(axis cs:11216000,0)
--(axis cs:12816000,0)
--(axis cs:14416000,0)
--(axis cs:16016000,0)
--(axis cs:17616000,0)
--(axis cs:19216000,0)
--(axis cs:20816000,0)
--(axis cs:22416000,0)
--(axis cs:24016000,0)
--(axis cs:25616000,0)
--(axis cs:27216000,0)
--(axis cs:28816000,0)
--(axis cs:30416000,0)
--(axis cs:32016000,0)
--(axis cs:33616000,0.0714285714285714)
--(axis cs:35216000,0)
--(axis cs:36816000,0)
--(axis cs:38416000,0)
--(axis cs:40016000,0)
--(axis cs:40016000,0.214285714285714)
--(axis cs:38416000,0.214285714285714)
--(axis cs:36816000,0.428571428571429)
--(axis cs:35216000,0.214285714285714)
--(axis cs:33616000,0.5)
--(axis cs:32016000,0.357142857142857)
--(axis cs:30416000,0)
--(axis cs:28816000,0)
--(axis cs:27216000,0)
--(axis cs:25616000,0)
--(axis cs:24016000,0.214285714285714)
--(axis cs:22416000,0.214285714285714)
--(axis cs:20816000,0.357142857142857)
--(axis cs:19216000,0.214285714285714)
--(axis cs:17616000,0.214285714285714)
--(axis cs:16016000,0)
--(axis cs:14416000,0)
--(axis cs:12816000,0.214285714285714)
--(axis cs:11216000,0.214285714285714)
--(axis cs:9616000,0)
--(axis cs:8016000,0)
--(axis cs:6416000,0)
--(axis cs:4816000,0)
--(axis cs:3216000,0)
--(axis cs:1616000,0)
--(axis cs:16000,0)
--cycle;

\addplot [thick, C2, mark=*, mark size=0, mark options={solid}]
table {%
16000 0
1616000 0
3216000 0
4816000 0
6416000 0
8016000 0
9616000 0
11216000 0.0714285714285714
12816000 0.0714285714285714
14416000 0
16016000 0
17616000 0.0714285714285714
19216000 0.0714285714285714
20816000 0.142857142857143
22416000 0.0714285714285714
24016000 0.0714285714285714
25616000 0
27216000 0
28816000 0
30416000 0
32016000 0.142857142857143
33616000 0.285714285714286
35216000 0.0714285714285714
36816000 0.214285714285714
38416000 0.0714285714285714
40016000 0.0714285714285714
};

\path [draw=C4, fill=C4, opacity=0.2]
(axis cs:0,0)
--(axis cs:0,0)
--(axis cs:1405200,0)
--(axis cs:2810400,0)
--(axis cs:4215600,0)
--(axis cs:5620800,0)
--(axis cs:7026000,0)
--(axis cs:8431200,0)
--(axis cs:9836400,0)
--(axis cs:11241600,0)
--(axis cs:12646800,0)
--(axis cs:14052000,0.03)
--(axis cs:15457200,0.03)
--(axis cs:16862400,0.04)
--(axis cs:18267600,0.04)
--(axis cs:19672800,0.03)
--(axis cs:21078000,0.04)
--(axis cs:22483200,0.11)
--(axis cs:23888400,0.02)
--(axis cs:25293600,0.1)
--(axis cs:26698800,0.11)
--(axis cs:28104000,0.16)
--(axis cs:29509200,0.12)
--(axis cs:30914400,0.12)
--(axis cs:32319600,0.16)
--(axis cs:33724800,0.14)
--(axis cs:35130000,0.14)
--(axis cs:36535200,0.14)
--(axis cs:37940400,0.16)
--(axis cs:39345600,0.18)
--(axis cs:39345600,0.45)
--(axis cs:37940400,0.48)
--(axis cs:36535200,0.42)
--(axis cs:35130000,0.4)
--(axis cs:33724800,0.45)
--(axis cs:32319600,0.45)
--(axis cs:30914400,0.33)
--(axis cs:29509200,0.31)
--(axis cs:28104000,0.36)
--(axis cs:26698800,0.36)
--(axis cs:25293600,0.3)
--(axis cs:23888400,0.3)
--(axis cs:22483200,0.24)
--(axis cs:21078000,0.19)
--(axis cs:19672800,0.15)
--(axis cs:18267600,0.16)
--(axis cs:16862400,0.18)
--(axis cs:15457200,0.13)
--(axis cs:14052000,0.11)
--(axis cs:12646800,0.03)
--(axis cs:11241600,0.06)
--(axis cs:9836400,0.1)
--(axis cs:8431200,0.03)
--(axis cs:7026000,0.03)
--(axis cs:5620800,0.01)
--(axis cs:4215600,0)
--(axis cs:2810400,0)
--(axis cs:1405200,0)
--(axis cs:0,0)
--cycle;

\addplot [thick, C4, mark=*, mark size=0, mark options={solid}]
table {%
0 0
1405200 0
2810400 0
4215600 0
5620800 0
7026000 0
8431200 0
9836400 0.02
11241600 0.02
12646800 0
14052000 0.07
15457200 0.07
16862400 0.09
18267600 0.08
19672800 0.08
21078000 0.09
22483200 0.17
23888400 0.14
25293600 0.17
26698800 0.2
28104000 0.26
29509200 0.18
30914400 0.23
32319600 0.29
33724800 0.27
35130000 0.25
36535200 0.27
37940400 0.33
39345600 0.31
};

\path [draw=C0, fill=C0, opacity=0.2]
(axis cs:0,0)
--(axis cs:0,0)
--(axis cs:1600000,0)
--(axis cs:3200000,0)
--(axis cs:4800000,0)
--(axis cs:6400000,0)
--(axis cs:8000000,0.02)
--(axis cs:9600000,0.11)
--(axis cs:11200000,0.33)
--(axis cs:12800000,0.48)
--(axis cs:14400000,0.54)
--(axis cs:16000000,0.55)
--(axis cs:17600000,0.58)
--(axis cs:19200000,0.7)
--(axis cs:20800000,0.65)
--(axis cs:22400000,0.73)
--(axis cs:24000000,0.75)
--(axis cs:25600000,0.71)
--(axis cs:27200000,0.76)
--(axis cs:28800000,0.72)
--(axis cs:30400000,0.74)
--(axis cs:32000000,0.77)
--(axis cs:33600000,0.79)
--(axis cs:35200000,0.8)
--(axis cs:36800000,0.78)
--(axis cs:38400000,0.77)
--(axis cs:40000000,0.81)
--(axis cs:40000000,0.94)
--(axis cs:38400000,0.85)
--(axis cs:36800000,0.94)
--(axis cs:35200000,0.97)
--(axis cs:33600000,0.9)
--(axis cs:32000000,0.89)
--(axis cs:30400000,0.88)
--(axis cs:28800000,0.86)
--(axis cs:27200000,0.9)
--(axis cs:25600000,0.830249999999996)
--(axis cs:24000000,0.85)
--(axis cs:22400000,0.87)
--(axis cs:20800000,0.83)
--(axis cs:19200000,0.82)
--(axis cs:17600000,0.770249999999996)
--(axis cs:16000000,0.71)
--(axis cs:14400000,0.71)
--(axis cs:12800000,0.59)
--(axis cs:11200000,0.5)
--(axis cs:9600000,0.2)
--(axis cs:8000000,0.1)
--(axis cs:6400000,0.03)
--(axis cs:4800000,0)
--(axis cs:3200000,0)
--(axis cs:1600000,0)
--(axis cs:0,0)
--cycle;

\addplot [thick, C0, mark=*, mark size=0, mark options={solid}]
table {%
0 0
1600000 0
3200000 0
4800000 0
6400000 0
8000000 0.06
9600000 0.16
11200000 0.42
12800000 0.54
14400000 0.62
16000000 0.62
17600000 0.68
19200000 0.78
20800000 0.74
22400000 0.8
24000000 0.8
25600000 0.77
27200000 0.84
28800000 0.79
30400000 0.82
32000000 0.84
33600000 0.88
35200000 0.89
36800000 0.86
38400000 0.81
40000000 0.86
};

\path [draw=C5, fill=C5, opacity=0.2]
(axis cs:0,0)
--(axis cs:0,0)
--(axis cs:4000000,0.06)
--(axis cs:8000000,0.29)
--(axis cs:12000000,0.75)
--(axis cs:16000000,0.71)
--(axis cs:20000000,0.74)
--(axis cs:24000000,0.78)
--(axis cs:28000000,0.83)
--(axis cs:32000000,0.73)
--(axis cs:36000000,0.65)
--(axis cs:40000000,0.71)
--(axis cs:40000000,0.85)
--(axis cs:36000000,0.86)
--(axis cs:32000000,0.88)
--(axis cs:28000000,0.99)
--(axis cs:24000000,0.94)
--(axis cs:20000000,0.95)
--(axis cs:16000000,0.93)
--(axis cs:12000000,0.92)
--(axis cs:8000000,0.64)
--(axis cs:4000000,0.26)
--(axis cs:0,0)
--cycle;

\addplot [thick, C5, mark=*, mark size=0, mark options={solid}]
table {%
0 0
4000000 0.13
8000000 0.47
12000000 0.84
16000000 0.85
20000000 0.89
24000000 0.86
28000000 0.93
32000000 0.83
36000000 0.79
40000000 0.78
};

\path [draw=C8, fill=C8, opacity=0.2]
(axis cs:0,0)
--(axis cs:0,0)
--(axis cs:50000,0)
--(axis cs:100000,0)
--(axis cs:150000,0)
--(axis cs:200000,0)
--(axis cs:250000,0)
--(axis cs:300000,0)
--(axis cs:350000,0)
--(axis cs:400000,0)
--(axis cs:450000,0)
--(axis cs:500000,0)
--(axis cs:550000,0)
--(axis cs:600000,0)
--(axis cs:650000,0)
--(axis cs:700000,0)
--(axis cs:750000,0)
--(axis cs:800000,0)
--(axis cs:850000,0)
--(axis cs:900000,0)
--(axis cs:950000,0)
--(axis cs:1000000,0)
--(axis cs:1050000,0)
--(axis cs:1100000,0)
--(axis cs:1150000,0)
--(axis cs:1200000,0)
--(axis cs:1250000,0)
--(axis cs:1300000,0)
--(axis cs:1350000,0)
--(axis cs:1400000,0)
--(axis cs:1450000,0)
--(axis cs:1500000,0)
--(axis cs:1550000,0)
--(axis cs:1600000,0)
--(axis cs:1650000,0)
--(axis cs:1700000,0)
--(axis cs:1750000,0)
--(axis cs:1800000,0)
--(axis cs:1850000,0)
--(axis cs:1900000,0)
--(axis cs:1950000,0)
--(axis cs:2000000,0)
--(axis cs:2050000,0)
--(axis cs:2100000,0)
--(axis cs:2150000,0)
--(axis cs:2200000,0)
--(axis cs:2250000,0)
--(axis cs:2300000,0)
--(axis cs:2350000,0)
--(axis cs:2400000,0)
--(axis cs:2450000,0)
--(axis cs:2500000,0)
--(axis cs:2550000,0)
--(axis cs:2600000,0)
--(axis cs:2650000,0)
--(axis cs:2700000,0)
--(axis cs:2750000,0)
--(axis cs:2800000,0)
--(axis cs:2850000,0)
--(axis cs:2900000,0)
--(axis cs:2950000,0)
--(axis cs:3000000,0)
--(axis cs:3050000,0)
--(axis cs:3100000,0)
--(axis cs:3150000,0)
--(axis cs:3200000,0.01)
--(axis cs:3250000,0)
--(axis cs:3300000,0)
--(axis cs:3350000,0)
--(axis cs:3400000,0)
--(axis cs:3450000,0)
--(axis cs:3500000,0)
--(axis cs:3550000,0)
--(axis cs:3600000,0)
--(axis cs:3650000,0)
--(axis cs:3700000,0)
--(axis cs:3750000,0)
--(axis cs:3800000,0)
--(axis cs:3850000,0.01)
--(axis cs:3900000,0)
--(axis cs:3950000,0)
--(axis cs:4000000,0)
--(axis cs:4050000,0)
--(axis cs:4100000,0)
--(axis cs:4150000,0)
--(axis cs:4200000,0)
--(axis cs:4250000,0)
--(axis cs:4300000,0)
--(axis cs:4350000,0)
--(axis cs:4400000,0)
--(axis cs:4450000,0)
--(axis cs:4500000,0)
--(axis cs:4550000,0.01)
--(axis cs:4600000,0)
--(axis cs:4650000,0)
--(axis cs:4700000,0)
--(axis cs:4750000,0)
--(axis cs:4800000,0)
--(axis cs:4850000,0)
--(axis cs:4900000,0)
--(axis cs:4950000,0)
--(axis cs:5000000,0)
--(axis cs:5050000,0)
--(axis cs:5100000,0)
--(axis cs:5150000,0)
--(axis cs:5200000,0)
--(axis cs:5250000,0)
--(axis cs:5300000,0)
--(axis cs:5350000,0)
--(axis cs:5400000,0)
--(axis cs:5450000,0)
--(axis cs:5500000,0)
--(axis cs:5550000,0)
--(axis cs:5600000,0)
--(axis cs:5650000,0)
--(axis cs:5700000,0)
--(axis cs:5750000,0)
--(axis cs:5800000,0)
--(axis cs:5850000,0.02)
--(axis cs:5900000,0)
--(axis cs:5950000,0)
--(axis cs:6000000,0)
--(axis cs:6050000,0)
--(axis cs:6100000,0)
--(axis cs:6150000,0)
--(axis cs:6200000,0)
--(axis cs:6250000,0.02)
--(axis cs:6300000,0.01)
--(axis cs:6350000,0)
--(axis cs:6400000,0)
--(axis cs:6450000,0)
--(axis cs:6500000,0)
--(axis cs:6550000,0)
--(axis cs:6600000,0)
--(axis cs:6650000,0)
--(axis cs:6700000,0)
--(axis cs:6750000,0)
--(axis cs:6800000,0)
--(axis cs:6850000,0)
--(axis cs:6900000,0)
--(axis cs:6950000,0)
--(axis cs:7000000,0)
--(axis cs:7050000,0)
--(axis cs:7100000,0)
--(axis cs:7150000,0.07)
--(axis cs:7200000,0.1)
--(axis cs:7250000,0.1)
--(axis cs:7300000,0)
--(axis cs:7350000,0)
--(axis cs:7400000,0)
--(axis cs:7450000,0)
--(axis cs:7500000,0)
--(axis cs:7550000,0.2)
--(axis cs:7600000,0)
--(axis cs:7650000,0)
--(axis cs:7700000,0)
--(axis cs:7750000,0)
--(axis cs:7800000,0)
--(axis cs:7850000,0)
--(axis cs:7900000,0)
--(axis cs:7950000,0)
--(axis cs:8000000,0)
--(axis cs:8050000,0)
--(axis cs:8100000,0)
--(axis cs:8150000,0)
--(axis cs:8200000,0)
--(axis cs:8250000,0)
--(axis cs:8300000,0)
--(axis cs:8350000,0)
--(axis cs:8400000,0)
--(axis cs:8450000,0)
--(axis cs:8500000,0)
--(axis cs:8550000,0.1)
--(axis cs:8600000,0)
--(axis cs:8650000,0.1)
--(axis cs:8700000,0)
--(axis cs:8750000,0.1)
--(axis cs:8800000,0.1)
--(axis cs:8850000,0)
--(axis cs:8900000,0)
--(axis cs:8950000,0)
--(axis cs:9000000,0)
--(axis cs:9050000,0)
--(axis cs:9100000,0)
--(axis cs:9150000,0)
--(axis cs:9200000,0.2)
--(axis cs:9250000,0)
--(axis cs:9300000,0.1)
--(axis cs:9350000,0.1)
--(axis cs:9400000,0.2)
--(axis cs:9450000,0)
--(axis cs:9500000,0)
--(axis cs:9550000,0)
--(axis cs:9600000,0)
--(axis cs:9650000,0)
--(axis cs:9700000,0)
--(axis cs:9750000,0)
--(axis cs:9800000,0.1)
--(axis cs:9850000,0)
--(axis cs:9900000,0)
--(axis cs:9950000,0)
--(axis cs:9950000,0)
--(axis cs:9950000,0)
--(axis cs:9900000,0)
--(axis cs:9850000,0)
--(axis cs:9800000,0.1)
--(axis cs:9750000,0)
--(axis cs:9700000,0)
--(axis cs:9650000,0)
--(axis cs:9600000,0)
--(axis cs:9550000,0)
--(axis cs:9500000,0)
--(axis cs:9450000,0)
--(axis cs:9400000,0.2)
--(axis cs:9350000,0.1)
--(axis cs:9300000,0.1)
--(axis cs:9250000,0)
--(axis cs:9200000,0.2)
--(axis cs:9150000,0)
--(axis cs:9100000,0)
--(axis cs:9050000,0)
--(axis cs:9000000,0)
--(axis cs:8950000,0)
--(axis cs:8900000,0)
--(axis cs:8850000,0)
--(axis cs:8800000,0.1)
--(axis cs:8750000,0.1)
--(axis cs:8700000,0)
--(axis cs:8650000,0.1)
--(axis cs:8600000,0)
--(axis cs:8550000,0.1)
--(axis cs:8500000,0)
--(axis cs:8450000,0)
--(axis cs:8400000,0)
--(axis cs:8350000,0)
--(axis cs:8300000,0)
--(axis cs:8250000,0)
--(axis cs:8200000,0)
--(axis cs:8150000,0)
--(axis cs:8100000,0)
--(axis cs:8050000,0)
--(axis cs:8000000,0)
--(axis cs:7950000,0)
--(axis cs:7900000,0)
--(axis cs:7850000,0)
--(axis cs:7800000,0)
--(axis cs:7750000,0)
--(axis cs:7700000,0)
--(axis cs:7650000,0)
--(axis cs:7600000,0)
--(axis cs:7550000,0.2)
--(axis cs:7500000,0)
--(axis cs:7450000,0)
--(axis cs:7400000,0)
--(axis cs:7350000,0)
--(axis cs:7300000,0)
--(axis cs:7250000,0.1)
--(axis cs:7200000,0.1)
--(axis cs:7150000,0.1)
--(axis cs:7100000,0.00524999999999991)
--(axis cs:7050000,0)
--(axis cs:7000000,0)
--(axis cs:6950000,0.01)
--(axis cs:6900000,0.01)
--(axis cs:6850000,0.03)
--(axis cs:6800000,0.01)
--(axis cs:6750000,0.01)
--(axis cs:6700000,0)
--(axis cs:6650000,0.01)
--(axis cs:6600000,0)
--(axis cs:6550000,0.02)
--(axis cs:6500000,0)
--(axis cs:6450000,0)
--(axis cs:6400000,0)
--(axis cs:6350000,0.0152499999999999)
--(axis cs:6300000,0.1)
--(axis cs:6250000,0.1)
--(axis cs:6200000,0)
--(axis cs:6150000,0.02)
--(axis cs:6100000,0.0152499999999999)
--(axis cs:6050000,0.01)
--(axis cs:6000000,0.04)
--(axis cs:5950000,0.01)
--(axis cs:5900000,0.03)
--(axis cs:5850000,0.1)
--(axis cs:5800000,0)
--(axis cs:5750000,0.05)
--(axis cs:5700000,0.05)
--(axis cs:5650000,0.03)
--(axis cs:5600000,0.0652499999999999)
--(axis cs:5550000,0.02)
--(axis cs:5500000,0.06)
--(axis cs:5450000,0.03)
--(axis cs:5400000,0.0704999999999998)
--(axis cs:5350000,0.08)
--(axis cs:5300000,0.0652499999999999)
--(axis cs:5250000,0.06)
--(axis cs:5200000,0.01)
--(axis cs:5150000,0.03)
--(axis cs:5100000,0.06)
--(axis cs:5050000,0.03)
--(axis cs:5000000,0.0752499999999999)
--(axis cs:4950000,0.06)
--(axis cs:4900000,0.06)
--(axis cs:4850000,0.06)
--(axis cs:4800000,0.03)
--(axis cs:4750000,0.06)
--(axis cs:4700000,0.05)
--(axis cs:4650000,0.02)
--(axis cs:4600000,0.08)
--(axis cs:4550000,0.0952499999999999)
--(axis cs:4500000,0.0352499999999999)
--(axis cs:4450000,0.09)
--(axis cs:4400000,0.10525)
--(axis cs:4350000,0.04)
--(axis cs:4300000,0.0552499999999999)
--(axis cs:4250000,0.06)
--(axis cs:4200000,0.08)
--(axis cs:4150000,0.1)
--(axis cs:4100000,0.0652499999999999)
--(axis cs:4050000,0.0352499999999999)
--(axis cs:4000000,0.09)
--(axis cs:3950000,0.0552499999999999)
--(axis cs:3900000,0.05)
--(axis cs:3850000,0.1)
--(axis cs:3800000,0.0652499999999999)
--(axis cs:3750000,0.0757499999999997)
--(axis cs:3700000,0.06)
--(axis cs:3650000,0.01)
--(axis cs:3600000,0.06)
--(axis cs:3550000,0.04)
--(axis cs:3500000,0.0704999999999998)
--(axis cs:3450000,0.08)
--(axis cs:3400000,0.04)
--(axis cs:3350000,0.05)
--(axis cs:3300000,0.0752499999999999)
--(axis cs:3250000,0.02)
--(axis cs:3200000,0.09)
--(axis cs:3150000,0.0252499999999999)
--(axis cs:3100000,0.0152499999999999)
--(axis cs:3050000,0.03)
--(axis cs:3000000,0.06)
--(axis cs:2950000,0.0352499999999999)
--(axis cs:2900000,0.01)
--(axis cs:2850000,0.06)
--(axis cs:2800000,0.06)
--(axis cs:2750000,0.03)
--(axis cs:2700000,0.04)
--(axis cs:2650000,0.0152499999999999)
--(axis cs:2600000,0)
--(axis cs:2550000,0.0604999999999998)
--(axis cs:2500000,0.05)
--(axis cs:2450000,0.02)
--(axis cs:2400000,0.01)
--(axis cs:2350000,0.0552499999999999)
--(axis cs:2300000,0.0352499999999999)
--(axis cs:2250000,0.0852499999999999)
--(axis cs:2200000,0.0452499999999999)
--(axis cs:2150000,0.02)
--(axis cs:2100000,0.0152499999999999)
--(axis cs:2050000,0)
--(axis cs:2000000,0.03)
--(axis cs:1950000,0.04)
--(axis cs:1900000,0.0252499999999999)
--(axis cs:1850000,0.0552499999999999)
--(axis cs:1800000,0.0452499999999999)
--(axis cs:1750000,0.03)
--(axis cs:1700000,0.01)
--(axis cs:1650000,0.03)
--(axis cs:1600000,0.00524999999999991)
--(axis cs:1550000,0.02)
--(axis cs:1500000,0.01)
--(axis cs:1450000,0.01)
--(axis cs:1400000,0)
--(axis cs:1350000,0)
--(axis cs:1300000,0.02)
--(axis cs:1250000,0.01)
--(axis cs:1200000,0)
--(axis cs:1150000,0)
--(axis cs:1100000,0)
--(axis cs:1050000,0)
--(axis cs:1000000,0)
--(axis cs:950000,0)
--(axis cs:900000,0)
--(axis cs:850000,0.00524999999999991)
--(axis cs:800000,0)
--(axis cs:750000,0)
--(axis cs:700000,0)
--(axis cs:650000,0)
--(axis cs:600000,0)
--(axis cs:550000,0)
--(axis cs:500000,0)
--(axis cs:450000,0)
--(axis cs:400000,0)
--(axis cs:350000,0)
--(axis cs:300000,0)
--(axis cs:250000,0)
--(axis cs:200000,0)
--(axis cs:150000,0)
--(axis cs:100000,0)
--(axis cs:50000,0)
--(axis cs:0,0)
--cycle;

\addplot [thick, C8, mark=*, mark size=0, mark options={solid}]
table {%
0 0
50000 0
100000 0
150000 0
200000 0
250000 0
300000 0
350000 0
400000 0
450000 0
500000 0
550000 0
600000 0
650000 0
700000 0
750000 0
800000 0
850000 0
900000 0
950000 0
1000000 0
1050000 0
1100000 0
1150000 0
1200000 0
1250000 0
1300000 0
1350000 0
1400000 0
1450000 0
1500000 0
1550000 0
1600000 0
1650000 0
1700000 0
1750000 0
1800000 0.01
1850000 0.01
1900000 0
1950000 0.01
2000000 0
2050000 0
2100000 0
2150000 0
2200000 0.01
2250000 0.04
2300000 0
2350000 0.02
2400000 0
2450000 0
2500000 0
2550000 0.02
2600000 0
2650000 0
2700000 0
2750000 0
2800000 0.02
2850000 0.01
2900000 0
2950000 0
3000000 0.02
3050000 0
3100000 0
3150000 0
3200000 0.04
3250000 0
3300000 0.02
3350000 0
3400000 0
3450000 0.03
3500000 0.02
3550000 0
3600000 0.02
3650000 0
3700000 0.02
3750000 0.02
3800000 0.02
3850000 0.05
3900000 0.02
3950000 0.02
4000000 0.03
4050000 0
4100000 0.02
4150000 0.03
4200000 0.03
4250000 0.01
4300000 0.01
4350000 0
4400000 0.04
4450000 0.02
4500000 0
4550000 0.05
4600000 0.02
4650000 0
4700000 0
4750000 0.02
4800000 0
4850000 0.02
4900000 0.01
4950000 0.02
5000000 0.02
5050000 0
5100000 0.02
5150000 0
5200000 0
5250000 0.02
5300000 0
5350000 0.03
5400000 0.01
5450000 0
5500000 0.03
5550000 0
5600000 0.01
5650000 0
5700000 0
5750000 0.01
5800000 0
5850000 0.06
5900000 0
5950000 0
6000000 0
6050000 0
6100000 0
6150000 0
6200000 0
6250000 0.06
6300000 0.06
6350000 0
6400000 0
6450000 0
6500000 0
6550000 0
6600000 0
6650000 0
6700000 0
6750000 0
6800000 0
6850000 0
6900000 0
6950000 0
7000000 0
7050000 0
7100000 0
7150000 0.1
7200000 0.1
7250000 0.1
7300000 0
7350000 0
7400000 0
7450000 0
7500000 0
7550000 0.2
7600000 0
7650000 0
7700000 0
7750000 0
7800000 0
7850000 0
7900000 0
7950000 0
8000000 0
8050000 0
8100000 0
8150000 0
8200000 0
8250000 0
8300000 0
8350000 0
8400000 0
8450000 0
8500000 0
8550000 0.1
8600000 0
8650000 0.1
8700000 0
8750000 0.1
8800000 0.1
8850000 0
8900000 0
8950000 0
9000000 0
9050000 0
9100000 0
9150000 0
9200000 0.2
9250000 0
9300000 0.1
9350000 0.1
9400000 0.2
9450000 0
9500000 0
9550000 0
9600000 0
9650000 0
9700000 0
9750000 0
9800000 0.1
9850000 0
9900000 0
9950000 0
};

\path [draw=C6, fill=C6, opacity=0.2]
(axis cs:52000,0)
--(axis cs:52000,0)
--(axis cs:1352000,0)
--(axis cs:2652000,0)
--(axis cs:3952000,0)
--(axis cs:5252000,0)
--(axis cs:6552000,0)
--(axis cs:7852000,0)
--(axis cs:9152000,0)
--(axis cs:10452000,0)
--(axis cs:11752000,0)
--(axis cs:13052000,0)
--(axis cs:14352000,0)
--(axis cs:15652000,0)
--(axis cs:16952000,0.00454545454545455)
--(axis cs:18252000,0)
--(axis cs:19552000,0)
--(axis cs:20852000,0)
--(axis cs:22152000,0)
--(axis cs:23452000,0)
--(axis cs:24752000,0)
--(axis cs:26052000,0)
--(axis cs:27352000,0.00454545454545455)
--(axis cs:28652000,0)
--(axis cs:29952000,0)
--(axis cs:31252000,0.0136363636363636)
--(axis cs:32552000,0)
--(axis cs:33852000,0)
--(axis cs:35152000,0.00454545454545455)
--(axis cs:36452000,0)
--(axis cs:37752000,0.00909090909090909)
--(axis cs:39052000,0.00909090909090909)
--(axis cs:39052000,0.0454545454545455)
--(axis cs:37752000,0.05)
--(axis cs:36452000,0.0363636363636364)
--(axis cs:35152000,0.0409090909090909)
--(axis cs:33852000,0.0318181818181818)
--(axis cs:32552000,0.0454545454545455)
--(axis cs:31252000,0.0545454545454545)
--(axis cs:29952000,0.0227272727272727)
--(axis cs:28652000,0.0454545454545455)
--(axis cs:27352000,0.0727272727272727)
--(axis cs:26052000,0.0272727272727273)
--(axis cs:24752000,0.05)
--(axis cs:23452000,0.0363636363636364)
--(axis cs:22152000,0.0363636363636364)
--(axis cs:20852000,0.0227272727272727)
--(axis cs:19552000,0.0454545454545455)
--(axis cs:18252000,0.0454545454545455)
--(axis cs:16952000,0.0545454545454546)
--(axis cs:15652000,0.0409090909090909)
--(axis cs:14352000,0.0409090909090909)
--(axis cs:13052000,0.0454545454545455)
--(axis cs:11752000,0.0318181818181818)
--(axis cs:10452000,0.0227272727272727)
--(axis cs:9152000,0.0227272727272727)
--(axis cs:7852000,0.00454545454545455)
--(axis cs:6552000,0.0227272727272727)
--(axis cs:5252000,0.0181818181818182)
--(axis cs:3952000,0.0227272727272727)
--(axis cs:2652000,0)
--(axis cs:1352000,0)
--(axis cs:52000,0)
--cycle;

\addplot [thick, solid, C6, mark=*, mark size=0, mark options={solid}]
table {%
52000 0
1352000 0
2652000 0
3952000 0.00454545454545455
5252000 0
6552000 0.00454545454545455
7852000 0
9152000 0.00454545454545455
10452000 0.00454545454545455
11752000 0.00454545454545455
13052000 0.0136363636363636
14352000 0.0181818181818182
15652000 0.0136363636363636
16952000 0.0227272727272727
18252000 0.0181818181818182
19552000 0.00909090909090909
20852000 0.00454545454545455
22152000 0.0136363636363636
23452000 0.0136363636363636
24752000 0.0181818181818182
26052000 0.00909090909090909
27352000 0.0363636363636364
28652000 0.0136363636363636
29952000 0.00454545454545455
31252000 0.0318181818181818
32552000 0.0181818181818182
33852000 0.00909090909090909
35152000 0.0227272727272727
36452000 0.0136363636363636
37752000 0.0272727272727273
39052000 0.0272727272727273
};
\end{axis}

\end{tikzpicture}

%% file: tikz_plots/box_pushing/box_pushing_sparse_iqm_sample_efficiency.tex
\begin{tikzpicture}

\begin{axis}[
legend cell align={left},
legend style={fill opacity=0.8, draw opacity=1, text opacity=1, draw=lightgray204, at={(0.03,0.03)},  anchor=north west},
title={Box Pushing - Sparse},
tick align=outside,
tick pos=left,
x grid style={darkgray176},
xlabel={Number Environment Interactions},
xmajorgrids,
xmin=-2500000, xmax=42000000.05,
xtick style={color=black},
y grid style={darkgray176},
ylabel={Success Rate},
ymajorgrids,
ymin=-0.05, ymax=1.05,
ytick style={color=black}
]
\path [draw=C0, fill=C0, opacity=0.2]
(axis cs:0,0)
--(axis cs:0,0)
--(axis cs:2000000,0)
--(axis cs:4000000,0)
--(axis cs:6000000,0)
--(axis cs:8000000,0)
--(axis cs:10000000,0.02)
--(axis cs:12000000,0.11)
--(axis cs:14000000,0.29)
--(axis cs:16000000,0.48)
--(axis cs:18000000,0.65)
--(axis cs:20000000,0.71)
--(axis cs:22000000,0.71)
--(axis cs:24000000,0.82)
--(axis cs:26000000,0.82)
--(axis cs:28000000,0.86)
--(axis cs:30000000,0.82)
--(axis cs:32000000,0.76)
--(axis cs:34000000,0.83)
--(axis cs:36000000,0.85)
--(axis cs:38000000,0.88)
--(axis cs:40000000,0.85)
--(axis cs:40000000,0.93)
--(axis cs:38000000,0.97)
--(axis cs:36000000,0.96)
--(axis cs:34000000,0.95)
--(axis cs:32000000,0.91)
--(axis cs:30000000,0.96)
--(axis cs:28000000,0.94)
--(axis cs:26000000,0.93)
--(axis cs:24000000,0.9)
--(axis cs:22000000,0.88)
--(axis cs:20000000,0.85)
--(axis cs:18000000,0.76)
--(axis cs:16000000,0.64)
--(axis cs:14000000,0.44)
--(axis cs:12000000,0.28)
--(axis cs:10000000,0.1)
--(axis cs:8000000,0)
--(axis cs:6000000,0)
--(axis cs:4000000,0)
--(axis cs:2000000,0)
--(axis cs:0,0)
--cycle;

\addplot [thick, solid, C0, mark=*, mark size=0, mark options={solid}]
table {%
0 0
2000000 0
4000000 0
6000000 0
8000000 0
10000000 0.06
12000000 0.2
14000000 0.37
16000000 0.57
18000000 0.7
20000000 0.78
22000000 0.82
24000000 0.87
26000000 0.87
28000000 0.9
30000000 0.89
32000000 0.84
34000000 0.89
36000000 0.91
38000000 0.93
40000000 0.89
};

\path [draw=C1, fill=C1, opacity=0.2]
(axis cs:0,0)
--(axis cs:0,0)
--(axis cs:1600000,0)
--(axis cs:3200000,0)
--(axis cs:4800000,0)
--(axis cs:6400000,0)
--(axis cs:8000000,0)
--(axis cs:9600000,0)
--(axis cs:11200000,0)
--(axis cs:12800000,0)
--(axis cs:14400000,0)
--(axis cs:16000000,0)
--(axis cs:17600000,0)
--(axis cs:19200000,0)
--(axis cs:20800000,0)
--(axis cs:22400000,0)
--(axis cs:24000000,0)
--(axis cs:25600000,0)
--(axis cs:27200000,0)
--(axis cs:28800000,0)
--(axis cs:30400000,0)
--(axis cs:32000000,0)
--(axis cs:32000000,0)
--(axis cs:32000000,0)
--(axis cs:30400000,0)
--(axis cs:28800000,0)
--(axis cs:27200000,0)
--(axis cs:25600000,0)
--(axis cs:24000000,0)
--(axis cs:22400000,0)
--(axis cs:20800000,0)
--(axis cs:19200000,0)
--(axis cs:17600000,0)
--(axis cs:16000000,0)
--(axis cs:14400000,0)
--(axis cs:12800000,0)
--(axis cs:11200000,0)
--(axis cs:9600000,0)
--(axis cs:8000000,0)
--(axis cs:6400000,0)
--(axis cs:4800000,0)
--(axis cs:3200000,0)
--(axis cs:1600000,0)
--(axis cs:0,0)
--cycle;

\addplot [thick, solid, C1, mark=*, mark size=0, mark options={solid}]
table {%
0 0
1600000 0
3200000 0
4800000 0
6400000 0
8000000 0
9600000 0
11200000 0
12800000 0
14400000 0
16000000 0
17600000 0
19200000 0
20800000 0
22400000 0
24000000 0
25600000 0
27200000 0
28800000 0
30400000 0
32000000 0
};

\path [draw=C0, fill=C0, opacity=0.2]
(axis cs:0,0)
--(axis cs:0,0)
--(axis cs:1600000,0)
--(axis cs:3200000,0)
--(axis cs:4800000,0)
--(axis cs:6400000,0)
--(axis cs:8000000,0)
--(axis cs:9600000,0)
--(axis cs:11200000,0)
--(axis cs:12800000,0)
--(axis cs:14400000,0)
--(axis cs:16000000,0.01)
--(axis cs:17600000,0.05)
--(axis cs:19200000,0.08)
--(axis cs:20800000,0.15)
--(axis cs:22400000,0.16)
--(axis cs:24000000,0.25)
--(axis cs:25600000,0.3)
--(axis cs:27200000,0.26)
--(axis cs:28800000,0.37)
--(axis cs:30400000,0.46)
--(axis cs:32000000,0.44)
--(axis cs:33600000,0.41)
--(axis cs:35200000,0.46)
--(axis cs:36800000,0.4)
--(axis cs:38400000,0.51)
--(axis cs:40000000,0.52)
--(axis cs:40000000,0.68)
--(axis cs:38400000,0.63)
--(axis cs:36800000,0.56)
--(axis cs:35200000,0.69)
--(axis cs:33600000,0.54)
--(axis cs:32000000,0.57)
--(axis cs:30400000,0.59)
--(axis cs:28800000,0.55)
--(axis cs:27200000,0.45)
--(axis cs:25600000,0.46)
--(axis cs:24000000,0.45)
--(axis cs:22400000,0.33)
--(axis cs:20800000,0.3)
--(axis cs:19200000,0.21)
--(axis cs:17600000,0.18)
--(axis cs:16000000,0.12)
--(axis cs:14400000,0.04)
--(axis cs:12800000,0.05)
--(axis cs:11200000,0)
--(axis cs:9600000,0)
--(axis cs:8000000,0)
--(axis cs:6400000,0)
--(axis cs:4800000,0)
--(axis cs:3200000,0)
--(axis cs:1600000,0)
--(axis cs:0,0)
--cycle;

\addplot [thick, dashed, C0, mark=*, mark size=0, mark options={solid}]
table {%
0 0
1600000 0
3200000 0
4800000 0
6400000 0
8000000 0
9600000 0
11200000 0
12800000 0.01
14400000 0
16000000 0.05
17600000 0.12
19200000 0.16
20800000 0.22
22400000 0.24
24000000 0.35
25600000 0.38
27200000 0.36
28800000 0.46
30400000 0.53
32000000 0.51
33600000 0.47
35200000 0.58
36800000 0.48
38400000 0.56
40000000 0.6
};

\path [draw=C1, fill=C1, opacity=0.2]
(axis cs:0,0)
--(axis cs:0,0)
--(axis cs:1600000,0)
--(axis cs:3200000,0)
--(axis cs:4800000,0)
--(axis cs:6400000,0)
--(axis cs:8000000,0)
--(axis cs:9600000,0)
--(axis cs:11200000,0)
--(axis cs:12800000,0)
--(axis cs:14400000,0)
--(axis cs:16000000,0)
--(axis cs:17600000,0)
--(axis cs:19200000,0)
--(axis cs:20800000,0)
--(axis cs:22400000,0)
--(axis cs:24000000,0)
--(axis cs:25600000,0)
--(axis cs:27200000,0)
--(axis cs:28800000,0)
--(axis cs:30400000,0)
--(axis cs:32000000,0)
--(axis cs:33600000,0)
--(axis cs:35200000,0)
--(axis cs:36800000,0)
--(axis cs:38400000,0)
--(axis cs:40000000,0)
--(axis cs:40000000,0)
--(axis cs:38400000,0)
--(axis cs:36800000,0)
--(axis cs:35200000,0)
--(axis cs:33600000,0)
--(axis cs:32000000,0)
--(axis cs:30400000,0)
--(axis cs:28800000,0)
--(axis cs:27200000,0)
--(axis cs:25600000,0)
--(axis cs:24000000,0)
--(axis cs:22400000,0)
--(axis cs:20800000,0)
--(axis cs:19200000,0)
--(axis cs:17600000,0)
--(axis cs:16000000,0)
--(axis cs:14400000,0)
--(axis cs:12800000,0)
--(axis cs:11200000,0)
--(axis cs:9600000,0)
--(axis cs:8000000,0)
--(axis cs:6400000,0)
--(axis cs:4800000,0)
--(axis cs:3200000,0)
--(axis cs:1600000,0)
--(axis cs:0,0)
--cycle;

\addplot [thick, dashed, C1, mark=*, mark size=0, mark options={solid}]
table {%
0 0
1600000 0
3200000 0
4800000 0
6400000 0
8000000 0
9600000 0
11200000 0
12800000 0
14400000 0
16000000 0
17600000 0
19200000 0
20800000 0
22400000 0
24000000 0
25600000 0
27200000 0
28800000 0
30400000 0
32000000 0
33600000 0
35200000 0
36800000 0
38400000 0
40000000 0
};

\path [draw=C4, fill=C4, opacity=0.2]
(axis cs:0,0)
--(axis cs:0,0)
--(axis cs:1526400,0)
--(axis cs:3052800,0)
--(axis cs:4579200,0)
--(axis cs:6105600,0)
--(axis cs:7632000,0.03)
--(axis cs:9158400,0.18)
--(axis cs:10684800,0.15)
--(axis cs:12211200,0.25)
--(axis cs:13737600,0.27)
--(axis cs:15264000,0.31)
--(axis cs:16790400,0.29)
--(axis cs:18316800,0.45)
--(axis cs:19843200,0.41)
--(axis cs:21369600,0.4)
--(axis cs:22896000,0.45)
--(axis cs:24422400,0.51)
--(axis cs:25948800,0.5)
--(axis cs:27475200,0.57)
--(axis cs:29001600,0.57)
--(axis cs:30528000,0.56)
--(axis cs:32054400,0.62)
--(axis cs:33580800,0.6)
--(axis cs:35107200,0.58)
--(axis cs:36633600,0.55)
--(axis cs:38160000,0.57)
--(axis cs:39686400,0.58)
--(axis cs:39686400,0.73)
--(axis cs:38160000,0.82)
--(axis cs:36633600,0.71)
--(axis cs:35107200,0.74)
--(axis cs:33580800,0.77)
--(axis cs:32054400,0.73)
--(axis cs:30528000,0.75)
--(axis cs:29001600,0.68)
--(axis cs:27475200,0.7)
--(axis cs:25948800,0.65)
--(axis cs:24422400,0.67)
--(axis cs:22896000,0.64)
--(axis cs:21369600,0.55)
--(axis cs:19843200,0.58)
--(axis cs:18316800,0.61)
--(axis cs:16790400,0.46)
--(axis cs:15264000,0.47)
--(axis cs:13737600,0.45)
--(axis cs:12211200,0.39)
--(axis cs:10684800,0.24)
--(axis cs:9158400,0.28)
--(axis cs:7632000,0.15)
--(axis cs:6105600,0.09)
--(axis cs:4579200,0.09)
--(axis cs:3052800,0)
--(axis cs:1526400,0)
--(axis cs:0,0)
--cycle;

\addplot [thick, solid, C4, mark=*, mark size=0, mark options={solid}]
table {%
0 0
1526400 0
3052800 0
4579200 0.04
6105600 0.03
7632000 0.08
9158400 0.23
10684800 0.19
12211200 0.32
13737600 0.37
15264000 0.38
16790400 0.39
18316800 0.52
19843200 0.52
21369600 0.47
22896000 0.54
24422400 0.6
25948800 0.57
27475200 0.64
29001600 0.63
30528000 0.65
32054400 0.67
33580800 0.69
35107200 0.65
36633600 0.64
38160000 0.71
39686400 0.65
};

\path [draw=C4, fill=C4, opacity=0.2]
(axis cs:0,0)
--(axis cs:0,0)
--(axis cs:1600000,0)
--(axis cs:3200000,0)
--(axis cs:4800000,0)
--(axis cs:6400000,0)
--(axis cs:8000000,0)
--(axis cs:9600000,0)
--(axis cs:11200000,0)
--(axis cs:12800000,0)
--(axis cs:14400000,0)
--(axis cs:16000000,0)
--(axis cs:17600000,0)
--(axis cs:19200000,0)
--(axis cs:20800000,0)
--(axis cs:22400000,0)
--(axis cs:24000000,0)
--(axis cs:25600000,0)
--(axis cs:27200000,0)
--(axis cs:28800000,0)
--(axis cs:30400000,0)
--(axis cs:32000000,0)
--(axis cs:33600000,0)
--(axis cs:35200000,0)
--(axis cs:36800000,0)
--(axis cs:38400000,0)
--(axis cs:40000000,0)
--(axis cs:40000000,0.333333333333333)
--(axis cs:38400000,0.4)
--(axis cs:36800000,0.3)
--(axis cs:35200000,0.35)
--(axis cs:33600000,0.316666666666667)
--(axis cs:32000000,0.383333333333333)
--(axis cs:30400000,0.316666666666667)
--(axis cs:28800000,0.216666666666667)
--(axis cs:27200000,0.333333333333333)
--(axis cs:25600000,0.266666666666667)
--(axis cs:24000000,0.233333333333333)
--(axis cs:22400000,0.233333333333333)
--(axis cs:20800000,0.2)
--(axis cs:19200000,0.0833333333333333)
--(axis cs:17600000,0.116666666666667)
--(axis cs:16000000,0.0833333333333333)
--(axis cs:14400000,0.1)
--(axis cs:12800000,0.133333333333333)
--(axis cs:11200000,0.05)
--(axis cs:9600000,0.0333333333333333)
--(axis cs:8000000,0)
--(axis cs:6400000,0)
--(axis cs:4800000,0)
--(axis cs:3200000,0)
--(axis cs:1600000,0)
--(axis cs:0,0)
--cycle;

\addplot [thick, dashed, C4, mark=*, mark size=0, mark options={solid}]
table {%
0 0
1600000 0
3200000 0
4800000 0
6400000 0
8000000 0
9600000 0
11200000 0
12800000 0.0333333333333333
14400000 0.0166666666666667
16000000 0.0166666666666667
17600000 0.0166666666666667
19200000 0.0166666666666667
20800000 0.0333333333333333
22400000 0.0333333333333333
24000000 0.0333333333333333
25600000 0.0333333333333333
27200000 0.0666666666666667
28800000 0.0333333333333333
30400000 0.0333333333333333
32000000 0.0666666666666667
33600000 0.0666666666666667
35200000 0.0333333333333333
36800000 0.05
38400000 0.0833333333333333
40000000 0.05
};

\path [draw=C5, fill=C5, opacity=0.2]
(axis cs:0,0)
--(axis cs:0,0)
--(axis cs:4000000,0)
--(axis cs:8000000,0)
--(axis cs:12000000,0)
--(axis cs:16000000,0)
--(axis cs:20000000,0)
--(axis cs:24000000,0)
--(axis cs:28000000,0)
--(axis cs:32000000,0)
--(axis cs:36000000,0)
--(axis cs:40000000,0)
--(axis cs:40000000,0)
--(axis cs:36000000,0)
--(axis cs:32000000,0)
--(axis cs:28000000,0)
--(axis cs:24000000,0)
--(axis cs:20000000,0)
--(axis cs:16000000,0)
--(axis cs:12000000,0)
--(axis cs:8000000,0)
--(axis cs:4000000,0)
--(axis cs:0,0)
--cycle;

\addplot [ultra thick, solid, C5, mark=*, mark size=0, mark options={solid}]
table {%
0 0
4000000 0
8000000 0
12000000 0
16000000 0
20000000 0
24000000 0
28000000 0
32000000 0
36000000 0
40000000 0
};

\path [draw=C5, fill=C5, opacity=0.2]
(axis cs:0,0)
--(axis cs:0,0)
--(axis cs:4444444.44444444,0)
--(axis cs:8888888.88888889,0)
--(axis cs:13333333.3333333,0)
--(axis cs:17777777.7777778,0)
--(axis cs:22222222.2222222,0)
--(axis cs:26666666.6666667,0)
--(axis cs:31111111.1111111,0)
--(axis cs:35555555.5555556,0)
--(axis cs:40000000,0)
--(axis cs:40000000,0)
--(axis cs:35555555.5555556,0)
--(axis cs:31111111.1111111,0)
--(axis cs:26666666.6666667,0)
--(axis cs:22222222.2222222,0)
--(axis cs:17777777.7777778,0)
--(axis cs:13333333.3333333,0)
--(axis cs:8888888.88888889,0)
--(axis cs:4444444.44444444,0)
--(axis cs:0,0)
--cycle;

\addplot [thick, dashed, C5, mark=*, mark size=0, mark options={solid}]
table {%
0 0
4444444.44444444 0
8888888.88888889 0
13333333.3333333 0
17777777.7777778 0
22222222.2222222 0
26666666.6666667 0
31111111.1111111 0
35555555.5555556 0
40000000 0
};

\path [draw=C8, fill=C8, opacity=0.2]
(axis cs:0,0)
--(axis cs:0,0)
--(axis cs:50000,0)
--(axis cs:100000,0)
--(axis cs:150000,0)
--(axis cs:200000,0)
--(axis cs:250000,0)
--(axis cs:300000,0)
--(axis cs:350000,0)
--(axis cs:400000,0)
--(axis cs:450000,0)
--(axis cs:500000,0)
--(axis cs:550000,0)
--(axis cs:600000,0)
--(axis cs:650000,0)
--(axis cs:700000,0)
--(axis cs:750000,0)
--(axis cs:800000,0)
--(axis cs:850000,0)
--(axis cs:900000,0)
--(axis cs:950000,0)
--(axis cs:1000000,0)
--(axis cs:1050000,0)
--(axis cs:1100000,0)
--(axis cs:1150000,0)
--(axis cs:1200000,0)
--(axis cs:1250000,0)
--(axis cs:1300000,0)
--(axis cs:1350000,0)
--(axis cs:1400000,0)
--(axis cs:1450000,0)
--(axis cs:1500000,0)
--(axis cs:1550000,0)
--(axis cs:1600000,0)
--(axis cs:1650000,0)
--(axis cs:1700000,0)
--(axis cs:1750000,0)
--(axis cs:1800000,0)
--(axis cs:1850000,0)
--(axis cs:1900000,0)
--(axis cs:1950000,0)
--(axis cs:2000000,0)
--(axis cs:2050000,0)
--(axis cs:2100000,0)
--(axis cs:2150000,0)
--(axis cs:2200000,0)
--(axis cs:2250000,0)
--(axis cs:2300000,0)
--(axis cs:2350000,0)
--(axis cs:2400000,0)
--(axis cs:2450000,0)
--(axis cs:2500000,0)
--(axis cs:2550000,0)
--(axis cs:2600000,0)
--(axis cs:2650000,0)
--(axis cs:2700000,0)
--(axis cs:2750000,0)
--(axis cs:2800000,0)
--(axis cs:2850000,0)
--(axis cs:2900000,0)
--(axis cs:2950000,0)
--(axis cs:3000000,0)
--(axis cs:3050000,0)
--(axis cs:3100000,0)
--(axis cs:3150000,0)
--(axis cs:3200000,0)
--(axis cs:3250000,0)
--(axis cs:3300000,0)
--(axis cs:3350000,0)
--(axis cs:3400000,0)
--(axis cs:3450000,0)
--(axis cs:3500000,0)
--(axis cs:3550000,0)
--(axis cs:3600000,0)
--(axis cs:3650000,0)
--(axis cs:3700000,0)
--(axis cs:3750000,0)
--(axis cs:3800000,0)
--(axis cs:3850000,0)
--(axis cs:3900000,0)
--(axis cs:3950000,0)
--(axis cs:4000000,0)
--(axis cs:4050000,0)
--(axis cs:4100000,0)
--(axis cs:4150000,0)
--(axis cs:4200000,0)
--(axis cs:4250000,0)
--(axis cs:4300000,0)
--(axis cs:4350000,0)
--(axis cs:4400000,0)
--(axis cs:4450000,0)
--(axis cs:4500000,0)
--(axis cs:4550000,0)
--(axis cs:4600000,0)
--(axis cs:4650000,0)
--(axis cs:4700000,0)
--(axis cs:4750000,0)
--(axis cs:4800000,0)
--(axis cs:4850000,0)
--(axis cs:4900000,0)
--(axis cs:4950000,0)
--(axis cs:5000000,0)
--(axis cs:5050000,0)
--(axis cs:5100000,0)
--(axis cs:5150000,0)
--(axis cs:5200000,0)
--(axis cs:5250000,0)
--(axis cs:5300000,0)
--(axis cs:5350000,0)
--(axis cs:5400000,0)
--(axis cs:5450000,0)
--(axis cs:5500000,0)
--(axis cs:5550000,0)
--(axis cs:5600000,0)
--(axis cs:5650000,0)
--(axis cs:5700000,0)
--(axis cs:5750000,0)
--(axis cs:5800000,0)
--(axis cs:5850000,0)
--(axis cs:5900000,0)
--(axis cs:5950000,0)
--(axis cs:6000000,0)
--(axis cs:6050000,0)
--(axis cs:6100000,0)
--(axis cs:6150000,0)
--(axis cs:6200000,0)
--(axis cs:6250000,0)
--(axis cs:6300000,0)
--(axis cs:6350000,0)
--(axis cs:6400000,0)
--(axis cs:6450000,0)
--(axis cs:6500000,0)
--(axis cs:6550000,0)
--(axis cs:6600000,0)
--(axis cs:6650000,0)
--(axis cs:6700000,0)
--(axis cs:6750000,0)
--(axis cs:6800000,0)
--(axis cs:6850000,0)
--(axis cs:6900000,0)
--(axis cs:6950000,0)
--(axis cs:7000000,0)
--(axis cs:7050000,0)
--(axis cs:7100000,0)
--(axis cs:7150000,0)
--(axis cs:7200000,0)
--(axis cs:7250000,0)
--(axis cs:7300000,0)
--(axis cs:7350000,0)
--(axis cs:7400000,0)
--(axis cs:7450000,0)
--(axis cs:7500000,0)
--(axis cs:7550000,0)
--(axis cs:7600000,0)
--(axis cs:7650000,0)
--(axis cs:7700000,0)
--(axis cs:7750000,0)
--(axis cs:7800000,0)
--(axis cs:7850000,0)
--(axis cs:7900000,0)
--(axis cs:7950000,0)
--(axis cs:8000000,0)
--(axis cs:8050000,0)
--(axis cs:8100000,0)
--(axis cs:8150000,0)
--(axis cs:8200000,0)
--(axis cs:8250000,0)
--(axis cs:8300000,0)
--(axis cs:8350000,0)
--(axis cs:8400000,0)
--(axis cs:8450000,0)
--(axis cs:8500000,0)
--(axis cs:8550000,0)
--(axis cs:8600000,0)
--(axis cs:8650000,0)
--(axis cs:8700000,0)
--(axis cs:8750000,0)
--(axis cs:8800000,0)
--(axis cs:8850000,0)
--(axis cs:8900000,0)
--(axis cs:8950000,0)
--(axis cs:9000000,0)
--(axis cs:9050000,0)
--(axis cs:9100000,0)
--(axis cs:9150000,0)
--(axis cs:9200000,0)
--(axis cs:9250000,0)
--(axis cs:9300000,0)
--(axis cs:9350000,0)
--(axis cs:9400000,0)
--(axis cs:9450000,0)
--(axis cs:9500000,0)
--(axis cs:9550000,0)
--(axis cs:9600000,0)
--(axis cs:9650000,0)
--(axis cs:9700000,0)
--(axis cs:9750000,0)
--(axis cs:9800000,0)
--(axis cs:9850000,0)
--(axis cs:9900000,0)
--(axis cs:9950000,0)
--(axis cs:10000000,0)
--(axis cs:10050000,0)
--(axis cs:10100000,0)
--(axis cs:10150000,0)
--(axis cs:10200000,0)
--(axis cs:10250000,0)
--(axis cs:10300000,0)
--(axis cs:10350000,0)
--(axis cs:10350000,0)
--(axis cs:10350000,0)
--(axis cs:10300000,0)
--(axis cs:10250000,0)
--(axis cs:10200000,0)
--(axis cs:10150000,0)
--(axis cs:10100000,0)
--(axis cs:10050000,0)
--(axis cs:10000000,0)
--(axis cs:9950000,0)
--(axis cs:9900000,0)
--(axis cs:9850000,0)
--(axis cs:9800000,0)
--(axis cs:9750000,0)
--(axis cs:9700000,0)
--(axis cs:9650000,0)
--(axis cs:9600000,0)
--(axis cs:9550000,0)
--(axis cs:9500000,0)
--(axis cs:9450000,0)
--(axis cs:9400000,0)
--(axis cs:9350000,0)
--(axis cs:9300000,0)
--(axis cs:9250000,0)
--(axis cs:9200000,0)
--(axis cs:9150000,0)
--(axis cs:9100000,0)
--(axis cs:9050000,0)
--(axis cs:9000000,0)
--(axis cs:8950000,0)
--(axis cs:8900000,0)
--(axis cs:8850000,0)
--(axis cs:8800000,0)
--(axis cs:8750000,0)
--(axis cs:8700000,0)
--(axis cs:8650000,0)
--(axis cs:8600000,0)
--(axis cs:8550000,0)
--(axis cs:8500000,0)
--(axis cs:8450000,0)
--(axis cs:8400000,0)
--(axis cs:8350000,0)
--(axis cs:8300000,0)
--(axis cs:8250000,0)
--(axis cs:8200000,0)
--(axis cs:8150000,0)
--(axis cs:8100000,0)
--(axis cs:8050000,0)
--(axis cs:8000000,0)
--(axis cs:7950000,0)
--(axis cs:7900000,0)
--(axis cs:7850000,0)
--(axis cs:7800000,0)
--(axis cs:7750000,0)
--(axis cs:7700000,0)
--(axis cs:7650000,0)
--(axis cs:7600000,0)
--(axis cs:7550000,0)
--(axis cs:7500000,0)
--(axis cs:7450000,0)
--(axis cs:7400000,0)
--(axis cs:7350000,0)
--(axis cs:7300000,0)
--(axis cs:7250000,0)
--(axis cs:7200000,0)
--(axis cs:7150000,0)
--(axis cs:7100000,0)
--(axis cs:7050000,0)
--(axis cs:7000000,0)
--(axis cs:6950000,0)
--(axis cs:6900000,0)
--(axis cs:6850000,0)
--(axis cs:6800000,0)
--(axis cs:6750000,0)
--(axis cs:6700000,0)
--(axis cs:6650000,0)
--(axis cs:6600000,0)
--(axis cs:6550000,0)
--(axis cs:6500000,0)
--(axis cs:6450000,0)
--(axis cs:6400000,0)
--(axis cs:6350000,0)
--(axis cs:6300000,0)
--(axis cs:6250000,0)
--(axis cs:6200000,0)
--(axis cs:6150000,0)
--(axis cs:6100000,0)
--(axis cs:6050000,0)
--(axis cs:6000000,0)
--(axis cs:5950000,0)
--(axis cs:5900000,0)
--(axis cs:5850000,0)
--(axis cs:5800000,0)
--(axis cs:5750000,0)
--(axis cs:5700000,0)
--(axis cs:5650000,0)
--(axis cs:5600000,0)
--(axis cs:5550000,0)
--(axis cs:5500000,0)
--(axis cs:5450000,0)
--(axis cs:5400000,0)
--(axis cs:5350000,0)
--(axis cs:5300000,0)
--(axis cs:5250000,0)
--(axis cs:5200000,0)
--(axis cs:5150000,0)
--(axis cs:5100000,0)
--(axis cs:5050000,0)
--(axis cs:5000000,0)
--(axis cs:4950000,0)
--(axis cs:4900000,0)
--(axis cs:4850000,0)
--(axis cs:4800000,0)
--(axis cs:4750000,0)
--(axis cs:4700000,0)
--(axis cs:4650000,0)
--(axis cs:4600000,0)
--(axis cs:4550000,0)
--(axis cs:4500000,0)
--(axis cs:4450000,0)
--(axis cs:4400000,0)
--(axis cs:4350000,0)
--(axis cs:4300000,0)
--(axis cs:4250000,0)
--(axis cs:4200000,0)
--(axis cs:4150000,0)
--(axis cs:4100000,0)
--(axis cs:4050000,0)
--(axis cs:4000000,0)
--(axis cs:3950000,0)
--(axis cs:3900000,0)
--(axis cs:3850000,0)
--(axis cs:3800000,0)
--(axis cs:3750000,0)
--(axis cs:3700000,0)
--(axis cs:3650000,0)
--(axis cs:3600000,0)
--(axis cs:3550000,0)
--(axis cs:3500000,0)
--(axis cs:3450000,0)
--(axis cs:3400000,0)
--(axis cs:3350000,0)
--(axis cs:3300000,0)
--(axis cs:3250000,0)
--(axis cs:3200000,0)
--(axis cs:3150000,0)
--(axis cs:3100000,0)
--(axis cs:3050000,0)
--(axis cs:3000000,0)
--(axis cs:2950000,0)
--(axis cs:2900000,0)
--(axis cs:2850000,0)
--(axis cs:2800000,0)
--(axis cs:2750000,0)
--(axis cs:2700000,0)
--(axis cs:2650000,0)
--(axis cs:2600000,0)
--(axis cs:2550000,0)
--(axis cs:2500000,0)
--(axis cs:2450000,0)
--(axis cs:2400000,0)
--(axis cs:2350000,0)
--(axis cs:2300000,0)
--(axis cs:2250000,0)
--(axis cs:2200000,0)
--(axis cs:2150000,0)
--(axis cs:2100000,0)
--(axis cs:2050000,0)
--(axis cs:2000000,0)
--(axis cs:1950000,0)
--(axis cs:1900000,0)
--(axis cs:1850000,0)
--(axis cs:1800000,0)
--(axis cs:1750000,0)
--(axis cs:1700000,0)
--(axis cs:1650000,0)
--(axis cs:1600000,0)
--(axis cs:1550000,0)
--(axis cs:1500000,0)
--(axis cs:1450000,0)
--(axis cs:1400000,0)
--(axis cs:1350000,0)
--(axis cs:1300000,0)
--(axis cs:1250000,0)
--(axis cs:1200000,0)
--(axis cs:1150000,0)
--(axis cs:1100000,0)
--(axis cs:1050000,0)
--(axis cs:1000000,0)
--(axis cs:950000,0)
--(axis cs:900000,0)
--(axis cs:850000,0)
--(axis cs:800000,0)
--(axis cs:750000,0)
--(axis cs:700000,0)
--(axis cs:650000,0)
--(axis cs:600000,0)
--(axis cs:550000,0)
--(axis cs:500000,0)
--(axis cs:450000,0)
--(axis cs:400000,0)
--(axis cs:350000,0)
--(axis cs:300000,0)
--(axis cs:250000,0)
--(axis cs:200000,0)
--(axis cs:150000,0)
--(axis cs:100000,0)
--(axis cs:50000,0)
--(axis cs:0,0)
--cycle;

\addplot [thick, solid, C8, mark=*, mark size=0, mark options={solid}]
table {%
0 0
50000 0
100000 0
150000 0
200000 0
250000 0
300000 0
350000 0
400000 0
450000 0
500000 0
550000 0
600000 0
650000 0
700000 0
750000 0
800000 0
850000 0
900000 0
950000 0
1000000 0
1050000 0
1100000 0
1150000 0
1200000 0
1250000 0
1300000 0
1350000 0
1400000 0
1450000 0
1500000 0
1550000 0
1600000 0
1650000 0
1700000 0
1750000 0
1800000 0
1850000 0
1900000 0
1950000 0
2000000 0
2050000 0
2100000 0
2150000 0
2200000 0
2250000 0
2300000 0
2350000 0
2400000 0
2450000 0
2500000 0
2550000 0
2600000 0
2650000 0
2700000 0
2750000 0
2800000 0
2850000 0
2900000 0
2950000 0
3000000 0
3050000 0
3100000 0
3150000 0
3200000 0
3250000 0
3300000 0
3350000 0
3400000 0
3450000 0
3500000 0
3550000 0
3600000 0
3650000 0
3700000 0
3750000 0
3800000 0
3850000 0
3900000 0
3950000 0
4000000 0
4050000 0
4100000 0
4150000 0
4200000 0
4250000 0
4300000 0
4350000 0
4400000 0
4450000 0
4500000 0
4550000 0
4600000 0
4650000 0
4700000 0
4750000 0
4800000 0
4850000 0
4900000 0
4950000 0
5000000 0
5050000 0
5100000 0
5150000 0
5200000 0
5250000 0
5300000 0
5350000 0
5400000 0
5450000 0
5500000 0
5550000 0
5600000 0
5650000 0
5700000 0
5750000 0
5800000 0
5850000 0
5900000 0
5950000 0
6000000 0
6050000 0
6100000 0
6150000 0
6200000 0
6250000 0
6300000 0
6350000 0
6400000 0
6450000 0
6500000 0
6550000 0
6600000 0
6650000 0
6700000 0
6750000 0
6800000 0
6850000 0
6900000 0
6950000 0
7000000 0
7050000 0
7100000 0
7150000 0
7200000 0
7250000 0
7300000 0
7350000 0
7400000 0
7450000 0
7500000 0
7550000 0
7600000 0
7650000 0
7700000 0
7750000 0
7800000 0
7850000 0
7900000 0
7950000 0
8000000 0
8050000 0
8100000 0
8150000 0
8200000 0
8250000 0
8300000 0
8350000 0
8400000 0
8450000 0
8500000 0
8550000 0
8600000 0
8650000 0
8700000 0
8750000 0
8800000 0
8850000 0
8900000 0
8950000 0
9000000 0
9050000 0
9100000 0
9150000 0
9200000 0
9250000 0
9300000 0
9350000 0
9400000 0
9450000 0
9500000 0
9550000 0
9600000 0
9650000 0
9700000 0
9750000 0
9800000 0
9850000 0
9900000 0
9950000 0
10000000 0
10050000 0
10100000 0
10150000 0
10200000 0
10250000 0
10300000 0
10350000 0
};

\path [draw=C8, fill=C8, opacity=0.2]
(axis cs:0,0)
--(axis cs:0,0)
--(axis cs:50000,0)
--(axis cs:100000,0)
--(axis cs:150000,0)
--(axis cs:200000,0)
--(axis cs:250000,0)
--(axis cs:300000,0)
--(axis cs:350000,0)
--(axis cs:400000,0)
--(axis cs:450000,0)
--(axis cs:500000,0)
--(axis cs:550000,0)
--(axis cs:600000,0)
--(axis cs:650000,0)
--(axis cs:700000,0)
--(axis cs:750000,0)
--(axis cs:800000,0)
--(axis cs:850000,0)
--(axis cs:900000,0)
--(axis cs:950000,0)
--(axis cs:1000000,0)
--(axis cs:1050000,0)
--(axis cs:1100000,0)
--(axis cs:1150000,0)
--(axis cs:1200000,0)
--(axis cs:1250000,0)
--(axis cs:1300000,0)
--(axis cs:1350000,0)
--(axis cs:1400000,0)
--(axis cs:1450000,0)
--(axis cs:1500000,0)
--(axis cs:1550000,0)
--(axis cs:1600000,0)
--(axis cs:1650000,0)
--(axis cs:1700000,0)
--(axis cs:1750000,0)
--(axis cs:1800000,0)
--(axis cs:1850000,0)
--(axis cs:1900000,0)
--(axis cs:1950000,0)
--(axis cs:2000000,0)
--(axis cs:2050000,0)
--(axis cs:2100000,0)
--(axis cs:2150000,0)
--(axis cs:2200000,0)
--(axis cs:2250000,0)
--(axis cs:2300000,0)
--(axis cs:2350000,0)
--(axis cs:2400000,0.02)
--(axis cs:2450000,0)
--(axis cs:2500000,0)
--(axis cs:2550000,0.03475)
--(axis cs:2600000,0)
--(axis cs:2650000,0)
--(axis cs:2700000,0)
--(axis cs:2750000,0)
--(axis cs:2800000,0)
--(axis cs:2850000,0)
--(axis cs:2900000,0.01475)
--(axis cs:2950000,0)
--(axis cs:3000000,0)
--(axis cs:3050000,0)
--(axis cs:3100000,0)
--(axis cs:3150000,0)
--(axis cs:3200000,0)
--(axis cs:3250000,0)
--(axis cs:3300000,0.01)
--(axis cs:3350000,0)
--(axis cs:3400000,0)
--(axis cs:3450000,0)
--(axis cs:3500000,0)
--(axis cs:3550000,0)
--(axis cs:3600000,0.02)
--(axis cs:3650000,0)
--(axis cs:3700000,0.02)
--(axis cs:3750000,0)
--(axis cs:3800000,0)
--(axis cs:3850000,0)
--(axis cs:3900000,0)
--(axis cs:3950000,0)
--(axis cs:4000000,0.02)
--(axis cs:4050000,0)
--(axis cs:4100000,0.02)
--(axis cs:4150000,0.02)
--(axis cs:4200000,0.01)
--(axis cs:4250000,0)
--(axis cs:4300000,0.02)
--(axis cs:4350000,0.01)
--(axis cs:4400000,0)
--(axis cs:4450000,0)
--(axis cs:4500000,0.01)
--(axis cs:4550000,0)
--(axis cs:4600000,0)
--(axis cs:4650000,0.03475)
--(axis cs:4700000,0)
--(axis cs:4750000,0)
--(axis cs:4800000,0)
--(axis cs:4850000,0.06)
--(axis cs:4900000,0.03)
--(axis cs:4950000,0.01475)
--(axis cs:5000000,0.03)
--(axis cs:5050000,0.02475)
--(axis cs:5100000,0.01475)
--(axis cs:5150000,0.01475)
--(axis cs:5200000,0.03475)
--(axis cs:5250000,0.07)
--(axis cs:5300000,0)
--(axis cs:5350000,0.00475000000000002)
--(axis cs:5400000,0.01475)
--(axis cs:5450000,0)
--(axis cs:5500000,0)
--(axis cs:5550000,0.04)
--(axis cs:5600000,0)
--(axis cs:5650000,0.01)
--(axis cs:5700000,0.02)
--(axis cs:5750000,0.02)
--(axis cs:5800000,0.03)
--(axis cs:5850000,0.05)
--(axis cs:5900000,0.02)
--(axis cs:5950000,0)
--(axis cs:6000000,0.02)
--(axis cs:6050000,0)
--(axis cs:6100000,0.01)
--(axis cs:6150000,0)
--(axis cs:6200000,0.02)
--(axis cs:6250000,0)
--(axis cs:6300000,0)
--(axis cs:6350000,0.03)
--(axis cs:6400000,0.00475000000000002)
--(axis cs:6450000,0)
--(axis cs:6500000,0)
--(axis cs:6550000,0)
--(axis cs:6600000,0)
--(axis cs:6650000,0.03)
--(axis cs:6700000,0)
--(axis cs:6750000,0.06)
--(axis cs:6800000,0.06)
--(axis cs:6850000,0)
--(axis cs:6900000,0.08)
--(axis cs:6950000,0)
--(axis cs:7000000,0.08)
--(axis cs:7050000,0.07475)
--(axis cs:7100000,0.09475)
--(axis cs:7150000,0.08)
--(axis cs:7200000,0.07)
--(axis cs:7250000,0.07475)
--(axis cs:7300000,0.06475)
--(axis cs:7350000,0.07)
--(axis cs:7400000,0.08475)
--(axis cs:7450000,0)
--(axis cs:7500000,0.08)
--(axis cs:7550000,0)
--(axis cs:7600000,0.09)
--(axis cs:7650000,0)
--(axis cs:7700000,0)
--(axis cs:7750000,0)
--(axis cs:7800000,0)
--(axis cs:7850000,0)
--(axis cs:7900000,0.1)
--(axis cs:7950000,0.2)
--(axis cs:8000000,0)
--(axis cs:8050000,0.1)
--(axis cs:8100000,0.1)
--(axis cs:8150000,0)
--(axis cs:8200000,0)
--(axis cs:8250000,0)
--(axis cs:8300000,0)
--(axis cs:8350000,0)
--(axis cs:8400000,0)
--(axis cs:8450000,0)
--(axis cs:8500000,0)
--(axis cs:8550000,0.1)
--(axis cs:8600000,0)
--(axis cs:8650000,0)
--(axis cs:8700000,0)
--(axis cs:8750000,0)
--(axis cs:8800000,0)
--(axis cs:8850000,0)
--(axis cs:8900000,0)
--(axis cs:8950000,0)
--(axis cs:9000000,0)
--(axis cs:9050000,0)
--(axis cs:9100000,0.1)
--(axis cs:9150000,0)
--(axis cs:9200000,0)
--(axis cs:9250000,0)
--(axis cs:9300000,0)
--(axis cs:9350000,0)
--(axis cs:9400000,0.1)
--(axis cs:9450000,0)
--(axis cs:9500000,0)
--(axis cs:9550000,0)
--(axis cs:9600000,0)
--(axis cs:9650000,0)
--(axis cs:9700000,0)
--(axis cs:9750000,0)
--(axis cs:9800000,0)
--(axis cs:9850000,0)
--(axis cs:9900000,0)
--(axis cs:9950000,0)
--(axis cs:10000000,0)
--(axis cs:10050000,0)
--(axis cs:10050000,0)
--(axis cs:10050000,0)
--(axis cs:10000000,0)
--(axis cs:9950000,0)
--(axis cs:9900000,0)
--(axis cs:9850000,0)
--(axis cs:9800000,0)
--(axis cs:9750000,0)
--(axis cs:9700000,0)
--(axis cs:9650000,0)
--(axis cs:9600000,0)
--(axis cs:9550000,0)
--(axis cs:9500000,0)
--(axis cs:9450000,0)
--(axis cs:9400000,0.1)
--(axis cs:9350000,0)
--(axis cs:9300000,0)
--(axis cs:9250000,0)
--(axis cs:9200000,0)
--(axis cs:9150000,0)
--(axis cs:9100000,0.1)
--(axis cs:9050000,0)
--(axis cs:9000000,0)
--(axis cs:8950000,0)
--(axis cs:8900000,0)
--(axis cs:8850000,0)
--(axis cs:8800000,0)
--(axis cs:8750000,0)
--(axis cs:8700000,0)
--(axis cs:8650000,0)
--(axis cs:8600000,0)
--(axis cs:8550000,0.1)
--(axis cs:8500000,0)
--(axis cs:8450000,0)
--(axis cs:8400000,0)
--(axis cs:8350000,0)
--(axis cs:8300000,0)
--(axis cs:8250000,0)
--(axis cs:8200000,0)
--(axis cs:8150000,0)
--(axis cs:8100000,0.1)
--(axis cs:8050000,0.1)
--(axis cs:8000000,0)
--(axis cs:7950000,0.2)
--(axis cs:7900000,0.1)
--(axis cs:7850000,0)
--(axis cs:7800000,0)
--(axis cs:7750000,0)
--(axis cs:7700000,0)
--(axis cs:7650000,0)
--(axis cs:7600000,0.1)
--(axis cs:7550000,0)
--(axis cs:7500000,0.1)
--(axis cs:7450000,0)
--(axis cs:7400000,0.1)
--(axis cs:7350000,0.1)
--(axis cs:7300000,0.1)
--(axis cs:7250000,0.1)
--(axis cs:7200000,0.1)
--(axis cs:7150000,0.1)
--(axis cs:7100000,0.1)
--(axis cs:7050000,0.1)
--(axis cs:7000000,0.1)
--(axis cs:6950000,0)
--(axis cs:6900000,0.1)
--(axis cs:6850000,0.01)
--(axis cs:6800000,0.12525)
--(axis cs:6750000,0.1)
--(axis cs:6700000,0.02)
--(axis cs:6650000,0.1)
--(axis cs:6600000,0.02)
--(axis cs:6550000,0)
--(axis cs:6500000,0.03)
--(axis cs:6450000,0.0352499999999999)
--(axis cs:6400000,0.09)
--(axis cs:6350000,0.11)
--(axis cs:6300000,0.0752499999999999)
--(axis cs:6250000,0.05)
--(axis cs:6200000,0.1405)
--(axis cs:6150000,0.04)
--(axis cs:6100000,0.09)
--(axis cs:6050000,0.07)
--(axis cs:6000000,0.1605)
--(axis cs:5950000,0.08)
--(axis cs:5900000,0.1205)
--(axis cs:5850000,0.12525)
--(axis cs:5800000,0.171)
--(axis cs:5750000,0.1505)
--(axis cs:5700000,0.0952499999999999)
--(axis cs:5650000,0.09)
--(axis cs:5600000,0.06)
--(axis cs:5550000,0.13)
--(axis cs:5500000,0.11)
--(axis cs:5450000,0.0552499999999999)
--(axis cs:5400000,0.12)
--(axis cs:5350000,0.1)
--(axis cs:5300000,0.0804999999999998)
--(axis cs:5250000,0.17525)
--(axis cs:5200000,0.13)
--(axis cs:5150000,0.1)
--(axis cs:5100000,0.09)
--(axis cs:5050000,0.19525)
--(axis cs:5000000,0.10525)
--(axis cs:4950000,0.15525)
--(axis cs:4900000,0.14)
--(axis cs:4850000,0.14)
--(axis cs:4800000,0.11)
--(axis cs:4750000,0.11)
--(axis cs:4700000,0.10525)
--(axis cs:4650000,0.14525)
--(axis cs:4600000,0.07)
--(axis cs:4550000,0.04)
--(axis cs:4500000,0.09)
--(axis cs:4450000,0.07)
--(axis cs:4400000,0.07)
--(axis cs:4350000,0.09)
--(axis cs:4300000,0.1)
--(axis cs:4250000,0.0852499999999999)
--(axis cs:4200000,0.09)
--(axis cs:4150000,0.13)
--(axis cs:4100000,0.10525)
--(axis cs:4050000,0.0852499999999999)
--(axis cs:4000000,0.1)
--(axis cs:3950000,0.0652499999999999)
--(axis cs:3900000,0.04)
--(axis cs:3850000,0.11)
--(axis cs:3800000,0.1)
--(axis cs:3750000,0.09)
--(axis cs:3700000,0.09)
--(axis cs:3650000,0.06)
--(axis cs:3600000,0.0952499999999999)
--(axis cs:3550000,0.0852499999999999)
--(axis cs:3500000,0.0652499999999999)
--(axis cs:3450000,0.0852499999999999)
--(axis cs:3400000,0.06)
--(axis cs:3350000,0.05)
--(axis cs:3300000,0.11)
--(axis cs:3250000,0.07)
--(axis cs:3200000,0.05)
--(axis cs:3150000,0.0752499999999999)
--(axis cs:3100000,0.08)
--(axis cs:3050000,0.04)
--(axis cs:3000000,0.07)
--(axis cs:2950000,0.03)
--(axis cs:2900000,0.11)
--(axis cs:2850000,0.0452499999999999)
--(axis cs:2800000,0.09)
--(axis cs:2750000,0.08)
--(axis cs:2700000,0.07)
--(axis cs:2650000,0.05)
--(axis cs:2600000,0.05)
--(axis cs:2550000,0.12)
--(axis cs:2500000,0.08)
--(axis cs:2450000,0.08)
--(axis cs:2400000,0.1)
--(axis cs:2350000,0.05)
--(axis cs:2300000,0.03)
--(axis cs:2250000,0.00524999999999991)
--(axis cs:2200000,0.03)
--(axis cs:2150000,0.08)
--(axis cs:2100000,0.03)
--(axis cs:2050000,0.0152499999999999)
--(axis cs:2000000,0)
--(axis cs:1950000,0)
--(axis cs:1900000,0.08)
--(axis cs:1850000,0.0552499999999999)
--(axis cs:1800000,0.0152499999999999)
--(axis cs:1750000,0)
--(axis cs:1700000,0.01)
--(axis cs:1650000,0)
--(axis cs:1600000,0)
--(axis cs:1550000,0)
--(axis cs:1500000,0)
--(axis cs:1450000,0)
--(axis cs:1400000,0.0252499999999999)
--(axis cs:1350000,0)
--(axis cs:1300000,0.01)
--(axis cs:1250000,0)
--(axis cs:1200000,0)
--(axis cs:1150000,0)
--(axis cs:1100000,0)
--(axis cs:1050000,0)
--(axis cs:1000000,0)
--(axis cs:950000,0)
--(axis cs:900000,0)
--(axis cs:850000,0)
--(axis cs:800000,0)
--(axis cs:750000,0)
--(axis cs:700000,0)
--(axis cs:650000,0)
--(axis cs:600000,0)
--(axis cs:550000,0)
--(axis cs:500000,0)
--(axis cs:450000,0)
--(axis cs:400000,0.02)
--(axis cs:350000,0)
--(axis cs:300000,0)
--(axis cs:250000,0)
--(axis cs:200000,0)
--(axis cs:150000,0)
--(axis cs:100000,0)
--(axis cs:50000,0)
--(axis cs:0,0)
--cycle;

\addplot [thick, dashed, C8, mark=*, mark size=0, mark options={solid}]
table {%
0 0
50000 0
100000 0
150000 0
200000 0
250000 0
300000 0
350000 0
400000 0
450000 0
500000 0
550000 0
600000 0
650000 0
700000 0
750000 0
800000 0
850000 0
900000 0
950000 0
1000000 0
1050000 0
1100000 0
1150000 0
1200000 0
1250000 0
1300000 0
1350000 0
1400000 0
1450000 0
1500000 0
1550000 0
1600000 0
1650000 0
1700000 0
1750000 0
1800000 0
1850000 0.01
1900000 0.03
1950000 0
2000000 0
2050000 0
2100000 0
2150000 0.03
2200000 0
2250000 0
2300000 0
2350000 0.01
2400000 0.06
2450000 0.04
2500000 0.04
2550000 0.08
2600000 0.01
2650000 0.01
2700000 0.03
2750000 0.02
2800000 0.04
2850000 0
2900000 0.06
2950000 0
3000000 0.01
3050000 0
3100000 0.03
3150000 0.02
3200000 0.02
3250000 0.02
3300000 0.04
3350000 0.01
3400000 0.03
3450000 0.03
3500000 0.02
3550000 0.04
3600000 0.06
3650000 0.02
3700000 0.06
3750000 0.04
3800000 0.04
3850000 0.04
3900000 0
3950000 0.01
4000000 0.07
4050000 0.03
4100000 0.06
4150000 0.06
4200000 0.05
4250000 0.04
4300000 0.06
4350000 0.05
4400000 0.02
4450000 0.01
4500000 0.05
4550000 0
4600000 0.03
4650000 0.08
4700000 0.03
4750000 0.04
4800000 0.03
4850000 0.1
4900000 0.07
4950000 0.06
5000000 0.06
5050000 0.11
5100000 0.05
5150000 0.06
5200000 0.08
5250000 0.13
5300000 0.03
5350000 0.05
5400000 0.06
5450000 0.01
5500000 0.05
5550000 0.07
5600000 0.03
5650000 0.05
5700000 0.05
5750000 0.07
5800000 0.09
5850000 0.08
5900000 0.06
5950000 0.03
6000000 0.06
6050000 0.04
6100000 0.05
6150000 0.01
6200000 0.06
6250000 0.01
6300000 0.03
6350000 0.06
6400000 0.05
6450000 0
6500000 0
6550000 0
6600000 0
6650000 0.07
6700000 0
6750000 0.09
6800000 0.1
6850000 0
6900000 0.1
6950000 0
7000000 0.1
7050000 0.1
7100000 0.1
7150000 0.1
7200000 0.1
7250000 0.1
7300000 0.1
7350000 0.1
7400000 0.1
7450000 0
7500000 0.1
7550000 0
7600000 0.1
7650000 0
7700000 0
7750000 0
7800000 0
7850000 0
7900000 0.1
7950000 0.2
8000000 0
8050000 0.1
8100000 0.1
8150000 0
8200000 0
8250000 0
8300000 0
8350000 0
8400000 0
8450000 0
8500000 0
8550000 0.1
8600000 0
8650000 0
8700000 0
8750000 0
8800000 0
8850000 0
8900000 0
8950000 0
9000000 0
9050000 0
9100000 0.1
9150000 0
9200000 0
9250000 0
9300000 0
9350000 0
9400000 0.1
9450000 0
9500000 0
9550000 0
9600000 0
9650000 0
9700000 0
9750000 0
9800000 0
9850000 0
9900000 0
9950000 0
10000000 0
10050000 0
};
\addplot [thick, solid, C8, mark=*, mark size=0, mark options={solid}]
table {%
0 0
50000 0
100000 0
150000 0
200000 0
250000 0
300000 0
350000 0
400000 0
450000 0
500000 0
550000 0
600000 0
650000 0
700000 0
750000 0
800000 0
850000 0
900000 0
950000 0
1000000 0
1050000 0
1100000 0
1150000 0
1200000 0
1250000 0
1300000 0
1350000 0
1400000 0
1450000 0
1500000 0
1550000 0
1600000 0
1650000 0
1700000 0
1750000 0
1800000 0
1850000 0
1900000 0
1950000 0
2000000 0
2050000 0
2100000 0
2150000 0
2200000 0
2250000 0
2300000 0
2350000 0
2400000 0
2450000 0
2500000 0
2550000 0
2600000 0
2650000 0
2700000 0
2750000 0
2800000 0
2850000 0
2900000 0
2950000 0
3000000 0
3050000 0
3100000 0
3150000 0
3200000 0
3250000 0
3300000 0
3350000 0
3400000 0
3450000 0
3500000 0
3550000 0
3600000 0
3650000 0
3700000 0
3750000 0
3800000 0
3850000 0
3900000 0
3950000 0
4000000 0
4050000 0
4100000 0
4150000 0
4200000 0
4250000 0
4300000 0
4350000 0
4400000 0
4450000 0
4500000 0
4550000 0
4600000 0
4650000 0
4700000 0
4750000 0
4800000 0
4850000 0
4900000 0
4950000 0
5000000 0
5050000 0
5100000 0
5150000 0
5200000 0
5250000 0
5300000 0
5350000 0
5400000 0
5450000 0
5500000 0
5550000 0
5600000 0
5650000 0
5700000 0
5750000 0
5800000 0
5850000 0
5900000 0
5950000 0
6000000 0
6050000 0
6100000 0
6150000 0
6200000 0
6250000 0
6300000 0
6350000 0
6400000 0
6450000 0
6500000 0
6550000 0
6600000 0
6650000 0
6700000 0
6750000 0
6800000 0
6850000 0
6900000 0
6950000 0
7000000 0
7050000 0
7100000 0
7150000 0
7200000 0
7250000 0
7300000 0
7350000 0
7400000 0
7450000 0
7500000 0
7550000 0
7600000 0
7650000 0
7700000 0
7750000 0
7800000 0
7850000 0
7900000 0
7950000 0
8000000 0
8050000 0
8100000 0
8150000 0
8200000 0
8250000 0
8300000 0
8350000 0
8400000 0
8450000 0
8500000 0
8550000 0
8600000 0
8650000 0
8700000 0
8750000 0
8800000 0
8850000 0
8900000 0
8950000 0
9000000 0
9050000 0
9100000 0
9150000 0
9200000 0
9250000 0
9300000 0
9350000 0
9400000 0
9450000 0
9500000 0
9550000 0
9600000 0
9650000 0
9700000 0
9750000 0
9800000 0
9850000 0
9900000 0
9950000 0
10000000 0
10050000 0
10100000 0
10150000 0
10200000 0
10250000 0
10300000 0
10350000 0
};

\path [draw=C6, fill=C6, opacity=0.2]
(axis cs:52000,0)
--(axis cs:52000,0)
--(axis cs:1352000,0)
--(axis cs:2652000,0)
--(axis cs:3952000,0)
--(axis cs:5252000,0)
--(axis cs:6552000,0.015)
--(axis cs:7852000,0.005)
--(axis cs:9152000,0)
--(axis cs:10452000,0.015)
--(axis cs:11752000,0.03)
--(axis cs:13052000,0)
--(axis cs:14352000,0.025)
--(axis cs:15652000,0.025)
--(axis cs:16952000,0.02)
--(axis cs:18252000,0.01)
--(axis cs:19552000,0.065)
--(axis cs:20852000,0.015)
--(axis cs:22152000,0.04)
--(axis cs:23452000,0.015)
--(axis cs:24752000,0.025)
--(axis cs:26052000,0.015)
--(axis cs:27352000,0.015)
--(axis cs:28652000,0.005)
--(axis cs:29952000,0.02)
--(axis cs:31252000,0.025)
--(axis cs:32552000,0.025)
--(axis cs:33852000,0.015)
--(axis cs:35152000,0.01)
--(axis cs:36452000,0.025)
--(axis cs:37752000,0.025)
--(axis cs:39052000,0.02)
--(axis cs:40352000,0.09)
--(axis cs:39052000,0.08)
--(axis cs:37752000,0.1)
--(axis cs:36452000,0.085)
--(axis cs:35152000,0.065)
--(axis cs:33852000,0.065)
--(axis cs:32552000,0.085)
--(axis cs:31252000,0.065)
--(axis cs:29952000,0.1)
--(axis cs:28652000,0.07)
--(axis cs:27352000,0.055)
--(axis cs:26052000,0.06)
--(axis cs:24752000,0.1)
--(axis cs:23452000,0.065)
--(axis cs:22152000,0.08)
--(axis cs:20852000,0.085)
--(axis cs:19552000,0.125)
--(axis cs:18252000,0.08)
--(axis cs:16952000,0.1)
--(axis cs:15652000,0.08)
--(axis cs:14352000,0.1)
--(axis cs:13052000,0.03)
--(axis cs:11752000,0.075)
--(axis cs:10452000,0.07)
--(axis cs:9152000,0.05)
--(axis cs:7852000,0.05)
--(axis cs:6552000,0.07)
--(axis cs:5252000,0.045)
--(axis cs:3952000,0.05)
--(axis cs:2652000,0.01)
--(axis cs:1352000,0)
--(axis cs:52000,0)
--cycle;

\addplot [thick, solid, C6, mark=*, mark size=0, mark options={solid}]
table {%
52000 0
1352000 0
2652000 0
3952000 0.02
5252000 0.005
6552000 0.035
7852000 0.025
9152000 0.02
10452000 0.035
11752000 0.05
13052000 0.005
14352000 0.05
15652000 0.055
16952000 0.06
18252000 0.04
19552000 0.09
20852000 0.04
22152000 0.06
23452000 0.035
24752000 0.06
26052000 0.035
27352000 0.035
28652000 0.03
29952000 0.06
31252000 0.045
32552000 0.05
33852000 0.035
35152000 0.03
36452000 0.045
37752000 0.065
39052000 0.045
};

\path [draw=C6, fill=C6, opacity=0.2]
(axis cs:52000,0)
--(axis cs:52000,0)
--(axis cs:1352000,0)
--(axis cs:2652000,0)
--(axis cs:3952000,0)
--(axis cs:5252000,0)
--(axis cs:6552000,0)
--(axis cs:7852000,0)
--(axis cs:9152000,0)
--(axis cs:10452000,0)
--(axis cs:11752000,0)
--(axis cs:13052000,0)
--(axis cs:14352000,0)
--(axis cs:15652000,0)
--(axis cs:16952000,0)
--(axis cs:18252000,0)
--(axis cs:19552000,0)
--(axis cs:20852000,0)
--(axis cs:22152000,0)
--(axis cs:23452000,0)
--(axis cs:24752000,0)
--(axis cs:26052000,0)
--(axis cs:27352000,0)
--(axis cs:28652000,0)
--(axis cs:29952000,0)
--(axis cs:31252000,0)
--(axis cs:32552000,0)
--(axis cs:33852000,0)
--(axis cs:35152000,0)
--(axis cs:36452000,0)
--(axis cs:37752000,0)
--(axis cs:39052000,0)
--(axis cs:39052000,0)
--(axis cs:37752000,0)
--(axis cs:36452000,0)
--(axis cs:35152000,0)
--(axis cs:33852000,0)
--(axis cs:32552000,0)
--(axis cs:31252000,0)
--(axis cs:29952000,0)
--(axis cs:28652000,0)
--(axis cs:27352000,0)
--(axis cs:26052000,0)
--(axis cs:24752000,0)
--(axis cs:23452000,0)
--(axis cs:22152000,0)
--(axis cs:20852000,0)
--(axis cs:19552000,0)
--(axis cs:18252000,0)
--(axis cs:16952000,0)
--(axis cs:15652000,0)
--(axis cs:14352000,0)
--(axis cs:13052000,0)
--(axis cs:11752000,0)
--(axis cs:10452000,0)
--(axis cs:9152000,0)
--(axis cs:7852000,0)
--(axis cs:6552000,0)
--(axis cs:5252000,0)
--(axis cs:3952000,0)
--(axis cs:2652000,0)
--(axis cs:1352000,0)
--(axis cs:52000,0)
--cycle;

\addplot [thick, dashed, C6, mark=*, mark size=0, mark options={solid}]
table {%
52000 0
1352000 0
2652000 0
3952000 0
5252000 0
6552000 0
7852000 0
9152000 0
10452000 0
11752000 0
13052000 0
14352000 0
15652000 0
16952000 0
18252000 0
19552000 0
20852000 0
22152000 0
23452000 0
24752000 0
26052000 0
27352000 0
28652000 0
29952000 0
31252000 0
32552000 0
33852000 0
35152000 0
36452000 0
37752000 0
39052000 0
};
\end{axis}

\end{tikzpicture}

%% file: tikz_plots/reacher/legend.tex
\begin{tikzpicture} 
    \begin{axis}[%
    hide axis,
    xmin=10,
    xmax=50,
    ymin=0,
    ymax=0.4,
    legend style={
        draw=white!15!black,
        legend cell align=left,
        legend columns=-1, 
        legend style={
            draw=none,
            column sep=1ex,
            line width=1pt
        }
    },
    ]
    \addlegendimage{C1}
    \addlegendentry{PPO};
    \addlegendimage{C5}
    \addlegendentry{TRPL};
    \addlegendimage{C2}
    \addlegendentry{NDP};
    \addlegendimage{C6}
    \addlegendentry{ES};
    \addlegendimage{C7}
    \addlegendentry{CMORE};
    \addlegendimage{C8}
    \addlegendentry{SAC};
    \addlegendimage{C4}
    \addlegendentry{BBRL-PPO};
    \addlegendimage{C0}
    \addlegendentry{BBRL-TRPL};
    \end{axis}
\end{tikzpicture}

%% file: tikz_plots/reacher/reacher_pareto.tex
\begin{tikzpicture}

\begin{axis}[
legend cell align={left},
legend style={fill opacity=0.8, draw opacity=1, text opacity=1, draw=lightgray204},
title={5D Reacher - Energy},
tick align=outside,
tick pos=left,
scaled ticks=false,
xmajorgrids=true,
ymajorgrids=true,
x grid style={darkgray176},
xlabel={Control Cost},
xmin=-0.25, xmax=4.2,
xtick style={color=black},
y tick label style=
y grid style={darkgray176},
ylabel={Goal Distance in [m]},
ymin=-0.00367045595988746, ymax=0.055,
ytick style={color=black},
yticklabel style={
    /pgf/number format/.cd,
        fixed,
        fixed zerofill,
        precision=2,
    /tikz/.cd
    },
name=ax1
]
\addplot [semithick, C0, mark=*, mark size=1, mark options={solid}, only marks]
table {%
0.0761166694301032 0.00623786380625012
0.0411478153793777 0.00619509787901616
0.031088966871415 0.00628147933958715
0.0251961040414871 0.00634073501763791
0.0214513908835172 0.00629758784836845
0.0184212210635465 0.00624427343893159
0.0173518556623966 0.00638589499080119
0.01603448353249 0.00647677607995042
0.0148634360948103 0.00663652402964895
0.0136176084619929 0.00633819598099539
0.0134472266595547 0.00637335626376129
0.0126815693038882 0.0065409688534958
0.0120948011780637 0.00642188957511391
0.0115717561073237 0.00672099153661967
0.01121387752385 0.00674908865800919
0.0107545200762243 0.00677682833080307
0.0104365622360233 0.0067526624030303
0.0101881087252347 0.0066776582073486
0.00995761586453003 0.00670390494515648
0.00982982525372756 0.00684677378427535
0.00864207984755357 0.00676497440959448
0.00812241622408941 0.00705769167792255
0.00744198995442658 0.00696257644232432
0.00690940418346233 0.00677424623385555
0.00669187797962886 0.00712548366863515
0.00640357019395304 0.00690846517266827
0.00620540754974669 0.00672489799913334
0.00592054160218229 0.00677735042878557
0.00581994379123407 0.00698476330228835
0.00565695619773785 0.00673265475110774
0.00558332282324088 0.00681844268259598
0.00543127728503315 0.0069693471752895
0.0052867827797334 0.00697669383096753
0.00525639717410556 0.00700346508879379
0.00516329085387839 0.00673125769100021
0.00512979855953202 0.006958519926207
};
\addplot [semithick, C1, mark=square*, mark size=1, mark options={solid}, only marks]
table {%
2.30136734979368 0.00588652261519048
1.27332182842184 0.00664409919433378
0.807665313558108 0.0126162869495419
0.585508823335014 0.0154168203163644
0.475020124694149 0.0188085550792419
0.41424913279561 0.024038040825114
0.372571646090098 0.025414768803837
0.330546979921109 0.0325025029728306
0.28412789313202 0.0262784864598892
0.275806702286927 0.0288070999612936
0.243942821192354 0.0353430405649638
0.236187284953288 0.0305203359443709
0.241589341581954 0.0399671290529157
0.219505133939767 0.0433266188950769
0.202733665419738 0.0389417708835401
0.202820029203675 0.0530862775640351
0.185785765188443 0.051339273415053
0.16865183737666 0.0443723265927089
0.167585348762573 0.0473135541341738
0.172533656009557 0.063451622913814
0.147160257158381 0.0417540033954945
0.0922615828703012 0.111748807856199
0.0642936777343629 0.175241275254231
};



\addplot [semithick, C4, mark=triangle*, mark size=1, mark options={solid}, only marks]
table {%
1.88809231595219 0.0419138091979824
0.488300004041918 0.0409058616718117
0.148885609873042 0.0424843978263697
0.092004522100548 0.0435045211904908
0.0654392334003469 0.0446367679207029
0.0519977063859921 0.0474636669253055
0.0439879058350099 0.0479194044035264
0.0381092402819536 0.0480859642954231
0.0366574065599025 0.0495564969060759
0.031566668295903 0.0502429973242361
0.0312272459135759 0.049879877632215
0.0273447185363163 0.0512717010595057
0.0269520525092751 0.0514918811941105
0.0250286300347991 0.0526169156793315
0.0240012931824594 0.0569464537598505
0.021574867622267 0.0552146779493538
0.0210599171136494 0.0570260946729313
0.0198102320234284 0.0562238041059758
0.0187341631885884 0.0574198876357938
};


\coordinate (c1) at (axis cs: 0.15,0.005);
\coordinate (c2) at (axis cs:-0.05,0.008);
\draw (c1) rectangle (c2);

\coordinate (insetSW) at (axis cs:\pgfkeysvalueof{/pgfplots/xmax},\pgfkeysvalueof{/pgfplots/ymax}); 
\end{axis}

\begin{axis}[
name=ax2,
at={(insetSW)}, anchor=north east, width=150,
scaled ticks=false,
axis background/.style={fill=white},
clip=false,
yticklabels={,,},
xticklabels={,,}
]
\addplot [semithick, C0, mark=*, mark size=1, mark options={solid}, only marks]
table {%
0.0411478153793777 0.00619509787901616
0.031088966871415 0.00628147933958715
0.0251961040414871 0.00634073501763791
0.0214513908835172 0.00629758784836845
0.0184212210635465 0.00624427343893159
0.0173518556623966 0.00638589499080119
0.01603448353249 0.00647677607995042
0.0148634360948103 0.00663652402964895
0.0136176084619929 0.00633819598099539
0.0134472266595547 0.00637335626376129
0.0126815693038882 0.0065409688534958
0.0120948011780637 0.00642188957511391
0.0115717561073237 0.00672099153661967
0.01121387752385 0.00674908865800919
0.0107545200762243 0.00677682833080307
0.0104365622360233 0.0067526624030303
0.0101881087252347 0.0066776582073486
0.00995761586453003 0.00670390494515648
0.00982982525372756 0.00684677378427535
0.00864207984755357 0.00676497440959448
0.00812241622408941 0.00705769167792255
0.00744198995442658 0.00696257644232432
0.00690940418346233 0.00677424623385555
0.00669187797962886 0.00712548366863515
0.00640357019395304 0.00690846517266827
0.00620540754974669 0.00672489799913334
0.00592054160218229 0.00677735042878557
0.00581994379123407 0.00698476330228835
0.00565695619773785 0.00673265475110774
0.00558332282324088 0.00681844268259598
0.00543127728503315 0.0069693471752895
0.0052867827797334 0.00697669383096753
0.00525639717410556 0.00700346508879379
0.00516329085387839 0.00673125769100021
0.00512979855953202 0.006958519926207
};
\end{axis}

\draw [dashed, opacity=0.2] (c1) -- (ax2.south east);
\draw [dashed, opacity=0.2] (c2) -- (ax2.north west);


\end{tikzpicture}

%% file: tikz_plots/box_pushing/box_pushing_pareto.tex
\begin{tikzpicture}

\begin{axis}[
legend cell align={left},
legend style={fill opacity=0.8, draw opacity=1, text opacity=1, draw=lightgray204},
title={Box Pushing - Energy},
tick align=outside,
tick pos=left,
x grid style={darkgray176},
xlabel={Control Cost},
xmajorgrids,
xmin=-500, xmax=10500,
xtick style={color=black},
y grid style={darkgray176},
ylabel={Success Rate},
ymajorgrids,
ymin=-0.05, ymax=1.05,
ytick style={color=black},
]
\addplot [draw=C0, fill=C0, mark=*, mark size=1, only marks]
table{%
x  y
13668.9782187624 0.904
12086.2467034566 0.925
9664.30419948519 0.921
8488.49515119024 0.928
6558.62511260695 0.934
4124.11519022654 0.938
3164.72114581158 0.931
2483.78145496797 0.921
1795.27888533972 0.904
1497.59629020714 0.918
1250.29004889935 0.915
915.592041823842 0.887
783.279365729972 0.835
692.854325586116 0.806
604.602725836629 0.674
};

\addplot [draw=C1, fill=C1, mark=square*, mark size=1, only marks]
table{%
x  y
36242.9563120944 0.923
11176.0713471751 0.967
7941.60221306806 0.965
7811.54141590076 0.963
7124.93820995149 0.959
5996.26589576516 0.96
5440.95930375086 0.938
4799.04647205061 0.952
4296.29281453872 0.906
2707.04020837567 0.81
3124.24833204274 0.852
1482.65692171329 0.523
971.730483101388 0.332
1048.87689528689 0.18
476.732573017216 0.194
};

\addplot [draw=C4, fill=C4, mark=triangle*, mark size=1, only marks]
table{%
x  y
5270.00758245373 0.837
3274.28637876125 0.865
2578.3765427498 0.876
2020.00002333305 0.879
1624.01703040571 0.865
1312.70799644618 0.82
959.323209226482 0.799
739.358036506305 0.762
624.188351409005 0.712
613.387146267677 0.591
};

\end{axis}

\end{tikzpicture}

%% file: tikz_plots/metaworld/metaworld_IQM_sample_efficiency.tex
\begin{tikzpicture}
\begin{axis}
[
legend cell align={left},
legend style={
  fill opacity=0.8,
  draw opacity=1,
  text opacity=1,
  at={(0.97,0.03)},
  anchor=south east,
  draw=lightgray204
},
title={Meta-World},
tick align=outside,
tick pos=left,
x grid style={darkgray176},
xlabel={Number Environment Interactions},
xmajorgrids,
xmin=-2097120, xmax=61912000,
xtick style={color=black},
y grid style={darkgray176},
ylabel={Success Rate},
ymajorgrids,
ymin=-0.05, ymax=1.05,
ytick style={color=black}
]
\path [draw=C1, fill=C1, opacity=0.2]
(axis cs:0,0)
--(axis cs:0,0)
--(axis cs:1638400,0.468168421052632)
--(axis cs:3276800,0.69183298245614)
--(axis cs:4915200,0.778182335526316)
--(axis cs:6553600,0.818580646198831)
--(axis cs:8192000,0.832787973684211)
--(axis cs:9830400,0.835387720760234)
--(axis cs:11468800,0.826697726328689)
--(axis cs:13107200,0.839031408023736)
--(axis cs:14745600,0.835226814155487)
--(axis cs:16384000,0.834370504815962)
--(axis cs:18022400,0.830095469341245)
--(axis cs:19660800,0.840157306501548)
--(axis cs:21299200,0.840023851393189)
--(axis cs:22937600,0.84480761631407)
--(axis cs:24576000,0.835523104618163)
--(axis cs:26214400,0.833989547987616)
--(axis cs:27852800,0.842478814069487)
--(axis cs:29491200,0.832776669225146)
--(axis cs:31129600,0.821613596491228)
--(axis cs:32768000,0.81430931755246)
--(axis cs:34406400,0.806732213966288)
--(axis cs:36044800,0.819968247985159)
--(axis cs:37683200,0.815765680801268)
--(axis cs:39321600,0.826026442921028)
--(axis cs:40960000,0.812478077516094)
--(axis cs:42598400,0.813185543819721)
--(axis cs:44236800,0.818765999840287)
--(axis cs:45875200,0.82738344410659)
--(axis cs:47513600,0.820765897768318)
--(axis cs:49152000,0.815562492668559)
--(axis cs:50790400,0.81212554727505)
--(axis cs:52428800,0.814960834743108)
--(axis cs:54067200,0.808573657867094)
--(axis cs:55705600,0.801811119628114)
--(axis cs:57344000,0.806944642436361)
--(axis cs:58982400,0.806800217793258)
--(axis cs:60020800,0.812897659507224)
--(axis cs:60020800,0.855728499419505)
--(axis cs:58982400,0.851088281067251)
--(axis cs:57344000,0.851350840213278)
--(axis cs:55705600,0.846061077838592)
--(axis cs:54067200,0.852306012350115)
--(axis cs:52428800,0.855928461398595)
--(axis cs:50790400,0.851354865417219)
--(axis cs:49152000,0.859762544053148)
--(axis cs:47513600,0.86014942588395)
--(axis cs:45875200,0.864419682054401)
--(axis cs:44236800,0.860218381916802)
--(axis cs:42598400,0.851845672106123)
--(axis cs:40960000,0.854415197184137)
--(axis cs:39321600,0.868503958818615)
--(axis cs:37683200,0.855870659479581)
--(axis cs:36044800,0.861580615042508)
--(axis cs:34406400,0.851798286549708)
--(axis cs:32768000,0.857476062972996)
--(axis cs:31129600,0.864240912280702)
--(axis cs:29491200,0.875180069809942)
--(axis cs:27852800,0.880866498968008)
--(axis cs:26214400,0.873946708462332)
--(axis cs:24576000,0.876108354833161)
--(axis cs:22937600,0.886302900326797)
--(axis cs:21299200,0.883717578947369)
--(axis cs:19660800,0.880377977812178)
--(axis cs:18022400,0.871606136007912)
--(axis cs:16384000,0.879360852855177)
--(axis cs:14745600,0.877021178362573)
--(axis cs:13107200,0.88056018249054)
--(axis cs:11468800,0.873009074109907)
--(axis cs:9830400,0.882814541666667)
--(axis cs:8192000,0.883401096491228)
--(axis cs:6553600,0.87127598245614)
--(axis cs:4915200,0.84395846125731)
--(axis cs:3276800,0.756958355263158)
--(axis cs:1638400,0.526537631578947)
--(axis cs:0,0)
--cycle;

\addplot [thick, C1, mark=*, mark size=0, mark options={solid}]
table {%
0 0
1638400 0.496968421052632
3276800 0.724806432748538
4915200 0.812263742690059
6553600 0.847197660818714
8192000 0.860414035087719
9830400 0.860820261437909
11468800 0.851340540075679
13107200 0.861282413140695
14745600 0.856919831441349
16384000 0.857382318541452
18022400 0.851665875472996
19660800 0.86140701754386
21299200 0.862810990712074
22937600 0.866576848985208
24576000 0.857320639834881
26214400 0.856167926556588
27852800 0.86279090127279
29491200 0.855824674922601
31129600 0.844543859649123
32768000 0.837244427244582
34406400 0.829768851049192
36044800 0.842023165757531
37683200 0.836431780431471
39321600 0.848948913951546
40960000 0.835067841171556
42598400 0.83395406899602
44236800 0.841332323455698
45875200 0.847124954543221
47513600 0.841569342473832
49152000 0.839487994496044
50790400 0.832488598948351
52428800 0.83636512605042
54067200 0.831498582239913
55705600 0.824716794437073
57344000 0.829731114551084
58982400 0.829570588235294
60020800 0.835893309253526
};

\path [draw=C2, fill=C2, opacity=0.2]
(axis cs:16000,0.0125105263157895)
--(axis cs:16000,0.0066)
--(axis cs:1616000,0.0623367105263158)
--(axis cs:3216000,0.137605263157895)
--(axis cs:4816000,0.200326315789474)
--(axis cs:6416000,0.245462894736842)
--(axis cs:8016000,0.292489473684211)
--(axis cs:9616000,0.322757763157895)
--(axis cs:11216000,0.362173552631579)
--(axis cs:12816000,0.391057894736842)
--(axis cs:14416000,0.401610394736842)
--(axis cs:16016000,0.432931578947369)
--(axis cs:17616000,0.4447525)
--(axis cs:19216000,0.4634)
--(axis cs:20816000,0.477805131578947)
--(axis cs:22416000,0.489715789473684)
--(axis cs:24016000,0.507342105263158)
--(axis cs:25616000,0.514768421052632)
--(axis cs:27216000,0.523952631578947)
--(axis cs:28816000,0.531089473684211)
--(axis cs:30416000,0.540736710526316)
--(axis cs:32016000,0.5419)
--(axis cs:33616000,0.552305263157895)
--(axis cs:35216000,0.560815789473684)
--(axis cs:36816000,0.572157894736842)
--(axis cs:38416000,0.573863157894737)
--(axis cs:40016000,0.571288947368421)
--(axis cs:41616000,0.588015789473684)
--(axis cs:43216000,0.587515789473684)
--(axis cs:44816000,0.592563157894737)
--(axis cs:46416000,0.603768289473684)
--(axis cs:48016000,0.607926184210526)
--(axis cs:49616000,0.613605263157895)
--(axis cs:51216000,0.620752631578947)
--(axis cs:52816000,0.612857894736842)
--(axis cs:54416000,0.622178815789474)
--(axis cs:56016000,0.623873684210526)
--(axis cs:57616000,0.628226315789474)
--(axis cs:59216000,0.636489342105263)
--(axis cs:59216000,0.672115789473684)
--(axis cs:59216000,0.672115789473684)
--(axis cs:57616000,0.663610526315789)
--(axis cs:56016000,0.660036973684211)
--(axis cs:54416000,0.657936842105263)
--(axis cs:52816000,0.648489473684211)
--(axis cs:51216000,0.6574475)
--(axis cs:49616000,0.649689605263158)
--(axis cs:48016000,0.645158157894737)
--(axis cs:46416000,0.639589473684211)
--(axis cs:44816000,0.628684342105263)
--(axis cs:43216000,0.623594868421053)
--(axis cs:41616000,0.623731578947369)
--(axis cs:40016000,0.607389605263158)
--(axis cs:38416000,0.610115789473684)
--(axis cs:36816000,0.608231710526316)
--(axis cs:35216000,0.596695)
--(axis cs:33616000,0.588042105263158)
--(axis cs:32016000,0.578389473684211)
--(axis cs:30416000,0.576773684210526)
--(axis cs:28816000,0.567136842105263)
--(axis cs:27216000,0.560626315789474)
--(axis cs:25616000,0.550305263157895)
--(axis cs:24016000,0.543126447368421)
--(axis cs:22416000,0.526057894736842)
--(axis cs:20816000,0.513226315789474)
--(axis cs:19216000,0.498710526315789)
--(axis cs:17616000,0.480194868421053)
--(axis cs:16016000,0.468773684210526)
--(axis cs:14416000,0.436421315789474)
--(axis cs:12816000,0.426000131578947)
--(axis cs:11216000,0.3968475)
--(axis cs:9616000,0.356352631578947)
--(axis cs:8016000,0.325110526315789)
--(axis cs:6416000,0.277173684210526)
--(axis cs:4816000,0.229968421052632)
--(axis cs:3216000,0.163447368421053)
--(axis cs:1616000,0.0809790789473685)
--(axis cs:16000,0.0125105263157895)
--cycle;

\addplot [thick, C2, mark=*, mark size=0, mark options={solid}]
table {%
16000 0.00941052631578947
1616000 0.0715684210526316
3216000 0.1504
4816000 0.215042105263158
6416000 0.261326315789474
8016000 0.308805263157895
9616000 0.339505263157895
11216000 0.379552631578947
12816000 0.408547368421053
14416000 0.419047368421053
16016000 0.451036842105263
17616000 0.462563157894737
19216000 0.481068421052632
20816000 0.495573684210526
22416000 0.507915789473684
24016000 0.525310526315789
25616000 0.532621052631579
27216000 0.542478947368421
28816000 0.549168421052632
30416000 0.558889473684211
32016000 0.560336842105263
33616000 0.570273684210526
35216000 0.578878947368421
36816000 0.590257894736842
38416000 0.59198947368421
40016000 0.589363157894737
41616000 0.605873684210526
43216000 0.605642105263158
44816000 0.610715789473684
46416000 0.621831578947368
48016000 0.626705263157895
49616000 0.631773684210526
51216000 0.639305263157895
52816000 0.630852631578947
54416000 0.640242105263158
56016000 0.642078947368421
57616000 0.645963157894737
59216000 0.654457894736842
};

\path [draw=C4, fill=C4, opacity=0.2]
(axis cs:0,0)
--(axis cs:0,0)
--(axis cs:800000,0)
--(axis cs:1600000,0.0532)
--(axis cs:2400000,0.1)
--(axis cs:3200000,0.1372)
--(axis cs:4000000,0.14919)
--(axis cs:4800000,0.1612)
--(axis cs:5600000,0.1784)
--(axis cs:6400000,0.19959)
--(axis cs:7200000,0.2164)
--(axis cs:8000000,0.2336)
--(axis cs:8800000,0.2656)
--(axis cs:9600000,0.26559)
--(axis cs:10400000,0.2756)
--(axis cs:11200000,0.2668)
--(axis cs:12000000,0.2896)
--(axis cs:12800000,0.3)
--(axis cs:13600000,0.3042)
--(axis cs:14400000,0.2952)
--(axis cs:15200000,0.3124)
--(axis cs:16000000,0.3216)
--(axis cs:16800000,0.3152)
--(axis cs:17600000,0.3208)
--(axis cs:18400000,0.3072)
--(axis cs:19200000,0.324)
--(axis cs:20000000,0.3404)
--(axis cs:20800000,0.3348)
--(axis cs:21600000,0.3432)
--(axis cs:22400000,0.3424)
--(axis cs:23200000,0.34839)
--(axis cs:24000000,0.352)
--(axis cs:24800000,0.3756)
--(axis cs:25600000,0.3812)
--(axis cs:26400000,0.3868)
--(axis cs:27200000,0.3864)
--(axis cs:28000000,0.3776)
--(axis cs:28800000,0.3912)
--(axis cs:29600000,0.3844)
--(axis cs:30400000,0.388)
--(axis cs:31200000,0.4048)
--(axis cs:32000000,0.4144)
--(axis cs:32800000,0.4208)
--(axis cs:33600000,0.4108)
--(axis cs:34400000,0.4068)
--(axis cs:35200000,0.42)
--(axis cs:36000000,0.423195)
--(axis cs:36800000,0.4236)
--(axis cs:37600000,0.4416)
--(axis cs:38400000,0.4364)
--(axis cs:39200000,0.4212)
--(axis cs:40000000,0.4344)
--(axis cs:40800000,0.438)
--(axis cs:41600000,0.4296)
--(axis cs:42400000,0.4472)
--(axis cs:43200000,0.4396)
--(axis cs:44000000,0.46719)
--(axis cs:44800000,0.4448)
--(axis cs:45600000,0.4712)
--(axis cs:46400000,0.4396)
--(axis cs:47200000,0.4444)
--(axis cs:48000000,0.4396)
--(axis cs:48800000,0.46959)
--(axis cs:49600000,0.43)
--(axis cs:50400000,0.4536)
--(axis cs:51200000,0.44919)
--(axis cs:52000000,0.4796)
--(axis cs:52800000,0.4536)
--(axis cs:53600000,0.4788)
--(axis cs:54400000,0.4624)
--(axis cs:55200000,0.4436)
--(axis cs:56000000,0.47599)
--(axis cs:56800000,0.46359)
--(axis cs:57600000,0.502)
--(axis cs:58400000,0.452)
--(axis cs:59200000,0.48159)
--(axis cs:60000000,0.4616)
--(axis cs:60000000,0.5256)
--(axis cs:60000000,0.5256)
--(axis cs:59200000,0.5456)
--(axis cs:58400000,0.516)
--(axis cs:57600000,0.5636)
--(axis cs:56800000,0.5264)
--(axis cs:56000000,0.5356)
--(axis cs:55200000,0.5092)
--(axis cs:54400000,0.524205)
--(axis cs:53600000,0.5396)
--(axis cs:52800000,0.5196)
--(axis cs:52000000,0.5396)
--(axis cs:51200000,0.5092)
--(axis cs:50400000,0.516)
--(axis cs:49600000,0.4924)
--(axis cs:48800000,0.53001)
--(axis cs:48000000,0.5036)
--(axis cs:47200000,0.5084)
--(axis cs:46400000,0.5022)
--(axis cs:45600000,0.5332)
--(axis cs:44800000,0.50561)
--(axis cs:44000000,0.52801)
--(axis cs:43200000,0.4996)
--(axis cs:42400000,0.5088)
--(axis cs:41600000,0.4904)
--(axis cs:40800000,0.5)
--(axis cs:40000000,0.494)
--(axis cs:39200000,0.4792)
--(axis cs:38400000,0.4964)
--(axis cs:37600000,0.5012)
--(axis cs:36800000,0.4832)
--(axis cs:36000000,0.4826)
--(axis cs:35200000,0.4784)
--(axis cs:34400000,0.4656)
--(axis cs:33600000,0.472)
--(axis cs:32800000,0.47921)
--(axis cs:32000000,0.4728)
--(axis cs:31200000,0.4632)
--(axis cs:30400000,0.4448)
--(axis cs:29600000,0.4404)
--(axis cs:28800000,0.4496)
--(axis cs:28000000,0.4368)
--(axis cs:27200000,0.4404)
--(axis cs:26400000,0.4448)
--(axis cs:25600000,0.4356)
--(axis cs:24800000,0.4292)
--(axis cs:24000000,0.40481)
--(axis cs:23200000,0.4036)
--(axis cs:22400000,0.3956)
--(axis cs:21600000,0.3916)
--(axis cs:20800000,0.384)
--(axis cs:20000000,0.392)
--(axis cs:19200000,0.374)
--(axis cs:18400000,0.3552)
--(axis cs:17600000,0.3672)
--(axis cs:16800000,0.362)
--(axis cs:16000000,0.368)
--(axis cs:15200000,0.3584)
--(axis cs:14400000,0.3412)
--(axis cs:13600000,0.3482)
--(axis cs:12800000,0.3444)
--(axis cs:12000000,0.3312)
--(axis cs:11200000,0.3092)
--(axis cs:10400000,0.3196)
--(axis cs:9600000,0.3092)
--(axis cs:8800000,0.3104)
--(axis cs:8000000,0.2748)
--(axis cs:7200000,0.2548)
--(axis cs:6400000,0.2356)
--(axis cs:5600000,0.2144)
--(axis cs:4800000,0.1964)
--(axis cs:4000000,0.1812)
--(axis cs:3200000,0.1704)
--(axis cs:2400000,0.1296)
--(axis cs:1600000,0.08)
--(axis cs:800000,0)
--(axis cs:0,0)
--cycle;

\addplot [thick, C4, mark=*, mark size=0, mark options={solid}]
table {%
0 0
800000 0
1600000 0.0664
2400000 0.1144
3200000 0.154
4000000 0.1644
4800000 0.1784
5600000 0.1964
6400000 0.2172
7200000 0.2356
8000000 0.2544
8800000 0.2876
9600000 0.2872
10400000 0.298
11200000 0.2876
12000000 0.31
12800000 0.3228
13600000 0.327
14400000 0.3184
15200000 0.336
16000000 0.3452
16800000 0.3384
17600000 0.344
18400000 0.3312
19200000 0.3492
20000000 0.3664
20800000 0.3592
21600000 0.368
22400000 0.3688
23200000 0.376
24000000 0.378
24800000 0.402
25600000 0.4076
26400000 0.4164
27200000 0.4136
28000000 0.408
28800000 0.4204
29600000 0.4124
30400000 0.4168
31200000 0.434
32000000 0.4428
32800000 0.45
33600000 0.4416
34400000 0.4356
35200000 0.4484
36000000 0.453
36800000 0.454
37600000 0.472
38400000 0.4668
39200000 0.4498
40000000 0.464
40800000 0.4696
41600000 0.46
42400000 0.4776
43200000 0.4696
44000000 0.498
44800000 0.4748
45600000 0.502
46400000 0.4706
47200000 0.4756
48000000 0.4712
48800000 0.4996
49600000 0.4616
50400000 0.4844
51200000 0.4788
52000000 0.5092
52800000 0.4868
53600000 0.5092
54400000 0.4934
55200000 0.476
56000000 0.506
56800000 0.4948
57600000 0.5334
58400000 0.4844
59200000 0.5132
60000000 0.4936
};

\path [draw=C0, fill=C0, opacity=0.2]
(axis cs:0,0)
--(axis cs:0,0)
--(axis cs:800000,0)
--(axis cs:1600000,0.0416)
--(axis cs:2400000,0.1172)
--(axis cs:3200000,0.2004)
--(axis cs:4000000,0.2688)
--(axis cs:4800000,0.3452)
--(axis cs:5600000,0.3928)
--(axis cs:6400000,0.4408)
--(axis cs:7200000,0.4784)
--(axis cs:8000000,0.528)
--(axis cs:8800000,0.54679)
--(axis cs:9600000,0.5932)
--(axis cs:10400000,0.6052)
--(axis cs:11200000,0.6456)
--(axis cs:12000000,0.6528)
--(axis cs:12800000,0.6584)
--(axis cs:13600000,0.6944)
--(axis cs:14400000,0.7148)
--(axis cs:15200000,0.7496)
--(axis cs:16000000,0.77039)
--(axis cs:16800000,0.76)
--(axis cs:17600000,0.756)
--(axis cs:18400000,0.778)
--(axis cs:19200000,0.8016)
--(axis cs:20000000,0.81)
--(axis cs:20800000,0.8028)
--(axis cs:21600000,0.7992)
--(axis cs:22400000,0.798)
--(axis cs:23200000,0.8052)
--(axis cs:24000000,0.824)
--(axis cs:24800000,0.8344)
--(axis cs:25600000,0.8184)
--(axis cs:26400000,0.82879)
--(axis cs:27200000,0.83719)
--(axis cs:28000000,0.83079)
--(axis cs:28800000,0.8428)
--(axis cs:29600000,0.8332)
--(axis cs:30400000,0.8244)
--(axis cs:31200000,0.8356)
--(axis cs:32000000,0.838)
--(axis cs:32800000,0.8528)
--(axis cs:33600000,0.8488)
--(axis cs:34400000,0.84519)
--(axis cs:35200000,0.8504)
--(axis cs:36000000,0.8664)
--(axis cs:36800000,0.86)
--(axis cs:37600000,0.8624)
--(axis cs:38400000,0.8532)
--(axis cs:39200000,0.8516)
--(axis cs:40000000,0.8532)
--(axis cs:40800000,0.8696)
--(axis cs:41600000,0.862)
--(axis cs:42400000,0.85959)
--(axis cs:43200000,0.862)
--(axis cs:44000000,0.8508)
--(axis cs:44800000,0.85479)
--(axis cs:45600000,0.8516)
--(axis cs:46400000,0.8588)
--(axis cs:47200000,0.8556)
--(axis cs:48000000,0.8556)
--(axis cs:48800000,0.8728)
--(axis cs:49600000,0.8688)
--(axis cs:50400000,0.8664)
--(axis cs:51200000,0.8572)
--(axis cs:52000000,0.8664)
--(axis cs:52800000,0.8588)
--(axis cs:53600000,0.8716)
--(axis cs:54400000,0.8648)
--(axis cs:55200000,0.8632)
--(axis cs:56000000,0.8612)
--(axis cs:56800000,0.8536)
--(axis cs:57600000,0.85759)
--(axis cs:58400000,0.856)
--(axis cs:59200000,0.8596)
--(axis cs:60000000,0.8664)
--(axis cs:60000000,0.9036)
--(axis cs:59200000,0.8992)
--(axis cs:58400000,0.892)
--(axis cs:57600000,0.892)
--(axis cs:56800000,0.89)
--(axis cs:56000000,0.8964)
--(axis cs:55200000,0.8992)
--(axis cs:54400000,0.9004)
--(axis cs:53600000,0.9064)
--(axis cs:52800000,0.8952)
--(axis cs:52000000,0.904)
--(axis cs:51200000,0.8936)
--(axis cs:50400000,0.9056)
--(axis cs:49600000,0.9028)
--(axis cs:48800000,0.9036)
--(axis cs:48000000,0.892)
--(axis cs:47200000,0.8932)
--(axis cs:46400000,0.894)
--(axis cs:45600000,0.8892)
--(axis cs:44800000,0.8936)
--(axis cs:44000000,0.8888)
--(axis cs:43200000,0.8968)
--(axis cs:42400000,0.896)
--(axis cs:41600000,0.8984)
--(axis cs:40800000,0.9008)
--(axis cs:40000000,0.8924)
--(axis cs:39200000,0.8872)
--(axis cs:38400000,0.8896)
--(axis cs:37600000,0.9004)
--(axis cs:36800000,0.8964)
--(axis cs:36000000,0.902)
--(axis cs:35200000,0.8868)
--(axis cs:34400000,0.8816)
--(axis cs:33600000,0.886)
--(axis cs:32800000,0.89)
--(axis cs:32000000,0.87601)
--(axis cs:31200000,0.87321)
--(axis cs:30400000,0.8632)
--(axis cs:29600000,0.8708)
--(axis cs:28800000,0.88)
--(axis cs:28000000,0.8656)
--(axis cs:27200000,0.8728)
--(axis cs:26400000,0.8652)
--(axis cs:25600000,0.8556)
--(axis cs:24800000,0.872)
--(axis cs:24000000,0.8604)
--(axis cs:23200000,0.8436)
--(axis cs:22400000,0.83721)
--(axis cs:21600000,0.84)
--(axis cs:20800000,0.8408)
--(axis cs:20000000,0.8484)
--(axis cs:19200000,0.8408)
--(axis cs:18400000,0.8204)
--(axis cs:17600000,0.79801)
--(axis cs:16800000,0.80361)
--(axis cs:16000000,0.814)
--(axis cs:15200000,0.79601)
--(axis cs:14400000,0.766)
--(axis cs:13600000,0.74361)
--(axis cs:12800000,0.7092)
--(axis cs:12000000,0.70161)
--(axis cs:11200000,0.69)
--(axis cs:10400000,0.648)
--(axis cs:9600000,0.6376)
--(axis cs:8800000,0.59)
--(axis cs:8000000,0.5708)
--(axis cs:7200000,0.5192)
--(axis cs:6400000,0.484)
--(axis cs:5600000,0.4324)
--(axis cs:4800000,0.3844)
--(axis cs:4000000,0.3068)
--(axis cs:3200000,0.2424)
--(axis cs:2400000,0.1508)
--(axis cs:1600000,0.0636)
--(axis cs:800000,0)
--(axis cs:0,0)
--cycle;

\addplot [thick, C0, mark=*, mark size=0, mark options={solid}]
table {%
0 0
800000 0
1600000 0.0524
2400000 0.134
3200000 0.2212
4000000 0.2876
4800000 0.3648
5600000 0.4128
6400000 0.4624
7200000 0.4988
8000000 0.5496
8800000 0.5684
9600000 0.6156
10400000 0.6268
11200000 0.6676
12000000 0.6776
12800000 0.684
13600000 0.7192
14400000 0.7412
15200000 0.774
16000000 0.7924
16800000 0.7824
17600000 0.7776
18400000 0.8
19200000 0.8216
20000000 0.8288
20800000 0.822
21600000 0.8204
22400000 0.8176
23200000 0.8244
24000000 0.8424
24800000 0.8532
25600000 0.8376
26400000 0.8468
27200000 0.8552
28000000 0.8484
28800000 0.8616
29600000 0.852
30400000 0.8436
31200000 0.8544
32000000 0.8572
32800000 0.8716
33600000 0.8672
34400000 0.8632
35200000 0.8688
36000000 0.8856
36800000 0.8792
37600000 0.8816
38400000 0.8716
39200000 0.87
40000000 0.8728
40800000 0.8868
41600000 0.8808
42400000 0.878
43200000 0.8808
44000000 0.8704
44800000 0.874
45600000 0.8712
46400000 0.8768
47200000 0.8756
48000000 0.8744
48800000 0.8892
49600000 0.888
50400000 0.8864
51200000 0.8752
52000000 0.886
52800000 0.878
53600000 0.8904
54400000 0.8836
55200000 0.882
56000000 0.8796
56800000 0.8728
57600000 0.8768
58400000 0.8756
59200000 0.8804
60000000 0.8856
};

\path [draw=C3, fill=C3, opacity=0.2]
(axis cs:0,0)
--(axis cs:0,0)
--(axis cs:800000,0)
--(axis cs:1600000,0.0188)
--(axis cs:2400000,0.106)
--(axis cs:3200000,0.178)
--(axis cs:4000000,0.2452)
--(axis cs:4800000,0.274)
--(axis cs:5600000,0.2964)
--(axis cs:6400000,0.33399)
--(axis cs:7200000,0.3456)
--(axis cs:8000000,0.3676)
--(axis cs:8800000,0.38)
--(axis cs:9600000,0.4052)
--(axis cs:10400000,0.4128)
--(axis cs:11200000,0.4164)
--(axis cs:12000000,0.4184)
--(axis cs:12800000,0.4368)
--(axis cs:13600000,0.4344)
--(axis cs:14400000,0.4516)
--(axis cs:15200000,0.4468)
--(axis cs:16000000,0.4464)
--(axis cs:16800000,0.4652)
--(axis cs:17600000,0.4664)
--(axis cs:18400000,0.46679)
--(axis cs:19200000,0.456)
--(axis cs:20000000,0.47119)
--(axis cs:20800000,0.472)
--(axis cs:21600000,0.4856)
--(axis cs:22400000,0.49119)
--(axis cs:23200000,0.4796)
--(axis cs:24000000,0.4724)
--(axis cs:24800000,0.4884)
--(axis cs:25600000,0.4904)
--(axis cs:26400000,0.4932)
--(axis cs:27200000,0.4972)
--(axis cs:28000000,0.4968)
--(axis cs:28800000,0.5108)
--(axis cs:29600000,0.4984)
--(axis cs:30400000,0.4984)
--(axis cs:31200000,0.5112)
--(axis cs:32000000,0.5008)
--(axis cs:32800000,0.5012)
--(axis cs:33600000,0.5156)
--(axis cs:34400000,0.52)
--(axis cs:35200000,0.5112)
--(axis cs:36000000,0.5156)
--(axis cs:36800000,0.53)
--(axis cs:37600000,0.5092)
--(axis cs:38400000,0.53)
--(axis cs:39200000,0.5348)
--(axis cs:40000000,0.5312)
--(axis cs:40800000,0.5384)
--(axis cs:41600000,0.5236)
--(axis cs:42400000,0.5428)
--(axis cs:43200000,0.5268)
--(axis cs:44000000,0.5292)
--(axis cs:44800000,0.5416)
--(axis cs:45600000,0.54079)
--(axis cs:46400000,0.5456)
--(axis cs:47200000,0.5556)
--(axis cs:48000000,0.544)
--(axis cs:48800000,0.5444)
--(axis cs:49600000,0.5484)
--(axis cs:50400000,0.5476)
--(axis cs:51200000,0.5476)
--(axis cs:52000000,0.5396)
--(axis cs:52800000,0.55)
--(axis cs:53600000,0.5384)
--(axis cs:54400000,0.5348)
--(axis cs:55200000,0.5572)
--(axis cs:56000000,0.5668)
--(axis cs:56800000,0.5512)
--(axis cs:57600000,0.548)
--(axis cs:58400000,0.5408)
--(axis cs:59200000,0.5464)
--(axis cs:59200000,0.58721)
--(axis cs:59200000,0.58721)
--(axis cs:58400000,0.584)
--(axis cs:57600000,0.588)
--(axis cs:56800000,0.5916)
--(axis cs:56000000,0.6084)
--(axis cs:55200000,0.598)
--(axis cs:54400000,0.5764)
--(axis cs:53600000,0.57761)
--(axis cs:52800000,0.5928)
--(axis cs:52000000,0.5844)
--(axis cs:51200000,0.59)
--(axis cs:50400000,0.58921)
--(axis cs:49600000,0.5896)
--(axis cs:48800000,0.586)
--(axis cs:48000000,0.5864)
--(axis cs:47200000,0.5972)
--(axis cs:46400000,0.5876)
--(axis cs:45600000,0.5832)
--(axis cs:44800000,0.5848)
--(axis cs:44000000,0.57241)
--(axis cs:43200000,0.568)
--(axis cs:42400000,0.5836)
--(axis cs:41600000,0.5668)
--(axis cs:40800000,0.5804)
--(axis cs:40000000,0.5728)
--(axis cs:39200000,0.576)
--(axis cs:38400000,0.5704)
--(axis cs:37600000,0.5512)
--(axis cs:36800000,0.5684)
--(axis cs:36000000,0.556)
--(axis cs:35200000,0.5524)
--(axis cs:34400000,0.5608)
--(axis cs:33600000,0.5544)
--(axis cs:32800000,0.5432)
--(axis cs:32000000,0.54)
--(axis cs:31200000,0.5504)
--(axis cs:30400000,0.5376)
--(axis cs:29600000,0.5352)
--(axis cs:28800000,0.54681)
--(axis cs:28000000,0.534)
--(axis cs:27200000,0.53481)
--(axis cs:26400000,0.534)
--(axis cs:25600000,0.5264)
--(axis cs:24800000,0.5268)
--(axis cs:24000000,0.5116)
--(axis cs:23200000,0.5156)
--(axis cs:22400000,0.52521)
--(axis cs:21600000,0.5216)
--(axis cs:20800000,0.5092)
--(axis cs:20000000,0.5052)
--(axis cs:19200000,0.4932)
--(axis cs:18400000,0.5016)
--(axis cs:17600000,0.50281)
--(axis cs:16800000,0.50201)
--(axis cs:16000000,0.4808)
--(axis cs:15200000,0.482)
--(axis cs:14400000,0.4852)
--(axis cs:13600000,0.4696)
--(axis cs:12800000,0.4704)
--(axis cs:12000000,0.4528)
--(axis cs:11200000,0.45)
--(axis cs:10400000,0.4444)
--(axis cs:9600000,0.4384)
--(axis cs:8800000,0.4152)
--(axis cs:8000000,0.4016)
--(axis cs:7200000,0.3796)
--(axis cs:6400000,0.3636)
--(axis cs:5600000,0.3324)
--(axis cs:4800000,0.3096)
--(axis cs:4000000,0.2812)
--(axis cs:3200000,0.2192)
--(axis cs:2400000,0.1388)
--(axis cs:1600000,0.0396)
--(axis cs:800000,0)
--(axis cs:0,0)
--cycle;

\addplot [thick, C3, mark=*, mark size=0, mark options={solid}]
table {%
0 0
800000 0
1600000 0.0292
2400000 0.1216
3200000 0.1988
4000000 0.2632
4800000 0.292
5600000 0.3148
6400000 0.3488
7200000 0.3624
8000000 0.3844
8800000 0.3976
9600000 0.422
10400000 0.4288
11200000 0.4336
12000000 0.4356
12800000 0.4532
13600000 0.4516
14400000 0.468
15200000 0.4644
16000000 0.4636
16800000 0.4836
17600000 0.4844
18400000 0.484
19200000 0.474
20000000 0.488
20800000 0.4904
21600000 0.5036
22400000 0.5084
23200000 0.4976
24000000 0.492
24800000 0.5076
25600000 0.5084
26400000 0.514
27200000 0.5156
28000000 0.5148
28800000 0.5292
29600000 0.5168
30400000 0.5176
31200000 0.5304
32000000 0.5204
32800000 0.5212
33600000 0.5352
34400000 0.5408
35200000 0.5316
36000000 0.5356
36800000 0.5488
37600000 0.5304
38400000 0.5504
39200000 0.5548
40000000 0.5516
40800000 0.5592
41600000 0.5456
42400000 0.5628
43200000 0.5472
44000000 0.5508
44800000 0.5628
45600000 0.562
46400000 0.566
47200000 0.5768
48000000 0.5648
48800000 0.5644
49600000 0.5684
50400000 0.5688
51200000 0.5692
52000000 0.5616
52800000 0.5712
53600000 0.558
54400000 0.5552
55200000 0.5776
56000000 0.5872
56800000 0.572
57600000 0.5676
58400000 0.5624
59200000 0.5672
};

\path [draw=C5, fill=C5, opacity=0.2]
(axis cs:0,0)
--(axis cs:0,0)
--(axis cs:1600000,0.4576)
--(axis cs:3200000,0.656)
--(axis cs:4800000,0.758595)
--(axis cs:6400000,0.8152)
--(axis cs:8000000,0.859195)
--(axis cs:9600000,0.882795)
--(axis cs:11200000,0.9018)
--(axis cs:12800000,0.903595)
--(axis cs:14400000,0.904795)
--(axis cs:16000000,0.914595)
--(axis cs:17600000,0.9176)
--(axis cs:19200000,0.9166)
--(axis cs:20800000,0.926195)
--(axis cs:22400000,0.9134)
--(axis cs:24000000,0.9288)
--(axis cs:25600000,0.9226)
--(axis cs:27200000,0.9214)
--(axis cs:28800000,0.9198)
--(axis cs:30400000,0.9172)
--(axis cs:32000000,0.9228)
--(axis cs:33600000,0.931)
--(axis cs:35200000,0.9318)
--(axis cs:36800000,0.9262)
--(axis cs:38400000,0.9268)
--(axis cs:40000000,0.9296)
--(axis cs:41600000,0.921)
--(axis cs:43200000,0.9292)
--(axis cs:44800000,0.9194)
--(axis cs:46400000,0.9148)
--(axis cs:48000000,0.9162)
--(axis cs:49600000,0.9246)
--(axis cs:51200000,0.9142)
--(axis cs:52800000,0.9202)
--(axis cs:54400000,0.9272)
--(axis cs:56000000,0.919)
--(axis cs:57600000,0.9176)
--(axis cs:59200000,0.9154)
--(axis cs:59200000,0.9438)
--(axis cs:59200000,0.9438)
--(axis cs:57600000,0.9438)
--(axis cs:56000000,0.945)
--(axis cs:54400000,0.9536)
--(axis cs:52800000,0.9468)
--(axis cs:51200000,0.9422)
--(axis cs:49600000,0.9482)
--(axis cs:48000000,0.9424)
--(axis cs:46400000,0.942005)
--(axis cs:44800000,0.946205)
--(axis cs:43200000,0.9524)
--(axis cs:41600000,0.9484)
--(axis cs:40000000,0.9554)
--(axis cs:38400000,0.953)
--(axis cs:36800000,0.9522)
--(axis cs:35200000,0.9558)
--(axis cs:33600000,0.9564)
--(axis cs:32000000,0.9488)
--(axis cs:30400000,0.9442)
--(axis cs:28800000,0.9452)
--(axis cs:27200000,0.947)
--(axis cs:25600000,0.9496)
--(axis cs:24000000,0.9534)
--(axis cs:22400000,0.9428)
--(axis cs:20800000,0.952805)
--(axis cs:19200000,0.9434)
--(axis cs:17600000,0.9444)
--(axis cs:16000000,0.9432)
--(axis cs:14400000,0.935)
--(axis cs:12800000,0.9352)
--(axis cs:11200000,0.9336)
--(axis cs:9600000,0.9188)
--(axis cs:8000000,0.8974)
--(axis cs:6400000,0.864605)
--(axis cs:4800000,0.8092)
--(axis cs:3200000,0.7116)
--(axis cs:1600000,0.5096)
--(axis cs:0,0)
--cycle;

\addplot [thick, C5, mark=*, mark size=0, mark options={solid}]
table {%
0 0
1600000 0.4832
3200000 0.684
4800000 0.785
6400000 0.841
8000000 0.8798
9600000 0.9026
11200000 0.92
12800000 0.9208
14400000 0.9206
16000000 0.9296
17600000 0.9326
19200000 0.931
20800000 0.9412
22400000 0.9298
24000000 0.9424
25600000 0.9374
27200000 0.9354
28800000 0.9334
30400000 0.932
32000000 0.9376
33600000 0.9452
35200000 0.9446
36800000 0.9408
38400000 0.9414
40000000 0.944
41600000 0.936
43200000 0.9412
44800000 0.934
46400000 0.9294
48000000 0.9308
49600000 0.9374
51200000 0.9292
52800000 0.935
54400000 0.9422
56000000 0.9334
57600000 0.932
59200000 0.93
};

\path [draw=C6, fill=C6, opacity=0.2]
(axis cs:100000,0.00221814870811348)
--(axis cs:100000,0.000842039088538437)
--(axis cs:2600000,0.211607473091712)
--(axis cs:5100000,0.319925367367366)
--(axis cs:7600000,0.388123271718104)
--(axis cs:10100000,0.429948372525566)
--(axis cs:12600000,0.450150062561687)
--(axis cs:15100000,0.469822778882938)
--(axis cs:17600000,0.483341520119652)
--(axis cs:20100000,0.495969619947825)
--(axis cs:22600000,0.507339945029742)
--(axis cs:25100000,0.525843860992088)
--(axis cs:27600000,0.537927574831664)
--(axis cs:30100000,0.546515443758716)
--(axis cs:32600000,0.559916268827681)
--(axis cs:35100000,0.567719635844476)
--(axis cs:37600000,0.563721078783142)
--(axis cs:40100000,0.570790339660801)
--(axis cs:42600000,0.573598879082049)
--(axis cs:45100000,0.574740563799375)
--(axis cs:47600000,0.581788516506468)
--(axis cs:50100000,0.584106478729909)
--(axis cs:52600000,0.58611596178173)
--(axis cs:55100000,0.585054113550992)
--(axis cs:57600000,0.584214432180829)
--(axis cs:57600000,0.607081679056769)
--(axis cs:57600000,0.607081679056769)
--(axis cs:55100000,0.607624596568082)
--(axis cs:52600000,0.608317676664529)
--(axis cs:50100000,0.606135679823467)
--(axis cs:47600000,0.603415987859587)
--(axis cs:45100000,0.59681285686519)
--(axis cs:42600000,0.595698812594446)
--(axis cs:40100000,0.593134201564171)
--(axis cs:37600000,0.585851091720125)
--(axis cs:35100000,0.590205701000964)
--(axis cs:32600000,0.582269934371604)
--(axis cs:30100000,0.568488010435101)
--(axis cs:27600000,0.560543552868433)
--(axis cs:25100000,0.549043923750483)
--(axis cs:22600000,0.529111746764561)
--(axis cs:20100000,0.517409424204293)
--(axis cs:17600000,0.504087313412314)
--(axis cs:15100000,0.489437489148768)
--(axis cs:12600000,0.470574567298497)
--(axis cs:10100000,0.450710154689826)
--(axis cs:7600000,0.410793633370074)
--(axis cs:5100000,0.34224631849274)
--(axis cs:2600000,0.23321005603829)
--(axis cs:100000,0.00221814870811348)
--cycle;

\addplot [thick, C6, mark=*, mark size=0, mark options={solid}]
table {%
100000 0.00145543462547605
2600000 0.222455570196498
5100000 0.331088510349511
7600000 0.399507800552251
10100000 0.440349918451076
12600000 0.460337762398081
15100000 0.479585683702716
17600000 0.493683885265684
20100000 0.506552894273
22600000 0.518143526366409
25100000 0.53731539761593
27600000 0.549258339820717
30100000 0.557468963934169
32600000 0.571075628176812
35100000 0.578963050521695
37600000 0.574812209885522
40100000 0.581970579830386
42600000 0.584609877768193
45100000 0.585734311039245
47600000 0.592557929213373
50100000 0.595039151154246
52600000 0.597116251475264
55100000 0.596293587399407
57600000 0.595648767359268
};

\end{axis}

\end{tikzpicture}

%% file: tikz_plots/metaworld/metaworld_performance_profile_6e7.tex
\begin{tikzpicture}

\begin{axis}[
legend cell align={left},
legend style={
  fill opacity=0.8,
  draw opacity=1,
  text opacity=1,
  at={(0.03,0.03)},
  anchor=south west,
  draw=lightgray204
},
title={Meta-World - Performance Profile},
tick align=outside,
tick pos=left,
x grid style={darkgray176},
xlabel={Success Rate \(\displaystyle \tau\)},
xmajorgrids,
xmin=-0.0495, xmax=1.0395,
xtick style={color=black},
y grid style={darkgray176},
ylabel={Fraction of Runs with Success Rate \(\displaystyle > \tau\)},
ymajorgrids,
ymin=-0.05, ymax=1.05,
ytick style={color=black}
]
\path [draw=C1, fill=C1, opacity=0.15]
(axis cs:0,0.827)
--(axis cs:0,0.791)
--(axis cs:0.1,0.78)
--(axis cs:0.2,0.758)
--(axis cs:0.3,0.748)
--(axis cs:0.4,0.721)
--(axis cs:0.5,0.701975)
--(axis cs:0.6,0.645975)
--(axis cs:0.7,0.607)
--(axis cs:0.8,0.513)
--(axis cs:0.9,0.412)
--(axis cs:0.99,0.412)
--(axis cs:0.99,0.462)
--(axis cs:0.99,0.462)
--(axis cs:0.9,0.462)
--(axis cs:0.8,0.563)
--(axis cs:0.7,0.656)
--(axis cs:0.6,0.692)
--(axis cs:0.5,0.746)
--(axis cs:0.4,0.764)
--(axis cs:0.3,0.788)
--(axis cs:0.2,0.799)
--(axis cs:0.1,0.818)
--(axis cs:0,0.827)
--cycle;

\addplot [thick, C1]
table {%
0 0.809
0.1 0.799
0.2 0.779
0.3 0.768
0.4 0.743
0.5 0.724
0.6 0.669
0.7 0.632
0.8 0.539
0.9 0.437
0.99 0.437
};

\path [draw=C5, fill=C5, opacity=0.15]
(axis cs:0,0.876)
--(axis cs:0,0.845)
--(axis cs:0.1,0.836)
--(axis cs:0.2,0.816)
--(axis cs:0.3,0.798975)
--(axis cs:0.4,0.775)
--(axis cs:0.5,0.749)
--(axis cs:0.6,0.714975)
--(axis cs:0.7,0.681)
--(axis cs:0.8,0.639)
--(axis cs:0.9,0.533)
--(axis cs:0.99,0.533)
--(axis cs:0.99,0.579)
--(axis cs:0.99,0.579)
--(axis cs:0.9,0.579)
--(axis cs:0.8,0.681)
--(axis cs:0.7,0.721)
--(axis cs:0.6,0.754)
--(axis cs:0.5,0.789)
--(axis cs:0.4,0.814)
--(axis cs:0.3,0.836)
--(axis cs:0.2,0.852)
--(axis cs:0.1,0.869)
--(axis cs:0,0.876)
--cycle;

\addplot [thick, C5]
table {%
0 0.86
0.1 0.852
0.2 0.834
0.3 0.817
0.4 0.794
0.5 0.769
0.6 0.734
0.7 0.701
0.8 0.66
0.9 0.555
0.99 0.555
};

\path [draw=C2, fill=C2, opacity=0.15]
(axis cs:0,0.645789473684211)
--(axis cs:0,0.639183552631579)
--(axis cs:0.1,0.603210526315789)
--(axis cs:0.2,0.577420394736842)
--(axis cs:0.3,0.553368421052632)
--(axis cs:0.4,0.529105263157895)
--(axis cs:0.5,0.499315789473684)
--(axis cs:0.6,0.462710526315789)
--(axis cs:0.7,0.419341447368421)
--(axis cs:0.8,0.355289473684211)
--(axis cs:0.9,0.249051973684211)
--(axis cs:0.99,0.249051973684211)
--(axis cs:0.99,0.255605263157895)
--(axis cs:0.99,0.255605263157895)
--(axis cs:0.9,0.255605263157895)
--(axis cs:0.8,0.361947368421053)
--(axis cs:0.7,0.426105263157895)
--(axis cs:0.6,0.469605921052632)
--(axis cs:0.5,0.506289473684211)
--(axis cs:0.4,0.535869078947368)
--(axis cs:0.3,0.560132236842105)
--(axis cs:0.2,0.584078947368421)
--(axis cs:0.1,0.609921052631579)
--(axis cs:0,0.645789473684211)
--cycle;

\addplot [thick, C2]
table {%
0 0.642552631578947
0.1 0.606684210526316
0.2 0.580763157894737
0.3 0.556684210526316
0.4 0.532394736842105
0.5 0.502815789473684
0.6 0.466236842105263
0.7 0.422736842105263
0.8 0.3585
0.9 0.252236842105263
0.99 0.252236842105263
};


\path [draw=C4, fill=C4, opacity=0.15]
(axis cs:0,0.655025)
--(axis cs:0,0.618)
--(axis cs:0.1,0.618)
--(axis cs:0.2,0.569)
--(axis cs:0.3,0.569)
--(axis cs:0.4,0.488)
--(axis cs:0.5,0.488)
--(axis cs:0.6,0.373)
--(axis cs:0.7,0.373)
--(axis cs:0.8,0.334)
--(axis cs:0.9,0.334)
--(axis cs:0.99,0.334)
--(axis cs:0.99,0.377)
--(axis cs:0.99,0.377)
--(axis cs:0.9,0.377)
--(axis cs:0.8,0.377)
--(axis cs:0.7,0.417)
--(axis cs:0.6,0.417)
--(axis cs:0.5,0.53)
--(axis cs:0.4,0.53)
--(axis cs:0.3,0.609)
--(axis cs:0.2,0.609)
--(axis cs:0.1,0.655025)
--(axis cs:0,0.655025)
--cycle;

\addplot [thick, C4]
table {%
0 0.636
0.1 0.636
0.2 0.588
0.3 0.588
0.4 0.509
0.5 0.509
0.6 0.395
0.7 0.395
0.8 0.356
0.9 0.356
0.99 0.356
};

\path [draw=C0, fill=C0, opacity=0.15]
(axis cs:0,0.841)
--(axis cs:0,0.814)
--(axis cs:0.1,0.814)
--(axis cs:0.2,0.786)
--(axis cs:0.3,0.786)
--(axis cs:0.4,0.722)
--(axis cs:0.5,0.722)
--(axis cs:0.6,0.615)
--(axis cs:0.7,0.615)
--(axis cs:0.8,0.556)
--(axis cs:0.9,0.556)
--(axis cs:0.99,0.556)
--(axis cs:0.99,0.599)
--(axis cs:0.99,0.599)
--(axis cs:0.9,0.599)
--(axis cs:0.8,0.599)
--(axis cs:0.7,0.655)
--(axis cs:0.6,0.655)
--(axis cs:0.5,0.753025)
--(axis cs:0.4,0.753025)
--(axis cs:0.3,0.814)
--(axis cs:0.2,0.814)
--(axis cs:0.1,0.841)
--(axis cs:0,0.841)
--cycle;

\addplot [thick, C0]
table {%
0 0.828
0.1 0.828
0.2 0.8
0.3 0.8
0.4 0.738
0.5 0.738
0.6 0.635
0.7 0.635
0.8 0.578
0.9 0.578
0.99 0.578
};

\path [draw=C3, fill=C3, opacity=0.15]
(axis cs:0,0.597)
--(axis cs:0,0.572)
--(axis cs:0.1,0.572)
--(axis cs:0.2,0.558)
--(axis cs:0.3,0.558)
--(axis cs:0.4,0.521)
--(axis cs:0.5,0.521)
--(axis cs:0.6,0.486)
--(axis cs:0.7,0.486)
--(axis cs:0.8,0.468)
--(axis cs:0.9,0.468)
--(axis cs:0.99,0.468)
--(axis cs:0.99,0.491)
--(axis cs:0.99,0.491)
--(axis cs:0.9,0.491)
--(axis cs:0.8,0.491)
--(axis cs:0.7,0.512)
--(axis cs:0.6,0.512)
--(axis cs:0.5,0.547)
--(axis cs:0.4,0.547)
--(axis cs:0.3,0.584)
--(axis cs:0.2,0.584)
--(axis cs:0.1,0.597)
--(axis cs:0,0.597)
--cycle;

\addplot [thick, C3]
table {%
0 0.585
0.1 0.585
0.2 0.571
0.3 0.571
0.4 0.534
0.5 0.534
0.6 0.499
0.7 0.499
0.8 0.479
0.9 0.479
0.99 0.479
};

\path [draw=C6, fill=C6, opacity=0.15]
(axis cs:0,0.67352675)
--(axis cs:0,0.672193333333333)
--(axis cs:0.1,0.63139)
--(axis cs:0.2,0.597621666666667)
--(axis cs:0.3,0.567559875)
--(axis cs:0.4,0.541719916666667)
--(axis cs:0.5,0.519565)
--(axis cs:0.6,0.497454791666667)
--(axis cs:0.7,0.466836625)
--(axis cs:0.8,0.425369958333333)
--(axis cs:0.9,0.37008)
--(axis cs:0.99,0.281124791666667)
--(axis cs:0.99,0.282495)
--(axis cs:0.99,0.282495)
--(axis cs:0.9,0.371301791666667)
--(axis cs:0.8,0.426550041666667)
--(axis cs:0.7,0.468005041666667)
--(axis cs:0.6,0.49864)
--(axis cs:0.5,0.520693666666667)
--(axis cs:0.4,0.542915)
--(axis cs:0.3,0.568843583333333)
--(axis cs:0.2,0.598966666666667)
--(axis cs:0.1,0.632730333333333)
--(axis cs:0,0.67352675)
--cycle;

\addplot [thick, C6]
table {%
0 0.672861666666667
0.1 0.632068333333333
0.2 0.598303333333333
0.3 0.568206666666667
0.4 0.542328333333333
0.5 0.520141666666667
0.6 0.498045
0.7 0.467426666666667
0.8 0.425975
0.9 0.370693333333333
0.99 0.281811666666667
};
\end{axis}

\end{tikzpicture}

%% file: tikz_plots/metaworld/legend.tex
\begin{tikzpicture} 
    \begin{axis}[%
    hide axis,
    xmin=10,
    xmax=50,
    ymin=0,
    ymax=0.4,
    legend style={
        draw=white!15!black,
        legend cell align=left,
        legend columns=-1, 
        legend style={
            draw=none,
            column sep=1ex,
            line width=1pt
        }
    },
    ]
    \addlegendimage{C1}
    \addlegendentry{PPO};
    \addlegendimage{C5}
    \addlegendentry{TRPL};
    \addlegendimage{C2}
    \addlegendentry{NDP};
    \addlegendimage{C6}
    \addlegendentry{ES};
    \addlegendimage{C4}
    \addlegendentry{BBRL-PPO};
    \addlegendimage{C0}
    \addlegendentry{BBRL-TRPL};
    \addlegendimage{C3}
    \addlegendentry{BBRL-TRPL sparse};
    \end{axis}
\end{tikzpicture}

%% file: tikz_plots/beer_pong/bp_iqm_sample_efficiency.tex
\begin{tikzpicture}
\begin{axis}[
legend cell align={left},
legend style={fill opacity=0.8, draw opacity=1, text opacity=1, draw=lightgray204, 
  at={(0.03,0.97)},  
  anchor=north west
 },
title={Beer Pong},
tick align=outside,
tick pos=left,
x grid style={darkgray176},
xlabel={Number Environment Interactions},
xmajorgrids,
xmin=-7333790.47619048, xmax=155000000,
xtick style={color=black},
y grid style={darkgray176},
ylabel={Success Rate},
ymajorgrids,
xtick={0, 50000000, 100000000, 150000000},
ymin=-0.05, ymax=1.05,
ytick style={color=black}
]
\path [draw=C1, fill=C1, opacity=0.2]
(axis cs:78019.0476190476,0)
--(axis cs:78019.0476190476,0)
--(axis cs:3978971.42857143,0)
--(axis cs:7879923.80952381,0)
--(axis cs:11780876.1904762,0)
--(axis cs:15681828.5714286,0)
--(axis cs:19582780.952381,0.0025)
--(axis cs:23483733.3333333,0)
--(axis cs:27384685.7142857,0)
--(axis cs:31285638.0952381,0.00375)
--(axis cs:35186590.4761905,0.0025)
--(axis cs:39087542.8571429,0.00375)
--(axis cs:42988495.2380952,0.00625)
--(axis cs:46889447.6190476,0.01)
--(axis cs:50790400,0.015)
--(axis cs:54691352.3809524,0.015)
--(axis cs:58592304.7619048,0.01)
--(axis cs:62493257.1428571,0.0175)
--(axis cs:66394209.5238095,0.02)
--(axis cs:70295161.9047619,0.0175)
--(axis cs:74196114.2857143,0.02125)
--(axis cs:78097066.6666667,0.02125)
--(axis cs:81998019.047619,0.015)
--(axis cs:85898971.4285714,0.02)
--(axis cs:89799923.8095238,0.03375)
--(axis cs:93700876.1904762,0.0175)
--(axis cs:97601828.5714286,0.03)
--(axis cs:101502780.952381,0.03)
--(axis cs:105403733.333333,0.045)
--(axis cs:109304685.714286,0.0375)
--(axis cs:113205638.095238,0.03875)
--(axis cs:117106590.47619,0.045)
--(axis cs:121007542.857143,0.06375)
--(axis cs:124908495.238095,0.06125)
--(axis cs:128809447.619048,0.0675)
--(axis cs:132710400,0.05625)
--(axis cs:136611352.380952,0.07625)
--(axis cs:140512304.761905,0.07)
--(axis cs:144413257.142857,0.0625)
--(axis cs:148314209.52381,0.0925)
--(axis cs:148314209.52381,0.22625)
--(axis cs:148314209.52381,0.22625)
--(axis cs:144413257.142857,0.1575)
--(axis cs:140512304.761905,0.16375)
--(axis cs:136611352.380952,0.19125)
--(axis cs:132710400,0.14375)
--(axis cs:128809447.619048,0.18125)
--(axis cs:124908495.238095,0.17375)
--(axis cs:121007542.857143,0.14375)
--(axis cs:117106590.47619,0.1375)
--(axis cs:113205638.095238,0.11625)
--(axis cs:109304685.714286,0.11375)
--(axis cs:105403733.333333,0.1125)
--(axis cs:101502780.952381,0.1225)
--(axis cs:97601828.5714286,0.0925)
--(axis cs:93700876.1904762,0.09375)
--(axis cs:89799923.8095238,0.06375)
--(axis cs:85898971.4285714,0.0775)
--(axis cs:81998019.047619,0.06125)
--(axis cs:78097066.6666667,0.08)
--(axis cs:74196114.2857143,0.055)
--(axis cs:70295161.9047619,0.05)
--(axis cs:66394209.5238095,0.05875)
--(axis cs:62493257.1428571,0.05125)
--(axis cs:58592304.7619048,0.03)
--(axis cs:54691352.3809524,0.05)
--(axis cs:50790400,0.04125)
--(axis cs:46889447.6190476,0.03125)
--(axis cs:42988495.2380952,0.035)
--(axis cs:39087542.8571429,0.02625)
--(axis cs:35186590.4761905,0.01875)
--(axis cs:31285638.0952381,0.0225)
--(axis cs:27384685.7142857,0.0225)
--(axis cs:23483733.3333333,0.0125)
--(axis cs:19582780.952381,0.0175)
--(axis cs:15681828.5714286,0.01)
--(axis cs:11780876.1904762,0.00375)
--(axis cs:7879923.80952381,0)
--(axis cs:3978971.42857143,0)
--(axis cs:78019.0476190476,0)
--cycle;

\addplot [thick, C1, mark=*, mark size=0, mark options={solid}]
table {%
78019.0476190476 0
3978971.42857143 0
7879923.80952381 0
11780876.1904762 0
15681828.5714286 0.005
19582780.952381 0.0075
23483733.3333333 0.0025
27384685.7142857 0.00375
31285638.0952381 0.0125
35186590.4761905 0.00875
39087542.8571429 0.01125
42988495.2380952 0.01625
46889447.6190476 0.02
50790400 0.025
54691352.3809524 0.02625
58592304.7619048 0.02
62493257.1428571 0.0325
66394209.5238095 0.03625
70295161.9047619 0.03
74196114.2857143 0.035
78097066.6666667 0.04125
81998019.047619 0.03375
85898971.4285714 0.04
89799923.8095238 0.04625
93700876.1904762 0.045
97601828.5714286 0.0525
101502780.952381 0.05375
105403733.333333 0.075
109304685.714286 0.0725
113205638.095238 0.0675
117106590.47619 0.0825
121007542.857143 0.0925
124908495.238095 0.105
128809447.619048 0.115
132710400 0.095
136611352.380952 0.1175
140512304.761905 0.1075
144413257.142857 0.10375
148314209.52381 0.15
};
\path [draw=C4, fill=C4, opacity=0.2]
(axis cs:48000,0)
--(axis cs:48000,0)
--(axis cs:2448000,0)
--(axis cs:4848000,0)
--(axis cs:7248000,0)
--(axis cs:9648000,0)
--(axis cs:12048000,0)
--(axis cs:14448000,0)
--(axis cs:16848000,0)
--(axis cs:19248000,0)
--(axis cs:21648000,0)
--(axis cs:24048000,0)
--(axis cs:26448000,0.00666666666666667)
--(axis cs:28848000,0.00666666666666667)
--(axis cs:31248000,0.06)
--(axis cs:33648000,0.06)
--(axis cs:36048000,0.12)
--(axis cs:38448000,0.106666666666667)
--(axis cs:40848000,0.12)
--(axis cs:43248000,0.14)
--(axis cs:45648000,0.14)
--(axis cs:48048000,0.153333333333333)
--(axis cs:50448000,0.126666666666667)
--(axis cs:52848000,0.226666666666667)
--(axis cs:55248000,0.153333333333333)
--(axis cs:57648000,0.2)
--(axis cs:60048000,0.2)
--(axis cs:62448000,0.213333333333333)
--(axis cs:64848000,0.273333333333333)
--(axis cs:67248000,0.32)
--(axis cs:69648000,0.28)
--(axis cs:72048000,0.32)
--(axis cs:74448000,0.333333333333333)
--(axis cs:76848000,0.38)
--(axis cs:79248000,0.433333333333333)
--(axis cs:81648000,0.406666666666667)
--(axis cs:84048000,0.426666666666667)
--(axis cs:86448000,0.506666666666667)
--(axis cs:88848000,0.36)
--(axis cs:91248000,0.446666666666667)
--(axis cs:93648000,0.5)
--(axis cs:96048000,0.453333333333333)
--(axis cs:98448000,0.466666666666667)
--(axis cs:100848000,0.466666666666667)
--(axis cs:103248000,0.446666666666667)
--(axis cs:105648000,0.506666666666667)
--(axis cs:108048000,0.466666666666667)
--(axis cs:110448000,0.506666666666667)
--(axis cs:112848000,0.48)
--(axis cs:115248000,0.426666666666667)
--(axis cs:117648000,0.493333333333333)
--(axis cs:120048000,0.533333333333333)
--(axis cs:122448000,0.466666666666667)
--(axis cs:124848000,0.52)
--(axis cs:127248000,0.513333333333333)
--(axis cs:129648000,0.4)
--(axis cs:132048000,0.513333333333333)
--(axis cs:134448000,0.48)
--(axis cs:136848000,0.533333333333333)
--(axis cs:139248000,0.506666666666667)
--(axis cs:141648000,0.506666666666667)
--(axis cs:144048000,0.546666666666667)
--(axis cs:146448000,0.52)
--(axis cs:148848000,0.5)
--(axis cs:148848000,0.813333333333333)
--(axis cs:148848000,0.813333333333333)
--(axis cs:146448000,0.76)
--(axis cs:144048000,0.766666666666667)
--(axis cs:141648000,0.793333333333333)
--(axis cs:139248000,0.733333333333333)
--(axis cs:136848000,0.806666666666667)
--(axis cs:134448000,0.8)
--(axis cs:132048000,0.753333333333333)
--(axis cs:129648000,0.7)
--(axis cs:127248000,0.773333333333333)
--(axis cs:124848000,0.713333333333333)
--(axis cs:122448000,0.753333333333333)
--(axis cs:120048000,0.746666666666667)
--(axis cs:117648000,0.76)
--(axis cs:115248000,0.72)
--(axis cs:112848000,0.726666666666667)
--(axis cs:110448000,0.713333333333333)
--(axis cs:108048000,0.746666666666667)
--(axis cs:105648000,0.753333333333333)
--(axis cs:103248000,0.746666666666667)
--(axis cs:100848000,0.766666666666667)
--(axis cs:98448000,0.753333333333333)
--(axis cs:96048000,0.746666666666667)
--(axis cs:93648000,0.78)
--(axis cs:91248000,0.693333333333333)
--(axis cs:88848000,0.72)
--(axis cs:86448000,0.733333333333333)
--(axis cs:84048000,0.633333333333333)
--(axis cs:81648000,0.68)
--(axis cs:79248000,0.7)
--(axis cs:76848000,0.686666666666667)
--(axis cs:74448000,0.586666666666667)
--(axis cs:72048000,0.54)
--(axis cs:69648000,0.486666666666667)
--(axis cs:67248000,0.493333333333333)
--(axis cs:64848000,0.426666666666667)
--(axis cs:62448000,0.386666666666667)
--(axis cs:60048000,0.413333333333333)
--(axis cs:57648000,0.42)
--(axis cs:55248000,0.32)
--(axis cs:52848000,0.36)
--(axis cs:50448000,0.246666666666667)
--(axis cs:48048000,0.266666666666667)
--(axis cs:45648000,0.273333333333333)
--(axis cs:43248000,0.3)
--(axis cs:40848000,0.293333333333333)
--(axis cs:38448000,0.24)
--(axis cs:36048000,0.306666666666667)
--(axis cs:33648000,0.18)
--(axis cs:31248000,0.173333333333333)
--(axis cs:28848000,0.0933333333333333)
--(axis cs:26448000,0.06)
--(axis cs:24048000,0.02)
--(axis cs:21648000,0)
--(axis cs:19248000,0)
--(axis cs:16848000,0)
--(axis cs:14448000,0)
--(axis cs:12048000,0)
--(axis cs:9648000,0)
--(axis cs:7248000,0)
--(axis cs:4848000,0)
--(axis cs:2448000,0)
--(axis cs:48000,0)
--cycle;

\addplot [thick, C4, mark=*, mark size=0, mark options={solid}]
table {%
48000 0
2448000 0
4848000 0
7248000 0
9648000 0
12048000 0
14448000 0
16848000 0
19248000 0
21648000 0
24048000 0
26448000 0.0333333333333333
28848000 0.04
31248000 0.12
33648000 0.126666666666667
36048000 0.22
38448000 0.186666666666667
40848000 0.2
43248000 0.226666666666667
45648000 0.22
48048000 0.213333333333333
50448000 0.193333333333333
52848000 0.293333333333333
55248000 0.226666666666667
57648000 0.313333333333333
60048000 0.306666666666667
62448000 0.293333333333333
64848000 0.34
67248000 0.406666666666667
69648000 0.4
72048000 0.453333333333333
74448000 0.473333333333333
76848000 0.553333333333333
79248000 0.586666666666667
81648000 0.58
84048000 0.54
86448000 0.68
88848000 0.566666666666667
91248000 0.613333333333333
93648000 0.72
96048000 0.633333333333333
98448000 0.66
100848000 0.706666666666667
103248000 0.646666666666667
105648000 0.68
108048000 0.653333333333333
110448000 0.646666666666667
112848000 0.673333333333333
115248000 0.593333333333333
117648000 0.68
120048000 0.68
122448000 0.66
124848000 0.653333333333333
127248000 0.686666666666667
129648000 0.596666666666667
132048000 0.68
134448000 0.713333333333333
136848000 0.72
139248000 0.66
141648000 0.713333333333333
144048000 0.713333333333333
146448000 0.68
148848000 0.72
};
\path [draw=C0, fill=C0, opacity=0.2]
(axis cs:48000,0)
--(axis cs:48000,0)
--(axis cs:2448000,0)
--(axis cs:4848000,0)
--(axis cs:7248000,0)
--(axis cs:9648000,0)
--(axis cs:12048000,0)
--(axis cs:14448000,0)
--(axis cs:16848000,0)
--(axis cs:19248000,0)
--(axis cs:21648000,0)
--(axis cs:24048000,0)
--(axis cs:26448000,0)
--(axis cs:28848000,0)
--(axis cs:31248000,0)
--(axis cs:33648000,0)
--(axis cs:36048000,0)
--(axis cs:38448000,0)
--(axis cs:40848000,0)
--(axis cs:43248000,0)
--(axis cs:45648000,0)
--(axis cs:48048000,0.02)
--(axis cs:50448000,0.0266666666666667)
--(axis cs:52848000,0.106666666666667)
--(axis cs:55248000,0.166666666666667)
--(axis cs:57648000,0.226666666666667)
--(axis cs:60048000,0.18)
--(axis cs:62448000,0.22)
--(axis cs:64848000,0.206666666666667)
--(axis cs:67248000,0.233333333333333)
--(axis cs:69648000,0.2)
--(axis cs:72048000,0.193333333333333)
--(axis cs:74448000,0.266666666666667)
--(axis cs:76848000,0.206666666666667)
--(axis cs:79248000,0.253333333333333)
--(axis cs:81648000,0.273333333333333)
--(axis cs:84048000,0.273333333333333)
--(axis cs:86448000,0.266666666666667)
--(axis cs:88848000,0.3)
--(axis cs:91248000,0.373333333333333)
--(axis cs:93648000,0.4)
--(axis cs:96048000,0.4)
--(axis cs:98448000,0.466666666666667)
--(axis cs:100848000,0.526666666666667)
--(axis cs:103248000,0.533333333333333)
--(axis cs:105648000,0.52)
--(axis cs:108048000,0.56)
--(axis cs:110448000,0.56)
--(axis cs:112848000,0.68)
--(axis cs:115248000,0.56)
--(axis cs:117648000,0.6)
--(axis cs:120048000,0.613333333333333)
--(axis cs:122448000,0.64)
--(axis cs:124848000,0.62)
--(axis cs:127248000,0.64)
--(axis cs:129648000,0.613333333333333)
--(axis cs:132048000,0.68)
--(axis cs:134448000,0.653333333333333)
--(axis cs:136848000,0.666666666666667)
--(axis cs:139248000,0.62)
--(axis cs:141648000,0.66)
--(axis cs:144048000,0.64)
--(axis cs:146448000,0.633333333333333)
--(axis cs:148848000,0.66)
--(axis cs:148848000,0.8)
--(axis cs:148848000,0.8)
--(axis cs:146448000,0.746666666666667)
--(axis cs:144048000,0.74)
--(axis cs:141648000,0.78)
--(axis cs:139248000,0.733333333333333)
--(axis cs:136848000,0.8)
--(axis cs:134448000,0.793333333333333)
--(axis cs:132048000,0.786666666666667)
--(axis cs:129648000,0.733333333333333)
--(axis cs:127248000,0.773333333333333)
--(axis cs:124848000,0.726666666666667)
--(axis cs:122448000,0.746666666666667)
--(axis cs:120048000,0.766666666666667)
--(axis cs:117648000,0.706666666666667)
--(axis cs:115248000,0.686666666666667)
--(axis cs:112848000,0.793333333333333)
--(axis cs:110448000,0.72)
--(axis cs:108048000,0.7)
--(axis cs:105648000,0.666666666666667)
--(axis cs:103248000,0.633333333333333)
--(axis cs:100848000,0.68)
--(axis cs:98448000,0.633333333333333)
--(axis cs:96048000,0.606666666666667)
--(axis cs:93648000,0.553333333333333)
--(axis cs:91248000,0.5)
--(axis cs:88848000,0.413333333333333)
--(axis cs:86448000,0.453333333333333)
--(axis cs:84048000,0.433333333333333)
--(axis cs:81648000,0.34)
--(axis cs:79248000,0.413333333333333)
--(axis cs:76848000,0.373333333333333)
--(axis cs:74448000,0.366666666666667)
--(axis cs:72048000,0.34)
--(axis cs:69648000,0.333333333333333)
--(axis cs:67248000,0.36)
--(axis cs:64848000,0.326666666666667)
--(axis cs:62448000,0.293333333333333)
--(axis cs:60048000,0.3)
--(axis cs:57648000,0.333333333333333)
--(axis cs:55248000,0.36)
--(axis cs:52848000,0.213333333333333)
--(axis cs:50448000,0.12)
--(axis cs:48048000,0.08)
--(axis cs:45648000,0.0466666666666667)
--(axis cs:43248000,0.02)
--(axis cs:40848000,0.0266666666666667)
--(axis cs:38448000,0.0266666666666667)
--(axis cs:36048000,0.04)
--(axis cs:33648000,0.0533333333333333)
--(axis cs:31248000,0.00666666666666667)
--(axis cs:28848000,0)
--(axis cs:26448000,0)
--(axis cs:24048000,0)
--(axis cs:21648000,0)
--(axis cs:19248000,0)
--(axis cs:16848000,0)
--(axis cs:14448000,0)
--(axis cs:12048000,0)
--(axis cs:9648000,0)
--(axis cs:7248000,0)
--(axis cs:4848000,0)
--(axis cs:2448000,0)
--(axis cs:48000,0)
--cycle;

\addplot [thick, C0, mark=*, mark size=0, mark options={solid}]
table {%
48000 0
2448000 0
4848000 0
7248000 0
9648000 0
12048000 0
14448000 0
16848000 0
19248000 0
21648000 0
24048000 0
26448000 0
28848000 0
31248000 0
33648000 0.0266666666666667
36048000 0.0133333333333333
38448000 0
40848000 0
43248000 0
45648000 0.02
48048000 0.0466666666666667
50448000 0.0733333333333333
52848000 0.153333333333333
55248000 0.26
57648000 0.266666666666667
60048000 0.24
62448000 0.26
64848000 0.266666666666667
67248000 0.3
69648000 0.266666666666667
72048000 0.26
74448000 0.326666666666667
76848000 0.28
79248000 0.333333333333333
81648000 0.313333333333333
84048000 0.333333333333333
86448000 0.373333333333333
88848000 0.346666666666667
91248000 0.44
93648000 0.46
96048000 0.506666666666667
98448000 0.56
100848000 0.613333333333333
103248000 0.58
105648000 0.593333333333333
108048000 0.613333333333333
110448000 0.633333333333333
112848000 0.733333333333333
115248000 0.62
117648000 0.653333333333333
120048000 0.7
122448000 0.693333333333333
124848000 0.673333333333333
127248000 0.693333333333333
129648000 0.66
132048000 0.74
134448000 0.726666666666667
136848000 0.74
139248000 0.673333333333333
141648000 0.72
144048000 0.686666666666667
146448000 0.68
148848000 0.733333333333333
};
\path [draw=C7, fill=C7, opacity=0.2]
(axis cs:78000,0)
--(axis cs:78000,0)
--(axis cs:978000,0)
--(axis cs:1878000,0)
--(axis cs:2778000,0)
--(axis cs:3678000,0)
--(axis cs:4578000,0)
--(axis cs:5478000,0)
--(axis cs:6378000,0)
--(axis cs:7278000,0)
--(axis cs:8178000,0)
--(axis cs:9078000,0.01)
--(axis cs:9978000,0)
--(axis cs:10878000,0.01)
--(axis cs:11778000,0.03)
--(axis cs:12678000,0.01)
--(axis cs:13578000,0.06)
--(axis cs:14478000,0.02)
--(axis cs:15378000,0)
--(axis cs:16278000,0.02)
--(axis cs:17178000,0)
--(axis cs:18078000,0.05)
--(axis cs:18978000,0.01)
--(axis cs:19878000,0.05)
--(axis cs:20778000,0.05)
--(axis cs:21678000,0.03)
--(axis cs:22578000,0.07)
--(axis cs:23478000,0.08)
--(axis cs:24378000,0.05)
--(axis cs:25278000,0.03)
--(axis cs:26178000,0.03)
--(axis cs:27078000,0.06)
--(axis cs:27978000,0.07)
--(axis cs:28878000,0.06)
--(axis cs:29778000,0.07)
--(axis cs:30678000,0.08)
--(axis cs:31578000,0.09)
--(axis cs:32478000,0.03)
--(axis cs:33378000,0.04)
--(axis cs:34278000,0.11)
--(axis cs:35178000,0.03)
--(axis cs:36078000,0.04)
--(axis cs:36978000,0.08)
--(axis cs:37878000,0.09)
--(axis cs:38778000,0.12)
--(axis cs:39678000,0.05)
--(axis cs:40578000,0.05)
--(axis cs:41478000,0.04)
--(axis cs:42378000,0.05)
--(axis cs:43278000,0.08)
--(axis cs:44178000,0.06)
--(axis cs:45078000,0.05)
--(axis cs:45978000,0.09)
--(axis cs:46878000,0.09)
--(axis cs:47778000,0.06)
--(axis cs:48678000,0.06)
--(axis cs:49578000,0.02)
--(axis cs:50478000,0.08)
--(axis cs:51378000,0.04)
--(axis cs:52278000,0.07)
--(axis cs:53178000,0.04)
--(axis cs:54078000,0.09)
--(axis cs:54978000,0.13)
--(axis cs:55878000,0.06)
--(axis cs:56778000,0.12)
--(axis cs:57678000,0.09)
--(axis cs:58578000,0.1)
--(axis cs:59478000,0.1)
--(axis cs:60378000,0.07)
--(axis cs:61278000,0.15)
--(axis cs:62178000,0.1)
--(axis cs:63078000,0.18)
--(axis cs:63978000,0.1)
--(axis cs:64878000,0.08)
--(axis cs:65778000,0.11)
--(axis cs:66678000,0.12)
--(axis cs:67578000,0.13)
--(axis cs:68478000,0.07)
--(axis cs:69378000,0.09)
--(axis cs:70278000,0.09)
--(axis cs:71178000,0.07)
--(axis cs:72078000,0.09)
--(axis cs:72978000,0.09)
--(axis cs:73878000,0.04)
--(axis cs:74778000,0.09)
--(axis cs:75678000,0.07)
--(axis cs:76578000,0.15)
--(axis cs:77478000,0.09)
--(axis cs:78378000,0.12)
--(axis cs:79278000,0.04)
--(axis cs:80178000,0.13)
--(axis cs:81078000,0.11)
--(axis cs:81978000,0.09)
--(axis cs:82878000,0.12)
--(axis cs:83778000,0.08)
--(axis cs:84678000,0.11)
--(axis cs:85578000,0.12)
--(axis cs:86478000,0.06)
--(axis cs:87378000,0.14)
--(axis cs:88278000,0.13)
--(axis cs:89178000,0.14)
--(axis cs:90078000,0.09)
--(axis cs:90978000,0.15)
--(axis cs:91878000,0.1)
--(axis cs:92778000,0.13)
--(axis cs:93678000,0.09)
--(axis cs:94578000,0.08)
--(axis cs:95478000,0.12)
--(axis cs:96378000,0.1)
--(axis cs:97278000,0.11)
--(axis cs:98178000,0.11)
--(axis cs:99078000,0.09)
--(axis cs:99978000,0.09)
--(axis cs:100878000,0.16)
--(axis cs:101778000,0.17)
--(axis cs:102678000,0.13)
--(axis cs:103578000,0.05)
--(axis cs:104478000,0.1)
--(axis cs:105378000,0.1)
--(axis cs:106278000,0.11)
--(axis cs:107178000,0.08)
--(axis cs:108078000,0.1)
--(axis cs:108978000,0.06)
--(axis cs:109878000,0.1)
--(axis cs:110778000,0.12)
--(axis cs:111678000,0.09)
--(axis cs:112578000,0.1)
--(axis cs:113478000,0.11)
--(axis cs:114378000,0.13)
--(axis cs:115278000,0.11)
--(axis cs:116178000,0.12)
--(axis cs:117078000,0.14)
--(axis cs:117978000,0.13)
--(axis cs:118878000,0.08)
--(axis cs:119778000,0.16)
--(axis cs:120678000,0.14)
--(axis cs:121578000,0.09)
--(axis cs:122478000,0.1)
--(axis cs:123378000,0.14)
--(axis cs:124278000,0.12)
--(axis cs:125178000,0.2)
--(axis cs:126078000,0.06)
--(axis cs:126978000,0.11)
--(axis cs:127878000,0.07)
--(axis cs:128778000,0.13)
--(axis cs:129678000,0.13)
--(axis cs:130578000,0.13)
--(axis cs:131478000,0.09)
--(axis cs:132378000,0.06)
--(axis cs:133278000,0.15)
--(axis cs:134178000,0.12)
--(axis cs:135078000,0.1)
--(axis cs:135978000,0.12)
--(axis cs:136878000,0.15)
--(axis cs:137778000,0.1)
--(axis cs:138678000,0.05)
--(axis cs:139578000,0.11)
--(axis cs:140478000,0.15)
--(axis cs:141378000,0.16)
--(axis cs:142278000,0.12)
--(axis cs:143178000,0.12)
--(axis cs:144078000,0.1)
--(axis cs:144978000,0.12)
--(axis cs:145878000,0.14)
--(axis cs:146778000,0.16)
--(axis cs:147678000,0.16)
--(axis cs:148578000,0.16)
--(axis cs:149478000,0.14)
--(axis cs:149478000,0.27)
--(axis cs:149478000,0.27)
--(axis cs:148578000,0.3)
--(axis cs:147678000,0.32)
--(axis cs:146778000,0.34)
--(axis cs:145878000,0.33)
--(axis cs:144978000,0.27)
--(axis cs:144078000,0.29)
--(axis cs:143178000,0.3)
--(axis cs:142278000,0.32)
--(axis cs:141378000,0.3)
--(axis cs:140478000,0.33)
--(axis cs:139578000,0.29)
--(axis cs:138678000,0.26)
--(axis cs:137778000,0.28)
--(axis cs:136878000,0.3)
--(axis cs:135978000,0.39)
--(axis cs:135078000,0.25)
--(axis cs:134178000,0.28)
--(axis cs:133278000,0.31)
--(axis cs:132378000,0.22)
--(axis cs:131478000,0.32)
--(axis cs:130578000,0.27)
--(axis cs:129678000,0.35)
--(axis cs:128778000,0.3)
--(axis cs:127878000,0.23)
--(axis cs:126978000,0.33)
--(axis cs:126078000,0.23)
--(axis cs:125178000,0.32)
--(axis cs:124278000,0.26)
--(axis cs:123378000,0.37)
--(axis cs:122478000,0.31)
--(axis cs:121578000,0.25)
--(axis cs:120678000,0.31)
--(axis cs:119778000,0.33)
--(axis cs:118878000,0.35)
--(axis cs:117978000,0.29)
--(axis cs:117078000,0.3)
--(axis cs:116178000,0.38)
--(axis cs:115278000,0.27)
--(axis cs:114378000,0.27)
--(axis cs:113478000,0.31)
--(axis cs:112578000,0.32)
--(axis cs:111678000,0.23)
--(axis cs:110778000,0.32)
--(axis cs:109878000,0.23)
--(axis cs:108978000,0.22)
--(axis cs:108078000,0.26)
--(axis cs:107178000,0.31)
--(axis cs:106278000,0.23)
--(axis cs:105378000,0.23)
--(axis cs:104478000,0.28)
--(axis cs:103578000,0.2)
--(axis cs:102678000,0.24)
--(axis cs:101778000,0.3)
--(axis cs:100878000,0.31)
--(axis cs:99978000,0.27)
--(axis cs:99078000,0.27)
--(axis cs:98178000,0.24)
--(axis cs:97278000,0.31)
--(axis cs:96378000,0.33)
--(axis cs:95478000,0.29)
--(axis cs:94578000,0.24)
--(axis cs:93678000,0.22)
--(axis cs:92778000,0.23)
--(axis cs:91878000,0.26)
--(axis cs:90978000,0.33)
--(axis cs:90078000,0.24)
--(axis cs:89178000,0.29)
--(axis cs:88278000,0.31)
--(axis cs:87378000,0.3)
--(axis cs:86478000,0.22)
--(axis cs:85578000,0.31)
--(axis cs:84678000,0.24)
--(axis cs:83778000,0.25)
--(axis cs:82878000,0.3)
--(axis cs:81978000,0.25)
--(axis cs:81078000,0.27)
--(axis cs:80178000,0.3)
--(axis cs:79278000,0.25)
--(axis cs:78378000,0.25)
--(axis cs:77478000,0.25)
--(axis cs:76578000,0.29)
--(axis cs:75678000,0.2)
--(axis cs:74778000,0.28)
--(axis cs:73878000,0.21)
--(axis cs:72978000,0.24)
--(axis cs:72078000,0.27)
--(axis cs:71178000,0.24)
--(axis cs:70278000,0.26)
--(axis cs:69378000,0.27)
--(axis cs:68478000,0.27)
--(axis cs:67578000,0.25)
--(axis cs:66678000,0.25)
--(axis cs:65778000,0.23)
--(axis cs:64878000,0.24)
--(axis cs:63978000,0.24)
--(axis cs:63078000,0.31)
--(axis cs:62178000,0.25)
--(axis cs:61278000,0.25)
--(axis cs:60378000,0.18)
--(axis cs:59478000,0.18)
--(axis cs:58578000,0.26)
--(axis cs:57678000,0.25)
--(axis cs:56778000,0.26)
--(axis cs:55878000,0.16)
--(axis cs:54978000,0.21)
--(axis cs:54078000,0.2)
--(axis cs:53178000,0.17)
--(axis cs:52278000,0.22)
--(axis cs:51378000,0.22)
--(axis cs:50478000,0.2)
--(axis cs:49578000,0.21)
--(axis cs:48678000,0.16)
--(axis cs:47778000,0.2)
--(axis cs:46878000,0.23)
--(axis cs:45978000,0.21)
--(axis cs:45078000,0.19)
--(axis cs:44178000,0.23)
--(axis cs:43278000,0.21)
--(axis cs:42378000,0.27)
--(axis cs:41478000,0.2)
--(axis cs:40578000,0.21)
--(axis cs:39678000,0.19)
--(axis cs:38778000,0.28)
--(axis cs:37878000,0.23)
--(axis cs:36978000,0.21)
--(axis cs:36078000,0.14)
--(axis cs:35178000,0.21)
--(axis cs:34278000,0.25)
--(axis cs:33378000,0.19)
--(axis cs:32478000,0.18)
--(axis cs:31578000,0.21)
--(axis cs:30678000,0.22)
--(axis cs:29778000,0.25)
--(axis cs:28878000,0.18)
--(axis cs:27978000,0.22)
--(axis cs:27078000,0.2)
--(axis cs:26178000,0.16)
--(axis cs:25278000,0.12)
--(axis cs:24378000,0.17)
--(axis cs:23478000,0.23)
--(axis cs:22578000,0.17)
--(axis cs:21678000,0.24)
--(axis cs:20778000,0.14)
--(axis cs:19878000,0.15)
--(axis cs:18978000,0.1)
--(axis cs:18078000,0.17)
--(axis cs:17178000,0.15)
--(axis cs:16278000,0.16)
--(axis cs:15378000,0.13)
--(axis cs:14478000,0.1)
--(axis cs:13578000,0.14)
--(axis cs:12678000,0.09)
--(axis cs:11778000,0.13)
--(axis cs:10878000,0.11)
--(axis cs:9978000,0.14)
--(axis cs:9078000,0.13)
--(axis cs:8178000,0.06)
--(axis cs:7278000,0.07)
--(axis cs:6378000,0.08)
--(axis cs:5478000,0.07)
--(axis cs:4578000,0.04)
--(axis cs:3678000,0.01)
--(axis cs:2778000,0)
--(axis cs:1878000,0)
--(axis cs:978000,0)
--(axis cs:78000,0)
--cycle;

\addplot [thick, C7, mark=*, mark size=0, mark options={solid}]
table {%
78000 0
978000 0
1878000 0
2778000 0
3678000 0
4578000 0
5478000 0.03
6378000 0.04
7278000 0.02
8178000 0.02
9078000 0.05
9978000 0.06
10878000 0.05
11778000 0.07
12678000 0.05
13578000 0.1
14478000 0.06
15378000 0.05
16278000 0.08
17178000 0.05
18078000 0.11
18978000 0.05
19878000 0.1
20778000 0.09
21678000 0.13
22578000 0.13
23478000 0.15
24378000 0.12
25278000 0.07
26178000 0.09
27078000 0.13
27978000 0.14
28878000 0.13
29778000 0.16
30678000 0.14
31578000 0.15
32478000 0.08
33378000 0.12
34278000 0.2
35178000 0.11
36078000 0.08
36978000 0.15
37878000 0.17
38778000 0.2
39678000 0.12
40578000 0.12
41478000 0.1
42378000 0.14
43278000 0.13
44178000 0.14
45078000 0.12
45978000 0.17
46878000 0.15
47778000 0.14
48678000 0.12
49578000 0.09
50478000 0.16
51378000 0.12
52278000 0.14
53178000 0.08
54078000 0.15
54978000 0.17
55878000 0.11
56778000 0.19
57678000 0.18
58578000 0.18
59478000 0.14
60378000 0.13
61278000 0.21
62178000 0.17
63078000 0.25
63978000 0.17
64878000 0.16
65778000 0.17
66678000 0.19
67578000 0.19
68478000 0.17
69378000 0.17
70278000 0.16
71178000 0.14
72078000 0.18
72978000 0.17
73878000 0.11
74778000 0.2
75678000 0.13
76578000 0.23
77478000 0.18
78378000 0.19
79278000 0.13
80178000 0.24
81078000 0.19
81978000 0.17
82878000 0.2
83778000 0.15
84678000 0.19
85578000 0.23
86478000 0.13
87378000 0.25
88278000 0.24
89178000 0.21
90078000 0.17
90978000 0.25
91878000 0.16
92778000 0.19
93678000 0.15
94578000 0.15
95478000 0.21
96378000 0.22
97278000 0.22
98178000 0.19
99078000 0.17
99978000 0.18
100878000 0.24
101778000 0.25
102678000 0.19
103578000 0.13
104478000 0.16
105378000 0.16
106278000 0.17
107178000 0.19
108078000 0.18
108978000 0.14
109878000 0.15
110778000 0.23
111678000 0.15
112578000 0.21
113478000 0.21
114378000 0.19
115278000 0.2
116178000 0.25
117078000 0.23
117978000 0.23
118878000 0.22
119778000 0.25
120678000 0.24
121578000 0.15
122478000 0.2
123378000 0.26
124278000 0.2
125178000 0.27
126078000 0.14
126978000 0.23
127878000 0.15
128778000 0.23
129678000 0.23
130578000 0.19
131478000 0.2
132378000 0.13
133278000 0.23
134178000 0.21
135078000 0.14
135978000 0.25
136878000 0.25
137778000 0.19
138678000 0.14
139578000 0.2
140478000 0.23
141378000 0.23
142278000 0.21
143178000 0.21
144078000 0.21
144978000 0.2
145878000 0.24
146778000 0.25
147678000 0.24
148578000 0.24
149478000 0.21
};
\end{axis}

\end{tikzpicture}

%% file: tikz_plots/hopper_jump/legend.tex
\begin{tikzpicture} 
    \begin{axis}[%
    hide axis,
    xmin=10,
    xmax=50,
    ymin=0,
    ymax=0.4,
    legend style={
        draw=white!15!black,
        legend cell align=left,
        legend columns=-1, 
        legend style={
            draw=none,
            column sep=1ex,
            line width=1pt
        }
    },
    ]
    \addlegendimage{C1}
    \addlegendentry{PPO};
    \addlegendimage{C5}
    \addlegendentry{TRPL};
    \addlegendimage{C6}
    \addlegendentry{ES};
    \addlegendimage{C7}
    \addlegendentry{CMORE};
    \addlegendimage{C8}
    \addlegendentry{SAC};
    \addlegendimage{C4}
    \addlegendentry{BBRL-PPO};
    \addlegendimage{C0}
    \addlegendentry{BBRL-TRPL};
    \end{axis}
\end{tikzpicture}

%% file: appendix/kl_trpl.tex
As already mentioned in the main text, TRPLs \cite{Otto2021} present a scalable and mathematically sound approach for enforcing trust regions in step-based deep RL.
The layer takes the output of a standard Gaussian policy as input in terms of mean $\bm{\mu}$ and variance $\bm{\Sigma}$ and projects it into the trust region if the given mean and variance violate their respective bounds. 
This projection is done for each input state individually. 
Subsequently, the projected Gaussian policy distribution with parameters $\tilde{\bm{\mu}}$, $\tilde{\bm{\Sigma}}$ is used for any further steps, e.\,g. for sampling and/or loss computation.
Formally, the layer solves the following two optimization problems for each state $\bm{s}$
    \begin{align}
        \argmin_{\til{\bm{\mu}}_s} d_\textrm{mean} \left(\til{\bm{\mu}}_s, \bm{\mu}(s) \right),  \quad &\st \quad d_\textrm{mean} \left(\til{\bm{\mu}}_s,  \old{\bm{\mu}}(s) \right) \leq \epsilon_{\bm{\mu}}, \quad \textrm{and} 
        \label{eq:apx:generic_mean}
        \\ 
        \argmin_{\til{\bm{\Sigma}}_s} d_\textrm{cov} \left(\til{\bm{\Sigma}}_s, \bm{\Sigma}(\bm{s}) \right), \quad &\st \quad d_\textrm{cov} \left(\til{\bm{\Sigma}}_s, \old{\bm{\Sigma}}(\bm{s}) \right) \leq \epsilon_\Sigma,`
        \label{eq:apx:generic_cov}
    \end{align}
where $\tilde{\bm{\mu}}_s$ and $\tilde{ \bm{\Sigma}}_s$ are the optimization variables for input state $\bm{s}$ and $\epsilon_\mu$ and $\epsilon_\Sigma$ are the trust region bounds for mean and covariance, respectively.
Finally, $\old{\mu}$ and $\old{\Sigma}$ are the reference mean and covariance for the trust region and $d_\textrm{mean}$ as well as $d_\textrm{cov}$ are the similarity metrics for the mean and covariance of a decomposable distance or divergence measure. 
As we only leverage the KL-divergence projection, we will provide only details for this particular projection below. 
For the other two projections we refer the reader to \citet{Otto2021}.

Inserting the mean part of the Gaussian KL divergence into \Eqref{eq:apx:generic_mean} yields
\begin{equation*}
    	\argmin_{\til{\mu}} \vecT{\left(\mu - \til{\mu}\right)} \oldInv{\Sigma} \left(\mu - \til{\mu}\right) 
    	\quad \st \quad \vecT{\left(\old{\mu} - \til{\mu}\right)} \oldInv{\Sigma} \left(\old{\mu} - \til{\mu}\right) \leq \epsilon_\mu.
\end{equation*}

After differentiating the dual w.r.t. $\til{\mu}$, we can solve for the projected mean 
\begin{align*}
\til{\mu} &= \frac{\mu + \omega  \old{\mu}}{1 + \omega} \quad \text{with} \quad \omega = \sqrt{\frac{\vecT{\left(\old{\mu} - \mu\right)} \oldInv{\Sigma} \left(\old{\mu} - \mu\right)}{\epsilon_\mu}} - 1,  
\end{align*}
leveraging the optimal Lagrange multiplier $\omega$.
Similarly, we can insert the covariance part of the Gaussian KL divergence into \Eqref{eq:apx:generic_cov}, which results in
\begin{align*}
	\argmin_{\til{\Sigma}}  \textrm{tr}\left(\inv{\Sigma}\til{\Sigma}\right) + \log \frac{|\Sigma|}{|\til{\Sigma}|}, \quad
	\st \quad \textrm{tr}\left(\oldInv{\Sigma} \til{\Sigma} \right) - d + \log\frac{|\old{\Sigma}|}{|\til{\Sigma}|} \leq \epsilon_\Sigma,
\end{align*}
where $d$ is the number of degrees of freedom (DoF).
Once again, differentiating and solving the dual $g(\eta)$ for the projected covariance yields
\begin{align*}
    \tilde{\Sigma} = \left( \dfrac{\eta^* \oldInv{\Sigma} + \Sigma^{-1}}{\eta^* + 1 } \right)^{-1} \quad \text{with} \quad \eta^* = \argmin_{\eta} g(\eta), \; \st \; \eta \ge 0.
    \label{eq:kl_proj_cov}
\end{align*}
Here, the the optimal Lagrange multiplier $\eta^*$ cannot be computed in closed form, however, a standard numerical optimizer, such as BFGS, is able to efficiently find it. 
This can be made differentiable by taking the differentials of the KKT conditions of the dual. 
For more details, we refer to the original work \citep{Otto2021}. 

%% file: appendix/environments.tex
\begin{figure}
    \centering
    \includegraphics[width=0.25\textwidth]{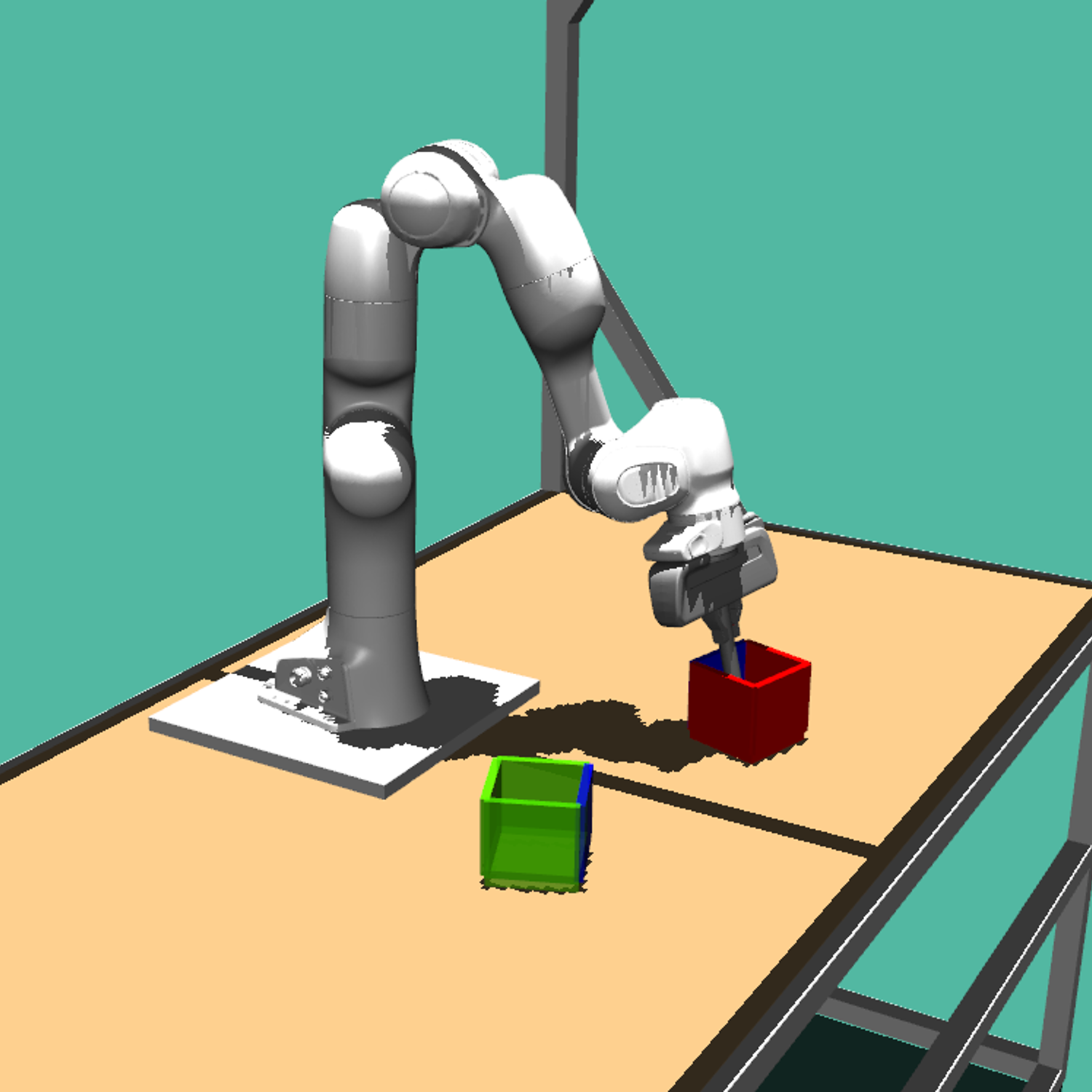}%
    \includegraphics[width=0.25\textwidth]{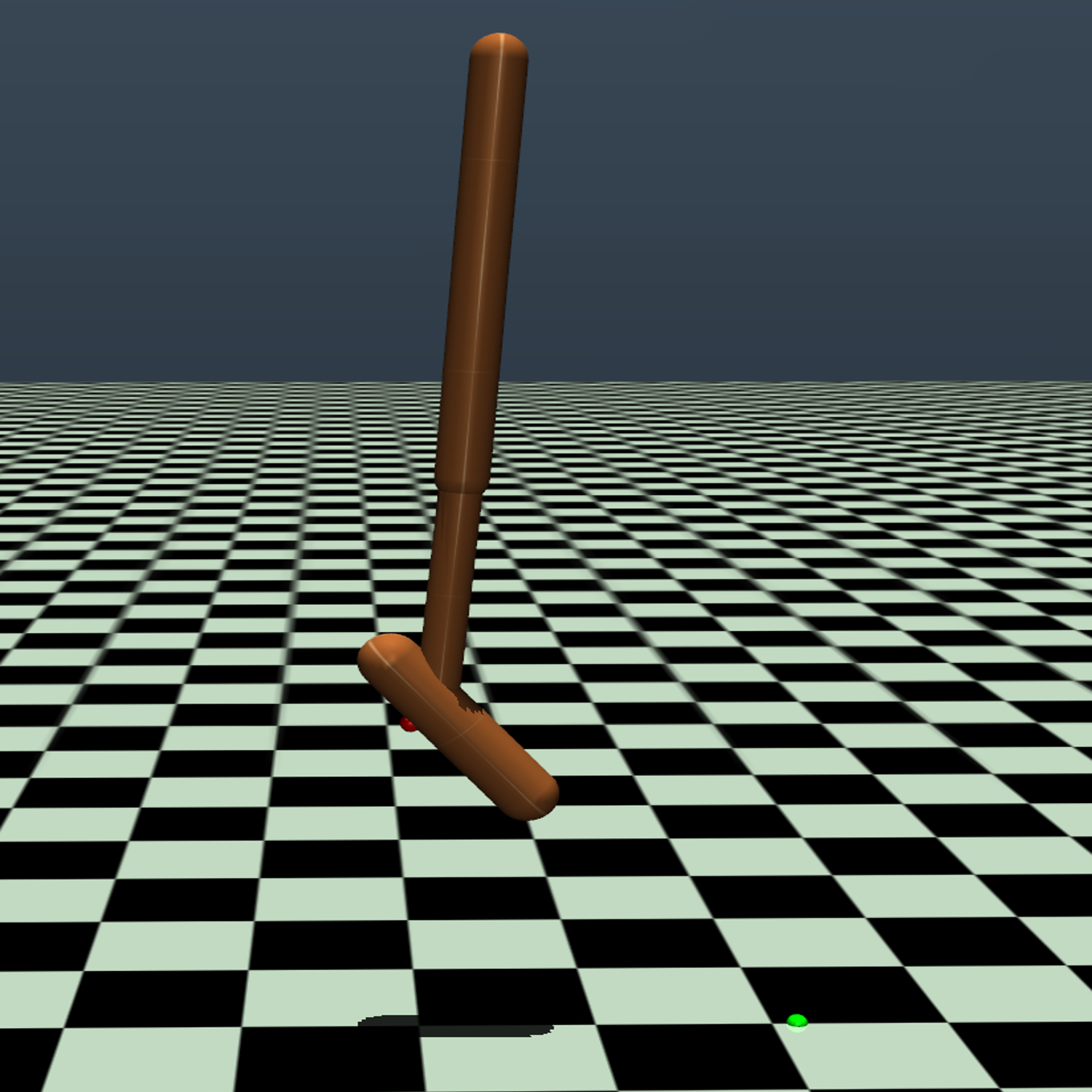}%
    \includegraphics[width=0.25\textwidth]{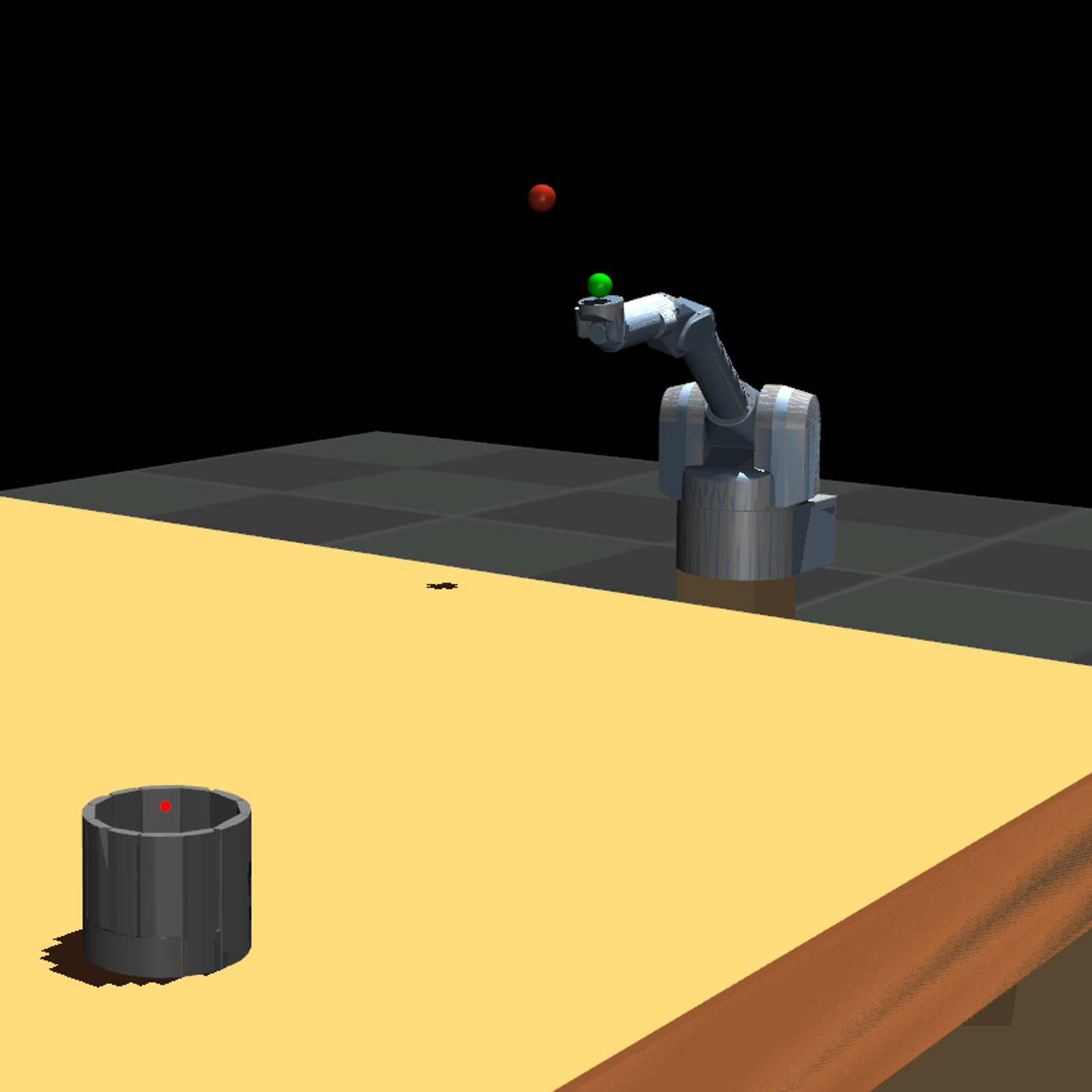}%
    \includegraphics[width=0.25\textwidth]{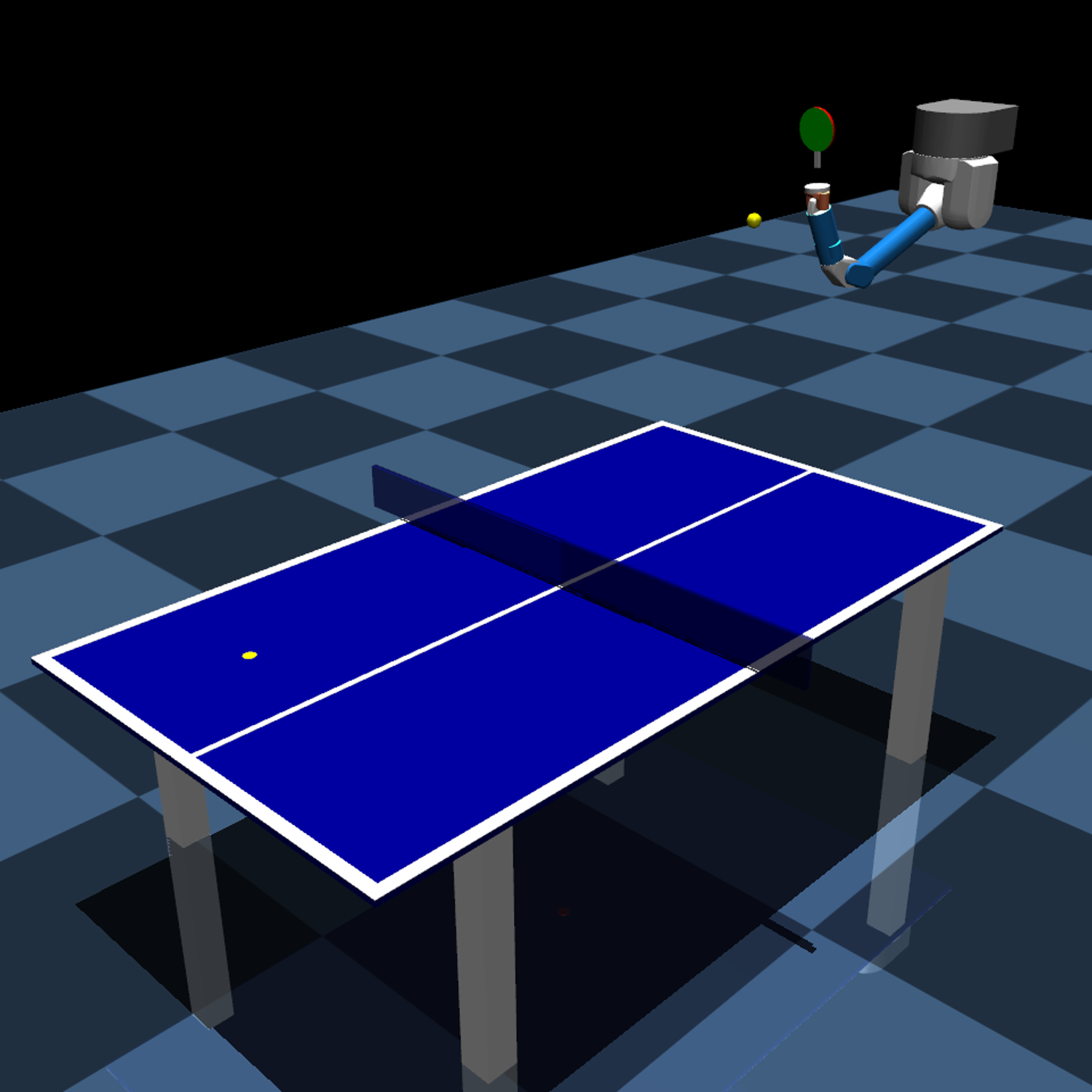}%
    \caption{Visualization of the four control tasks box pushing, hopper jumping, beer pong, and table tennis.}
    \label{fig:environments}
\end{figure}

{\color{black}
\subsection{Reacher5d}
\label{app:reacher}
For the Reacher task we modify the original OpenAI gym Reacher-v2 by adding three additional joints, resulting in a total of five joints. 
The task goal is still to minimize the distance between the goal point $\mathbf{p}_{goal}$ and the end-effector $\mathbf{p}$.
We, however, only sample the goal point for $y\geq0$, i.\,e. in the first two quadrants, to slightly reduce task complexity while maintaining the increased control complexity.
The observation space remains unchanged, unless for the sparse reward where we additionally add the current step value to make learning possible for step-based methods.
The context space only contains the coordinates of the goal position. 
The action space is the 5d equivalent to the original version.

For the reward the original setting leverages the goal distance
\[
 R_\text{goal} = \lVert \mathbf{p}-\mathbf{p}_{goal}\rVert_2
\]
and the action cost
\begin{align*}
    \tau_t = \sum_i^K (a^i_t)^2,
\end{align*}

\textbf{Dense Reward.} The dense reward in the 5d setting, hence, stays the same and the agent receives in each time step $t$
\begin{equation*}
    R_\text{tot} = - \tau_t - R_{\text{goal}}
\end{equation*}
\textbf{Sparse Reward.} The sparse reward only returns the task reward in the last time step $T$ and additionally adds a velocity penalty $R_\text{vel} = \sum_i^K (\dot{q}_T^i)^2$, where $\mathbf{\dot{q}}$ are the joint velocities, to avoid overshooting
\begin{equation*}
    R_\text{tot} = \begin{cases}
        - \tau_t 
        &t<T, 
        \\
        - \tau_t 
        -200 R_\text{goal}
        -10 R_\text{vel}
        &t=T.
        \end{cases}
\end{equation*}

}

\subsection{Box Pushing}
\label{app:box_pushing}
The goal of the box pushing task is to move a box to a specified goal location and orientation using the seven DoF Franka Emika Panda.
Hence, the context space for this task is the goal position $x \in [0.3, 0.6]$, $y \in [-0.45, 0.45]$ and the goal orientation $\theta \in [0, 2\pi]$. 
{\color{black}In addition to the contexts, the observation space for the step-based algorithms contains the sine and cosine of the joint angles and their velocities as well as position and orientation quaternions for both end-effector and the box. 
For the action space we use the torques per joint and additionally add gravity compensation in each time step, that does not have to be learnt.}
The task is considered successfully solved if the position distance $\leq 0.05$m and  the orientation error $\leq 0.5$rad.
For the total reward we consider different sub-rewards. 
First, the distance to the goal
\[
 R_\text{goal} = \lVert\mathbf{p}-\mathbf{p}_{goal}\rVert,
\]
where $\mathbf{p}$ is the box position and $\mathbf{p}_{goal}$ the goal position itself.
Second, the rotation distance
\[
R_\text{rotation} =\frac{1}{\pi} \arccos{|\mathbf{r}\cdot\mathbf{r}_{goal}|},
\]
where $\mathbf{r}$ and $\mathbf{r}_{goal}$ are the box orientation and goal orientation in quaternion, respectively.
Third, an incentive to keep the rod within the box
\[
R_\text{rod} = \text{clip}(||\mathbf{p}-\mathbf{h}_{pos}||, 0.05, 10)
\]
where $\mathbf{h}_{pos}$ is the position of the rod tip.
Fourth, a similar incentive that encourages to maintain the rod in a desired rotation
\[
R_\text{rod\_rotation} = \text{clip} (\frac{2}{\pi} \arccos{|\mathbf{h}_{rot}\cdot\mathbf{h}_0|} , 0.25, 2),
\]
where $\mathbf{h}_{rot}$ and $\mathbf{h}_{0}=(0.0, 1.0, 0.0, 0.0)$ are the current and desired rod orientation in quaternion, respectively.
And lastly, we utilize the following error
\[\text{err}(\mathbf{q}, \mathbf{\dot{q}}) = \sum_{i \in \{i | |q_i| > |q_i^{b}|\}}{(|q_i|-|q^{b}_i|)}
+ \sum_{j \in \{j | |\dot{q}_j| >|\dot{q}_j^{b}|\}}{(|\dot{q}_j|-|\dot{q}^{b}_j|)}.
\]
Here, $\mathbf{q}$, $\mathbf{\dot{q}}$, $\mathbf{q}^{b}$, and $\mathbf{\dot{q}}^{b}$ are the robot joint's position and velocity as well as their respective bounds.
Additionally, we consider an action cost in each time step $t$
\begin{align*}
    \tau_t = 5\cdot10^{-4}\sum_i^K (a^i_t)^2,
\end{align*}
where $K=7$ is the number of DoF.
In total we consider three different rewards. 

\textbf{Dense Reward.} The dense reward provides information about the goal and rotation distance in each time step $t$ on top of the utility rewards
\begin{equation*}
    R_\text{tot} = 
    - R_{\text{rod}}
    - R_{\text{rod\_rotation}}
    - \tau_t 
    - \text{err}(\bm{q}, \bm{\dot{q}})
    -3.5 R_{\text{goal}}
    - 2 R_{\text{rotation}}.
\end{equation*}
\textbf{Time-Dependent Sparse Reward.} The time-dependent sparse reward is similar to the dense reward, but only returns the goal and rotation distance in the last time step $T$
\begin{equation*}
    R_\text{tot} = \begin{cases}
        - R_\text{rod}
        - R_\text{rod\_rotation}
        - \tau_t 
        - \text{err}(\mathbf{q}, \mathbf{\dot{q}}),
        &t<T, 
        \\
        - R_\text{rod}
        - R_\text{rod\_rotation}
        - \tau_t 
        - \text{err}(\mathbf{q}, \mathbf{\dot{q}})
        -350 R_\text{goal}
        -200R_\text{rotation},
        &t=T.
        \end{cases}
\end{equation*}
\textbf{Time- and Space-Dependent Sparse Reward.} The second sparse reward additionally adds sparsity based on the position and only returns goal and rotation distance in the last time step when the box is near the goal location 
\begin{equation*}
    R_\text{tot} = \begin{cases}
        - R_\text{rod}
        - R_\text{rod\_rotation}
        - \tau_t 
        - \text{err}(\mathbf{q}, \mathbf{\dot{q}}) \cdots\\
        \cdots - \text{clip}(1050 R_\text{goal}, 0, 100)
        - \text{clip}(15 R_\text{rotation}, 0, 100) 
        + 300,
        & t=T \text{ and } R_\text{goal} \leq 0.1,\\
        - R_\text{rod}
        - R_\text{rod\_rotation}
        - \tau_t 
        - \text{err}(\mathbf{q}, \mathbf{\dot{q}}),
        &\text{else}.
        \\
        \end{cases}
\end{equation*}

\subsection{Hopper Jump}
\label{app:hopper_jump}
In the Hopper jump task the agent has to learn to jump as high as possible and land on a certain goal position at the same time. We consider five basis functions per joint resulting in an 15 dimensional weight space. The context is four-dimensional consisting of the initial joint angles $\theta \in [-0.5, 0]$, $\gamma \in [-0.2, 0]$, $\phi \in [0, 0.785]$ and the goal landing position $x \in [0.3, 1.35]$. 
{\color{black}The full observation space extends the original observation space from the OpenAI gym Hopper by adding the x-value of the goal position and the x-y-z difference between the goal point and the reference point at the Hopper's foot.
The action space is the same as for the original Hopper task.}
We consider a non-Markovian reward function for the episode-based algorithms and a step-based reward for PPO, which we have extensively designed to obtain the highest possible jump.

\textbf{Non-Markovian Reward.} In each time-step $t$ we provide an action cost 
\begin{align*}
    \tau_t = 10^{-3}\sum_i^K (a^i_t)^2,
\end{align*}
where $K=3$ is the number of DoF. In the last time-step $T$ of the episode we provide a reward which contains information about the whole episode as
\begin{align*}
    R_{height} &= 10h_{max}, \\
    R_{gdist} &= ||p_{foot, T} - p_{goal}||_2,\\
    R_{cdist} &= ||p_{foot,contact} - p_{goal}||_2,\\
    R_{healthy} &= \left\{\begin{array}{ll} 2 & \textrm{if } z_T \in [0.5, \infty] \textrm{and } \theta, \gamma, \phi \in [-\infty, \infty]\\
                                           0 & \textrm{else} \end{array}\right.,
\end{align*}
where $h_{max}$ is the maximum jump height in z-direction of the center of mass reached during the whole episode, $p_{foot, t}$ is the x-y-z position of the foot's heel at time step $t$, $p_{foot,contact}$ is the foot's heel position when having a contact with the ground after the first jump, $p_{goal}$ is the goal landing position of the heel. $R_{healthy}$ is a slightly modified reward of the healthy reward defined in the original hopper task. The hopper is considered as 'healthy' if the z position of the center of mass is within the range $[0.5m, \infty]$. This encourages the hopper to stand at the end of the episode. Note that all states need to be within the range $[-100, 100]$ for $R_{healthy}$. Since this is defined in the hopper task from OpenAI already, we haven't mentioned it here. The total reward at the end of an episode is given as
\begin{align*}
    R_{tot} = -\sum_{t=0}^T\tau_t + R_{height} + R_{gdist} + R_{cdist} + R_{healthy}.
\end{align*}

\textbf{Step-Based Reward.} We consider a step-based alternative reward such that PPO is also able to learn a meaningful behavior on this task. We have tuned the reward such that we can obtain the best performance. The observation space is the same as in the original hopper task from OpenAI extended with the goal landing position and the current distance of the foot's heel and the goal landing postion.  
We again consider the action cost in each time-step $t$
\begin{align*}
    \tau_t = 10^{-3}\sum_i^K (a^i_t)^2,
\end{align*}
and additionally consider the rewards
\begin{align*}
    R_{height, t} &= 3h_t\\
    R_{gdist, t} &= 3 ||p_{foot, t} - p_{goal}||_2 \\
    R_{healthy, t} &= \left\{\begin{array}{ll} 1 & \textrm{if } z_t \in [0.5, \infty] \textrm{and } \theta, \gamma, \phi \in [-\infty, \infty]\\
                                           0 & \textrm{else} \end{array}\right.,
\end{align*}
where these rewards are now returned to the agent in each time-step $t$, resulting in the reward per time-step 
\begin{align*}
    r_t(s_t, a_t) = -\tau_t + R_{height, t} + R_{gdist, t} + R_{healthy, t}.
\end{align*}

\subsection{Beer Pong}
\label{app:beer_pong}
In the Beer Pong task the $K=7$ Degrees of Freedom (DoF) robot has to throw a ball into a cup on a big table. The context is defined by the cup's two dimensional position on the table which lies in the range $x\in [-1.42, 1.42]$, $y\in [-4.05, -1.25]$. 
{\color{black}
For the step-based algorithms we consider cosine and sine of the robot's joint angles, the angle velocities, the ball's distance to the bottom of the cup, the ball's distance to the top of the cup, the cup position and the current time step.}
The action space for the step-based algorithms is defined as the torques for each joint, the parameter space for the episode-based methods is 15 dimensional which consists of the two weights for the basis functions per joint and the duration of the throwing trajectory, i.e. the ball release time.

We generally consider action penalties
\begin{align*}
    \tau_t = \frac{1}{K}\sum_i^K (a^i_t)^2,
\end{align*}
consisting of the sum of squared torques per joint.
For $t<T$ we consider the reward
\begin{align*}
    r_t(s_t,a_t) = -\alpha_t \tau_t,
\end{align*}
with $\alpha_t = 10^{-2}$.
For $t=T$ we consider the non-Markovian reward

\begin{equation*}
    R_{task} = \begin{cases}
    -4 
    - min(||p_{c,top}-p_{b,1:T}||_2^2) 
    - 0.5||p_{c,bottom}-p_{b,T}||_2^2 \cdots \\
    \cdots - 2||p_{c,bottom}-p_{b,k}||_2^2
    -\alpha_T\tau, 
    & \textrm{if cond. 1} \\
    -4 
    - min(||p_{c,top}-p_{b,1:T}||_2^2) 
    - 0.5||p_{c,bottom}-p_{b,T}||_2^2 
    -\alpha_T\tau, 
    & \textrm{if cond. 2} \\
    -2 
    - min(||p_{c,top}-p_{b,1:T}||_2^2) 
    - 0.5||p_{c,bottom}-p_{b,T}||_2^2 
    -\alpha_T\tau, 
    & \textrm{if cond. 3}\\
    -||p_{c,bottom}-p_{b,T}||_2^2 
    - \alpha_T \tau, 
    & \textrm{if cond. 4}
    \end{cases}
\end{equation*}

\begin{align*}
    R_{task} = \left\{
    \begin{array}{llll} -4 - min(||p_{c,top}-p_{b,1:T}||_2^2) - 0.5||p_{c,bottom}-p_{b,T}||_2^2 \cdots \\
                                    \cdots - 2||p_{c,bottom}-p_{b,k}||_2^2-\alpha_T\tau, & \textrm{if cond. 1} \\
                                    -4 - min(||p_{c,top}-p_{b,1:T}||_2^2) - 0.5||p_{c,bottom}-p_{b,T}||_2^2 -\alpha_T\tau, & \textrm{if cond. 2} \\
                                    -2 - min(||p_{c,top}-p_{b,1:T}||_2^2) - 0.5||p_{c,bottom}-p_{b,T}||_2^2 -\alpha_T\tau, & \textrm{if cond. 3} \\
                                                         -||p_{c,bottom}-p_{b,T}||_2^2 - \alpha_T \tau, & \textrm{if cond. 4}
                                                         \end{array}\right.,
\end{align*}
where $p_{c, top}$ is the position of the top edge of the cup, $p_{c, bottom}$ is the ground position of the cup, $p_{b,t}$ is the position of the ball at time point $t$, and $\tau$ is the squared mean torque over all joints during one rollout and $\alpha_T=10^{-4}$.  
The different conditions are:
\begin{itemize}
    \item cond. 1: The ball had a contact with the ground before having a contact with the table.
    \item cond. 2: The ball is not in the cup and had no table contact
    \item cond. 3: The ball is not in the cup and had table contact
    \item cond. 4: The ball is in the cup.
\end{itemize}
Note that $p_{b,k}$ is the ball's and the ground's contact position and is only given, if the ball had a contact with the ground first.

At time step $t=T$ we also give information whether the agent's chosen ball release time $B$ was reasonable
\begin{align*}
    R_{release} = \left\{\begin{array}{l}-30-10(B-B_{min})^2, ~~ \textrm{if } B < B_{min}\\
                                           -30-10(B-B_{max})^2, ~~ \textrm{if } B < B_{max} \end{array}\right.,
\end{align*}
where we define $B_{min} = 0.1s$ and $B_{max} = 1s$, such that the agent is encouraged to throw the ball within the time range $[B_{min}, B_{max}]$.

The total return over the whole episode is therefore given as 
\begin{align*}
    R_{tot} = \sum_{t=1}^{T-1} r_t(s_t,a_t) + R_{task} + R_{release}
\end{align*}

A throw is considered as successful if the ball is in the cup at the end of an episode.

\subsection{Table Tennis}
\label{app:table_tennis}
We consider table tennis for the entire table, i.\,e. incoming balls are anywhere on the side of the robot and goal locations anywhere on the opponents side. 
The goal is to use the seven degree of freedom robotic arm to hit the incoming ball based on its landing position and return it as close as possible to the specified goal location.
As context space we consider the initial ball position $x \in [-1, -0.2]$, $y \in [-0.65, 0.65]$ and the goal position $x \in [-1.2, -0.2]$, $y \in [-0.6, 0.6]$. 
{\color{black}The full observation space again contains cosine and sine of the joint space and the joint velocities as well as the ball velocity on top of the above context information.
The torques of the joints make up the action space.}
For this experiment, we do not use any gravity compensation and allow in the episode-based setting to learn the start time $t_0$ and the trajectory duration $T$.
The task is considered successful if the returned ball lands on the opponent's side of table and within $\leq 0.2$m to the goal location. 
The reward is defined as 
\begin{equation*}
    r_{task} = \begin{cases}
        0, & \text{if cond. 1} \\
        0.2-0.2\tanh{(\min{||\mathbf{p}_{r}-\mathbf{p}_{b}||^2})}, & \text{if cond. 2} \\
        3-2\tanh{(\min{||\mathbf{p}_{r}-\mathbf{p}_{b}||^2})}-\tanh{(||\mathbf{p}_{l}-\mathbf{p}_{goal}||^2)}, &\text{if cond. 3}\\
        6-2\tanh{(\min{||\mathbf{p}_{r}-\mathbf{p}_{b}||^2})}-4\tanh{(||\mathbf{p}_{l}-\mathbf{p}_{goal}||}^2), &\text{if cond. 4}\\
        7-2\tanh{(\min{||\mathbf{p}_{r}-\mathbf{p}_{b}||^2})}-4\tanh{(||\mathbf{p}_{l}-\mathbf{p}_{goal}||}^2), &\text{if cond. 5}
        \end{cases}
\end{equation*}
where $\mathbf{p}_r$ is the position of racket center, $\mathbf{p}_b$ is the position of the ball, $\mathbf{p}_{l}$ is the ball landing position, $\mathbf{p}_{goal}$ is the target position. The different conditions are
\begin{itemize}
    \item cond. 1: the end of episode is not reached,
    \item cond. 2: the end of episode is reached,
    \item cond. 3: cond.2 is satisfied and robot did hit the ball,
    \item cond. 4: cond.3 is satisfied and the returned ball landed on the table,
    \item cond. 5: cond.4 is satisfied and the landing position is at the opponent's side.
\end{itemize}
The episode ends when any of the following conditions are met
\begin{itemize}
    \item the maximum horizon length is reached
    \item ball did land on the floor without hitting
    \item ball did land on the floor or table after hitting
\end{itemize}

For BBRL-PPO and BBRL-TRPL, the whole desired trajectory is obtained ahead of environment interaction, making use of this property we can collect some samples without physical simulation. The reward function based on this desired trajectory is defined as
\begin{equation*}
    r_{traj} = -\sum_{(i,j)}{|\tau_{ij}^d| - |q^b_j|} , \quad 
    (i,j) \in \{(i,j)\mid |\tau_{ij}^d| > |q^b_j|\}
\end{equation*}
where $\tau^d$ is the desired trajectory, $i$ is the time index, $j$ is the joint index, $q^b$ is the joint position upperbound. The desired trajectory is considered as invalid if $r_{traj} < 0$, an invalid trajectory will not be executed by robot. The overall reward for BBRL is defined as:
\begin{equation*}
    r = \begin{cases}
    r_{traj}, & r_{traj} < 0 \\
    r_{task}, & \text{otherwise}
    \end{cases}
\end{equation*}

%% file: appendix/evaluations.tex
\begin{figure}[ht]
    \centering
    \begin{minipage}{0.25\textwidth}
        \resizebox{\textwidth}{!}{\input{tikz_plots/hopper_jump/hj_iqm_height_traj}}%
    \end{minipage}%
    \begin{minipage}{0.25\textwidth}
        \resizebox{\textwidth}{!}{\input{tikz_plots/reacher/reacher_sac_sparse}}%
    \end{minipage}%
    \caption{(Left) The improved performance on the Hopper Jump task is also demonstrated on the jumping profile for a fixed context. While BBRL-TRPL jumps once as high as possible, PPO constantly tries to maximize the height at each time step which leads to several jumps throughout the episode and consequently to a lower maximum height.
    (Right) Learning curve of SAC for the sparse reward of the 5D Reacher task.}
    \label{fig:hopper_jump_traj}
\end{figure}
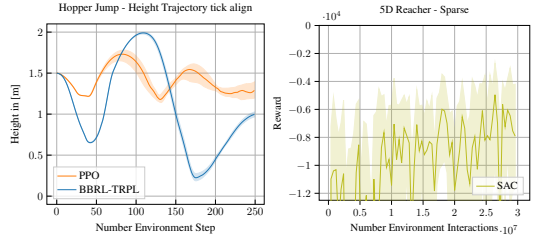

%% file: tikz_plots/hopper_jump/hj_iqm_height_traj.tex
\begin{tikzpicture}

\begin{axis}[
legend cell align={left},
legend style={fill opacity=0.8, draw opacity=1, text opacity=1, draw=lightgray204,at={(0.03,0.03)},  
  anchor=south west},
 title={Hopper Jump - Height Trajectory}
tick align=outside,
tick pos=left,
x grid style={darkgray176},
xlabel={Number Environment Step},
xmajorgrids,
xmin=-12.45, xmax=261.45,
xtick style={color=black},
y grid style={darkgray176},
ylabel={Height in [m]},
ymajorgrids,
ymin=-0.1, ymax=2.1,
ytick style={color=black}
]
\path [draw=C1, fill=C1, opacity=0.2]
(axis cs:0,1.49979866585973)
--(axis cs:0,1.49974285872014)
--(axis cs:1,1.49877016601333)
--(axis cs:2,1.49679637377694)
--(axis cs:3,1.49370489795399)
--(axis cs:4,1.48947470620233)
--(axis cs:5,1.48422876530949)
--(axis cs:6,1.47831778009126)
--(axis cs:7,1.47186368249624)
--(axis cs:8,1.46469451780223)
--(axis cs:9,1.45687539377264)
--(axis cs:10,1.44841982839606)
--(axis cs:11,1.43931298324306)
--(axis cs:12,1.42962309349231)
--(axis cs:13,1.41949741902882)
--(axis cs:14,1.40959045122909)
--(axis cs:15,1.39954326201896)
--(axis cs:16,1.38915740708021)
--(axis cs:17,1.37823330810942)
--(axis cs:18,1.366886836866)
--(axis cs:19,1.35484305008295)
--(axis cs:20,1.34209961131478)
--(axis cs:21,1.32864444044949)
--(axis cs:22,1.31448103419736)
--(axis cs:23,1.29972703461905)
--(axis cs:24,1.28451527335351)
--(axis cs:25,1.26904357971753)
--(axis cs:26,1.2556252942788)
--(axis cs:27,1.24509146026215)
--(axis cs:28,1.2374068691474)
--(axis cs:29,1.23207935185508)
--(axis cs:30,1.22855672666995)
--(axis cs:31,1.2263832915634)
--(axis cs:32,1.22496051049701)
--(axis cs:33,1.22387554138042)
--(axis cs:34,1.222852302452)
--(axis cs:35,1.22176111644547)
--(axis cs:36,1.2205624103631)
--(axis cs:37,1.2187387917463)
--(axis cs:38,1.21629725781301)
--(axis cs:39,1.2137289698726)
--(axis cs:40,1.21416524505497)
--(axis cs:41,1.21599032538898)
--(axis cs:42,1.22204731932572)
--(axis cs:43,1.23317169900505)
--(axis cs:44,1.24852190230563)
--(axis cs:45,1.26724470044295)
--(axis cs:46,1.2884517037204)
--(axis cs:47,1.31060060185633)
--(axis cs:48,1.33322237459452)
--(axis cs:49,1.35535316973263)
--(axis cs:50,1.37556427166442)
--(axis cs:51,1.39431960070734)
--(axis cs:52,1.41191880493663)
--(axis cs:53,1.42841889569092)
--(axis cs:54,1.4447653079312)
--(axis cs:55,1.45991766298667)
--(axis cs:56,1.47563285525182)
--(axis cs:57,1.49055307772218)
--(axis cs:58,1.50484006097632)
--(axis cs:59,1.5183379971383)
--(axis cs:60,1.53115231568527)
--(axis cs:61,1.54319307581066)
--(axis cs:62,1.55518102421179)
--(axis cs:63,1.56618879704031)
--(axis cs:64,1.57600563911415)
--(axis cs:65,1.58580805559681)
--(axis cs:66,1.59464952315137)
--(axis cs:67,1.60270745118188)
--(axis cs:68,1.61068203199823)
--(axis cs:69,1.61745189224339)
--(axis cs:70,1.62435845415918)
--(axis cs:71,1.62974203986693)
--(axis cs:72,1.63428436994606)
--(axis cs:73,1.63866078250095)
--(axis cs:74,1.64251241605615)
--(axis cs:75,1.64617455872368)
--(axis cs:76,1.64745683713827)
--(axis cs:77,1.6502355948603)
--(axis cs:78,1.65079401191089)
--(axis cs:79,1.65126538746464)
--(axis cs:80,1.64985166056767)
--(axis cs:81,1.64867375535955)
--(axis cs:82,1.64784779341868)
--(axis cs:83,1.64614432218239)
--(axis cs:84,1.64363394376844)
--(axis cs:85,1.63876464867201)
--(axis cs:86,1.63506019539709)
--(axis cs:87,1.63152549101906)
--(axis cs:88,1.62535439662109)
--(axis cs:89,1.61891104214268)
--(axis cs:90,1.61232089680307)
--(axis cs:91,1.6041781532347)
--(axis cs:92,1.59756714203172)
--(axis cs:93,1.58934383818073)
--(axis cs:94,1.57986454696043)
--(axis cs:95,1.56879757788675)
--(axis cs:96,1.55932243080024)
--(axis cs:97,1.54618143768635)
--(axis cs:98,1.53738264315073)
--(axis cs:99,1.52305267179769)
--(axis cs:100,1.50837603728533)
--(axis cs:101,1.49383038253807)
--(axis cs:102,1.47965607171859)
--(axis cs:103,1.46575123046124)
--(axis cs:104,1.44748308327532)
--(axis cs:105,1.43037969712116)
--(axis cs:106,1.4140867063565)
--(axis cs:107,1.39635915291112)
--(axis cs:108,1.38022775208176)
--(axis cs:109,1.36427681911699)
--(axis cs:110,1.34853804281982)
--(axis cs:111,1.33452659294528)
--(axis cs:112,1.32107877174574)
--(axis cs:113,1.30909666144926)
--(axis cs:114,1.29527333040455)
--(axis cs:115,1.28329607970933)
--(axis cs:116,1.27096884183586)
--(axis cs:117,1.26040328146516)
--(axis cs:118,1.25066370356817)
--(axis cs:119,1.23862026688635)
--(axis cs:120,1.22709759211458)
--(axis cs:121,1.21670835111184)
--(axis cs:122,1.20762510550615)
--(axis cs:123,1.19886425300932)
--(axis cs:124,1.18986995609633)
--(axis cs:125,1.18052999703187)
--(axis cs:126,1.16974550367235)
--(axis cs:127,1.16072382534513)
--(axis cs:128,1.1570239189053)
--(axis cs:129,1.15102182340607)
--(axis cs:130,1.14439836257897)
--(axis cs:131,1.14081161474566)
--(axis cs:132,1.14171987466153)
--(axis cs:133,1.14958960585437)
--(axis cs:134,1.16111323042177)
--(axis cs:135,1.17634342135801)
--(axis cs:136,1.1928560089598)
--(axis cs:137,1.2103577906521)
--(axis cs:138,1.22706029634914)
--(axis cs:139,1.24207289056192)
--(axis cs:140,1.25549244950628)
--(axis cs:141,1.2669872670453)
--(axis cs:142,1.27777158326442)
--(axis cs:143,1.28753509462072)
--(axis cs:144,1.29727569828925)
--(axis cs:145,1.30692130755638)
--(axis cs:146,1.31711947111994)
--(axis cs:147,1.32692420807177)
--(axis cs:148,1.33810020412403)
--(axis cs:149,1.34986224100855)
--(axis cs:150,1.36028039704362)
--(axis cs:151,1.36909408540126)
--(axis cs:152,1.3782052428303)
--(axis cs:153,1.39045366760756)
--(axis cs:154,1.40001148865232)
--(axis cs:155,1.40878515367695)
--(axis cs:156,1.41629605720397)
--(axis cs:157,1.42108338676484)
--(axis cs:158,1.42489228599394)
--(axis cs:159,1.42729834774432)
--(axis cs:160,1.42726434708174)
--(axis cs:161,1.42547736490498)
--(axis cs:162,1.42365063159935)
--(axis cs:163,1.42087419855101)
--(axis cs:164,1.41723125146209)
--(axis cs:165,1.41329351180249)
--(axis cs:166,1.40986154416342)
--(axis cs:167,1.4025442849025)
--(axis cs:168,1.39685412278649)
--(axis cs:169,1.39583849764517)
--(axis cs:170,1.39302172193064)
--(axis cs:171,1.39121571435872)
--(axis cs:172,1.39132417423791)
--(axis cs:173,1.38821209885217)
--(axis cs:174,1.38831095555757)
--(axis cs:175,1.38593592325745)
--(axis cs:176,1.38446991036263)
--(axis cs:177,1.38283981806662)
--(axis cs:178,1.37984053048565)
--(axis cs:179,1.37669925293871)
--(axis cs:180,1.37213109762576)
--(axis cs:181,1.36724961035286)
--(axis cs:182,1.36052078399363)
--(axis cs:183,1.35439562515155)
--(axis cs:184,1.34704803631162)
--(axis cs:185,1.33911231163479)
--(axis cs:186,1.33392306510587)
--(axis cs:187,1.32869122006831)
--(axis cs:188,1.3251184369848)
--(axis cs:189,1.3206185098962)
--(axis cs:190,1.31502557135333)
--(axis cs:191,1.30991565370694)
--(axis cs:192,1.30713357127565)
--(axis cs:193,1.30526023538357)
--(axis cs:194,1.30178988854094)
--(axis cs:195,1.29725253010112)
--(axis cs:196,1.29356707686094)
--(axis cs:197,1.29234729154524)
--(axis cs:198,1.29028359724234)
--(axis cs:199,1.28800961001294)
--(axis cs:200,1.28433462915803)
--(axis cs:201,1.28097470084007)
--(axis cs:202,1.27630539784962)
--(axis cs:203,1.26881783350349)
--(axis cs:204,1.26098331308878)
--(axis cs:205,1.25413257540649)
--(axis cs:206,1.24773667621677)
--(axis cs:207,1.24079031421556)
--(axis cs:208,1.23540526440975)
--(axis cs:209,1.23187024022682)
--(axis cs:210,1.22795928745624)
--(axis cs:211,1.22425790731119)
--(axis cs:212,1.22022866072575)
--(axis cs:213,1.2173483755809)
--(axis cs:214,1.21626161914822)
--(axis cs:215,1.21579716764399)
--(axis cs:216,1.21509813746791)
--(axis cs:217,1.21464860893691)
--(axis cs:218,1.21362214498291)
--(axis cs:219,1.21067455067891)
--(axis cs:220,1.20741754894478)
--(axis cs:221,1.20467168923188)
--(axis cs:222,1.20222380264227)
--(axis cs:223,1.20164360911505)
--(axis cs:224,1.20195037954616)
--(axis cs:225,1.2026837381899)
--(axis cs:226,1.20227866926806)
--(axis cs:227,1.20008081397959)
--(axis cs:228,1.19761431364473)
--(axis cs:229,1.19316915152471)
--(axis cs:230,1.18715969530102)
--(axis cs:231,1.18269438803748)
--(axis cs:232,1.17749901331129)
--(axis cs:233,1.17589976370693)
--(axis cs:234,1.17756384438809)
--(axis cs:235,1.17794498650964)
--(axis cs:236,1.17853430620413)
--(axis cs:237,1.18087086486925)
--(axis cs:238,1.18246826189897)
--(axis cs:239,1.18516101045859)
--(axis cs:240,1.1868864657937)
--(axis cs:241,1.18665664249221)
--(axis cs:242,1.18830330472422)
--(axis cs:243,1.18942791552174)
--(axis cs:244,1.1903353504597)
--(axis cs:245,1.18895802122722)
--(axis cs:246,1.19069535874082)
--(axis cs:247,1.19178234645326)
--(axis cs:248,1.19197186542507)
--(axis cs:249,1.19516214838798)
--(axis cs:249,1.39335218275064)
--(axis cs:249,1.39335218275064)
--(axis cs:248,1.38571159879674)
--(axis cs:247,1.38264479976921)
--(axis cs:246,1.37839106460384)
--(axis cs:245,1.37589050190132)
--(axis cs:244,1.37292493549224)
--(axis cs:243,1.37309368011497)
--(axis cs:242,1.37242918177677)
--(axis cs:241,1.37153700155409)
--(axis cs:240,1.36990450211508)
--(axis cs:239,1.37219279760948)
--(axis cs:238,1.36814707360276)
--(axis cs:237,1.36729321383123)
--(axis cs:236,1.36427694258371)
--(axis cs:235,1.36073673323124)
--(axis cs:234,1.35780125677463)
--(axis cs:233,1.35458413161987)
--(axis cs:232,1.35227279817266)
--(axis cs:231,1.35029365835559)
--(axis cs:230,1.34731157145857)
--(axis cs:229,1.34451259334961)
--(axis cs:228,1.34260220560033)
--(axis cs:227,1.33668680401527)
--(axis cs:226,1.33143698996417)
--(axis cs:225,1.32935056327684)
--(axis cs:224,1.32630521687356)
--(axis cs:223,1.32424212984415)
--(axis cs:222,1.3242150011985)
--(axis cs:221,1.32201825968672)
--(axis cs:220,1.32230640530758)
--(axis cs:219,1.32057417175988)
--(axis cs:218,1.319172026214)
--(axis cs:217,1.3176482778867)
--(axis cs:216,1.3169573434395)
--(axis cs:215,1.31610285604786)
--(axis cs:214,1.31649791906791)
--(axis cs:213,1.3159598580015)
--(axis cs:212,1.31540274107105)
--(axis cs:211,1.31604192182475)
--(axis cs:210,1.31556536362376)
--(axis cs:209,1.31588231760185)
--(axis cs:208,1.31798606328693)
--(axis cs:207,1.32045443988293)
--(axis cs:206,1.32465305990136)
--(axis cs:205,1.32617475291598)
--(axis cs:204,1.32918725512246)
--(axis cs:203,1.33417558936755)
--(axis cs:202,1.34304556337077)
--(axis cs:201,1.35402928611964)
--(axis cs:200,1.36677293748834)
--(axis cs:199,1.37973299610133)
--(axis cs:198,1.39476046362204)
--(axis cs:197,1.4086066340354)
--(axis cs:196,1.42291837974016)
--(axis cs:195,1.43630643076699)
--(axis cs:194,1.44886471022204)
--(axis cs:193,1.46314972480837)
--(axis cs:192,1.47446199305068)
--(axis cs:191,1.48539755509549)
--(axis cs:190,1.49818067826122)
--(axis cs:189,1.50889715508683)
--(axis cs:188,1.52024217473362)
--(axis cs:187,1.5306346603927)
--(axis cs:186,1.54043583301936)
--(axis cs:185,1.55029337511496)
--(axis cs:184,1.55889072675124)
--(axis cs:183,1.5660721924201)
--(axis cs:182,1.57485687030644)
--(axis cs:181,1.58030198305288)
--(axis cs:180,1.58788365499688)
--(axis cs:179,1.59292451220481)
--(axis cs:178,1.59792980253685)
--(axis cs:177,1.60182438183619)
--(axis cs:176,1.60628741240119)
--(axis cs:175,1.60835928400937)
--(axis cs:174,1.61096766313807)
--(axis cs:173,1.61266470343752)
--(axis cs:172,1.61254993956551)
--(axis cs:171,1.61119220525664)
--(axis cs:170,1.6092903297153)
--(axis cs:169,1.60759563866596)
--(axis cs:168,1.60226866457818)
--(axis cs:167,1.59826648137493)
--(axis cs:166,1.59405199243619)
--(axis cs:165,1.58764258671549)
--(axis cs:164,1.58156435893185)
--(axis cs:163,1.57465317920708)
--(axis cs:162,1.56851996845021)
--(axis cs:161,1.56216952812189)
--(axis cs:160,1.55523830234929)
--(axis cs:159,1.54713745347642)
--(axis cs:158,1.5381704785414)
--(axis cs:157,1.52878513522625)
--(axis cs:156,1.51762325752223)
--(axis cs:155,1.5057038155094)
--(axis cs:154,1.49336265260854)
--(axis cs:153,1.48004386343626)
--(axis cs:152,1.4662030584317)
--(axis cs:151,1.45366040841484)
--(axis cs:150,1.44187861239433)
--(axis cs:149,1.43081044500691)
--(axis cs:148,1.42152284740199)
--(axis cs:147,1.41277807578469)
--(axis cs:146,1.40448836324486)
--(axis cs:145,1.39569386621952)
--(axis cs:144,1.38698306475469)
--(axis cs:143,1.3783756542841)
--(axis cs:142,1.36787158898882)
--(axis cs:141,1.35895835994625)
--(axis cs:140,1.34808168029134)
--(axis cs:139,1.33587578962524)
--(axis cs:138,1.32347233673205)
--(axis cs:137,1.31007313233564)
--(axis cs:136,1.29810830439283)
--(axis cs:135,1.28581826934376)
--(axis cs:134,1.27386565685332)
--(axis cs:133,1.26156105687896)
--(axis cs:132,1.24921577823212)
--(axis cs:131,1.24013986799764)
--(axis cs:130,1.23271449108366)
--(axis cs:129,1.23266697459959)
--(axis cs:128,1.23889060716743)
--(axis cs:127,1.2513238572678)
--(axis cs:126,1.26834237497397)
--(axis cs:125,1.28531786922121)
--(axis cs:124,1.30155071962304)
--(axis cs:123,1.31674105800674)
--(axis cs:122,1.33024176669645)
--(axis cs:121,1.34619580080681)
--(axis cs:120,1.36581299872337)
--(axis cs:119,1.38691555461323)
--(axis cs:118,1.40846201313973)
--(axis cs:117,1.43121589610661)
--(axis cs:116,1.45376274238387)
--(axis cs:115,1.47508479053626)
--(axis cs:114,1.49726264681803)
--(axis cs:113,1.51745430965541)
--(axis cs:112,1.53825884702973)
--(axis cs:111,1.55707824929947)
--(axis cs:110,1.57510614349391)
--(axis cs:109,1.59175072338562)
--(axis cs:108,1.60843397699881)
--(axis cs:107,1.62406035619364)
--(axis cs:106,1.63762202091457)
--(axis cs:105,1.65147072614122)
--(axis cs:104,1.66464465274311)
--(axis cs:103,1.67660166134747)
--(axis cs:102,1.68805580115602)
--(axis cs:101,1.69837181546673)
--(axis cs:100,1.70893298404253)
--(axis cs:99,1.717573018289)
--(axis cs:98,1.72633096564497)
--(axis cs:97,1.73426697152948)
--(axis cs:96,1.7414023410677)
--(axis cs:95,1.74770282906304)
--(axis cs:94,1.75383141707484)
--(axis cs:93,1.75907755813732)
--(axis cs:92,1.76361769736712)
--(axis cs:91,1.7675105135525)
--(axis cs:90,1.77139313974222)
--(axis cs:89,1.77434464330097)
--(axis cs:88,1.7771089355636)
--(axis cs:87,1.77898090917116)
--(axis cs:86,1.78028196317255)
--(axis cs:85,1.78094528408799)
--(axis cs:84,1.78058039473623)
--(axis cs:83,1.78038685503459)
--(axis cs:82,1.77905490592896)
--(axis cs:81,1.77721424629719)
--(axis cs:80,1.77500920409884)
--(axis cs:79,1.77192655450521)
--(axis cs:78,1.76853060781524)
--(axis cs:77,1.76452456918836)
--(axis cs:76,1.75967466101487)
--(axis cs:75,1.75430924400972)
--(axis cs:74,1.74816081980184)
--(axis cs:73,1.7415442147171)
--(axis cs:72,1.73411357106101)
--(axis cs:71,1.7258861321781)
--(axis cs:70,1.71711375416025)
--(axis cs:69,1.70785382295875)
--(axis cs:68,1.6975904571635)
--(axis cs:67,1.68697600650967)
--(axis cs:66,1.67516824233527)
--(axis cs:65,1.66307350206786)
--(axis cs:64,1.65002122766317)
--(axis cs:63,1.63617608315017)
--(axis cs:62,1.62182424633212)
--(axis cs:61,1.60666052842183)
--(axis cs:60,1.59099257370048)
--(axis cs:59,1.57470183614536)
--(axis cs:58,1.55797194385661)
--(axis cs:57,1.5405272106774)
--(axis cs:56,1.52257318691812)
--(axis cs:55,1.50414984209507)
--(axis cs:54,1.4853198014542)
--(axis cs:53,1.46541271109292)
--(axis cs:52,1.44456709856876)
--(axis cs:51,1.42228367960701)
--(axis cs:50,1.39949250089842)
--(axis cs:49,1.37642060354322)
--(axis cs:48,1.35450403570591)
--(axis cs:47,1.33254878826767)
--(axis cs:46,1.31002674807134)
--(axis cs:45,1.28840873917719)
--(axis cs:44,1.26786216602992)
--(axis cs:43,1.24910632102834)
--(axis cs:42,1.23417634882666)
--(axis cs:41,1.22377624074416)
--(axis cs:40,1.21916307491674)
--(axis cs:39,1.2203531454605)
--(axis cs:38,1.22224768074453)
--(axis cs:37,1.22368696676898)
--(axis cs:36,1.22477212479489)
--(axis cs:35,1.22561012006022)
--(axis cs:34,1.22630524040061)
--(axis cs:33,1.22704072803391)
--(axis cs:32,1.2279312661037)
--(axis cs:31,1.22911684500329)
--(axis cs:30,1.23104462894193)
--(axis cs:29,1.2342153202427)
--(axis cs:28,1.23916952290499)
--(axis cs:27,1.24633059586959)
--(axis cs:26,1.25616060224076)
--(axis cs:25,1.26935645924524)
--(axis cs:24,1.28515342859114)
--(axis cs:23,1.30043953991771)
--(axis cs:22,1.3150831330586)
--(axis cs:21,1.32930245062002)
--(axis cs:20,1.34308838808705)
--(axis cs:19,1.35647466602255)
--(axis cs:18,1.36941454713439)
--(axis cs:17,1.38174596836382)
--(axis cs:16,1.39365878997504)
--(axis cs:15,1.40521623700051)
--(axis cs:14,1.41623066637369)
--(axis cs:13,1.42666098358103)
--(axis cs:12,1.43652960256976)
--(axis cs:11,1.44567684187365)
--(axis cs:10,1.45381628343549)
--(axis cs:9,1.46135848971813)
--(axis cs:8,1.46823605696367)
--(axis cs:7,1.47464561349478)
--(axis cs:6,1.48056518078329)
--(axis cs:5,1.48593590469908)
--(axis cs:4,1.49062200186852)
--(axis cs:3,1.49439910142877)
--(axis cs:2,1.4971973438971)
--(axis cs:1,1.49897089461649)
--(axis cs:0,1.49979866585973)
--cycle;

\addplot [thick, C1, mark=*, mark size=0, mark options={solid}]
table {%
0 1.49976712133287
1 1.49885854202257
2 1.49698751569336
3 1.49405701980262
4 1.49003616421148
5 1.48505508255841
6 1.47935545392114
7 1.47296405572401
8 1.46607873767317
9 1.45875271830735
10 1.45081160165094
11 1.44221556983755
12 1.43284127049199
13 1.42280080505137
14 1.41243789708973
15 1.40165998571963
16 1.39046756572729
17 1.37900555992189
18 1.36736241423526
19 1.35516180909321
20 1.3423469367448
21 1.32890241544776
22 1.31469993841332
23 1.30006345392436
24 1.28481054069766
25 1.26912711913647
26 1.25581518980829
27 1.24564222314891
28 1.23823455319852
29 1.23308737541604
30 1.22970940383418
31 1.22764803015233
32 1.22641257574332
33 1.22552455462803
34 1.22475163235329
35 1.22399917973553
36 1.22284995861301
37 1.22128181746031
38 1.21917625123045
39 1.21694405523605
40 1.2170303662759
41 1.22002740889508
42 1.22731881393126
43 1.2399725484887
44 1.25659603309864
45 1.27632170021974
46 1.29825314771232
47 1.32149539006176
48 1.34537048650551
49 1.36843198173416
50 1.38984752630749
51 1.41042455505704
52 1.43008687323676
53 1.44861572250306
54 1.46669380449844
55 1.48444300333052
56 1.50148066692649
57 1.51798277862937
58 1.53451304049849
59 1.55040940842164
60 1.56565758105019
61 1.58024246556699
62 1.59415899318125
63 1.60740723123549
64 1.61997961271552
65 1.63187324361037
66 1.64308907275299
67 1.65359823124607
68 1.66341155502048
69 1.67242191505134
70 1.68081823986066
71 1.68858531834488
72 1.69571692485091
73 1.70219373921886
74 1.70799661201057
75 1.71312281235223
76 1.71758032047864
77 1.72139198991589
78 1.72458233773333
79 1.72716584665929
80 1.72914898646104
81 1.73036486052715
82 1.73093886479178
83 1.73089488606227
84 1.7302413935387
85 1.72898702590827
86 1.727129647147
87 1.72467172406651
88 1.72161651674549
89 1.71797212009406
90 1.7137514379222
91 1.70895637203804
92 1.70358362589236
93 1.69762057677204
94 1.69105231541499
95 1.68386582123036
96 1.67604848123218
97 1.66757909045894
98 1.658420974601
99 1.64856158774681
100 1.63797212273579
101 1.6266061242813
102 1.61449841445762
103 1.60163480712152
104 1.58798105424047
105 1.57351651254604
106 1.55824496705046
107 1.54218892709792
108 1.52531458563336
109 1.50757765461303
110 1.48889750273972
111 1.46881474309728
112 1.44734544949499
113 1.42680743178385
114 1.40690533202247
115 1.38600463922284
116 1.36420922644248
117 1.34743773337867
118 1.33095878549261
119 1.31785752284722
120 1.30587941764975
121 1.29355468306588
122 1.28082670181616
123 1.26607493875133
124 1.24851752781123
125 1.23523392736578
126 1.21749659090091
127 1.20276131306719
128 1.19504173064514
129 1.1844590709043
130 1.17915373784704
131 1.17956421481905
132 1.18518723986773
133 1.1929689148887
134 1.20322468684429
135 1.21485203929821
136 1.22693482098994
137 1.23931806844584
138 1.25115421319719
139 1.26266098486311
140 1.27656403352625
141 1.28947775180006
142 1.30201574334362
143 1.31570290552154
144 1.32919115240463
145 1.3425955728519
146 1.35602721437673
147 1.36953579945521
148 1.38326393112319
149 1.39704566752673
150 1.41097465472332
151 1.42349691905946
152 1.43314461116838
153 1.44266951278417
154 1.45384385614478
155 1.46420417249949
156 1.47431648498477
157 1.48681323809774
158 1.4981146024085
159 1.50835554099825
160 1.51752617182314
161 1.52427520692029
162 1.52958473170068
163 1.53409677344205
164 1.53786907629143
165 1.54084848152417
166 1.54308206106523
167 1.54455188938653
168 1.54392524398547
169 1.54252962740169
170 1.54053456167619
171 1.53798657907459
172 1.53488427122763
173 1.53115137165125
174 1.52674966012528
175 1.52161346937264
176 1.51580607964469
177 1.50931326051346
178 1.50212841283216
179 1.49422563442216
180 1.4855579045673
181 1.47813645485618
182 1.47017271065514
183 1.46167520410473
184 1.45282739834211
185 1.44376806623976
186 1.43716434751307
187 1.43022442698431
188 1.4228858090249
189 1.41507354011424
190 1.40694103101071
191 1.39885727512644
192 1.39005065544156
193 1.38208160731477
194 1.37430837477437
195 1.36591345445525
196 1.35686837943445
197 1.34934119163447
198 1.34274775904017
199 1.3353520882507
200 1.32719756331064
201 1.3205822217191
202 1.31598317866067
203 1.31034557695981
204 1.30022046223517
205 1.29021232161813
206 1.28022811873067
207 1.27399620869852
208 1.27128011117204
209 1.26824186568386
210 1.26481502282536
211 1.2604868753112
212 1.25782827722555
213 1.25512875303854
214 1.25445278002714
215 1.25332676760645
216 1.25163224209921
217 1.25022375812
218 1.25005107441338
219 1.24972530601409
220 1.25048961888997
221 1.25280658655499
222 1.25642460496663
223 1.2605433299117
224 1.26251375369956
225 1.26443132784111
226 1.26537433203282
227 1.26503588236537
228 1.26689264284059
229 1.27251823702155
230 1.27837810886643
231 1.28198915760395
232 1.28496828477537
233 1.28763412286761
234 1.28848417429016
235 1.28592563804912
236 1.28267212369626
237 1.27866604506434
238 1.27395351254431
239 1.27174500792731
240 1.26900527711982
241 1.26599700254553
242 1.26363964510191
243 1.26183878657347
244 1.26129317033499
245 1.26306104746716
246 1.26689230982831
247 1.27276337167813
248 1.28051450512548
249 1.28982721035953
};
\addlegendentry{PPO}
\path [draw=C0, fill=C0, opacity=0.2]
(axis cs:0,1.49955256501752)
--(axis cs:0,1.49955072138009)
--(axis cs:1,1.49818179352727)
--(axis cs:2,1.49582080650785)
--(axis cs:3,1.49233729327896)
--(axis cs:4,1.48754887181825)
--(axis cs:5,1.48123618696159)
--(axis cs:6,1.47317438115472)
--(axis cs:7,1.46316109835359)
--(axis cs:8,1.45104152509894)
--(axis cs:9,1.43670961420658)
--(axis cs:10,1.42010738098833)
--(axis cs:11,1.40125824164267)
--(axis cs:12,1.38022617384235)
--(axis cs:13,1.3570701403388)
--(axis cs:14,1.33195166857515)
--(axis cs:15,1.30508357814745)
--(axis cs:16,1.27666980468387)
--(axis cs:17,1.24688157212662)
--(axis cs:18,1.21612491013412)
--(axis cs:19,1.18453200699303)
--(axis cs:20,1.15245650126832)
--(axis cs:21,1.12020540096213)
--(axis cs:22,1.08797706472305)
--(axis cs:23,1.05610926260899)
--(axis cs:24,1.02477118242216)
--(axis cs:25,0.994414674563089)
--(axis cs:26,0.964932983168278)
--(axis cs:27,0.936466342645752)
--(axis cs:28,0.909096404838535)
--(axis cs:29,0.88287033699262)
--(axis cs:30,0.858429708800055)
--(axis cs:31,0.835481508623902)
--(axis cs:32,0.813088736252146)
--(axis cs:33,0.790720068051701)
--(axis cs:34,0.768130349436676)
--(axis cs:35,0.745194666942674)
--(axis cs:36,0.721869209184738)
--(axis cs:37,0.698291765316902)
--(axis cs:38,0.677393302594327)
--(axis cs:39,0.663294618925308)
--(axis cs:40,0.65505471971059)
--(axis cs:41,0.650386465295045)
--(axis cs:42,0.64848913098364)
--(axis cs:43,0.64875446970968)
--(axis cs:44,0.651141413187205)
--(axis cs:45,0.655511886869239)
--(axis cs:46,0.661903393980079)
--(axis cs:47,0.670382248565477)
--(axis cs:48,0.680864016175065)
--(axis cs:49,0.693654401127286)
--(axis cs:50,0.708791937771705)
--(axis cs:51,0.72641995211484)
--(axis cs:52,0.74672721880513)
--(axis cs:53,0.769780084141753)
--(axis cs:54,0.79580146309524)
--(axis cs:55,0.825016708463279)
--(axis cs:56,0.857532952035312)
--(axis cs:57,0.887801895183219)
--(axis cs:58,0.920044341996622)
--(axis cs:59,0.959652378844421)
--(axis cs:60,1.00472125824275)
--(axis cs:61,1.05255260203674)
--(axis cs:62,1.10274533850578)
--(axis cs:63,1.15474724935709)
--(axis cs:64,1.20792526596951)
--(axis cs:65,1.26106142121185)
--(axis cs:66,1.31336559951269)
--(axis cs:67,1.36307301759672)
--(axis cs:68,1.40904023979216)
--(axis cs:69,1.44964735538418)
--(axis cs:70,1.48440481806207)
--(axis cs:71,1.5131309357913)
--(axis cs:72,1.53711385069762)
--(axis cs:73,1.55963065826761)
--(axis cs:74,1.58191845470302)
--(axis cs:75,1.60445078898145)
--(axis cs:76,1.62680981194307)
--(axis cs:77,1.64891023452577)
--(axis cs:78,1.67033628148486)
--(axis cs:79,1.69079062752358)
--(axis cs:80,1.71047238671029)
--(axis cs:81,1.72928333681266)
--(axis cs:82,1.74731828033878)
--(axis cs:83,1.76483767755846)
--(axis cs:84,1.78155683805634)
--(axis cs:85,1.79756982966769)
--(axis cs:86,1.81269221085302)
--(axis cs:87,1.8271208741836)
--(axis cs:88,1.84084909556976)
--(axis cs:89,1.8539930760462)
--(axis cs:90,1.86613828638989)
--(axis cs:91,1.8779459422577)
--(axis cs:92,1.88892816187795)
--(axis cs:93,1.89934853302566)
--(axis cs:94,1.90897783146465)
--(axis cs:95,1.91810558614041)
--(axis cs:96,1.92635926443904)
--(axis cs:97,1.93414736930873)
--(axis cs:98,1.9412397982348)
--(axis cs:99,1.94753071720582)
--(axis cs:100,1.95364806108532)
--(axis cs:101,1.95865771506376)
--(axis cs:102,1.96324434037033)
--(axis cs:103,1.96700156361602)
--(axis cs:104,1.97030063042518)
--(axis cs:105,1.9727174679599)
--(axis cs:106,1.97469528253584)
--(axis cs:107,1.97595385573004)
--(axis cs:108,1.97646197746879)
--(axis cs:109,1.9759729654956)
--(axis cs:110,1.97543582999048)
--(axis cs:111,1.97383107814607)
--(axis cs:112,1.97161461070019)
--(axis cs:113,1.96783783093893)
--(axis cs:114,1.96410465576622)
--(axis cs:115,1.95899400766864)
--(axis cs:116,1.95272038567132)
--(axis cs:117,1.94578103558856)
--(axis cs:118,1.93715340404174)
--(axis cs:119,1.92745082036702)
--(axis cs:120,1.91639708909734)
--(axis cs:121,1.90373602869949)
--(axis cs:122,1.88945972345681)
--(axis cs:123,1.8735723997253)
--(axis cs:124,1.85695371893136)
--(axis cs:125,1.83774510226579)
--(axis cs:126,1.81707122928616)
--(axis cs:127,1.79432688892125)
--(axis cs:128,1.76949789650663)
--(axis cs:129,1.74311028021168)
--(axis cs:130,1.71508760504033)
--(axis cs:131,1.68521392921333)
--(axis cs:132,1.65431660245099)
--(axis cs:133,1.62199816254941)
--(axis cs:134,1.58810830146299)
--(axis cs:135,1.55365444535317)
--(axis cs:136,1.51723628944168)
--(axis cs:137,1.48059914268614)
--(axis cs:138,1.44164510373046)
--(axis cs:139,1.40283147466497)
--(axis cs:140,1.36400261482434)
--(axis cs:141,1.32352900067635)
--(axis cs:142,1.28283634602433)
--(axis cs:143,1.2428787844314)
--(axis cs:144,1.20159320060181)
--(axis cs:145,1.16004276745428)
--(axis cs:146,1.11843345353551)
--(axis cs:147,1.07644184913438)
--(axis cs:148,1.03449753398243)
--(axis cs:149,0.993015765593022)
--(axis cs:150,0.953170530960292)
--(axis cs:151,0.915586646865111)
--(axis cs:152,0.879480261296826)
--(axis cs:153,0.845859465448329)
--(axis cs:154,0.811469412100292)
--(axis cs:155,0.777085434894402)
--(axis cs:156,0.742732123897935)
--(axis cs:157,0.708668736244363)
--(axis cs:158,0.673469684642979)
--(axis cs:159,0.637028173678976)
--(axis cs:160,0.600846450339618)
--(axis cs:161,0.563350825755693)
--(axis cs:162,0.526013910505303)
--(axis cs:163,0.488603942649261)
--(axis cs:164,0.453681853162923)
--(axis cs:165,0.417208037655175)
--(axis cs:166,0.381997723641452)
--(axis cs:167,0.350591957808472)
--(axis cs:168,0.319641877037421)
--(axis cs:169,0.28901372567936)
--(axis cs:170,0.259145709425979)
--(axis cs:171,0.23395036729011)
--(axis cs:172,0.214919464958152)
--(axis cs:173,0.199800665563576)
--(axis cs:174,0.191208965694794)
--(axis cs:175,0.187015794174953)
--(axis cs:176,0.187513619006719)
--(axis cs:177,0.190090913778787)
--(axis cs:178,0.193359336754016)
--(axis cs:179,0.197336815566017)
--(axis cs:180,0.20154694375883)
--(axis cs:181,0.206781670763419)
--(axis cs:182,0.212114064705427)
--(axis cs:183,0.218477193094833)
--(axis cs:184,0.226075801714157)
--(axis cs:185,0.235578575923623)
--(axis cs:186,0.245331492205938)
--(axis cs:187,0.254823906235162)
--(axis cs:188,0.26496950174823)
--(axis cs:189,0.27541223345211)
--(axis cs:190,0.286297691636888)
--(axis cs:191,0.297450836642882)
--(axis cs:192,0.309270216361034)
--(axis cs:193,0.321491402777522)
--(axis cs:194,0.334284218516159)
--(axis cs:195,0.34757169316712)
--(axis cs:196,0.362523783798368)
--(axis cs:197,0.377576317078266)
--(axis cs:198,0.393566471875594)
--(axis cs:199,0.40958364083063)
--(axis cs:200,0.425010433050861)
--(axis cs:201,0.440741218027672)
--(axis cs:202,0.456735068082841)
--(axis cs:203,0.472645035128981)
--(axis cs:204,0.488651766090701)
--(axis cs:205,0.504641152653162)
--(axis cs:206,0.520773423998024)
--(axis cs:207,0.5368341371039)
--(axis cs:208,0.552570182719592)
--(axis cs:209,0.567590577955063)
--(axis cs:210,0.582749339938484)
--(axis cs:211,0.596549672588268)
--(axis cs:212,0.610036061619217)
--(axis cs:213,0.622966291633427)
--(axis cs:214,0.635484445421398)
--(axis cs:215,0.647668085581339)
--(axis cs:216,0.660577832907197)
--(axis cs:217,0.673912600613537)
--(axis cs:218,0.687571031163351)
--(axis cs:219,0.700823058463132)
--(axis cs:220,0.714423378594579)
--(axis cs:221,0.72664361176813)
--(axis cs:222,0.739679742155925)
--(axis cs:223,0.753556997154677)
--(axis cs:224,0.767487568283558)
--(axis cs:225,0.78066393811449)
--(axis cs:226,0.793586803233212)
--(axis cs:227,0.805754591484879)
--(axis cs:228,0.818020365396638)
--(axis cs:229,0.829608618115435)
--(axis cs:230,0.840718252496045)
--(axis cs:231,0.851698637855096)
--(axis cs:232,0.861533778993299)
--(axis cs:233,0.871144156549273)
--(axis cs:234,0.880468715148347)
--(axis cs:235,0.888769445207507)
--(axis cs:236,0.896753113727989)
--(axis cs:237,0.904557417035556)
--(axis cs:238,0.911837276931234)
--(axis cs:239,0.918430414457332)
--(axis cs:240,0.924958233174126)
--(axis cs:241,0.931657449415138)
--(axis cs:242,0.937039257888956)
--(axis cs:243,0.942273428029676)
--(axis cs:244,0.947111064332367)
--(axis cs:245,0.951399256399126)
--(axis cs:246,0.955188594126376)
--(axis cs:247,0.958959159118238)
--(axis cs:248,0.962330185588134)
--(axis cs:249,0.965336879423462)
--(axis cs:249,1.02533589637238)
--(axis cs:249,1.02533589637238)
--(axis cs:248,1.02153281893559)
--(axis cs:247,1.01793397420392)
--(axis cs:246,1.01400232126947)
--(axis cs:245,1.0098653402666)
--(axis cs:244,1.00492216536718)
--(axis cs:243,0.999613733140369)
--(axis cs:242,0.993782751861255)
--(axis cs:241,0.987519252600607)
--(axis cs:240,0.980800611964967)
--(axis cs:239,0.973662319011068)
--(axis cs:238,0.966025722927402)
--(axis cs:237,0.957813456812657)
--(axis cs:236,0.949156104424467)
--(axis cs:235,0.940447460138931)
--(axis cs:234,0.93101935442434)
--(axis cs:233,0.921095893755607)
--(axis cs:232,0.911003211827834)
--(axis cs:231,0.900015861340277)
--(axis cs:230,0.888824473600892)
--(axis cs:229,0.877388081230007)
--(axis cs:228,0.86536523139714)
--(axis cs:227,0.853390936309786)
--(axis cs:226,0.841171985884762)
--(axis cs:225,0.828346131926558)
--(axis cs:224,0.815297628782504)
--(axis cs:223,0.802085831604707)
--(axis cs:222,0.788604603480766)
--(axis cs:221,0.77473434106441)
--(axis cs:220,0.760281844606104)
--(axis cs:219,0.745348570076858)
--(axis cs:218,0.729848245916649)
--(axis cs:217,0.714763245124131)
--(axis cs:216,0.699713387945313)
--(axis cs:215,0.684615893893189)
--(axis cs:214,0.66967716771111)
--(axis cs:213,0.654991698413917)
--(axis cs:212,0.64047311943051)
--(axis cs:211,0.626639050058841)
--(axis cs:210,0.612837588653243)
--(axis cs:209,0.599032492544841)
--(axis cs:208,0.585611343457854)
--(axis cs:207,0.572170967237465)
--(axis cs:206,0.558771829632859)
--(axis cs:205,0.545629640912408)
--(axis cs:204,0.532544988183833)
--(axis cs:203,0.519536745852051)
--(axis cs:202,0.505949193385404)
--(axis cs:201,0.490905675363015)
--(axis cs:200,0.475898604752907)
--(axis cs:199,0.460260110182751)
--(axis cs:198,0.445937338435519)
--(axis cs:197,0.4313433792808)
--(axis cs:196,0.417371701657875)
--(axis cs:195,0.403600115944273)
--(axis cs:194,0.39154920467643)
--(axis cs:193,0.379791918964801)
--(axis cs:192,0.368927324102877)
--(axis cs:191,0.35859861310227)
--(axis cs:190,0.348665516225564)
--(axis cs:189,0.338996239596765)
--(axis cs:188,0.331043637417634)
--(axis cs:187,0.322704721127621)
--(axis cs:186,0.315500868206604)
--(axis cs:185,0.309163574353697)
--(axis cs:184,0.30381524259424)
--(axis cs:183,0.297758247594637)
--(axis cs:182,0.293463358681434)
--(axis cs:181,0.289695669491175)
--(axis cs:180,0.285483678745939)
--(axis cs:179,0.281527716995314)
--(axis cs:178,0.2792125414944)
--(axis cs:177,0.276795809447827)
--(axis cs:176,0.276016077993475)
--(axis cs:175,0.275636901858727)
--(axis cs:174,0.278739698451849)
--(axis cs:173,0.286091066962111)
--(axis cs:172,0.299236276540234)
--(axis cs:171,0.317524815116073)
--(axis cs:170,0.33867031125386)
--(axis cs:169,0.365587326257339)
--(axis cs:168,0.397241760924529)
--(axis cs:167,0.430884045268683)
--(axis cs:166,0.468933244683122)
--(axis cs:165,0.508208186958992)
--(axis cs:164,0.548468624885539)
--(axis cs:163,0.588299380683808)
--(axis cs:162,0.626011416105176)
--(axis cs:161,0.664728762383942)
--(axis cs:160,0.702137396867016)
--(axis cs:159,0.739179117763711)
--(axis cs:158,0.775097196169943)
--(axis cs:157,0.810058919218323)
--(axis cs:156,0.844517739132421)
--(axis cs:155,0.879190940073413)
--(axis cs:154,0.91379521747542)
--(axis cs:153,0.949457344934031)
--(axis cs:152,0.986198356187749)
--(axis cs:151,1.02423228677019)
--(axis cs:150,1.06266609006614)
--(axis cs:149,1.10307508459982)
--(axis cs:148,1.1425553138723)
--(axis cs:147,1.18276577352156)
--(axis cs:146,1.22293676829809)
--(axis cs:145,1.26264914402001)
--(axis cs:144,1.30233914037754)
--(axis cs:143,1.34171143332059)
--(axis cs:142,1.37956330332355)
--(axis cs:141,1.41772498692948)
--(axis cs:140,1.45539337801216)
--(axis cs:139,1.49217501469217)
--(axis cs:138,1.52807513264112)
--(axis cs:137,1.5637178167214)
--(axis cs:136,1.59809095100573)
--(axis cs:135,1.63119247903473)
--(axis cs:134,1.66314721402824)
--(axis cs:133,1.69387704954578)
--(axis cs:132,1.72341132416701)
--(axis cs:131,1.75142538048809)
--(axis cs:130,1.77798719173943)
--(axis cs:129,1.80304149791459)
--(axis cs:128,1.82640354564874)
--(axis cs:127,1.84837066464843)
--(axis cs:126,1.86856811090317)
--(axis cs:125,1.88629813929301)
--(axis cs:124,1.90288919909968)
--(axis cs:123,1.9176200617318)
--(axis cs:122,1.93115357172513)
--(axis cs:121,1.94281786787093)
--(axis cs:120,1.95310568380746)
--(axis cs:119,1.96214771521423)
--(axis cs:118,1.96993627096275)
--(axis cs:117,1.97657688557931)
--(axis cs:116,1.9821203956435)
--(axis cs:115,1.98670826122472)
--(axis cs:114,1.99070190048774)
--(axis cs:113,1.99361997875108)
--(axis cs:112,1.99575517134972)
--(axis cs:111,1.99709886129166)
--(axis cs:110,1.99774181032287)
--(axis cs:109,1.99772933067286)
--(axis cs:108,1.99709730405745)
--(axis cs:107,1.99565134722246)
--(axis cs:106,1.99360025911885)
--(axis cs:105,1.99086485270215)
--(axis cs:104,1.9876845997714)
--(axis cs:103,1.98368923050509)
--(axis cs:102,1.97902407221633)
--(axis cs:101,1.97375449147097)
--(axis cs:100,1.96782704517873)
--(axis cs:99,1.96122074640498)
--(axis cs:98,1.95408145715982)
--(axis cs:97,1.94618422571517)
--(axis cs:96,1.9376042074473)
--(axis cs:95,1.92845510712498)
--(axis cs:94,1.91864168030187)
--(axis cs:93,1.90826711688835)
--(axis cs:92,1.8972123450826)
--(axis cs:91,1.88548870852207)
--(axis cs:90,1.87312374478133)
--(axis cs:89,1.86010957552586)
--(axis cs:88,1.84648842066622)
--(axis cs:87,1.83224237960524)
--(axis cs:86,1.81732913778534)
--(axis cs:85,1.80179195034402)
--(axis cs:84,1.78563577028843)
--(axis cs:83,1.76890797232736)
--(axis cs:82,1.75150633706699)
--(axis cs:81,1.73344068373393)
--(axis cs:80,1.71476331043223)
--(axis cs:79,1.69545318722696)
--(axis cs:78,1.67554825245706)
--(axis cs:77,1.65499196037317)
--(axis cs:76,1.63363712704349)
--(axis cs:75,1.61180026294743)
--(axis cs:74,1.58947799872875)
--(axis cs:73,1.56701545281616)
--(axis cs:72,1.54454849178075)
--(axis cs:71,1.52111931821687)
--(axis cs:70,1.49370603394675)
--(axis cs:69,1.46049854967206)
--(axis cs:68,1.42124478148103)
--(axis cs:67,1.37694237020182)
--(axis cs:66,1.32819148703552)
--(axis cs:65,1.27673053272088)
--(axis cs:64,1.22394245867031)
--(axis cs:63,1.17103284935909)
--(axis cs:62,1.11916495909175)
--(axis cs:61,1.06870537161659)
--(axis cs:60,1.02077235891309)
--(axis cs:59,0.975507627733479)
--(axis cs:58,0.934675346135002)
--(axis cs:57,0.901564063110455)
--(axis cs:56,0.874196627151254)
--(axis cs:55,0.844392088606292)
--(axis cs:54,0.814269619535362)
--(axis cs:53,0.787826572375717)
--(axis cs:52,0.763648885987994)
--(axis cs:51,0.742626012676923)
--(axis cs:50,0.724067120198423)
--(axis cs:49,0.708340787053256)
--(axis cs:48,0.69482547407148)
--(axis cs:47,0.683304828736442)
--(axis cs:46,0.67403305935216)
--(axis cs:45,0.667097571709367)
--(axis cs:44,0.66193522187657)
--(axis cs:43,0.658920644591134)
--(axis cs:42,0.657970964675302)
--(axis cs:41,0.659450089198795)
--(axis cs:40,0.663369611235489)
--(axis cs:39,0.670766412574577)
--(axis cs:38,0.68369663882511)
--(axis cs:37,0.70338197180594)
--(axis cs:36,0.726198512070546)
--(axis cs:35,0.749449207155972)
--(axis cs:34,0.772335648919472)
--(axis cs:33,0.795069346624314)
--(axis cs:32,0.817914396911607)
--(axis cs:31,0.841290351569074)
--(axis cs:30,0.865427259767817)
--(axis cs:29,0.891052888116025)
--(axis cs:28,0.91793021965345)
--(axis cs:27,0.945727768734531)
--(axis cs:26,0.974449343334419)
--(axis cs:25,1.00420320539799)
--(axis cs:24,1.03459671249547)
--(axis cs:23,1.06568408175035)
--(axis cs:22,1.09723502085317)
--(axis cs:21,1.12897474225503)
--(axis cs:20,1.16058927323627)
--(axis cs:19,1.19203571440437)
--(axis cs:18,1.22284223712829)
--(axis cs:17,1.25278885906126)
--(axis cs:16,1.28171997415946)
--(axis cs:15,1.3094368607635)
--(axis cs:14,1.33557499725437)
--(axis cs:13,1.35997207270193)
--(axis cs:12,1.38249536722082)
--(axis cs:11,1.40299841855048)
--(axis cs:10,1.42136746126465)
--(axis cs:9,1.43756962295907)
--(axis cs:8,1.45160305021834)
--(axis cs:7,1.46350058916125)
--(axis cs:6,1.473363746879)
--(axis cs:5,1.48134291627185)
--(axis cs:4,1.48760981307672)
--(axis cs:3,1.49237138901733)
--(axis cs:2,1.49583835137798)
--(axis cs:1,1.49818933880518)
--(axis cs:0,1.49955256501752)
--cycle;

\addplot [thick, C0, mark=*, mark size=0, mark options={solid}]
table {%
0 1.4995517173791
1 1.49818580190853
2 1.49582972863573
3 1.49235391486178
4 1.4875771574501
5 1.48128115353224
6 1.47324967517758
7 1.46329502386876
8 1.45126887148465
9 1.4370586018998
10 1.4206132081598
11 1.40196791123082
12 1.38115471798198
13 1.35826944447738
14 1.33345448698191
15 1.30688880446821
16 1.27878164800371
17 1.24936634728311
18 1.21889423408574
19 1.18762859170054
20 1.15583838308177
21 1.12379147610018
22 1.09176893825395
23 1.06003111146693
24 1.02880665213586
25 0.998331740621817
26 0.968786513728701
27 0.940222479624829
28 0.912610499939346
29 0.886052861128479
30 0.861042689253853
31 0.837559198438135
32 0.814759634991104
33 0.792285524084576
34 0.769707059007311
35 0.746823374434987
36 0.723544449955751
37 0.70021703543232
38 0.679790059290946
39 0.666163455216148
40 0.658258927549436
41 0.653890576915501
42 0.652163477937193
43 0.652684303325065
44 0.655295424813073
45 0.65993240901763
46 0.666591484159224
47 0.675308423857973
48 0.686144729624796
49 0.69925108900057
50 0.714750646702645
51 0.732770254590037
52 0.753443353859279
53 0.776940186143972
54 0.803425422983458
55 0.83304369175493
56 0.8652914390754
57 0.894728021168108
58 0.926724213120361
59 0.966978772767031
60 1.01221716968139
61 1.06034558444297
62 1.11079404857188
63 1.16296789713808
64 1.21609263988449
65 1.2691984178445
66 1.3209744414846
67 1.37015679483928
68 1.41529392916431
69 1.45518637376222
70 1.48937515948925
71 1.51745125045531
72 1.54115135674405
73 1.56375158893624
74 1.58617118118467
75 1.60856044512864
76 1.63056992198161
77 1.65226185606592
78 1.67325578403333
79 1.69339675213671
80 1.71279955379935
81 1.73144468283479
82 1.74943131237667
83 1.76700589046068
84 1.78388980873355
85 1.80013359599835
86 1.81556047070105
87 1.83028407149785
88 1.84435118446163
89 1.85777100349319
90 1.87054546953383
91 1.8827124757188
92 1.89423759769444
93 1.90507235037812
94 1.91523949690164
95 1.92476575781829
96 1.93365131939803
97 1.9418963019436
98 1.94949472444407
99 1.95640841282228
100 1.96268142388954
101 1.96831376726388
102 1.97328997915514
103 1.97757817533376
104 1.9812252261391
105 1.98423053020359
106 1.98659163319938
107 1.98830342640816
108 1.98935709376066
109 1.9897389934066
110 1.98942966510089
111 1.98840296969404
112 1.98662552411781
113 1.98405641194942
114 1.98064709213588
115 1.976329566338
116 1.97104111976551
117 1.96472614236432
118 1.95731321345756
119 1.94872906800604
120 1.93885074189644
121 1.92755995604969
122 1.91484991180554
123 1.90065851651886
124 1.88490863726027
125 1.86719694586964
126 1.84781019582262
127 1.82647226549049
128 1.80304854959938
129 1.77788416402802
130 1.75103326385594
131 1.72254910377652
132 1.69250222193584
133 1.66097809729959
134 1.62807371065839
135 1.59389431996903
136 1.55855052511316
137 1.52215543755825
138 1.48482246718002
139 1.44666340071112
140 1.40778673154852
141 1.36829642184647
142 1.32828981513272
143 1.28785517749236
144 1.24707236041047
145 1.20601382147587
146 1.16474292311401
147 1.12331463058514
148 1.08177742819934
149 1.04047064108471
150 1.00013890761979
151 0.961706497952162
152 0.924881148381443
153 0.88921070864023
154 0.854396069146712
155 0.820021528094503
156 0.785797940022776
157 0.751470961231123
158 0.716657761868462
159 0.681167016835052
160 0.64503934165608
161 0.608246337503269
162 0.570709452368908
163 0.532448218162897
164 0.493507534701695
165 0.453943286635664
166 0.415447853046879
167 0.380047505184117
168 0.347310299982979
169 0.316296476445972
170 0.287761519194508
171 0.266292792722999
172 0.248976075848303
173 0.235902201820505
174 0.227306347082867
175 0.224236566510142
176 0.224699437439571
177 0.227340264074383
178 0.230783175726391
179 0.235060960077275
180 0.240057408663324
181 0.245705257572025
182 0.250624354987151
183 0.254957070863364
184 0.260835790416605
185 0.267974038430926
186 0.275818979008108
187 0.284345860584176
188 0.293492400933104
189 0.303220754379853
190 0.313507242327459
191 0.324368923190517
192 0.335783178743965
193 0.347785740427175
194 0.360310628877115
195 0.373291828868138
196 0.387727413366814
197 0.402543105703443
198 0.417674692460182
199 0.433285710579278
200 0.449212735747843
201 0.465116859474284
202 0.48019420989442
203 0.49450980167957
204 0.508765809093983
205 0.523241610726545
206 0.537882617897389
207 0.552575476274906
208 0.567275267442396
209 0.581957266123611
210 0.596282480924883
211 0.610255170552148
212 0.623999030359345
213 0.637645728225203
214 0.651585731978132
215 0.665422331999898
216 0.679352126332645
217 0.693850468768046
218 0.708964434507415
219 0.723798449893041
220 0.73815298469162
221 0.751687271018153
222 0.765207997554826
223 0.778695716860107
224 0.791816383236463
225 0.805002994561672
226 0.817906567624001
227 0.830402796673184
228 0.842478147928807
229 0.854120129994471
230 0.86531526564109
231 0.876055632728627
232 0.886333294652734
233 0.896141901028535
234 0.905475503011649
235 0.914332753542556
236 0.922713013798837
237 0.930619185327432
238 0.938181181216476
239 0.945514411504069
240 0.952595432004501
241 0.959272852312643
242 0.965490356578877
243 0.971132262068408
244 0.976246240816608
245 0.980939165885052
246 0.985227714882203
247 0.989129733229155
248 0.99275840872855
249 0.996072829422528
};
\addlegendentry{BBRL-TRPL}
\end{axis}

\end{tikzpicture}

%% file: tikz_plots/reacher/reacher_sac_sparse.tex
\begin{tikzpicture}
\begin{axis}[
legend cell align={left},
legend style={
  fill opacity=0.8,
  draw opacity=1,
  text opacity=1,
  at={(0.97,0.03)},
  anchor=south east,
  draw=lightgray204
},
title={5D Reacher - Sparse},
tick align=outside,
tick pos=left,
x grid style={darkgray176},
xlabel={Number Environment Interactions},
xmajorgrids,
xmin=-1472000, xmax=31912000,
xtick style={color=black},
y grid style={darkgray176},
ylabel={Reward},
ymajorgrids,
ymin=-12500, ymax=5,
ytick style={color=black}
]
\path [draw=C8, fill=C8, opacity=0.2]
(axis cs:0,-804930.130645931)
--(axis cs:0,-1278894.09449712)
--(axis cs:400000,-16572.5208399925)
--(axis cs:800000,-27382.2598870615)
--(axis cs:1200000,-20781.0124473907)
--(axis cs:1600000,-45148.294311903)
--(axis cs:2000000,-20082.5248534612)
--(axis cs:2400000,-17448.479055946)
--(axis cs:2800000,-103206.15146879)
--(axis cs:3200000,-23192.3845824757)
--(axis cs:3600000,-58283.9074550735)
--(axis cs:4000000,-28636.3524705853)
--(axis cs:4400000,-32545.1620167667)
--(axis cs:4800000,-14407.48670961)
--(axis cs:5200000,-33564.2880822662)
--(axis cs:5600000,-22920.4363017282)
--(axis cs:6000000,-19840.4275005392)
--(axis cs:6400000,-26495.5362260327)
--(axis cs:6800000,-45718.811172274)
--(axis cs:7200000,-19548.5917888007)
--(axis cs:7600000,-36637.7472059447)
--(axis cs:8000000,-34488.2164246332)
--(axis cs:8400000,-17110.9504357945)
--(axis cs:8800000,-17693.3063968742)
--(axis cs:9200000,-14500.766995704)
--(axis cs:9600000,-17235.33149568)
--(axis cs:10000000,-19298.5469884752)
--(axis cs:10400000,-19243.3937680765)
--(axis cs:10800000,-16074.3957659797)
--(axis cs:11200000,-12364.5777672952)
--(axis cs:11600000,-20262.4735354012)
--(axis cs:12000000,-14126.63987363)
--(axis cs:12400000,-10643.4230272382)
--(axis cs:12800000,-16298.3035389125)
--(axis cs:13200000,-21273.3543641582)
--(axis cs:13600000,-14533.929151352)
--(axis cs:14000000,-14561.7585231272)
--(axis cs:14400000,-15066.2369797982)
--(axis cs:14800000,-11133.473057644)
--(axis cs:15200000,-14078.298017307)
--(axis cs:15600000,-15110.2952657327)
--(axis cs:16000000,-14872.3282812033)
--(axis cs:16400000,-16716.7497312435)
--(axis cs:16800000,-16139.973035384)
--(axis cs:17200000,-11400.5383604987)
--(axis cs:17600000,-10265.1840167385)
--(axis cs:18000000,-9981.72245764575)
--(axis cs:18400000,-9668.51891450025)
--(axis cs:18800000,-10541.4607907837)
--(axis cs:19200000,-16455.3922856197)
--(axis cs:19600000,-12449.0732760778)
--(axis cs:20000000,-16840.8545616757)
--(axis cs:20400000,-20849.0079780733)
--(axis cs:20800000,-13149.3912218525)
--(axis cs:21200000,-12774.3004152557)
--(axis cs:21600000,-16794.0186884842)
--(axis cs:22000000,-20000.8526771405)
--(axis cs:22400000,-20411.4860795577)
--(axis cs:22800000,-20245.3742300272)
--(axis cs:23200000,-20906.334386445)
--(axis cs:23600000,-22086.8820392477)
--(axis cs:24000000,-17500.7618157068)
--(axis cs:24400000,-8922.24153647975)
--(axis cs:24800000,-17873.7208003277)
--(axis cs:25200000,-12244.3871920452)
--(axis cs:25600000,-14638.9756391048)
--(axis cs:26000000,-11518.5623958585)
--(axis cs:26400000,-10327.412604866)
--(axis cs:26800000,-12112.9449201955)
--(axis cs:27200000,-27753.0977247535)
--(axis cs:27600000,-9231.7304188495)
--(axis cs:28000000,-12700.4111832012)
--(axis cs:28400000,-9233.65771457675)
--(axis cs:28800000,-12827.1825733297)
--(axis cs:29200000,-11917.257162812)
--(axis cs:29600000,-11621.0647866475)
--(axis cs:29600000,-4200.86796382875)
--(axis cs:29600000,-4200.86796382875)
--(axis cs:29200000,-4236.228010763)
--(axis cs:28800000,-3026.36408900125)
--(axis cs:28400000,-3809.1742956085)
--(axis cs:28000000,-4616.44681985225)
--(axis cs:27600000,-4063.27269159875)
--(axis cs:27200000,-7258.8986200075)
--(axis cs:26800000,-4051.84929265751)
--(axis cs:26400000,-2690.218663658)
--(axis cs:26000000,-4407.34407152625)
--(axis cs:25600000,-3517.62984714)
--(axis cs:25200000,-5596.52211195225)
--(axis cs:24800000,-3570.20632473925)
--(axis cs:24400000,-4177.52918310525)
--(axis cs:24000000,-4162.5543497175)
--(axis cs:23600000,-5147.1715048255)
--(axis cs:23200000,-4299.765578122)
--(axis cs:22800000,-3818.4895706375)
--(axis cs:22400000,-6412.125237483)
--(axis cs:22000000,-6994.67180463925)
--(axis cs:21600000,-5377.79959075225)
--(axis cs:21200000,-4000.334978613)
--(axis cs:20800000,-5432.06228207025)
--(axis cs:20400000,-5747.3403836475)
--(axis cs:20000000,-7664.5289134105)
--(axis cs:19600000,-6105.94830406025)
--(axis cs:19200000,-6310.594683876)
--(axis cs:18800000,-5017.9308956285)
--(axis cs:18400000,-3684.50408930725)
--(axis cs:18000000,-3482.55558418375)
--(axis cs:17600000,-4781.000901916)
--(axis cs:17200000,-5191.92466328226)
--(axis cs:16800000,-8259.04373392325)
--(axis cs:16400000,-4999.81107279025)
--(axis cs:16000000,-5940.2032437075)
--(axis cs:15600000,-5369.2695458555)
--(axis cs:15200000,-6451.91963178251)
--(axis cs:14800000,-5693.99233834325)
--(axis cs:14400000,-5764.1338813055)
--(axis cs:14000000,-4572.95982016425)
--(axis cs:13600000,-6966.8997944495)
--(axis cs:13200000,-8992.787871566)
--(axis cs:12800000,-5078.26902438575)
--(axis cs:12400000,-5591.434639889)
--(axis cs:12000000,-4982.9118448405)
--(axis cs:11600000,-5331.08390882225)
--(axis cs:11200000,-5700.426595168)
--(axis cs:10800000,-5618.57333190925)
--(axis cs:10400000,-3770.04573435775)
--(axis cs:10000000,-5696.608614398)
--(axis cs:9600000,-5722.98982227325)
--(axis cs:9200000,-7165.0507746175)
--(axis cs:8800000,-4952.021905382)
--(axis cs:8400000,-5077.9693937625)
--(axis cs:8000000,-4514.38141022476)
--(axis cs:7600000,-6174.05420301075)
--(axis cs:7200000,-6431.02442594125)
--(axis cs:6800000,-12637.3795918443)
--(axis cs:6400000,-9344.06129204726)
--(axis cs:6000000,-7043.449038697)
--(axis cs:5600000,-6866.70914288075)
--(axis cs:5200000,-10298.964978899)
--(axis cs:4800000,-5325.96123530275)
--(axis cs:4400000,-5626.63362145)
--(axis cs:4000000,-7713.0795632195)
--(axis cs:3600000,-8772.63975426075)
--(axis cs:3200000,-8827.36879114651)
--(axis cs:2800000,-10754.1427100465)
--(axis cs:2400000,-5348.1663490705)
--(axis cs:2000000,-6949.42007200225)
--(axis cs:1600000,-7823.32898501725)
--(axis cs:1200000,-6636.08707317825)
--(axis cs:800000,-5108.77938039)
--(axis cs:400000,-7201.46020312325)
--(axis cs:0,-804930.130645931)
--cycle;

\addplot [thick, C8, mark=*, mark size=0, mark options={solid}]
table {%
0 -1064468.61451158
400000 -10978.49556842
800000 -10397.16034207
1200000 -10310.74732736
1600000 -14283.00320912
2000000 -11712.0594854
2400000 -10599.77788235
2800000 -22525.18933141
3200000 -14629.81311805
3600000 -17725.4117261
4000000 -16969.98937502
4400000 -10515.49615643
4800000 -8531.91742166
5200000 -16970.42190651
5600000 -13532.43151365
6000000 -12149.82933308
6400000 -15352.28316259
6800000 -23782.43799484
7200000 -10980.40945137
7600000 -15620.00137169
8000000 -13242.45653198
8400000 -8729.51952771
8800000 -7655.02173237
9200000 -10152.67378537
9600000 -9581.22727557
10000000 -11430.87293246
10400000 -7051.42121806
10800000 -9584.09670305
11200000 -8154.64006133
11600000 -9832.93436525
12000000 -7329.39464119
12400000 -7865.76107402
12800000 -8402.81601146
13200000 -14842.8480109
13600000 -10036.91055484
14000000 -7089.97142322
14400000 -8963.71806051
14800000 -8278.69872128
15200000 -9232.47638948
15600000 -8921.87787325
16000000 -8311.4958769
16400000 -8487.37746984
16800000 -11838.81749544
17200000 -8488.89542955
17600000 -7164.89055584
18000000 -5989.47279515
18400000 -6113.92729756
18800000 -7004.23499418
19200000 -9249.12027928
19600000 -8358.71472537
20000000 -11789.78151165
20400000 -10369.59041593
20800000 -8889.2402405
21200000 -6528.47439316
21600000 -7354.27993123
22000000 -10326.02595758
22400000 -11319.33403145
22800000 -8731.52957495
23200000 -7952.08222538
23600000 -10282.05082861
24000000 -8043.50565888
24400000 -6085.83881868
24800000 -8203.46478492
25200000 -7881.35099742
25600000 -8168.60751249
26000000 -6707.59894717
26400000 -4952.25825964
26800000 -6989.63463294
27200000 -14470.64041691
27600000 -5626.3505008
28000000 -7238.83479514
28400000 -6050.25242169
28800000 -6453.87455266
29200000 -7532.2779253
29600000 -7916.66048917
};
\addlegendentry{SAC}
\end{axis}
\end{tikzpicture}

%% file: appendix/hyperparameters.tex
For all methods, where applicable, we optimized the learning rate, sample size, batch size, number of layers, and the number of epochs. 
For all BBRL methods and NDP, we additionally optimized the number of basis functions. 
Moreover, we found that NDP requires tuning of the scale of the predicted DMP weights, which was hard-coded to 100 in the original code base.
However, this value only worked for the meta-world tasks, but not for the other tasks, hence we adjusted it to allow for a fair comparison.
The population size of ES is always half the number of samples because two function evaluations are used per parameter vector.

\begin{table}[ht]
\centering
\caption{Hyperparameters for the modified reacher experiments.}
\label{tab:reacher-HP}
\begin{adjustbox}{max width=\textwidth}
\begin{tabular}{lcccccccc}
                                 & PPO        & NDP        & TRPL       & SAC         & CMORE & ES         & BBRL-PPO   & BBRL-TRPL   \\ 
\hline
\multicolumn{9}{l}{}                                                                                                                  \\
number samples                   & 16000      & 16000      & 16000      & 1000        & 120   & 200        & 64         & 64          \\
GAE $\lambda$                    & 0.95       & 0.95       & 0.95       & 0.95        & n.a.  & n.a.       & n.a.       & n.a.        \\
discount factor                  & 0.99       & 0.99       & 0.99       & 0.99        & n.a.  & n.a.       & n.a.       & n.a.        \\
\multicolumn{9}{l}{}                                                                                                                  \\ 
\hline
\multicolumn{9}{l}{}                                                                                                                  \\
$\epsilon_\mu$/$\epsilon$        & n.a.       & n.a.       & 0.005      & n.a.        & 0.1   & n.a.       & n.a.       & 0.05        \\
$\epsilon_\Sigma$                & n.a.       & n.a.       & 0.0005     & n.a.        & n.a.  & n.a.       & n.a.       & 0.0005      \\
\multicolumn{9}{l}{}                                                                                                                  \\ 
\hline
\multicolumn{9}{l}{}                                                                                                                  \\
optimizer                        & adam       & adam       & adam       & adam        & n.a.  & adam       & adam       & adam        \\
epochs                           & 10         & 10         & 20         & 1000        & n.a.  & n.a.       & 100        & 100         \\
learning rate                    & 3e-4       & 3e-4       & 5e-5       & 3e-4        & n.a.  & 1e-2       & 3e-4       & 3e-4        \\
use critic                       & True       & True       & True       & True        & False & False      & False      & False       \\
epochs critic                    & 10         & 10         & 10         & 1000        & n.a.  & n.a.       & n.a.       & n.a.        \\
learning rate critic (and alpha) & 3e-4       & 3e-4       & 3e-4       & 3e-4        & n.a.  & n.a.       & n.a.       & n.a.        \\
number minibatches               & 32         & 32         & 64         & n.a.        & n.a.  & n.a.       & 1          & 1           \\
batch size                       & n.a.       & n.a.       & n.a.       & 256         & n.a.  & n.a.       & n.a.       & n.a.        \\
buffer size                      & n.a.       & n.a.       & n.a.       & 1e6         & n.a.  & n.a.       & n.a.       & n.a.        \\
learning starts                  & 0          & 0          & 0          & 10000       & 0     & 0          & 0          & 0           \\
polyak\_weight                   & n.a.       & n.a.       & n.a.       & 5e-3        & n.a.  & n.a.       & n.a.       & n.a.        \\
trust region loss weight         & n.a.       & n.a.       & 10         & n.a.        & n.a.  & n.a.       & n.a.       & 10          \\
\multicolumn{9}{l}{}                                                                                                                  \\ 
\hline
\multicolumn{9}{l}{}                                                                                                                  \\
normalized observations          & True       & True       & True       & False       & False & False      & False      & False       \\
normalized rewards               & True       & True       & False      & False       & False & False      & False      & False       \\
observation clip                 & 10.0       & 10.0       & n.a.       & n.a.        & n.a.  & n.a.       & n.a.       & n.a.        \\
reward clip                      & 10.0       & 10.0       & n.a.       & n.a.        & n.a.  & n.a.       & n.a.       & n.a.        \\
critic clip                      & 0.2        & 0.2        & n.a.       & n.a.        & n.a.  & n.a.       & 0.2        & n.a.        \\
importance ratio clip            & 0.2        & 0.2        & n.a.       & n.a.        & n.a.  & n.a.       & 0.2        & n.a.        \\
\multicolumn{9}{l}{}                                                                                                                  \\ 
\hline
\multicolumn{9}{l}{}                                                                                                                  \\
hidden layers                    & {[}32, 32] & {[}32, 32] & {[}32, 32] & {[}128,128] & n.a.  & {[}32, 32] & {[}32, 32] & {[}32, 32]  \\
hidden layers critic             & {[}32, 32] & {[}32, 32] & {[}32, 32] & {[}128,128] & n.a.  & n.a.       & n.a.       & n.a.        \\
hidden activation                & tanh       & tanh       & tanh       & relu        & n.a.  & tanh       & tanh       & tanh        \\
initial std                      & 1.0        & 1.0        & 1.0        & 1.0         & 1.0   & 1.0        & 1.0        & 1.0         \\
\multicolumn{9}{l}{}                                                                                                                  \\ 
\hline
\multicolumn{9}{l}{}                                                                                                                  \\
number basis functions           & n.a.       & 5          & n.a.       & n.a.        & 5     & n.a.       & 5          & 5           \\
number zero basis                & n.a.       & n.a.       & n.a.       & n.a.        & 1     & n.a.       & 1          & 1           \\
weight scale                     & n.a.       & 20         & n.a.       & n.a.        & n.a.  & n.a.       & n.a.       & n.a.       
\end{tabular}
\end{adjustbox}
\end{table}

\begin{table}[ht]
\centering
\caption{Hyperparameters for the box pushing experiments.}
\label{tab:boxpushing-HP}
\begin{adjustbox}{max width=\textwidth}
\begin{tabular}{lccccccc}
                                 & PPO          & NDP          & TRPL  & SAC   & ES    & BBRL-PPO     & BBRL-TRPL     \\ 
\hline
\multicolumn{8}{l}{}                                                                                                  \\
number samples                   & 16000        & 16000        & 16000 & 1000  & 250   & 160          & 160           \\
GAE $\lambda$                    & 0.95         & 0.95         & 0.95  & 0.95  & n.a.  & n.a.         & n.a.          \\
discount factor                  & 0.99         & 0.99         & 0.99  & 0.99  & n.a.  & n.a.         & n.a.          \\
\multicolumn{8}{l}{}                                                                                                  \\ 
\hline
\multicolumn{8}{l}{}                                                                                                  \\
$\epsilon_\mu$                   & n.a.         & n.a.         & 0.005      & n.a.  & n.a.  & n.a.         & 0.005         \\
$\epsilon_\Sigma$                & n.a.         & n.a.         & 0.00005      & n.a.  & n.a.  & n.a.         & 0.0005        \\
\multicolumn{8}{l}{}                                                                                                  \\ 
\hline
\multicolumn{8}{l}{}                                                                                                  \\
optimizer                        & adam         & adam         & adam  & adam  & adam  & adam         & adam          \\
epochs                           & 10           & 10           & 20      & 1000  & n.a.  & 100          & 100           \\
learning rate                    & 1e-4         & 1e-4         & 5e-5      & 1e-4  & 1e-2       & 1e-4         & 1e-4          \\
use critic                       & True         & True         & True  & True  & False & True         & True          \\
epochs critic                    & 10           & 10           & 10    & 1000  & n.a.  & 100          & 100           \\
learning rate critic (and alpha) & 1e-4         & 1e-4         & 2e-4      & 1e-4       & n.a.  & 1e-4         & 1e-4          \\
number minibatches               & 40           & 32           & 40      & n.a.  & n.a.  & 1            & 1             \\
batch size                       & n.a.         & n.a.         & n.a.  & 256   & n.a.  & n.a.         & n.a.          \\
buffer size                      & n.a.         & n.a.         & n.a.  & 1e6   & n.a.  & n.a.         & n.a.          \\
learning starts                  & 0            & 0            & 0     & 10000 & 0     & 0            & 0             \\
polyak\_weight                   & n.a.         & n.a.         & n.a.  & 5e-3  & n.a.  & n.a.         & n.a.          \\
trust region loss weight         & n.a.         & n.a.         & 10      & n.a.  & n.a.  & n.a.         & 25            \\
\multicolumn{8}{l}{}                                                                                                  \\ 
\hline
\multicolumn{8}{l}{}                                                                                                  \\
normalized observations          & True         & True         & True  & False & False & False        & False         \\
normalized rewards               & True         & True         & False & False & False & False        & False         \\
observation clip                 & 10.0         & 10.0         & n.a.  & n.a.  & n.a.  & n.a.         & n.a.          \\
reward clip                      & 10.0         & 10.0         & n.a.  & n.a.  & n.a.  & n.a.         & n.a.          \\
critic clip                      & 0.2          & 0.2          & n.a.  & n.a.  & n.a.  & 0.2          & n.a.          \\
importance ratio clip            & 0.2          & 0.2          & n.a.  & n.a.  & n.a.  & 0.2          & n.a.          \\
\multicolumn{8}{l}{}                                                                                                  \\ 
\hline
\multicolumn{8}{l}{}                                                                                                  \\
hidden layers                    & {[}256, 256] & {[}256, 256] & {[}256, 256]  &{[}256, 256]  &{[}256, 256]       & {[}128, 128] & {[}128, 128]  \\
hidden layers critic             & {[}256, 256] & {[}256, 256] & {[}256, 256]      &  {[}256, 256]     & n.a.  & {[}32, 32]   & {[}32, 32]    \\
hidden activation                & tanh         & tanh         & tanh  & relu  & tanh  & tanh         & relu          \\
initial std                      & 1.0          & 1.0          & 1.0   & 1.0   & 1.0   & 1.0          & 1.0           \\
\multicolumn{8}{l}{}                                                                                                  \\ 
\hline
\multicolumn{8}{l}{}                                                                                                  \\
number basis functions           & n.a.         & 5            & n.a.  & n.a.  & n.a.  & 5            & 5             \\
number zero basis                & n.a.         & n.a.         & n.a.  & n.a.  & n.a.  & 1            & 1             \\
weight scale                     & n.a.         & 10           & n.a.  & n.a.  & n.a.  & n.a.         & n.a.         
\end{tabular}
\end{adjustbox}
\end{table}

\begin{table}[h]
\centering
\centering
\caption{Hyperparameters for the Meta-World experiments.}
\label{tab:metaworld-HP}
\begin{adjustbox}{max width=\textwidth}
\begin{tabular}{lcccccc}
                                 & PPO          & NDP          & TRPL         & ES           & BBRL-PPO   & BBRL-TRPL   \\ 
\hline
\multicolumn{7}{l}{}                                                                                                    \\
number samples                   & 16000        & 16000        & 16000        & 200          & 16         & 16          \\
GAE $\lambda$                    & 0.95         & 0.95         & 0.95         & n.a.         & n.a.       & n.a.        \\
discount factor                  & 0.99         & 0.99         & 0.99         & n.a.         & n.a.       & n.a.        \\
\multicolumn{7}{l}{}                                                                                                    \\ 
\hline
\multicolumn{7}{l}{}                                                                                                    \\
$\epsilon_\mu$                   & n.a.         & n.a.         & 0.005        & n.a.         & n.a.       & 0.005       \\
$\epsilon_\Sigma$                & n.a.         & n.a.         & 0.0005       & n.a.         & n.a.       & 0.0005      \\
\multicolumn{7}{l}{}                                                                                                    \\ 
\hline
\multicolumn{7}{l}{}                                                                                                    \\
optimizer                        & adam         & adam         & adam         & adam         & adam       & adam        \\
epochs                           & 10           & 10           & 20           & n.a.         & 100        & 100         \\
learning rate                    & 3e-4         & 3e-4         & 5e-5         & 1e-2         & 3e-4       & 3e-4        \\
use critic                       & True         & True         & True         & False        & False      & False       \\
epochs critic                    & 10           & 10           & 10           & n.a.         & n.a.       & n.a.        \\
learning rate critic (and alpha) & 3e-4         & 3e-4         & 3e-4         & n.a.         & n.a.       & n.a.        \\
number minibatches               & 32           & 32           & 64           & n.a.         & 1          & 1           \\
trust region loss weight         & n.a.         & n.a.         & 10.0         & n.a.         & n.a.       & 10          \\
\multicolumn{7}{l}{}                                                                                                    \\ 
\hline
\multicolumn{7}{l}{}                                                                                                    \\
normalized observations          & True         & True         & True         & False        & False      & False       \\
normalized rewards               & True         & True         & False        & False        & False      & False       \\
observation clip                 & 10.0         & 10.0         & n.a.         & n.a.         & n.a.       & n.a.        \\
reward clip                      & 10.0         & 10.0         & n.a.         & n.a.         & n.a.       & n.a.        \\
critic clip                      & 0.2          & 0.2          & n.a.         & n.a.         & 0.2        & n.a.        \\
importance ratio clip            & 0.2          & 0.2          & n.a.         & n.a.         & 0.2        & n.a.        \\
\multicolumn{7}{l}{}                                                                                                    \\ 
\hline
\multicolumn{7}{l}{}                                                                                                    \\
hidden layers                    & {[}128, 128] & {[}128, 128] & {[}128, 128] & {[}128, 128] & {[}32, 32] & {[}32, 32]  \\
hidden layers critic             & {[}128, 128] & {[}128, 128] & {[}128, 128] & n.a.         & n.a.       & n.a.        \\
hidden activation                & tanh         & tanh         & tanh         & tanh         & tanh       & relu        \\
initial std                      & 1.0          & 1.0          & 1.0          & 1.0          & 1.0        & 10.0        \\
\multicolumn{7}{l}{}                                                                                                    \\ 
\hline
\multicolumn{7}{l}{}                                                                                                    \\
number basis functions           & n.a.         & 5            & n.a.         & n.a.         & 5          & 5           \\
number zero basis                & n.a.         & n.a.         & n.a.         & n.a.         & 1          & 1           \\
weight scale                     & n.a.         & 100          & n.a.         & n.a.         & n.a.       & n.a.       
\end{tabular}
\end{adjustbox}
\end{table}

\begin{table}[h]
\centering
\centering
\caption{Hyperparameters for the hopper jumping experiments.}
\label{tab:hopper-HP}
\begin{adjustbox}{max width=\textwidth}
\begin{tabular}{lccccccc}
                                 & PPO          & TRPL         & SAC          & CMORE & ES           & BBRL-PPO   & BBRL-TRPL   \\ 
\hline
\multicolumn{8}{l}{}                                                                                              \\
number samples                   & 16384        & 16384        & 1000         & 60    & 200          & 64         & 64          \\
GAE $\lambda$                    & 0.95         & 0.95         & 0.95         & n.a.  & n.a.         & n.a.       & n.a.        \\
discount factor                  & 0.99         & 0.99         & 0.99         & n.a.  & n.a.         & n.a.       & n.a.        \\
\multicolumn{8}{l}{}                                                                                              \\ 
\hline
\multicolumn{8}{l}{}                                                                                              \\
$\epsilon_\mu$/$\epsilon$        & n.a.         & 0.005        & n.a.         & 0.1   & n.a.         & n.a.       & 0.005       \\
$\epsilon_\Sigma$                & n.a.         & 0.00005      & n.a.         & n.a.  & n.a.         & n.a.       & 0.0005      \\
\multicolumn{8}{l}{}                                                                                              \\ 
\hline
\multicolumn{8}{l}{}                                                                                              \\
optimizer                        & adam         & adam         & adam         & n.a.  & adam         & adam       & adam        \\
epochs                           & 10           & 20           & 1000         & n.a.  & n.a.         & 100        & 100         \\
learning rate                    & 3e-4         & 5e-5         & 1e-4         & n.a.  & 0.01         & 1e-4       & 5e-5        \\
use critic                       & True         & True         & True         & False & False        & False      & False       \\
epochs critic                    & 10           & 10           & 1000         & n.a.  & n.a.         & n.a.       & n.a.        \\
learning rate critic (and alpha) & 3e-4         & 3e-4         & 1e-4         & n.a.  & n.a.         & n.a.       & n.a.        \\
number minibatches               & 32           & 64           & n.a.         & n.a.  & n.a.         & 1          & 1           \\
batch size                       & n.a.         & n.a.         & 256          & n.a.  & n.a.         & n.a.       & n.a.        \\
buffer size                      & n.a.         & n.a.         & 1e6          & n.a.  & n.a.         & n.a.       & n.a.        \\
learning starts                  & 0            & 0            & 10000        & 0     & 0            & 0          & 0           \\
polyak\_weight                   & n.a.         & n.a.         & 5e-3         & n.a.  & n.a.         & n.a.       & n.a.        \\
trust region loss weight         & n.a.         & 10           & n.a.         & n.a.  & n.a.         & n.a.       & 25          \\
\multicolumn{8}{l}{}                                                                                              \\ 
\hline
\multicolumn{8}{l}{}                                                                                              \\
normalized observations          & True         & True         & False        & False & False        & False      & False       \\
normalized rewards               & True         & False        & False        & False & False        & False      & False       \\
observation clip                 & 10.0         & n.a.         & n.a.         & n.a.  & n.a.         & n.a.       & n.a.        \\
reward clip                      & 10.0         & n.a.         & n.a.         & n.a.  & n.a.         & n.a.       & n.a.        \\
critic clip                      & 0.2          & n.a.         & n.a.         & n.a.  & n.a.         & 0.2        & n.a.        \\
importance ratio clip            & 0.2          & n.a.         & n.a.         & n.a.  & n.a.         & 0.2        & n.a.        \\
\multicolumn{8}{l}{}                                                                                              \\ 
\hline
\multicolumn{8}{l}{}                                                                                              \\
hidden layers                    & {[}128, 128] & {[}128, 128] & {[}128, 128] & n.a   & {[}128, 128] & {[}32, 32] & {[}32, 32]  \\
hidden layers critic             & {[}128, 128] & {[}128, 128] & {[}128, 128] & n.a   & n.a          & n.a        & n.a         \\
hidden activation                & tanh         & tanh         & relu         & n.a.  & tanh         & tanh       & tanh        \\
initial std                      & 1.0          & 1.0          & 1.0          & 1.0   & 1.0          & 1.0        & 1.0         \\
\multicolumn{8}{l}{}                                                                                              \\ 
\hline
\multicolumn{8}{l}{}                                                                                              \\
number basis functions           & n.a.         & n.a.         & n.a.         & 5    & n.a.         & 5          & 5           \\
number zero basis                & n.a.         & n.a.         & n.a.         & 1    & n.a.         & 1          & 1           \\
\end{tabular}
\end{adjustbox}
\end{table}

\begin{table}[h]
\centering
\centering
\caption{Hyperparameters for the Beer Pong experiments.}
\label{tab:beerpong}
\begin{adjustbox}{max width=\textwidth}
\begin{tabular}{lcccc}
                                 & PPO          & CMORE & BBRL-PPO   & BBRL-TRPL   \\ 
\hline
\multicolumn{5}{l}{}                                                               \\
number samples                   & 16384        & 60    & 160        & 160         \\
GAE $\lambda$                    & 0.95         & n.a.  & n.a.       & n.a.        \\
discount factor                  & 0.99         & n.a.  & n.a.       & n.a.        \\
\multicolumn{5}{l}{}                                                               \\ 
\hline
\multicolumn{5}{l}{}                                                               \\
$\epsilon_\mu$/$\epsilon$        & n.a.         & 0.1   & n.a.       & 0.005       \\
$\epsilon_\Sigma$                & n.a.         & n.a.  & n.a.       & 0.0005      \\
\multicolumn{5}{l}{}                                                               \\ 
\hline
\multicolumn{5}{l}{}                                                               \\
optimizer                        & adam         & n.a.  & adam       & adam        \\
epochs                           & 10           & n.a.  & 100        & 100         \\
learning rate                    & 3e-4         & n.a.  & 1e-4       & 5e-5        \\
use critic                       & True         & False & False      & False       \\
epochs critic                    & 10           & n.a.  & n.a.       & n.a.        \\
learning rate critic (and alpha) & 3e-4         & n.a.  & n.a.       & n.a.        \\
number minibatches               & 32           & n.a.  & 1          & 1           \\
trust region loss weight         & n.a.         & n.a.  & n.a.       & 25          \\
\multicolumn{5}{l}{}                                                               \\ 
\hline
\multicolumn{5}{l}{}                                                               \\
normalized observations          & True         & False & False      & False       \\
normalized rewards               & True         & False & False      & False       \\
observation clip                 & 10.0         & n.a.  & n.a.       & n.a.        \\
reward clip                      & 10.0         & n.a.  & n.a.       & n.a.        \\
critic clip                      & 0.2          & n.a.  & 0.2        & n.a.        \\
importance ratio clip            & 0.2          & n.a.  & 0.2        & n.a.        \\
\multicolumn{5}{l}{}                                                               \\ 
\hline
\multicolumn{5}{l}{}                                                               \\
hidden layers                    & {[}128, 128] & n.a.  & {[}32, 32] & {[}32, 32]  \\
hidden layers critic             & {[}128, 128] & n.a.  & n.a.       & n.a.        \\
hidden activation                & tanh         & n.a.  & tanh       & tanh        \\
initial std                      & 1.0          & 1.0   & 1.0        & 1.0         \\
\multicolumn{5}{l}{}                                                               \\ 
\hline
\multicolumn{5}{l}{}                                                               \\
number basis functions           & n.a.         & 2     & 2          & 2           \\
number zero basis                & n.a.         & 2     & 2          & 2           \\
\end{tabular}
\end{adjustbox}
\end{table}

\begin{table}[h]
\centering
\centering
\caption{Hyperparameters for the Table Tennis experiments.}
\label{tab:table tennis}
\begin{adjustbox}{max width=\textwidth}
\begin{tabular}{lcccc}
                                 & PPO          & TRPL         & BBRL-PPO & BBRL-TRPL  \\ 
\hline
\multicolumn{5}{l}{}                                                                   \\
number samples                   & 16000        & 16000        & 200      & 200        \\
GAE $\lambda$                    & 0.95         & 0.95         & n.a.     & n.a.       \\
discount factor                  & 0.99         & 0.99         & n.a.     & n.a.       \\
\multicolumn{5}{l}{}                                                                   \\ 
\hline
\multicolumn{5}{l}{}                                                                   \\
$\epsilon_\mu$                   & n.a.         & 0.0005       & n.a.     & 0.0005     \\
$\epsilon_\Sigma$                & n.a.         & 0.00005      & n.a.     & 0.00005    \\
\multicolumn{5}{l}{}                                                                   \\ 
\hline
\multicolumn{5}{l}{}                                                                   \\
optimizer                        & adam         & adam         & adam     & adam       \\
epochs                           & 10           & 20           & 100      & 100        \\
learning rate                    & 1e-4         & 5e-5         & 1e-4     & 3e-4       \\
use critic                       & True         & True         & True     & True       \\
epochs critic                    & 10           & 10           & 100      & 100        \\
learning rate critic (and alpha) & 1e-4         & 1e-4         & 1e-4     & 3e-4       \\
number minibatches               & 40           & 40           & 1        & 1          \\
trust region loss weight         & n.a.         & 10.0         & n.a.     & 25         \\
\multicolumn{5}{l}{}                                                                   \\ 
\hline
\multicolumn{5}{l}{}                                                                   \\
normalized observations          & True         & True         & False    & False      \\
normalized rewards               & True         & False        & False    & False      \\
observation clip                 & 10.0         & n.a.         & n.a.     & n.a.       \\
reward clip                      & 10.0         & n.a.         & n.a.     & n.a.       \\
critic clip                      & 0.2          & n.a.         & 0.2      & n.a.       \\
importance ratio clip            & 0.2          & n.a.         & 0.2      & n.a.       \\
\multicolumn{5}{l}{}                                                                   \\ 
\hline
\multicolumn{5}{l}{}                                                                   \\
hidden layers                    & {[}256, 256] & {[}256, 256] & {[}256]  & {[}256]    \\
hidden layers critic             & {[}256, 256] & {[}256, 256] & n.a.     & n.a.       \\
hidden activation                & tanh         & tanh         & tanh     & relu       \\
initial std                      & 1.0          & 1.0          & 1.0      & 1.0        \\
\multicolumn{5}{l}{}                                                                   \\ 
\hline
\multicolumn{5}{l}{}                                                                   \\
number basis functions           & n.a.         & n.a.         & 3        & 3          \\
number zero basis                & n.a.         & n.a.         & 1        & 1          \\
\end{tabular}
\end{adjustbox}
\end{table}